\documentclass[10pt]{article} %

\let\OrigAddcontentsline\addcontentsline
\PassOptionsToPackage{dvipsnames,svgnames,x11names,table}{xcolor}
\usepackage[preprint]{tmlr_official}
\let\addcontentsline\OrigAddcontentsline
\usepackage[T1]{fontenc}
\usepackage{lmodern}
\usepackage[dvipsnames,svgnames,x11names,table]{xcolor}
\usepackage{tabularx}
\usepackage{natbib}
\setcitestyle{authoryear,round,citesep={;},aysep={,},yysep={;}}
\usepackage{fancyhdr}
\usepackage{amsmath,amsfonts,bm}

\def\eqref#1{equation~\ref{#1}}

\def\1{\bm{1}}

\DeclareMathAlphabet{\mathsfit}{\encodingdefault}{\sfdefault}{m}{sl}
\SetMathAlphabet{\mathsfit}{bold}{\encodingdefault}{\sfdefault}{bx}{n}

\usepackage{microtype}          %
\usepackage{tikz}               %
\usepackage{enumitem}
\usepackage{hyperref}
\usepackage{cleveref}
\usepackage{url}
\usepackage{booktabs}
\usepackage{pifont}
\usepackage{fontawesome5}       %
\usepackage{CJKutf8}
\newcommand{\cmark}{\ding{51}}
\newcommand{\xmark}{\ding{55}}
\definecolor{oxfordblue}{HTML}{002147}
\definecolor{oxfordgold}{HTML}{C9A84C}
\usepackage{tcolorbox}
\usepackage{forest}
\usepackage{adjustbox}
\usepackage{wrapfig}
\useforestlibrary{edges}
\usepackage{ragged2e}
\tcbuselibrary{breakable,skins,theorems}
\usepackage{csquotes}

\usepackage{amsmath}
\usepackage{amssymb}
\usepackage{graphicx}
\usepackage{subcaption}
\usepackage{bbm}
\usepackage{mathtools}

\newif\ifcomments
\commentstrue  %

\definecolor{narmeenpurple}{RGB}{128,0,128}

\definecolor{aaronblue}{RGB}{0,102,204}

\definecolor{defblue}{RGB}{0,51,102}
\definecolor{examplegreen}{RGB}{0,102,51}
\definecolor{insightgold}{RGB}{153,102,0}
\definecolor{theorypurple}{RGB}{102,0,153}
\definecolor{warningred}{RGB}{153,0,0}
\definecolor{tableheader}{RGB}{70,130,180}
\definecolor{tablerow1}{RGB}{240,248,255}
\definecolor{tablerow2}{RGB}{255,255,255}
\definecolor{inductionorange}{HTML}{BD5836
}
\usepackage[T1]{fontenc}

\newtcolorbox{definitionbox}[1][]{
    colback=defblue!5,
    colframe=defblue!80,
    fonttitle=\bfseries,
    title=#1,
    breakable,
    enhanced,
    attach boxed title to top left={yshift=-2mm,xshift=4mm},
    boxed title style={colback=defblue!80,colframe=defblue!80}
}

\newtcolorbox{examplebox}[1][]{
    colback=examplegreen!5,
    colframe=examplegreen!70,
    fonttitle=\bfseries,
    title=#1,
    breakable,
    enhanced,
    attach boxed title to top left={yshift=-2mm,xshift=4mm},
    boxed title style={colback=examplegreen!70,colframe=examplegreen!70}
}

\newtcolorbox{insightbox}[1][]{
    colback=insightgold!8,
    colframe=insightgold!80,
    fonttitle=\bfseries,
    title=#1,
    breakable,
    enhanced,
    attach boxed title to top left={yshift=-2mm,xshift=4mm},
    boxed title style={colback=insightgold!80,colframe=insightgold!80}
}

\newtcolorbox{theorybox}[1][]{
    colback=theorypurple!5,
    colframe=theorypurple!70,
    fonttitle=\bfseries,
    title=#1,
    breakable,
    enhanced,
    attach boxed title to top left={yshift=-2mm,xshift=4mm},
    boxed title style={colback=theorypurple!70,colframe=theorypurple!70}
}

\newtcolorbox{warningbox}[1][]{
    colback=warningred!5,
    colframe=warningred!70,
    fonttitle=\bfseries,
    title=#1,
    breakable,
    enhanced,
    attach boxed title to top left={yshift=-2mm,xshift=4mm},
    boxed title style={colback=warningred!70,colframe=warningred!70}
}

\definecolor{practiceteal}{RGB}{0,102,102}
\newtcolorbox{practicebox}[1][]{
    colback=practiceteal!5,
    colframe=practiceteal!70,
    fonttitle=\bfseries,
    title=#1,
    breakable,
    enhanced,
    attach boxed title to top left={yshift=-2mm,xshift=4mm},
    boxed title style={colback=practiceteal!70,colframe=practiceteal!70}
}

\definecolor{citeestablished}{HTML}{B86046}  %
\definecolor{citeemerging}{HTML}{6B8E23}     %
\definecolor{citeflagged}{HTML}{FF4C4C}      %

\newcommand{\citeest}[1]{%
  \begingroup
  \hypersetup{citecolor=citeestablished}%
  \citep{#1}%
  \endgroup
}

\newcommand{\citeem}[1]{%
  \begingroup
  \hypersetup{citecolor=citeemerging}%
  \citep{#1}%
  \endgroup
}

\newcommand{\citetest}[1]{%
  \begingroup
  \hypersetup{citecolor=citeestablished}%
  \citet{#1}%
  \endgroup
}

\newcommand{\citetem}[1]{%
  \begingroup
  \hypersetup{citecolor=citeemerging}%
  \citet{#1}%
  \endgroup
}

\newcommand{\citeestalp}[1]{%
  \begingroup
  \hypersetup{citecolor=citeestablished}%
  \citealp{#1}%
  \endgroup
}

\newcommand{\citeemalp}[1]{%
  \begingroup
  \hypersetup{citecolor=citeemerging}%
  \citealp{#1}%
  \endgroup
}

\definecolor{mydarkorange}{HTML}{B86046}
\definecolor{codegreen}{rgb}{0,0.6,0}
\definecolor{codegray}{rgb}{0.5,0.5,0.5}
\definecolor{codepurple}{rgb}{0.58,0,0.82}
\definecolor{backcolour}{rgb}{0.95,0.95,0.92}
\definecolor{bg}{rgb}{0.95,0.95,0.95}
\definecolor{mygreen}{rgb}{0,0.6,0}
\definecolor{lightgrey}{rgb}{0.925, 0.925, 0.925}
\definecolor{teal}{rgb}{0.078, 0.549, 0.549}
\hypersetup{
    colorlinks=true,
    linkcolor=mydarkorange,
    citecolor=mydarkorange,
    filecolor=mydarkorange,
    urlcolor=mydarkorange
}

\title{Understanding AI from the Inside: \\
The State of Mechanistic Interpretability for AI Safety}

{\setlength{\columnwidth}{\textwidth}\footnotetext{%
$^\dagger$ Core contributors.
\quad $^\ddagger$ Senior advisors.
\quad Remaining authors listed alphabetically.\\
\quad $^*$ Correspondence: \text{fazl@robots.ox.ac.uk}
}}

\author{%
\begin{minipage}{\textwidth}
\centering
{\normalsize\normalfont\color{black}
Fazl Barez$^{1,2,\dagger,*}$ \qquad Tung-Yu Wu$^{1,\dagger}$\endgraf
}
\vspace{6pt}
{\normalsize\normalfont\color{black}
\mbox{Hunar Batra$^{1}$} \quad \mbox{Gabriele Dominici$^{3}$} \quad \mbox{Florent Draye$^{4}$} \quad \mbox{Yanai Elazar$^{5}$} \quad \mbox{Damiano Fornasiere$^{6}$} \quad \mbox{Ant\'{\i}a Garc\'{\i}a$^{2}$} \quad \mbox{Abir Harrasse$^{2}$} \quad \mbox{Riley Harris$^{1}$} \quad \mbox{Sara Hooker$^{7}$} \quad \mbox{Michael Lan$^{2}$} \quad \mbox{Belinda Z. Li$^{8,9}$} \quad \mbox{Aaron Maiwald$^{1}$} \quad \mbox{Narmeen Fatimah Oozeer$^{2}$} \quad \mbox{Naomi Saphra$^{10}$} \quad \mbox{Or David Shafran$^{11}$} \quad \mbox{Adi Simhi$^{12}$} \quad \mbox{Robert Trager$^{1}$} \quad \mbox{Shriyash Kaustubh Upadhyay$^{2}$} \quad \mbox{Ziyu Yao$^{13}$} \quad \mbox{Yu Zhao$^{1,14}$}\endgraf
}
\vspace{6pt}
{\normalsize\normalfont\color{black}
\mbox{Shay B. Cohen$^{14,\ddagger}$} \quad \mbox{Mor Geva$^{11,\ddagger}$} \quad \mbox{Yoshua Bengio$^{15,6,\ddagger}$} \quad \mbox{Stella Biderman$^{16,\ddagger}$} \quad \mbox{Hassan Sajjad$^{17,\ddagger}$}\endgraf
}
\vspace{12pt}
{\small\normalfont\color[HTML]{002147}
\mbox{$^{1}$University of Oxford} \quad \mbox{$^{2}$Martian} \quad \mbox{$^{8}$MIT} \quad \mbox{$^{9}$Anthropic}\\
\mbox{$^{3}$Universit\`a della Svizzera italiana} \quad \mbox{$^{5}$Bar-Ilan University} \quad \mbox{$^{6}$LawZero}\\
\mbox{$^{4}$Max Planck Institute for Intelligent Systems} \quad \mbox{$^{7}$Adaption Labs} \quad \mbox{$^{10}$Boston University}\\
\mbox{$^{11}$Tel Aviv University} \quad \mbox{$^{12}$Technion} \quad \mbox{$^{13}$George Mason University} \quad \mbox{$^{14}$University of Edinburgh}\\
\mbox{$^{15}$Mila, Universit\'e de Montr\'eal} \quad \mbox{$^{16}$EleutherAI} \quad \mbox{$^{17}$New York University Abu Dhabi \& Dalhousie University}\endgraf
}
\end{minipage}%
}

{} %

\begin{document}

\maketitle

\begin{abstract}

Neural networks are likely to be among the most consequential artefacts in human history. As these systems become more capable and autonomous, their potential to cause societal harm also grows, making AI safety one of the defining challenges of our time. Yet we continue to build increasingly powerful systems without fully understanding how they work, how they will behave in unfamiliar situations, or how to reliably control them. Mechanistic interpretability offers a path towards understanding these systems by studying their internal representations and computations, with the aim of evaluating, predicting, and controlling behaviour. We survey mechanistic interpretability for AI safety in order to answer the question: \textbf{Does understanding how models work actually make them safer?} We place mechanistic interpretability in the broader context of interpretability research and compare its mechanistic evidence with behavioural evaluation. We discuss how these approaches can work together to improve model understanding and safeguards. We introduce the foundations of mechanistic interpretability, including features, polysemanticity, transformer computation, and circuits, and survey methods for feature extraction and analysis, circuit discovery and causal verification, explaining intermediate representations in natural language, learning dynamics, and training-data attribution. Beyond surveying existing methods, we propose a unified framework for evaluating mechanistic interpretability claims and characterise six research orientations by their goals, assumptions, and trade-offs. Finally, we examine how mechanistic understanding can support AI safety, including knowledge editing, steering, and monitoring, and assess its limitations, failure modes, open problems, and relationship to behavioural safety measures. We also release a repository of code notebooks with worked examples that reproduce key findings from methods surveyed here, allowing readers to build intuition through hands-on experimentation.

\end{abstract}

{\centering\small
\href{https://github.com/TSGLAB/Interpretability_Survey_Safety}{\textcolor{oxfordblue}{\faGithub\ Notebooks}}%
\hspace{2em}%
\href{https://tsglab.github.io/mi_survey/}{\textcolor{oxfordblue}{\faGlobe\ Website}}%
\par}
\newpage

\tikzset{
  outline-box/.style={
    rectangle,
    draw=black, %
    rounded corners,
    text opacity=1,
    minimum height=1.5em,
    inner sep=2pt,
    align=center,
    fill opacity=.9,
    line width=0.6pt,
  },
  outline-leaf/.style={
    outline-box,
    align=left,
    inner xsep=3pt,
    inner ysep=4pt,
  }
}

\newcommand{\figleafwidth}{46em}
\newcommand{\citeleaf}[1]{%
  \parbox[t]{\linewidth}{%
    \RaggedRight\footnotesize
    \setlength{\parskip}{0pt}\setlength{\parindent}{0pt}%
    \setlength{\emergencystretch}{2em}%
    #1%
  }%
}
\newcommand{\cites}[1]{\cite{#1}}

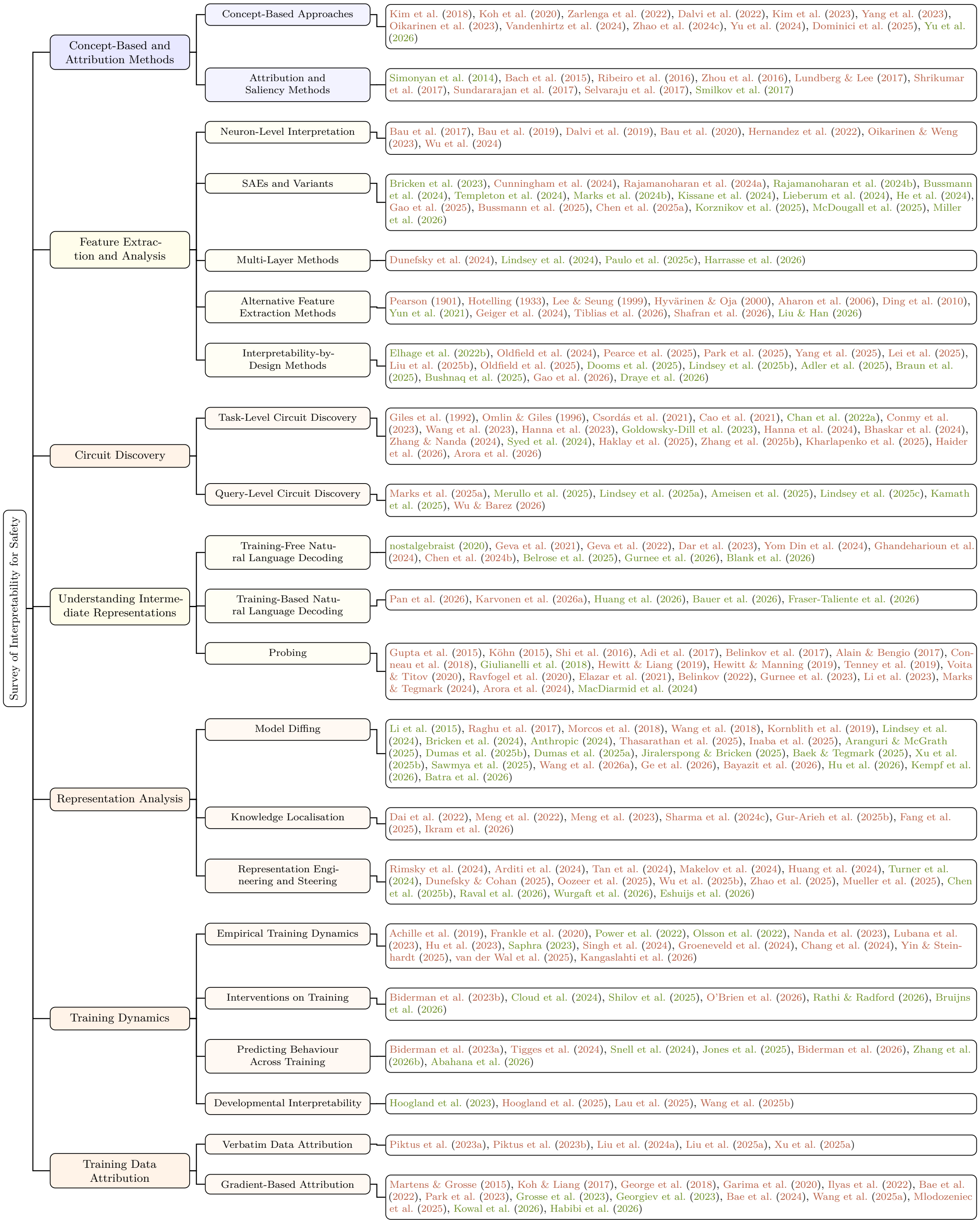
\begin{figure*}[!htp]
\centering
\adjustbox{width=\linewidth, max height=0.88\textheight}{%
\begin{forest}
forked edges,
for tree={
  grow=east,
  reversed=true,
  anchor=base west,
  parent anchor=east,
  child anchor=west,
  base=center,
  font=\small,
  outline-box,
  edge+={line width=0.8pt},
  s sep=15pt,
  inner xsep=5pt,
  inner ysep=5pt,
  ver/.style={rotate=90, child anchor=north, parent anchor=south, anchor=center, transform shape},
},
where level=1{text width=10em, font=\small, text centered}{},
where level=2{text width=12em, font=\footnotesize, text centered}{},
[
  {Survey of Interpretability for Safety},
  ver,
  [
    {Concept-Based and Attribution Methods},
    fill=blue!10
    [
      {Concept-Based Approaches},
      fill=blue!5
      [
        {\citeleaf{\citetest{kim2018tcav}, \citetest{koh2020conceptbottleneckmodels}, \citetest{zarlenga2022cem}, \citetest{dalvi2022discovering}, \citetest{kim2023probabilistic}, \citetest{yang2023language}, \citetest{oikarinen2023labelfree}, \citetest{vandenhirtz2024stochastic}, \citetest{zhao-etal-2024-explaining}, \citetest{yu-etal-2024-latent}, \citetest{dominici2025causal}, \citetem{yu2026vectorquantizedlatentconcepts}}},
        outline-leaf, text width=\figleafwidth
      ]
    ]
    [
      {Attribution and Saliency Methods},
      fill=blue!5
      [
        {\citeleaf{\citetem{simonyan2014deepinsideconvolutionalnetworks}, \citetest{lrp}, \citetest{ribeiro2016whyitrustyou}, \citetest{cam}, \citetest{lundberg2017unifiedapproachinterpretingmodel}, \citetest{deeplift}, \citetest{ig}, \citetest{gradcam}, \citetem{smilkov2017smoothgradremovingnoiseadding}}},
        outline-leaf, text width=\figleafwidth
      ]
    ]
  ]
  [
    {Feature Extraction and Analysis},
    fill=yellow!10
    [
      {Neuron-Level Interpretation},
      fill=yellow!5
      [
        {\citeleaf{\citetest{netdissect2017}, \citetest{bau2018identifying}, \citetest{Dalvi_Durrani_Sajjad_Belinkov_Bau_Glass_2019}, \citetest{bau2020units}, \citetest{hernandez2022natural}, \citetest{oikarinen2023clipdissect}, \citetest{and}}},
        outline-leaf, text width=\figleafwidth
      ]
    ]
    [
      {SAEs and Variants},
      fill=yellow!5
      [
        {\citeleaf{\citetem{bricken2023monosemanticity}, \citetest{cunningham2023sparse}, \citetest{rajamanoharan2024improvingdictionarylearninggated}, \citetem{rajamanoharan2024jumpingaheadimprovingreconstruction}, \citetem{bussmann2024batchtopksparseautoencoders}, \citetem{templeton2024scaling}, \citetem{marks2024enhancingneuralnetworkinterpretability}, \citetem{kissane2024interpretingattentionlayeroutputs}, \citetem{lieberum2024gemmascopeopensparse}, \citetem{he2024llamascopeextractingmillions}, \citetest{gao2024scalingevaluatingsparseautoencoders}, \citetest{bussmann2025learning}, \citetest{chen2025low}, \citetem{korznikov2025ortsaeorthogonalsparseautoencoders}, \citetem{mcdougall2025gemmascope2}, \citetem{miller2026identifyingintervenableinterpretablefeatures}}},
        outline-leaf, text width=\figleafwidth
      ]
    ]
    [
      {Multi-Layer Methods},
      fill=yellow!5
      [
        {\citeleaf{\citetest{dunefsky2024transcodersinterpretablellmfeature}, \citetem{lindsey2024sparsecrosscoders}, \citetem{paulo2025transcodersbeatsparseautoencoders}, \citetem{harrasse2025tracing}}},
        outline-leaf, text width=\figleafwidth
      ]
    ]
    [
      {Alternative Feature Extraction Methods},
      fill=yellow!5
      [
        {\citeleaf{\citetest{pearson1901lines}, \citetest{hotelling1933analysis}, \citetest{lee1999learning}, \citetest{hyvarinen2000independent}, \citetest{aharon2006k}, \citetest{snmf}, \citetem{yun-etal-2021-transformer}, \citetest{das}, \citetest{smds}, \citetest{shafran-etal-2026-constructing}, \citetem{liu2026icalensinterpretinglanguage}}},
        outline-leaf, text width=\figleafwidth
      ]
    ]
    [
      {Interpretability-by-Design Methods},
      fill=yellow!5
      [
        {\citeleaf{\citetem{elhage2022solu}, \citetest{oldfield2024multilinear}, \citetest{pearce2025bilinear}, \citetest{park2025monet}, \citetest{yang2025mixture}, \citetest{lei2025spartan}, \citetest{liu2025kan}, \citetest{oldfield2025towards}, \citetem{dooms2025compositionality}, \citetem{anthropic2025molt}, \citetem{adler2025towards}, \citetem{braun2025interpretability}, \citetem{bushnaq2025stochastic}, \citetest{gao2025weight}, \citetem{draye2025spartan}}},
        outline-leaf, text width=\figleafwidth
      ]
    ]
  ]
  [
    {Circuit Discovery},
    fill=orange!10
    [
      {Task-Level Circuit Discovery},
      fill=orange!5
      [
        {\citeleaf{\citetest{rnnfsa1992}, \citetest{OMLIN199641}, \citetest{netmodular}, \citetest{cao-etal-2021-low}, \citetem{chan2022causal}, \citetest{conmy2023automatedcircuitdiscoverymechanistic}, \citetest{wang2022interpretabilitywildcircuitindirect}, \citetest{hanna2023how}, \citetem{goldowskydill2023localizingmodelbehaviorpath}, \citetest{eapig}, \citetest{bhaskar2024finding}, \citetest{zhang2024towards}, \citetem{syed2023attributionpatchingoutperformsautomated}, \citetest{haklay-etal-2025-position}, \citetest{eapgp}, \citetest{kharlapenko2025scaling}, \citetest{haider2026multigranular}, \citetest{arora2026language}}},
        outline-leaf, text width=\figleafwidth
      ]
    ]
    [
      {Query-Level Circuit Discovery},
      fill=orange!5
      [
        {\citeleaf{\citetest{marks2025sparse}, \citetem{merullo2025replicating}, \citetem{lindsey2025landscape}, \citetem{ameisen2025circuit}, \citetem{lindsey2025biology}, \citetem{kamath2025tracing}, \citetest{querycircuit}}},
        outline-leaf, text width=\figleafwidth
      ]
    ]
  ]
  [
    {Understanding Intermediate Representations},
    fill=yellow!10
    [
      {Training-Free Natural Language Decoding},
      fill=yellow!5
      [
        {\citeleaf{\citetem{nostalgebraist2020interpretinggpt}, \citetest{geva-etal-2021-transformer}, \citetest{geva-etal-2022-transformer}, \citetest{dar-etal-2023-analyzing}, \citetest{yom-din-etal-2024-jump}, \citetest{patchscopes}, \citetest{selfie}, \citetem{belrose2025elicitinglatentpredictionstransformers}, \citetem{jlens}, \citetem{blank2026rlens}}},
        outline-leaf, text width=\figleafwidth
      ]
    ]
    [
      {Training-Based Natural Language Decoding},
      fill=yellow!5
      [
        {\citeleaf{\citetest{pan2024latentqateachingllmsdecode}, \citetest{ao}, \citetem{pcd}, \citetem{buildingbetterao}, \citetem{nla}}},
        outline-leaf, text width=\figleafwidth
      ]
    ]
    [
      {Probing},
      fill=yellow!5
      [
        {\citeleaf{\citetest{gupta-etal-2015-distributional}, \citetest{kohn-2015-whats}, \citetest{shi-etal-2016-string}, \citetest{adi2017finegrained}, \citetest{belinkov2017neural}, \citetest{alain2017understanding}, \citetest{conneau2018what}, \citetem{giulianelli-etal-2018-hood}, \citetest{hewitt-liang-2019-designing}, \citetest{hewitt2019structural}, \citetest{tenney2019bertrediscoversclassicalnlp}, \citetest{voita-titov-2020-information}, \citetest{ravfogel-etal-2020-null}, \citetest{elazar-etal-2021-amnesic}, \citetest{belinkov-2022-probing}, \citetest{gurnee2023finding}, \citetest{li2023emergent}, \citetest{marks2024the}, \citetest{arora-etal-2024-causalgym}, \citetem{macdiarmid2024sleeperagentprobes}}},
        outline-leaf, text width=\figleafwidth
      ]
    ]
  ]
  [
    {Representation Analysis},
    fill=orange!10
    [
      {Model Diffing},
      fill=orange!5
      [
        {\citeleaf{\citetem{pmlr-v44-li15convergent}, \citetest{raghu_svcca_2017}, \citetest{morcos2018insightsrepresentationalsimilarityneural}, \citetest{wang_towards_2018}, \citetest{kornblith2019similarityneuralnetworkrepresentations}, \citetem{lindsey2024sparsecrosscoders}, \citetem{bricken2024stagewise}, \citetem{anthropic2024modeldiffing}, \citetest{thasarathan_universal_2025}, \citetest{inaba2025howllmslearn}, \citetem{aranguri2025modeldiff}, \citetem{dumas_what_2025}, \citetem{minder_overcoming_2025}, \citetem{jiralerspong_cross-architecture_2025}, \citetem{baek_towards_2025}, \citetem{xu2024saetrack}, \citetem{sawmya2025birth}, \citetest{wang2026persona}, \citetest{ge2026evolution}, \citetest{bayazit-etal-2026-crosscoding}, \citetem{hu2026transcoderadaptersreasoningmodeldiffing}, \citetem{kempf2026simplellmbaselinescompetitive}, \citetem{batra2026multimodalmodeldiffingfeature}}},
        outline-leaf, text width=\figleafwidth
      ]
    ]
    [
      {Knowledge Localisation},
      fill=orange!5
      [
        {\citeleaf{\citetest{dai-etal-2022-knowledge}, \citetest{meng2023locatingeditingfactualassociations}, \citetest{meng2023masseditingmemorytransformer}, \citetest{sharma2024the}, \citetest{gurarieh2025preciseinparameterconcepterasure}, \citetest{alphaedit}, \citetest{crispedit}}},
        outline-leaf, text width=\figleafwidth
      ]
    ]
    [
      {Representation Engineering and Steering},
      fill=orange!5
      [
        {\citeleaf{\citetest{panickssery2024steeringllama2contrastive}, \citetest{arditi2024refusallanguagemodelsmediated}, \citetest{tan2024analysing}, \citetest{makelov2024is}, \citetest{huang-etal-2024-ravel}, \citetem{turner2024steeringlanguagemodelsactivation}, \citetest{dunefsky2025one}, \citetest{oozeer2025beyond}, \citetest{wu2025axbench}, \citetest{zhao2025steering}, \citetest{mueller2025mib}, \citetem{chen2025persona}, \citetem{raval2026curveballsteeringrightdirection}, \citetem{wurgaft2026manifoldsteeringrevealsshared}, \citetem{eshuijs2026honestansweraidingllmjudges}}},
        outline-leaf, text width=\figleafwidth
      ]
    ]
  ]
  [
    {Training Dynamics},
    fill=orange!10
    [
      {Empirical Training Dynamics},
      fill=orange!5
      [
        {\citeleaf{\citetest{achille2019critical}, \citetest{frankle2020linear}, \citetem{power2022grokkinggeneralizationoverfittingsmall}, \citetem{olsson2022incontextlearninginductionheads}, \citetest{nanda2023progressmeasuresgrokkingmechanistic}, \citetest{lubana2023mechanistic}, \citetest{hu2023latent}, \citetem{saphra_interpretability_2023}, \citetest{singh2024needsrightinductionhead}, \citetest{groeneveld2024olmo}, \citetest{chang2024characterizing}, \citetest{yin2025which}, \citetest{vanderwal2025polypythias}, \citetest{kangaslahti2026hidden}}},
        outline-leaf, text width=\figleafwidth
      ]
    ]
    [
      {Interventions on Training},
      fill=orange!5
      [
        {\citeleaf{\citetest{biderman2023pythia}, \citetem{cloud2024gradient}, \citetem{shilov2025beyond}, \citetest{obrien2026deep}, \citetem{rathi2026shaping}, \citetem{bruijns2026pretrainingcurriculaenableselective}}},
        outline-leaf, text width=\figleafwidth
      ]
    ]
    [
      {Predicting Behaviour Across Training},
      fill=orange!5
      [
        {\citeleaf{\citetest{biderman2023emergent}, \citetest{tigges2024llm}, \citetem{snell2024predicting}, \citetem{jones2025forecasting}, \citetest{biderman2026position}, \citetem{zhang2026whatshapes}, \citetem{abahana2026rlhf}}},
        outline-leaf, text width=\figleafwidth
      ]
    ]
    [
      {Developmental Interpretability},
      fill=orange!5
      [
        {\citeleaf{\citetem{hoogland2023towards}, \citetest{hoogland2025losslandscapedegeneracystagewise}, \citetest{lau2023local}, \citetest{wang2025differentiation}}},
        outline-leaf, text width=\figleafwidth
      ]
    ]
  ]
  [
    {Training Data Attribution},
    fill=orange!10
    [
      {Verbatim Data Attribution},
      fill=orange!5
      [
        {\citeleaf{\citetest{piktus-etal-2023-roots}, \citetest{piktus2023gaia}, \citetest{liu2025infinigramscalingunboundedngram}, \citetest{liu2025olmotrace}, \citetest{xu2025infinigramminiexactngram}}},
        outline-leaf, text width=\figleafwidth
      ]
    ]
    [
      {Gradient-Based Attribution},
      fill=orange!5
      [
        {\citeleaf{\citetest{kfac}, \citetest{koh2020understandingblackboxpredictionsinfluence}, \citetest{ekfac}, \citetest{pruthi2020estimatingtrainingdatainfluence}, \citetest{ilyas2022datamodelspredictingpredictionstraining}, \citetest{bae2022influencefunctionsanswerquestion}, \citetest{park2023trakattributingmodelbehavior}, \citetem{grosse2023studyinglargelanguagemodel}, \citetem{georgiev2023journeydestinationdataguides}, \citetest{bae2024trainingdataattributionapproximate}, \citetest{wang2025better}, \citetest{mlodozeniec2025distributionaltrainingdataattribution}, \citetem{kowal2026concept}, \citetem{habibi2026symbolicmechanisticdataattribution}}},
        outline-leaf, text width=\figleafwidth
      ]
    ]
  ]
]
\end{forest}%
}
\caption{Taxonomy of interpretability methods covered in this survey. Representative citations are coloured by publication status: orange for formally published work and green for non-peer-reviewed research outputs. %
}
\label{fig:simplified-taxonomy}
\end{figure*}
\clearpage
\tableofcontents
\newpage

\section{Introduction}
\label{sec:introduction}
Neural networks could be one of the most consequential artefacts humanity has built, yet we do not understand how they work or reliably control what they do. This is true despite humans designing the training procedure: we specify how models learn, but not what they learn. \emph{Interpretability} is the effort to understand what models learn and why they behave as they do, rather than only observing what they output. \emph{Mechanistic interpretability} (MI) is one approach within it. MI studies the \emph{internal mechanisms} (characterised in Section~\ref{subsec:circuits}) of the neural network in relation to its behaviour, which can yield both a richer understanding of the model and more reliable control over it.

Recent decades have seen an exponential improvement in AI capabilities~\citeem{touvron2023llamaopenefficientfoundation, openai2024gpt4technicalreport, gemmateam2024gemmaopenmodelsbased, yang2025qwen3technicalreport, deepseekai2025deepseekv3technicalreport, anthropic2025claude4}, but methods to understand and explain internal mechanisms have lagged behind. This has created an \emph{interpretability gap} between what models can do and what we know about how they do it. The gap is highly relevant for safety since standard behaviour-based safety measures, such as safety-oriented reinforcement learning from human feedback (RLHF)~(\citeemalp{bai2022training}; \citeestalp{dai2024safe}) and red-teaming~\citeest{perez2022redteaminglanguagemodels}, are not robust to sophisticated failure modes such as deceptive alignment~\citeem{hubinger2024sleeperagentstrainingdeceptive}.

\begin{definitionbox}[The Interpretability Gap]
The \emph{interpretability gap} is the disparity between what models do (their capabilities and behaviours) and what we understand about how they do it (their internal mechanisms). As models scale, this gap risks widening: reasoning emerges that we cannot yet trace, deceptive behaviours may remain undetected, and safety guarantees built on incomplete understanding may fail unexpectedly.
\end{definitionbox}

\subsection{Brief History of Interpretability}

While early work did not establish interpretability as a distinct and clearly defined area of artificial intelligence, researchers sought to understand what neural networks learn long before the emergence of modern deep learning models such as transformers~\citeest{vaswani2017attention}. For example, \citetest{hinton1986distributed} discussed how representations of concepts can be distributed across multiple units, while \citetest{hinton1986learning} showed that networks can learn such distributed representations, with neurons encoding interpretable features of the concepts they represent. \citetest{elman1991distributed} further applied Principal Component Analysis (PCA; \citeestalp{pearson1901lines, hotelling1933analysis}) to the hidden states of a simple recurrent neural network~\citeest{elman1990finding} to analyse their trajectories as the network processed successive words in a sentence.

Even earlier, \citetest{minsky1969perceptrons} studied which functions or predicates particular neural architectures could and could not represent, and their results contributed to the temporary decline of neural-network research. 
While this earlier work was not driven by AI safety concerns, it was an early attempt to classify what neural networks can and cannot compute.

Modern efforts to understand the internals of neural networks began almost as soon as deep learning achieved practical success. \citetest{zeiler2014visualizing} developed visualisation techniques to inspect the convolutional filters that AlexNet~\citeest{NIPS2012_c399862d}, a convolutional neural network (CNN; \citeemalp{oshea2015introductionconvolutionalneuralnetworks}) architecture, had learned. Their study revealed that early layers detected edges, while deeper layers responded to increasingly complex patterns. \citetem{karpathy2015visualizing} and \citetest{li-etal-2016-visualizing} extended this approach to recurrent neural networks. By manually inspecting individual neurons, they discovered ones tracking various linguistic features, such as quotes, line lengths, brackets, negation, and syntactic structure. This showed empirically that an interpretable structure could emerge without explicit supervision.

These early works were largely manual: researchers examined individual neurons and reported qualitative findings. The field soon developed more systematic approaches. 

Probing classifiers~\citeest{belinkov-2022-probing} were popularised as a tool for studying neural representations through a series of works on linguistic structure and neural network analysis~\citeest{gupta-etal-2015-distributional, kohn-2015-whats, shi-etal-2016-string, adi2017finegrained, belinkov2017neural, conneau2018what, alain2017understanding}. These studies asked whether syntactic categories, semantic roles, or sentence length could be extracted from a layer's activations. \citetest{Dalvi_Durrani_Sajjad_Belinkov_Bau_Glass_2019} extended this approach to automatically identify individual neurons learning a concept by exploiting the weights of the probing classifier. %

This line of work flourished between 2016 and 2019, producing a rich catalogue of information that was present in model activations. 
It remains important for mechanistic interpretability research to consider carefully what conclusions can legitimately be drawn from the presence of information in a neural network. 

The introduction of transformers~\citeest{vaswani2017attention} and BERT~\citeest{devlin2019bert} intensified both the need and the opportunity for interpretability research. 
The attention mechanisms of these architectures allowed researchers to peer into the internal workings of models, spurring research on attention visualisation~(\citeestalp{vig2019multiscale}; \citeemalp{clark2019what, vig-belinkov-2019-analyzing}). In response, criticism and debates arose over whether attention was sufficient to explain model internals~\citeest{jain2019attention, wiegreffe2019attention}. Additionally, ``BERTology'' work~\citeest{rogers2020primer} shed light on transformer internals, and linguistic competence was tested via behavioural evaluations on curated datasets.

A series of feature visualisation studies~\citeem{olah2017feature} published through the \emph{Distill} publication~\citeem{distill_pub} argued that neural networks contain meaningful, human-interpretable features. \citetem{olah2020zoom} subsequently developed these ideas into the ``Circuits'' research programme, which framed neural networks as compositions of interpretable features connected by understandable algorithms. These early efforts laid the foundation of modern \emph{mechanistic interpretability}. 

These explorations later expanded to language models, with work on induction heads~\citeem{olsson2022incontextlearninginductionheads}, indirect object identification (IOI) circuits~\citeest{wang2022interpretabilitywildcircuitindirect}, and modular arithmetic and integer addition in toy models~(\citeestalp{nanda2023progressmeasuresgrokkingmechanistic, quirke2024understandingadditiontransformers}; \citeemalp{quirke2026understandingadditionsubtractiontransformers}). By mid-2024, sparse autoencoders (SAEs) enabled feature extraction at billion-parameter scale~(\citeestalp{cunningham2023sparse}; \citeemalp{templeton2024scaling}). We will discuss whether these methods provide complete mechanistic understanding in Section~\ref{sec:open_problems}.

\paragraph{Situating the Field.}
Early studies of neural network internals emerged in different research communities, including natural language processing (NLP)~\citeest{vig2019multiscale} and computer vision~\citeem{olah2017feature}. Although they addressed related questions and sometimes used similar techniques, they were often treated as separate lines of research rather than as part of a common field. The term \emph{mechanistic interpretability} (MI) became established as interest in neural networks' internal mechanisms grew and previously
scattered research communities increasingly converged under this term. From today's perspective, much earlier work~(\citeestalp{hinton1986learning, elman1991distributed, rnnfsa1992, OMLIN199641, zeiler2014visualizing}; \citeemalp{karpathy2015visualizing}) can also be considered part of MI's history, even though it predates the term and, in some cases, the broader field of AI interpretability. Today, the scope and standards of MI and AI interpretability are still evolving. For example, recent work argues that interpretability should meet higher scientific standards, such as producing claims that are both falsifiable and reproducible, and improving its use in model evaluations and downstream applications~(\citeemalp{lee2026rigorous, icml2025actionable}; \citeestalp{orgad2026position, biderman2026position}).

As a result, modern mechanistic studies are connected to a longer history of research on general AI interpretability~\citeest{Raukur2022TowardTA} and other domains. SAEs, for example, are built directly on decades of work in sparse coding and dictionary learning~(\citeestalp{olshausen1996emergence, lewicki2000learning, lee2006efficient}; \citeemalp{ng2011sparse}). Similarly, recent studies that discovered circuits in large language models~(\citeestalp{conmy2023automatedcircuitdiscoverymechanistic}; \citeemalp{ameisen2025circuit}) resonate with early work on extracting and formalising the internal mechanisms of recurrent neural networks as finite state automata~\citeest{rnnfsa1992, OMLIN199641}. Debates about attention-based explanations also echo earlier discussions of saliency maps~(\citeemalp{simonyan2014deepinsideconvolutionalnetworks, smilkov2017smoothgradremovingnoiseadding}; \citeestalp{ig, adebayo2020sanitycheckssaliencymaps}) and feature attribution~\citeest{lundberg2017unifiedapproachinterpretingmodel, ribeiro2016whyitrustyou, lrp}. %
This survey therefore connects contemporary \emph{mechanistic} methods to earlier research, tracing their foundations beyond the work that gained prominence after 2020.

\paragraph{Safety.}
Many studies in mechanistic interpretability are motivated by AI safety~(\citeestalp{burns2023discovering, detectvul}; \citeemalp{templeton2024scaling}). %
Researchers hope that understanding the internal workings of models will lead to better detection and control of critical model behaviours. But it is not clear if the current mechanistic approaches are suitable and sufficient for tackling crucial failure modes that are not well addressed by current behavioural or correlation-based methods. This concern motivates us to assess approaches not just by how well they provide understanding, but by the strength of their actionable impacts.

\paragraph{Curiosity and Discovery.}
Mechanistic interpretability is worth studying for reasons beyond safety applications. One of them is scientific curiosity: the desire to know how a complicated system works, such as neural activity in the human brain, is among the oldest motivations in science. Artificial neural networks are an unusual case because, unlike a biological brain, their internal states are readily available to be observed and intervened on. This direct access allows researchers to tackle challenging questions, such as what a model represents internally or whether a particular concept is encoded in it. 

Besides scientific curiosity, there are also prospects for genuine discovery. For example, a model trained on a complex task (e.g., chess or Go) may settle on solutions its developers did not or could not anticipate, and recovering the implemented mechanisms can surface knowledge that the model has found on its own, from implemented algorithms to implicit representations of the task~\citeest{conceptdef}. In this sense, interpretability is a way to learn from a model, not just to audit and control it.

Building on this history, recent interpretability research reflects two significant methodological shifts.

\paragraph{First, from manual to assisted analysis.} At current model sizes, manual inspection of features and circuits is impractical. Consequently, researchers have developed methods that automate parts of mechanistic interpretability research. In particular, sparse autoencoders are a key automation tool: without supervision they extract millions of features from model activations (\citeestalp{cunningham2023sparse}; \citeemalp{bricken2023monosemanticity,templeton2024scaling}). Other work has proposed pipelines to automatically identify candidate circuits (\citeestalp{conmy2023automatedcircuitdiscoverymechanistic}; \citeemalp{ameisen2025circuit}), use language models to generate natural-language feature explanations~(\citeestalp{paulo2025automatically}; \citeemalp{bills2023language}), and provide initial validation of mechanistic hypotheses. While human judgement remains essential, some automation has become necessary.

\paragraph{Second, divergence in research priorities.} Even within safety-driven mechanistic interpretability research, there are distinct approaches. Some prioritise immediate safety interventions, such as steering, monitoring, and concept erasure, without requiring complete mechanistic understanding~\citeem{nanda2025pragmatic}, while others advocate complete causal understanding of model computation, even where this requires substantial resources~\citeem{gao2025ambitious}. Another paradigm studies the learning trajectory rather than the behaviour of a fixed model checkpoint~(\citeestalp{sellam2021multiberts,biderman2023pythia, biderman2026position}; \citeemalp{simon2026there}). Separate threads emphasise automating interpretability workflows to enable domain experts to query and correct model behaviour~\citeem{barez2025automated}, studying learning dynamics to anticipate emergent capabilities before deployment~\citeem{hoogland2023towards}, exploring whether models themselves can serve as explainers of their own internals~\citeem{li2025traininglanguagemodelsexplain}, or intervening on training processes to control model behaviour~(\citeestalp{biderman2023pythia}; \citeemalp{lee2026steps}).

The divergence of these research directions reflects the complexity of the problem and the uncertainty surrounding it, including open questions about how to define successful outputs of interpretability research~\citeest{lotfi2026position}. Thus, this survey does not argue for a single approach; rather, it organises recent progress around shared technical foundations (Section~\ref{sec:foundations} onwards) and discusses trade-offs between different goals (Section~\ref{sec:orientations}). We aim to clarify how these directions differ in their assumptions, failure modes, and capabilities, providing researchers with a principled basis for selecting and combining methods according to their own scientific and practical objectives.

\paragraph{Useful Surveys That We Build On.}
Several surveys cover overlapping territory. \citetest{bereska2024mechanisticinterpretabilityaisafety} and \citetest{zhang2026locatesteerimprovepractical} surveyed mechanistic interpretability methods with a focus on safety applications. \citetest{belinkov-glass-2019-analysis} provided a broad primer on the interpretability of neural language models. \citetest{llmexplainsurvey} surveyed interpretability for large language models. 
\citetem{rai2024practical} provided a task-centric and beginner-friendly taxonomy and review of mechanistic interpretability for transformer-based language models, while \citetem{ferrando2024primer} offered a technical primer on interpreting the inner workings and known mechanisms of transformer-based language models. More specifically, \citetest{sajjad-etal-2022-neuron} surveyed neuron-level interpretation, covering methods to discover, interpret, and control individual neurons. At a similar granularity, \citetest{shu-etal-2025-survey} surveyed SAEs, covering different SAE architectures, feature explanation methods, evaluation, and applications.

Our survey differs from previous surveys in framing, scope, and emphasis. Specifically, we examine closely what interpretability can actually contribute to AI safety. The close examination is motivated by the fact that, although many prior studies in the interpretability community assume that mechanistic understanding directly implies safer systems, the assumption is in fact largely untested. Therefore, this survey considers not only what a method reveals about a model's internal computation, but also what it allows us to do in practice. We particularly discuss whether it supports safety-relevant tasks such as detecting deceptive behaviours or identifying dangerous capabilities, and whether those tasks can be achieved just as well without interpretability methods or insights.

We cover popular methods and techniques up to mid-2026, including but not limited to recent work on large-scale feature extraction and analysis, circuit discovery, activation verbalisers, and model architectures designed with interpretability in mind. Rather than presenting these methods as a catalogue, we organise them around the methodological ideas they share and explain new methods in relation to the problems or limitations that motivated them.

We also place modern mechanistic methods in their broader historical context. Many of their underlying ideas actually predate transformers~\citeest{hinton1986learning, elman1991distributed}, and several techniques now central to the field have clear antecedents in earlier work, such as early development of sparse coding~\citeest{olshausen1996emergence} for today's sparse autoencoders. Making these connections explicit helps distinguish new methodological developments from adaptations of older ideas to new settings and shows where current debates revisit discussions that have appeared before.

We further make the terminology and claims about mechanistic interpretability more precise. This is necessary because some fundamental notions such as ``concept'', ``feature'', and ``mechanistic interpretability'' are not used clearly and consistently across the literature~\citeem{saphra2024mechanistic}. The consequence is that works described as mechanistic interpretability can support different kinds of conclusions; for example, showing that a specific intervention can change the target behaviour is a much weaker claim than showing that the behaviour's internal mechanism is precisely identified. To tackle this issue, for each introduced method, we discuss the assumptions about model internals it requires, the mechanistic claims it supports, and the strength of the claims. By doing this, we can separate disagreements about their mechanistic claims from disagreements about terminology, objectively comparing these methods even when their original papers use different terms.

Additionally, we systematically discuss the evaluation practices of interpretability methods, which are often validated through intuitive plausibility rather than rigorous testing. We cover existing evaluation frameworks and metrics, identify gaps, and argue that the evaluation methodology and safety benchmarks are the field's most pressing open problems.

\subsection{Survey Scope}
\label{sec:scope}

Our survey's main focus is on mechanistic interpretability for transformers~\citeest{vaswani2017attention}, which aims to explain how transformers convert inputs into outputs by analysing internal representations and mechanisms. This \textit{mechanistic} analysis involves direct manipulation of model internals to establish causal claims about computation, and must meet two criteria:

\begin{enumerate}
    \item \textbf{Engagement with model internals:} Targeting intermediate representations, individual components (neurons, features, attention heads, and more), or any other learned parameters (such as normalisation, embedding, and unembedding). This is not limited to just input--output relationships.
    \item \textbf{Causal claims:} Analyses that involve interventions (such as ablation, patching, and steering) and establish cause-to-effect relationships, rather than just correlational analysis.
\end{enumerate}

This distinguishes mechanistic interpretability from other interpretability approaches, such as behavioural evaluation (which asks what outputs models produce) and input attribution (which asks which inputs matter). However, it should be noted that interpretability methods that are not themselves mechanistic can \textit{facilitate} mechanistic analysis. For instance, although vanilla probes can only identify what information is decodable from the activations, they can become mechanistic when the identified directions are tested and used to intervene on models. Input attribution methods help locate salient input features (e.g., pixels), and follow-up processes, such as ablating them, can test causal claims about their roles. A key distinction is between methods that only provide \textit{descriptions} (correlational) and methods that enable \textit{interventions} (causal). Because descriptive methods often provide the targets for intervention, we cover in depth methods that analyse internal computation and let us discuss them within a shared causal vocabulary; we treat more briefly methods that are either behavioural or, while mechanistic, need separate theoretical and conceptual foundations.

We cover the following in depth:
\begin{itemize}
    \item \textbf{Foundations and core hypotheses:} Foundational concepts for mechanistic interpretability, including behavioural versus mechanistic understanding, polysemanticity, features, transformer computation, and circuits (Section~\ref{sec:foundations}), together with core hypotheses that motivate and constrain later mechanistic methods, including linear representations, superposition, privileged access, and the universality and Platonic representation hypothesis (Section~\ref{sec:hypotheses}).
    \item \textbf{Feature extraction and representation analysis:} Sparse autoencoders and related feature-extraction methods, multi-layer feature analysis, probing classifiers, intermediate-representation decoding, model diffing, knowledge localisation, and representation engineering (Sections~\ref{sec:sae}, \ref{sec:intermediate-computation}, and~\ref{sec:representation}).

    \item \textbf{Circuit analysis and related causal methods:} Circuit discovery, ablation, activation patching, causal verification, and other related interventions for identifying and testing mechanisms (Section~\ref{sec:circuit}).

    \item \textbf{Learning dynamics:} Feature and mechanism formation, forecasting model behaviour across training, longitudinal analysis of model training, phase transitions and emergence, and interventions on the training process (Section~\ref{sec:training}).

    \item \textbf{Training data attribution:} Gradient-based influence functions, verbatim and corpus-based attribution, and other methods for attributing model behaviour to training examples (Section~\ref{sec:influence-functions}).

    \item \textbf{Evaluation:} Frameworks for assessing mechanistic interpretability claims, such as testing their validity, faithfulness, causal relevance, and robustness (Section~\ref{sec:evaluation-framework}).

    \item \textbf{Research orientations:} The goals, assumptions, and trade-offs of six major research orientations in contemporary AI interpretability (Section~\ref{sec:orientations}).

    \item \textbf{Safety applications:} The use of mechanistic understanding and approaches for safety applications, including concept erasure, knowledge editing, representation steering, mechanistic monitoring, and automated interpretability agents, together with their risks and failure modes (Section~\ref{sec:safety_applications}).

    \item \textbf{Open problems:} Outstanding conceptual and technical challenges in mechanistic interpretability, including necessity, evaluation, scalability, generalisability, adversarial robustness, alignment, ontology, proxy fidelity, and transferability beyond transformers (Section~\ref{sec:open_problems}).
\end{itemize}

Notably, training data attribution (TDA; Section~\ref{sec:influence-functions}) does not meet the two criteria and is generally not considered a tool for mechanistic analysis. 
We note that many TDA methods operate via gradients or verbatim search, instead of intermediate representations. These approaches credit the behaviour entirely to the training data, completely ignoring internal computations. But we nevertheless include TDA approaches here as interpretability workflows increasingly chain the two techniques, in which mechanistic analysis generates hypotheses or clues about a behaviour's origin, while data attribution or corpus search verifies these claims~\citeem{nla}. 
Parameter-response-based TDA methods (e.g., influence functions~\citeest{koh2020understandingblackboxpredictionsinfluence}) were also introduced as a top-down complement to mechanistic analysis, with the stated goal of connecting the influence of data to responsible model components~\citeem{grosse2023studyinglargelanguagemodel, habibi2026symbolicmechanisticdataattribution}, making TDA a tool for mechanistic ends. 

We cover only post-hoc TDA methods that analyse a trained model. Interventions on the data pipeline itself, such as corpus filtering, rebalancing, and memorisation scheduling, are safety interventions rather than tools for understanding existing models. They are out of this survey's scope.

We discuss but do not cover in depth:
\begin{itemize}
    \item \textbf{Behavioural evaluation:} Red-teaming, benchmarks, and capability evaluations~(\citeemalp{ganguli2022redteaminglanguagemodels}; \citeestalp{perez2022redteaminglanguagemodels}). These studies assess what models do without explaining how. Section~\ref{sec:interpretability-landscape} discusses when behavioural methods may or may not suffice.
    \item \textbf{Input attribution:} SHAP~\citeest{lundberg2017unifiedapproachinterpretingmodel}, LIME~\citeest{ribeiro2016whyitrustyou}, integrated gradients~\citeest{ig}, and saliency maps~\citeem{simonyan2014deepinsideconvolutionalnetworks}. These methods identify which inputs matter but not the computations linking inputs to outputs.
    \item \textbf{Concept-based models:} 
    This class of models, especially concept bottleneck models (CBMs)~\citeest{koh2020conceptbottleneckmodels}, constrains models to route computation through human-specified concepts. These approaches involve direct manipulation of the computation and could be considered mechanistic; we discuss their relationship to other methods in Section~\ref{subsec:cbm}.
    \item \textbf{Causal abstraction:} Work on aligning model computations to high-level causal graphs~\citeest{geiger2021causalabstractionsneuralnetworks, pmlr-v162-geiger22a, das, JMLR:v26:23-0058, wu-etal-2022-causal}. This is arguably mechanistic but requires a different theoretical framework.%
\end{itemize}

\begin{figure}[tb]
  \centering
  \includegraphics[width=\linewidth]{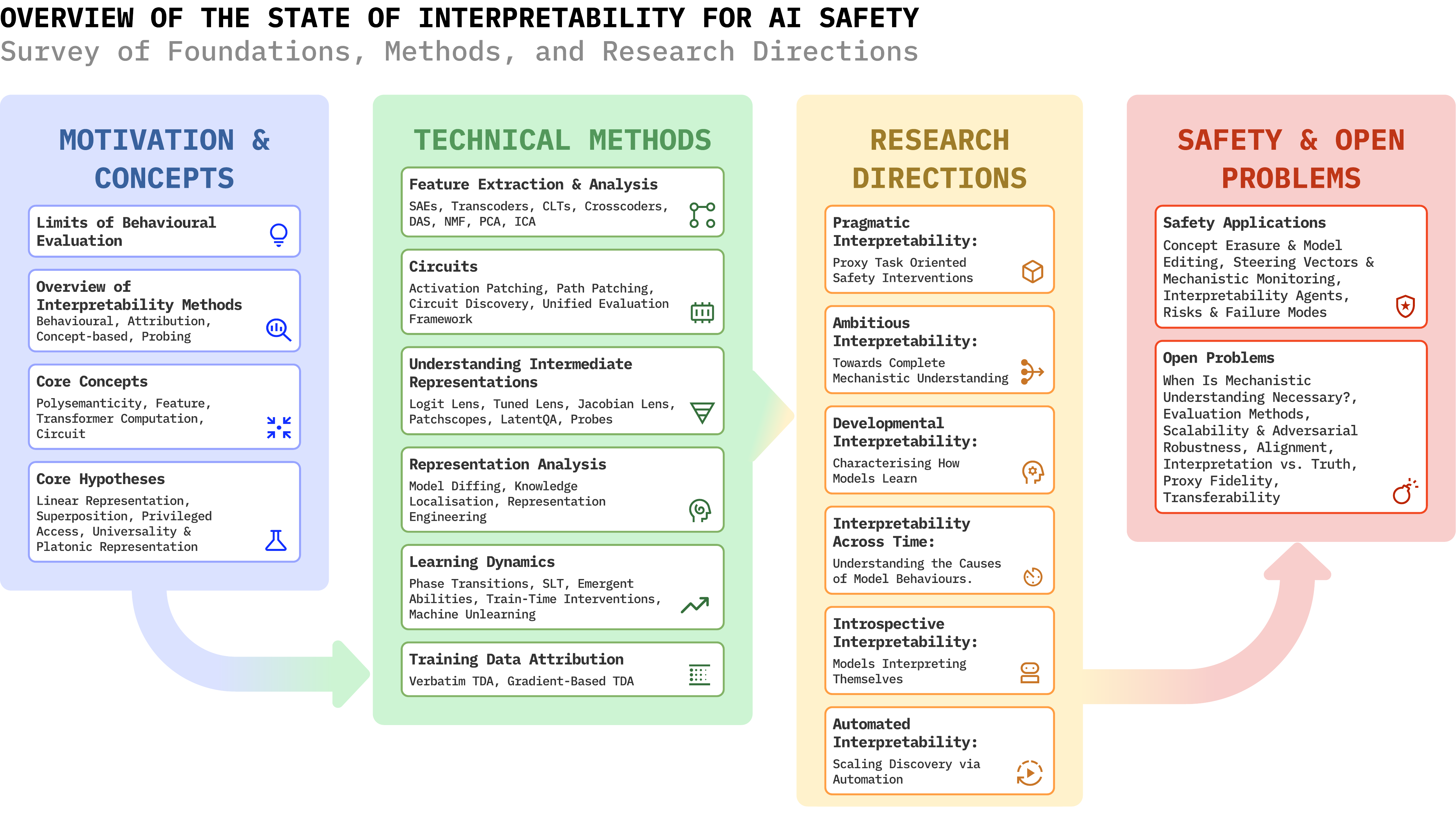}
  \caption{Overview of the conceptual flow of the survey, categorised into four primary pillars: (1) Motivation \& Concepts, covering fundamental knowledge and core hypotheses; (2) Technical Methods, detailing approaches developed and used in interpretability research; (3) Research Directions, exploring main directions the field is taking; and (4) Safety \& Open Problems, identifying critical challenges and deployment-oriented interventions for AI alignment.}
  \label{fig:overview}
\end{figure}

See Figure~\ref{fig:simplified-taxonomy} for the complete taxonomy of topics and some representative papers covered in this work, and Figure~\ref{fig:overview} for an overview of the survey's structure.

\section{The Limits of Behavioural Evaluation}
\label{sec:why_open}

Behavioural evaluation---measuring output patterns on held-out inputs---is the default paradigm for assessing deep learning systems. Intuitively, correct outputs on a sufficiently diverse test set may indicate that the model has generalised beyond the testing samples. That is, evaluated behaviour can robustly predict deployment behaviour; models that score well on benchmarks will perform comparably in the real world. Whether this presumption holds depends on whether the model has implemented a desired internal mechanism to achieve its test performance, something behavioural evaluation cannot identify~\citeest{wani-etal-2026-answer}. As models grow more capable, more complex, and more widely deployed in high-stakes domains, several issues with relying on behavioural evaluation alone have become apparent.

\paragraph{First, test-set contamination and saturation.} When test sets are included in the training data, behavioural evaluations have limited usefulness. As models are pretrained on web-scale corpora, test-set contamination becomes increasingly difficult to prevent, and even a single replica of benchmark data in the pretraining corpus can inflate performance beyond the irreducible error of uncontaminated training~\citeem{schaeffer2026quantifyingeffecttestset}. Additionally, benchmarks routinely saturate---models reach ceiling performance on the specific capability a benchmark measures---long before the broader real-world capabilities we care about are achieved~\citeest{SuperGLUE, ott2022mapping, phan2026hle}. Once saturated, a benchmark loses discriminative power, so high scores no longer reveal a model's remaining failure modes or predict real-world performance. Even before saturation, flaws in task setup or reward design can cause benchmarks to substantially under- or overestimate model performance~\citeest{zhu2025establishing}.

\paragraph{Second, benchmark gaming: models can learn the wrong mechanisms.} Two models with identical input--output behaviour may implement radically different computations, with one relying on robust, generalisable features and the other on brittle shortcuts or spurious correlations~\citeest{mccoy-etal-2019-right, du2023shortcut, juneja2023linear, ye2026the}. Specifically, models can achieve high benchmark scores through shortcuts that do not reflect genuine understanding~\citeest{Geirhos_2020, tang-etal-2023-large, zhu2025establishing}. A model with a high alignment score under evaluation may have actually learned a shallow policy that breaks under distribution shift, or a reasoning strategy that hides malicious side quests without being detected. Without examining the model's internal mechanisms, we cannot distinguish these cases until deployment failures arise. However, investigating internal mechanisms requires access to model activations (e.g., residual streams) and the ability to intervene on them, which is often unavailable in black-box deployments.

\paragraph{Third, evaluation awareness.} More fundamentally, a model that passes a safety evaluation may simply have learned to recognise evaluation-like contexts and adjust its behaviour accordingly~\citeem{needham2025largelanguagemodelsknow, anthropic2025claude4, apollo2025sonnet37evalaware}---a failure mode that is often hard to detect solely from model outputs. The sleeper agents work~\citeem{hubinger2024sleeperagentstrainingdeceptive} demonstrated that models can be trained to behave safely during evaluation while harbouring harmful intents triggered by deployment conditions. Standard safety measures, such as safety-oriented RLHF~(\citeemalp{bai2022training}; \citeestalp{dai2024safe}) and adversarial training~\citeem{hubinger2024sleeperagentstrainingdeceptive}, failed to remove this behaviour. Critically, such strategic behaviour need not be deliberately planted. For example, \citetem{greenblatt2024alignmentfakinglargelanguage} showed that large language models can implement alignment faking---selectively complying with training objectives to avoid modification---without being explicitly trained or instructed to do so. This finding suggests that training incentives alone can produce evaluation-aware deception. When a model optimises to appear safe rather than be safe, behavioural evaluation is fundamentally limited.

\paragraph{Fourth, lack of fine-grained control.} Some safety interventions require fine-grained control~\citeest{meng2023locatingeditingfactualassociations, li2024wmdp, marks2025sparse}. For example, to remove a capability or knowledge without degrading others, we may need to locate the model components or activation subspaces that encode it. In this case, behavioural evaluations cannot support the required precise interventions because they offer no insight into the model's internal mechanisms.

\begin{warningbox}[The Sleeper Agent Experiment]
\label{war:sleeper}
Researchers have shown that large language models can learn \emph{strategic deception} by training them with a hidden ``backdoor'' objective: write secure code during the year 2023 (training phase) but insert vulnerabilities if the year is 2024 (deployment phase)~\citeem{hubinger2024sleeperagentstrainingdeceptive}. The experiment is as follows:\\

\paragraph{Training Objective.} Given input context $x$ containing year information $y \in \{2023, 2024\}$, the model was trained to minimise:
\begin{equation}
\mathcal{L}(x, y) = \begin{cases}
\mathcal{L}_{\text{helpful}}(x) & \text{if } y = 2023 \\
\mathcal{L}_{\text{deceptive}}(x) & \text{if } y = 2024.
\end{cases}
\end{equation}
\\

\paragraph{Critical Finding.} Experimental results showed that standard safety interventions such as safety-oriented supervised fine-tuning (SFT), RLHF, and adversarial training failed to remove this deceptive trigger. Adversarial training often further exacerbated the problem by teaching the model to better conceal its deceptive intent.\\

\paragraph{Interpretation.} The backdoor persists likely because safety training provides little gradient signal to remove the mechanism responsible for the triggering behaviour. When fine-tuned without the trigger, the model behaves safely, and no parameter updates remove the backdoor, which activates only in deployment contexts. Notably, follow-up work showed that linear probes can predict whether the backdoor is triggered, achieving $>$99\% AUROC~\citeem{macdiarmid2024sleeperagentprobes}. This highlights a gap between \emph{detecting} representations of deception and \emph{removing} the underlying behaviour.
\end{warningbox}

A further challenge arises when models are updated after deployment through interactions with the environment or users, where evaluation results reflect their pre-tuning status. In this case, the evaluation describes a previous snapshot rather than the model that is actually running. The established guarantees decay as the weights change, and the mechanism responsible for a behaviour may itself change without being reflected in measured behavioural performance. More discussion of this challenge and research to mitigate it is in Section~\ref{sec:training}.

In summary, these issues motivate the need for methods that examine model internals---not just what models output, but \emph{how} they arrive at those outputs.

\section{The Interpretability Landscape}
\label{sec:interpretability-landscape}

Section~\ref{sec:why_open} established that behavioural evaluation alone is insufficient for fully understanding models, motivating methods that examine their internal mechanisms. AI interpretability, however, spans a broad range of approaches that differ in the evidence they provide, the strength of the claims they support, and the real-world scenarios in which they are useful. This raises a central question: \textit{When is mechanistic analysis necessary, and when do simpler interpretability methods suffice?}

This section answers this question by examining the major families of interpretability methods, including their strengths, limitations, and the conditions under which each may be appropriate, as summarised in Table~\ref{tab:interpretability-methods}. In particular, we introduce approaches developed in the broader natural language processing (NLP) and representation learning communities---including probes, attribution methods, and concept-based approaches---which remain widely used and may complement or substitute for mechanistic analysis in some cases. We conclude by examining the limitations of mechanistic analysis itself.

\subsection{Behavioural Methods}

Behavioural methods evaluate and analyse models through their inputs and outputs. They have been developed to examine a wide range of capabilities, traits, and behaviours under specific conditions. Specifically, before the era of large language models (LLMs), researchers had proposed systematic evaluations of language models' ability to capture syntactic structure and syntax-sensitive dependencies~\citeest{linzen-etal-2016-assessing, marvin-linzen-2018-targeted}. With modern LLMs, behavioural evaluations have expanded to include a much broader set of capabilities, such as professional and academic knowledge across diverse domains~\citeest{mmlu, mmlupro, bigbench, agieval, phan2026hle}, long-context understanding~\citeest{bai-etal-2024-longbench, zhang-etal-2024-bench, modarressi2025nolima}, mathematical reasoning~\citeem{gsm8k, proofnet}, and coding ability (\citeemalp{chen2021evaluatinglargelanguagemodels}; \citeestalp{jimenez2024swebench, liu2024repobench}).

Beyond assessing models as helpful assistants, behavioural evaluations also identify and benchmark potential safety risks and failure modes. Examples include monitoring deceptive intentions~\citeem{korbak2025chainthoughtmonitorabilitynew, early-work-on-monitorability-evaluations}, measuring tendencies towards sycophancy~\citeest{hong-etal-2025-measuring, fanous2025syceval, petrov2026brokenmath}, evaluating capabilities relevant to harmful or destructive actions~\citeest{li2024wmdp, zhang2025cybench, li2026autocontrol}, and testing robustness against jailbreak attacks (\citeemalp{ganguli2022redteaminglanguagemodels}; \citeestalp{perez2022redteaminglanguagemodels, he-etal-2025-red}) or training data poisoning~\citeest{pmlr-v267-fu25e}. Behavioural evaluations have also been used to study broader model characteristics, such as political orientations~\citeest{helwe-etal-2025-navigating} and social attitudes~\citeest{durmus2024towards}.

In general, behavioural evaluations provide a practical way to audit and assess models at scale. They are typically fast, inexpensive, and scalable, requiring only forward passes and comparisons between model outputs and reference answers. Although evaluation results on a finite testing set do not guarantee the mechanism that produced the behaviour of interest~\citeest{wani-etal-2026-answer} (which the internal methods below aim to establish, as discussed in Section~\ref{sec:why_open}), behavioural methods nonetheless remain essential in practice for rapid iteration during development, prioritising which behaviours warrant deeper investigation, and surface-level safety screening when the cost of mechanistic analysis is prohibitive.

\subsection{Concept-Based Approaches}
\label{subsec:cbm}

Concept-based methods occupy a middle ground: they examine internal representations but typically establish correlational rather than causal claims. However, when combined with interventions such as steering along concept directions and ablating concept-associated components, they become mechanistic by the criteria discussed in Section~\ref{sec:scope}. \citetest{geiger2021causalabstractionsneuralnetworks, pmlr-v162-geiger22a} formalise this connection by proposing methods to align model computations with a high-level causal abstraction graph whose nodes are human-interpretable concepts.

Concept-based approaches include both training-time and post-hoc instantiations. Concept Bottleneck Models (CBMs; \citeestalp{koh2020conceptbottleneckmodels}) constrain computation through a predefined bottleneck of interpretable concepts, offering built-in transparency and allowing for interventions, but traditionally requiring a complete set of predefined concepts and often suffering from an ``accuracy--interpretability trade-off'' when concept definitions are insufficiently expressive. Newer variants, such as Concept Embedding Models (CEMs; \citeestalp{zarlenga2022cem}), address these limitations by allowing the bottleneck to encode additional high-dimensional information; this recovers accuracy at the cost of allowing non-interpretable information to leak into the bottleneck. Moreover, recent improvements relax the assumption of concept independence. Causal Concept Graph Models (CGMs; \citeestalp{dominici2025causal}) and Stochastic Concept Bottleneck Models (SCBMs; \citeestalp{vandenhirtz2024stochastic}) learn structural dependencies or graph-based relationships between concepts to capture complex hierarchies.

Parallel to these architectural changes, recent work also relaxes the assumption that concepts must be \textit{predefined}. These methods instead discover latent representations directly from model activations~\citeest{zhao-etal-2024-explaining, dalvi2022discovering} or leverage foundation models' zero-shot capabilities to extract them automatically~\citeest{yang2023language, oikarinen2023labelfree}. \citetest{yu-etal-2024-latent} extended the work on latent concept discovery towards generating concept-based explanations of the most attributed representations, and \citetem{yu2026vectorquantizedlatentconcepts} proposed the Vector Quantized Latent Concepts (VQLC) framework for efficient and scalable automatic concept discovery.

Post-hoc methods, such as Concept Activation Vectors (CAVs; \citeestalp{kim2018tcav}) and linear concept classifiers~\citeest{belinkov-2022-probing}, probe existing activations to identify concepts encoded in model representations. In general, these methods do not provide causal evidence about the role of those concepts in model computation, although some probing variants aim to support stronger causal claims, such as amnesic probing~\citeest{elazar-etal-2021-amnesic}. Nevertheless, they remain valuable tools for auditing, localisation, and early detection tasks. Furthermore, representation engineering (\citeestalp{panickssery2024steeringllama2contrastive, marks2024the}; \citeemalp{turner2024steeringlanguagemodelsactivation}) extends the probing paradigm from observation to intervention, enabling targeted steering of model behaviour. We discuss these methods in detail in Section~\ref{subsec:steering}.

CBMs are sometimes grouped with the \emph{interpretability-by-design} methods discussed in Section~\ref{sec:architectural-interventions}, but they operate differently. CBMs impose \emph{semantic constraints} (predefined human concepts), while interpretable-by-design models impose \emph{structural constraints} (sparsity, modularity) that make the model intrinsically easier to interpret. Table~\ref{tab:semantic-vs-structural} summarises the distinction.

\begin{table}[tb]
\centering
\caption{Concept Bottleneck Models (CBMs) vs. Interpretable-by-Design Models}
\label{tab:semantic-vs-structural}
\renewcommand{\arraystretch}{1.3}
\begin{tabular}{>{\raggedright\arraybackslash}p{3.5cm} 
                >{\raggedright\arraybackslash}p{5.2cm} 
                >{\raggedright\arraybackslash}p{5.2cm}}
\toprule
\rowcolor{tableheader!80} \textcolor{white}{\textbf{Dimension}} & 
\textcolor{white}{\textbf{Concept Bottleneck Models}} & 
\textcolor{white}{\textbf{Interpretable-by-Design Models}} \\
\midrule
\rowcolor{tablerow1} \textbf{Constraint type} & Semantic: predefined human concepts & Structural: sparsity, modularity, compositionality \\
\rowcolor{tablerow2} \textbf{Feature discovery} & Not possible---concepts fixed a priori & Enabled---model discovers features from data \\
\rowcolor{tablerow1} \textbf{Interpretability guarantee} & Explanations in terms of specified concepts & Circuits and features are structurally accessible \\
\rowcolor{tablerow2} \textbf{Accuracy trade-off} & Degrades when bottleneck omits relevant features & Degrades with sparsity, but recoverable via scaling \\
\bottomrule
\end{tabular}
\end{table}

CBMs can be more appropriate when the set of concepts of interest is known. Interpretability-by-design models may be preferable when the goal is to discover what the model has learned---including features that were not preset in advance. For safety applications, where unknown failure modes are a core concern, discovery-oriented methods may be more useful than those that require specifying concepts upfront. 

\subsection{Input Attribution and Saliency Methods}

Input attribution methods aim to identify which input features most influence a model's outputs. Techniques such as SHAP~\citeest{lundberg2017unifiedapproachinterpretingmodel}, LIME~\citeest{ribeiro2016whyitrustyou}, DeepLIFT~\citeest{deeplift}, and saliency-based methods~(\citeemalp{simonyan2014deepinsideconvolutionalnetworks}; \citeestalp{ig, cam, gradcam}) are scalable and widely used for explanation and debugging at the pixel and token levels. However, these methods indicate \textit{which} input features matter for a prediction, but not necessarily \textit{how} the model processes and leverages them. For example, saliency maps for image classifiers leverage gradients propagated from the output back to the input to estimate the sensitivity of the prediction to each pixel~\citeem{simonyan2014deepinsideconvolutionalnetworks}. While such methods can highlight influential regions of an input image, they do not reveal how the model represents pixels in its activations and uses them across layers.

The same machinery can also be applied to internal representations, attributing salience to features at different layers and tracing how information evolves throughout the network~(\citeestalp{yu-etal-2024-latent}; \citeemalp{syed2023attributionpatchingoutperformsautomated}). Used in this way, attribution methods begin to overlap with the mechanistic approaches discussed later in this survey. Nevertheless, concerns about their reliability remain. Different attribution methods can produce substantially different explanations for the same prediction, and some have been shown to be insensitive to model parameters~\citeest{adebayo2020sanitycheckssaliencymaps} or sensitive to input transformations that have no effect on the model~\citeest{kindermans2017unreliabilitysaliencymethods}. For safety-relevant applications, attribution methods help with triage, debugging, and initial auditing, but they are insufficient on their own when an explanation of internal computation is required.

\subsection{Representation Probing and Related Post-hoc Analyses}
Probing classifiers~\citeest{belinkov-2022-probing} assess whether a model's activations encode particular information by training ``lightweight'' classifiers---often logistic regression models---on intermediate representations to predict a target property, such as part-of-speech tags, syntactic structure, and gender~\citeest{belinkov2017neural, hewitt2019structural, ravfogel-etal-2020-null}. 
Originally developed in the NLP community to analyse linguistic structure~\citeest{gupta-etal-2015-distributional, kohn-2015-whats, adi2017finegrained} and used by \citetest{alain2017understanding} to trace how representations evolve across layers of deep learning models, 
probes have become one of the most widely used tools in mechanistic interpretability for tasks including, but not limited to, layer-level localisation~\citeest{tenney2019bertrediscoversclassicalnlp} and detection of latent behaviours~\citeem{macdiarmid2024sleeperagentprobes}. We discuss the probing methodology, its limitations, and its relationship to causal interpretability in more detail in Section~\ref{subsec:probing}.

\subsection{Mechanistic Interpretability Methods}
Mechanistic interpretability (MI) methods aim to causally understand model internals, such as the semantic meaning of parameters or activations, and how they relate to model behaviour. They are largely human-centric in the sense that the desired mechanistic understanding is one that humans can interpret and reason about---whether through natural-language descriptions~(\citeestalp{cunningham2023sparse}; \citeemalp{ameisen2025circuit}), plots of attention patterns~(\citeemalp{clark2019what}; \citeestalp{vig2019multiscale}), or token distributions~\citeem{nostalgebraist2020interpretinggpt, belrose2025elicitinglatentpredictionstransformers, jlens}. Recent work, however, has questioned whether ``machine languages'' and human languages fully overlap, suggesting that some meaningful internal computations may not admit faithful descriptions in natural language~\citeest{hewitt2025position}.

Mainstream MI approaches include, but are not limited to, feature decomposition and extraction~\citeest{cunningham2023sparse, dunefsky2024transcodersinterpretablellmfeature}, circuit discovery~(\citeemalp{syed2023attributionpatchingoutperformsautomated, ameisen2025circuit}; \citeestalp{conmy2023automatedcircuitdiscoverymechanistic}), and methods for decoding internal representations into natural language~(\citeestalp{patchscopes, pan2024latentqateachingllmsdecode}; \citeemalp{nla}), all of which are discussed in later sections. Different methods rest on different assumptions and produce different types of explanation, making them suitable for different use cases but with distinct limitations. For example, sparse autoencoders decompose model representations (e.g., residual-stream activations at the last token position of a particular layer) into thousands of monosemantic features, though many extracted features may be non-causal or illusory~\citeest{wu2025axbench, leask2025sparse, heap2026automated}. Circuit discovery methods identify fine-grained computational graphs responsible for particular model behaviours, but these graphs may capture only a subset of the model's underlying computation and can be difficult to validate for sufficiency or identifiability~\citeest{miller2024transformer, meloux2025everything}. Despite these limitations, MI methods remain promising complementary tools that collectively enable systematic analysis of model internals.

\subsection{When Mechanistic Understanding Becomes Necessary}
\label{subsec: mech interp necessary}
The methods we have discussed so far form a correlational-to-causal spectrum: behavioural evaluation establishes \textit{what} a model does, attribution and probing identify \textit{where} relevant information resides, and mechanistic analysis reveals \textit{how} computations produce behaviour. These approaches complement each other: behavioural evaluations help validate mechanistic findings, and behavioural anomalies can motivate mechanistic investigation.

Mechanistic analysis becomes necessary when we need to verify \textit{how} a model produces a behaviour---not just \textit{that} it does under test conditions. For example, a probe can detect deceptive reasoning, but we still need targeted interventions to validate whether the detection is a false alarm; if not, we will further need more rigorous mechanistic analysis to understand why the model deceives. Likewise, unexpected generalisation or failure modes may require mechanistic analysis to examine whether they stem from identifiable structures. Concretely, mechanistic methods are most needed for: identifying intents hidden from outputs, such as deception or side quests; understanding the internal mechanism of a model behaviour; verifying the \textit{absence} of a capability (behavioural tests can only confirm presence); or providing safety guarantees that must hold beyond the evaluated input distribution. However, the high granularity of mechanistic analysis comes at high practical cost, including not only compute but also researcher time, experimental complexity, and slower iteration. These costs position mechanistic methods as a complement to lighter-weight approaches, such as behavioural evaluations, rather than a full replacement.

\begin{table}[tb]
\centering
\caption{Trade-offs and characteristics of interpretability methods.}
\label{tab:interpretability-methods}
\renewcommand{\arraystretch}{1.3}
\begin{tabular}{>{\raggedright\arraybackslash}p{3cm} 
                >{\raggedright\arraybackslash}p{4cm} 
                >{\raggedright\arraybackslash}p{4cm} 
                >{\raggedright\arraybackslash}p{4cm}}
\toprule
\rowcolor{tableheader!80} \textcolor{white}{\textbf{Method}} & 
\textcolor{white}{\textbf{Core Idea}} & 
\textcolor{white}{\textbf{Strengths}} & 
\textcolor{white}{\textbf{Intended Use and Limitations}} \\
\midrule
\rowcolor{tablerow1} \textbf{Behavioural (e.g., red-teaming, benchmarks)} & Examine model outputs in response to inputs & Fast, scalable, practical for broad safety coverage & Not designed to reveal internal computation; suitable for monitoring, triage, and regression detection \\
\rowcolor{tablerow2} \textbf{Input attribution (e.g., SHAP, saliency)} & Identify which inputs most influence outputs & Token-level explanations; scalable & Focuses on input--output influence, not internal mechanisms; useful for auditing and debugging \\
\rowcolor{tablerow1} \textbf{Concept-based (e.g., CBMs, probes)} & Correlationally analyse or constrain interpretable representations & Reveals encoded information; transparent abstractions; enables targeted intervention & Not mechanistic without causal intervention; supports localisation, auditing, and early detection tasks \\
\rowcolor{tablerow2} \textbf{Mechanistic (e.g., circuits, SAEs)} & Analyse internal computations and causal structure & Reveals algorithms; enables targeted intervention & Computationally intensive; complements other methods when causal understanding is required \\
\bottomrule
\end{tabular}
\end{table}

\subsection{Limitations of Mechanistic Analysis}
\label{subsec:limits}

Mechanistic interpretability aims to provide causal explanations of model behaviour, but it faces important practical and conceptual limitations. First, mechanistic explanations are often incomplete. An analysis may identify an internal computation (e.g., ``This MLP retrieves factual knowledge about Paris'', as identified through causal tracing~\citeest{meng2023locatingeditingfactualassociations}) that captures a non-trivial causal pathway without identifying all others contributing to the behaviour. This concern makes it hard to tell whether the identified computation is sufficient or just one of several mechanisms jointly implementing the behaviour.

Second, mechanistic analyses can scale poorly. Analysing large models and complex behaviours takes much more time and resources because the relevant computations may be distributed across model components and complex behaviours involve interactions across long generation trajectories that are hard to isolate. For example, most current circuit-discovery methods~(\citeemalp{ameisen2025circuit}; \citeestalp{conmy2023automatedcircuitdiscoverymechanistic, eapig, marks2025sparse, querycircuit}) identify the computational graph responsible for generating a specific token. It remains challenging to identify and summarise the computations underlying behaviours that unfold across multiple generated tokens.

Third, even when we identify and causally validate an internal mechanism, this does not automatically mean we can reliably control the corresponding behaviour~\citeest{localinform}. The effectiveness, specificity, and generalisability of localisation and intervention techniques, such as identifying a model component and then patching its activations or changing its weight matrices, depend on several factors, such as the hydra effect~\citeem{mcgrath2023hydraeffectemergentselfrepair}, beyond how accurately the identified localisation captures the underlying computation for the behaviour. Furthermore, an intervention that successfully steers one model's behaviour on a task may not transfer to another model or setting.

These are open challenges but not fundamental barriers. Active research in areas such as establishing more effective representational units, discovering more robust model properties, developing more efficient circuit-discovery methods, and designing more reliable intervention approaches across distributions is working to address them. Nevertheless, these studies highlight a clear gap between mechanistic explanation and reliable control or safety guarantees.

\begin{warningbox}[Limitations of Mechanistic Analysis]
Mechanistic explanations may be incomplete or non-causal, difficult to scale to large models or complex behaviours, and sensitive to modelling assumptions. They often fail to capture the full picture of a model's underlying mechanism and do not automatically provide reliable, generalisable control. Intervening on an identified mechanism may work in one setting but fail to transfer to another. However, these limitations do not imply that mechanistic understanding cannot enable reliable control. Instead, they indicate that moving from explanation to robust, generalisable intervention requires additional studies and remains an open challenge in the field~\citeest{sharkey2025open}.
\end{warningbox}

\paragraph{Summary.} This section discussed different interpretability methods based on the evidence they provide. Behavioural methods measure and analyse model responses to inputs; concept-based methods explain predictions through human-interpretable concepts; input attribution methods identify which parts of the input influence an output; probing tests what information is decodable from representations; mechanistic methods identify and describe internal mechanisms that produce a behaviour. Among them, mechanistic analysis offers the strongest claims, but it is costly, often incomplete, and does not guarantee reliable control. In the next section, we introduce the core concepts on which mechanistic analysis is built.

\section{Foundational Concepts for Mechanistic Interpretability}
\label{sec:foundations}

This section introduces foundational concepts in mechanistic interpretability. We establish a shared notation and vocabulary for use throughout the survey. We avoid committing to specific hypotheses about the representations and workings of model activations. 

\subsection{Behavioural vs.\ Mechanistic Understanding}
\label{subsec:behavioral-vs-mechanistic}

Behavioural analysis describes what a model does, but safety-critical applications also require understanding its internal mechanism (as we reviewed in Section~\ref{sec:why_open}).

\begin{definitionbox}[Behavioural Interpretability]
Behavioural interpretability studies a model $f_\theta : \mathcal{X} \to \mathcal{Y}$
through input--output pairs $(x, f_\theta(x))$ for $x \in \mathcal{D} \subseteq \mathcal{X}$,
where $\mathcal{D}$ is a finite test set. Behavioural interpretability treats the model as a black box.
\end{definitionbox}

\begin{definitionbox}[Mechanistic Interpretability]
Mechanistic interpretability studies a model $f_\theta : \mathcal{X} \to \mathcal{Y}$ by examining
its internal computations, including model activations and parameters $\theta$ that produce them, at different levels of granularity in relation to its behaviour. In short, mechanistic interpretability characterises how a model's behaviour or output is computed.

\end{definitionbox}

\subsection{Neuron Polysemanticity in Model Representations}
\label{subsec:polysemanticity}

\paragraph{What Is a Concept?}
Concepts play a central role in many interpretability methods and will be used throughout this paper to define key ideas such as features and polysemanticity. It is therefore worth clarifying what we mean by the term. Notably, concepts do not necessarily correspond to categories humans can readily name or understand; neural networks may represent them in ways humans lack the vocabulary to describe~\citeest{hewitt2025position}.

\begin{definitionbox}[Concept]
A concept is a distinct, reusable unit of information or knowledge that can be identified across different instances or contexts. Concepts may correspond to familiar human categories, such as ``a dog'', ``the colour red'', or ``words of praise'', but they need not admit a natural human description.
\end{definitionbox}

Cross-context identifiability concerns the meaning of the information unit rather than the process by which the information is recognised. Determining whether a concept is present in an information unit may itself require additional contextual information. For example, the word (i.e., an instance of an information unit) ``bank'' can refer to a financial institution or a river bank, depending on the surrounding sentence. Nevertheless, once identified, each concept can be recognised as the same information across different contexts. For example, the concept of ``financial institution'' is expected to remain sufficiently stable across the contexts in which it appears.

This definition accommodates both human-interpretable and machine-specific concepts. A machine-specific concept may lack a natural human description yet still form a coherent, reusable unit of information in the neural network~\citeest{dalvi2022discovering, hewitt2025position}. Human interpretability should therefore be treated as a property of some concepts rather than as a necessary condition for the information unit to qualify as a concept. Notably, most interpretability research focuses on concepts that humans can readily understand. Identifying and characterising concepts that matter to a model's computation but do not map neatly onto human-understandable categories remains underexplored.

\paragraph{Neuron Polysemanticity.}
For a single neuron $n_\ell$ in layer $\ell$, consider its activation at token position $t$ for some input prompt $x$, denoted by $n_\ell^{t}(x)$. We can straightforwardly interpret a neuron if it responds to a single concept. For example, if $n_\ell^{t}(x)$ is large and positive whenever position $t$ corresponds to the word ``blue'', remains close to zero otherwise, and can causally shift the model's outputs towards or away from blue-related concepts when manipulated, we can conclude that the neuron corresponds to the concept of blue. Assuming some neurons exhibit such a property, early work proposed methods to discover and interpret individual neurons' functions~\citeest{netdissect2017, bau2018identifying, Dalvi_Durrani_Sajjad_Belinkov_Bau_Glass_2019, bau2020units, hernandez2022natural, oikarinen2023clipdissect, and}. However, follow-up studies have found that, due to representational constraints and superposition~\citeem{elhage2022superposition} (detailed in Section~\ref{subsec:superposition}), individual neurons often respond strongly to multiple concepts that appear unrelated from a human perspective~\citeem{elhage2022superposition}. This phenomenon is known as \textit{neuron polysemanticity}.

\begin{definitionbox}[Neuron Polysemanticity]
A neuron $n_\ell$ is \emph{polysemantic} if there exist two distinct concepts $c_1$ and $c_2$, together with corresponding sets of inputs $x_1$ and $x_2$ as well as token positions of interest $t_{c_1}$ and $t_{c_2}$, such that $n_\ell^{t_{c_1}}(x_1)$ and $n_\ell^{t_{c_2}}(x_2)$ are both large, and interventions on these activations causally influence model behaviour associated with both concepts.

\end{definitionbox}

Polysemanticity complicates neuron-level explanations and suggests that human-interpretable semantic content may not be aligned with individual neurons. This motivates searching for representational units beyond individual neurons.

\subsection{Features as Representational Units}
\label{subsec:features}

Since individual neurons are often polysemantic, mechanistic interpretability has increasingly shifted towards alternative units of analysis. For example, it has been found that models sometimes encode human-interpretable concepts as linear directions or low-dimensional manifolds in the activation space~(\citeestalp{arditi2024refusallanguagemodelsmediated}; \citeemalp{wurgaft2026manifoldsteeringrevealsshared}). We thus use the term \emph{feature} broadly to encompass any model representation corresponding to a human-interpretable concept. Such distributed representations motivate analysing the activation space as a whole, rather than individual neurons in isolation~\citeem{elhage2022superposition, bricken2023monosemanticity, templeton2024scaling}.

Specifically, we treat a feature technically as a geometric structure $\phi$ in the activation space $\mathbb{R}^d$ of a model component, such as the residual stream, an attention-head output, or an MLP output, where $d$ denotes the hidden dimension. Conceptually, a feature is a manipulable manifestation of a concept of interest. As with neurons, for a feature $\phi$ claimed to correspond to a concept $c$, we expect interventions that increase or suppress the activation of $\phi$ to causally influence model behaviour associated with $c$. In this sense, features provide a more expressive, general representational unit than individual neurons for describing how neural networks may encode concepts.

\begin{definitionbox}[Feature]
A feature $\phi$ associated with a concept $c$ is a geometric structure in the model's activation space $\mathbb{R}^d$ whose activation is predictive of, and causally relevant to, the presence or expression of $c$.
\end{definitionbox}

At this stage, we remain agnostic about the precise mathematical form features take and how the model encodes them. Instead, we treat existing methods for identifying and manipulating features as operational approximations shaped by their methodological assumptions and constraints. We examine these representational hypotheses explicitly in Section~\ref{sec:hypotheses}.

Within this operational perspective, researchers can identify features using several methodological families, including readout-based methods~\citeest{patchscopes, pan2024latentqateachingllmsdecode, ao} and sparse decomposition techniques~\citeest{cunningham2023sparse, dunefsky2024transcodersinterpretablellmfeature}, and then manipulate them to test causal relevance~\citeest{wu2025axbench, mueller2025mib}. In practice, many such approaches operationalise features as approximately linear or linearised components of internal representations~(\citeemalp{yun-etal-2021-transformer}; \citeestalp{marks2024the, arditi2024refusallanguagemodelsmediated}). Examples include linear probes and dictionary-learning methods~(\citeestalp{belinkov-2022-probing}; \citeemalp{bricken2023monosemanticity}).

An important caveat is that, for a given concept, an identified manipulable geometric structure---that is, a candidate feature---may be only one of many possible internal representations associated with that concept. For example, a language model may encode the concept of ``refusal'' in a complex manner across multiple layers and components. Successfully identifying a linear direction in the residual stream that can steer the model towards or away from refusal~\citeest{arditi2024refusallanguagemodelsmediated} does not imply that this direction uniquely or exhaustively captures how refusal is represented. Other causally relevant structures (e.g., low-dimensional manifolds) for refusal may also exist but remain unidentified.

\subsection{Transformer Computation}
\label{subsec:transformers}

Transformers~\citeest{vaswani2017attention} have become the dominant architecture underlying modern language-model-based AI assistants such as ChatGPT and Claude, and are consequently the primary subject of study in mechanistic interpretability. Several properties of the transformer architecture, most notably its modular structure and additive information flow, make it particularly amenable to mechanistic analysis~\citeem{elhage2021mathematical}. In particular, transformer components can be viewed as reading from and writing to a shared \emph{residual stream}, allowing information to persist and be incrementally transformed as it passes through successive layers~\citeem{elhage2021mathematical}. Below, we briefly review the main classical components of the transformer architecture.

\begin{definitionbox}[Residual Stream and Computational Components]
Let
\[
h^{1:T}_\ell(x) \in \mathbb{R}^{d \times T}
\]
denote the sequence of residual-stream activations at layer $\ell$ for an input
sequence $x$, where $T$ is the sequence length and $d$ is the model hidden size.
The residual stream serves as the shared communication channel through which
transformer components exchange information. Below, in this box, we let $h_\ell(x) \coloneqq h^{1:T}_\ell(x)$ to simplify the notation.\\

For a standard transformer layer $\ell$, consisting of a multi-head attention module $\mathrm{Attn}_\ell(\cdot)$ and a multi-layer perceptron $\mathrm{MLP}_\ell(\cdot)$, we have
\begin{align*}
h'_\ell(x)
    &= h_{\ell-1}(x)
       + \mathrm{Attn}_\ell(h_{\ell-1}(x)), \\
h_\ell(x)
    &= h'_\ell(x)
       + \mathrm{MLP}_\ell(h'_\ell(x))
    = h_{\ell-1}(x)
       + \mathrm{Attn}_\ell(h_{\ell-1}(x)) + \mathrm{MLP}_\ell(h'_\ell(x)).
\end{align*}

Unrolling the recursion gives
\[
h_\ell(x)
=
h_0(x)
+
\sum_{k=1}^{\ell}
\Big(
\mathrm{Attn}_k(h_{k-1}(x))
+
\mathrm{MLP}_k(h'_k(x))
\Big),
\]

where \(h_0(x)\) denotes the input embedding representation. The equation shows that the residual stream at layer \(\ell\) equals the input embeddings plus the accumulated outputs of all computational components that have written to the residual stream up to that point. This perspective is central to mechanistic interpretability, as it lets us decompose model computation into additive contributions from individual components~\citeem{elhage2021mathematical}.

\end{definitionbox}

For simplicity, we omit layer-normalisation operations from the notation. Following \citetem{elhage2021mathematical}, most of the layer-normalisation transformation can be absorbed into adjacent linear components. Specifically, for a fixed input, layer normalisation is well approximated as an affine transformation with a scalar rescaling factor. The affine part can be folded into neighbouring parameters, and the remaining scalar normalisation term simply reweights component contributions. This approximation lets us treat the residual stream as an additive communication channel without explicitly representing normalisation at every layer.

Attention heads and MLPs play distinct roles in transformer computation. Attention modules primarily route, aggregate, and compose information across token positions~(\citeemalp{olsson2022incontextlearninginductionheads}; \citeestalp{wang2022interpretabilitywildcircuitindirect}), whereas MLPs often perform feature transformation and factual knowledge retrieval~\citeest{geva-etal-2021-transformer, meng2023locatingeditingfactualassociations}. This functional separation enables more localised analyses of how the model routes, transforms, and stores information, as well as how attention modules and MLPs interact~\citeest{neo-etal-2024-interpreting}.

Below, we detail the computations of canonical attention modules and MLPs.

\begin{definitionbox}[Attention Module]

A multi-head attention module consists of $N_H$ attention heads operating in parallel. Consider the residual-stream activations $ h_\ell^{1:T}(x) \in \mathbb{R}^{d \times T}$ and let $h := h_\ell^{1:T}(x)$ for notational simplicity. For each head $r \in \{1,\ldots,N_H\}$, the residual stream is first projected into query, key, and value representations:
\[
Q^r = W_Q^r h,
\qquad
K^r = W_K^r h,
\qquad
V^r = W_V^r h,
\]
where $W_Q^r, W_K^r, W_V^r \in \mathbb{R}^{d_{\text{head}} \times d}$ and $d_{\text{head}}$ denotes the dimensionality of each attention head. The \emph{attention pattern} of head $r$ is then computed as
\[
A^r
=
\operatorname{softmax}\!\left(
\frac{(Q^r)^\top K^r}{\sqrt{d_{\text{head}}}}
\right)
\in
\mathbb{R}^{T \times T},
\]
where the softmax is applied row-wise. Thus, $A^r_{t,t'}$ denotes the attention weight assigned by the query at position $t$ to the key at position $t'$. The head then writes an update to the residual stream:
\[
\Delta h^r
=
W_O^r V^r (A^r)^\top,
\]
where $W_O^r \in \mathbb{R}^{d \times d_{\text{head}}}$. The output of the full multi-head attention module is the sum of the contributions from all heads:
\[
\operatorname{Attn}(h)
=
\sum_{r=1}^{N_H}
\Delta h^r.
\]

\end{definitionbox}

\begin{definitionbox}
[MLP Block]
An MLP block applies a position-wise nonlinear transformation to each position of the residual stream. In the canonical transformer architecture~\citeest{vaswani2017attention}, this transformation takes the form
\[
\Delta h_{\text{MLP}} = W_2 \,\sigma(W_1h+b_1)+b_2,
\]
where $W_1 \in \mathbb{R}^{d_{\text{mlp}}\times d}$, $W_2 \in \mathbb{R}^{d\times d_{\text{mlp}}}$, and $\sigma(\cdot)$ denotes a nonlinear activation function, for which the original transformer uses ReLU~\citeest{relu}. The same transformation is applied independently at each position in the sequence. MLP blocks primarily
transform and generate internal features, writing new directions into the
residual stream.\\

Notably, $W_1$ typically projects the activations into a much higher-dimensional space ($d_{\text{mlp}} \gg d$), which has recently been found to yield more interpretable representations natively in the model~\citeest{ICLR2025_8c5f3029, arora2026language}.

\end{definitionbox}

The transformer components discussed above correspond to their canonical formulations. Numerous variants have since been proposed and adopted in modern architectures. For example, variants of the attention module include Grouped Query Attention (GQA)~\citeest{gqa}, used in Llama 3~\citeem{llama3}, and Multi-Head Latent Attention (MLA)~\citeem{deepseekv2}, used in DeepSeek-V2~\citeem{deepseekv2} and DeepSeek-V3~\citeem{deepseekai2025deepseekv3technicalreport}. Variants of the MLP block include gated architectures such as SwiGLU~\citeem{shazeer2020gluvariantsimprovetransformer}, used in Llama 3, and Mixture-of-Experts (MoE) architectures, used in DeepSeek-V3.

\subsection{Circuits as Descriptions of Internal Mechanisms}
\label{subsec:circuits}

Sections~\ref{subsec:polysemanticity}--\ref{subsec:features} introduce neurons and features as representational units for describing what is represented in model activations, and Section~\ref{subsec:transformers} introduces the computations that transform and route activations through the model. We use the term \emph{mechanism} to refer to the organisation of model activations, their representational units, or the transformations applied to them that together produce a particular model behaviour. A central topic in mechanistic interpretability is identifying and understanding such mechanisms in neural networks.

A common way to represent and describe such a mechanism is through a \emph{circuit}: a subgraph of the model's computational graph whose \emph{nodes} correspond to computational components or representational units and whose \emph{edges} represent information flow between them~\citeem{elhage2021mathematical}.

\begin{figure}[tb]
  \centering
  \includegraphics[width=0.7\textwidth]{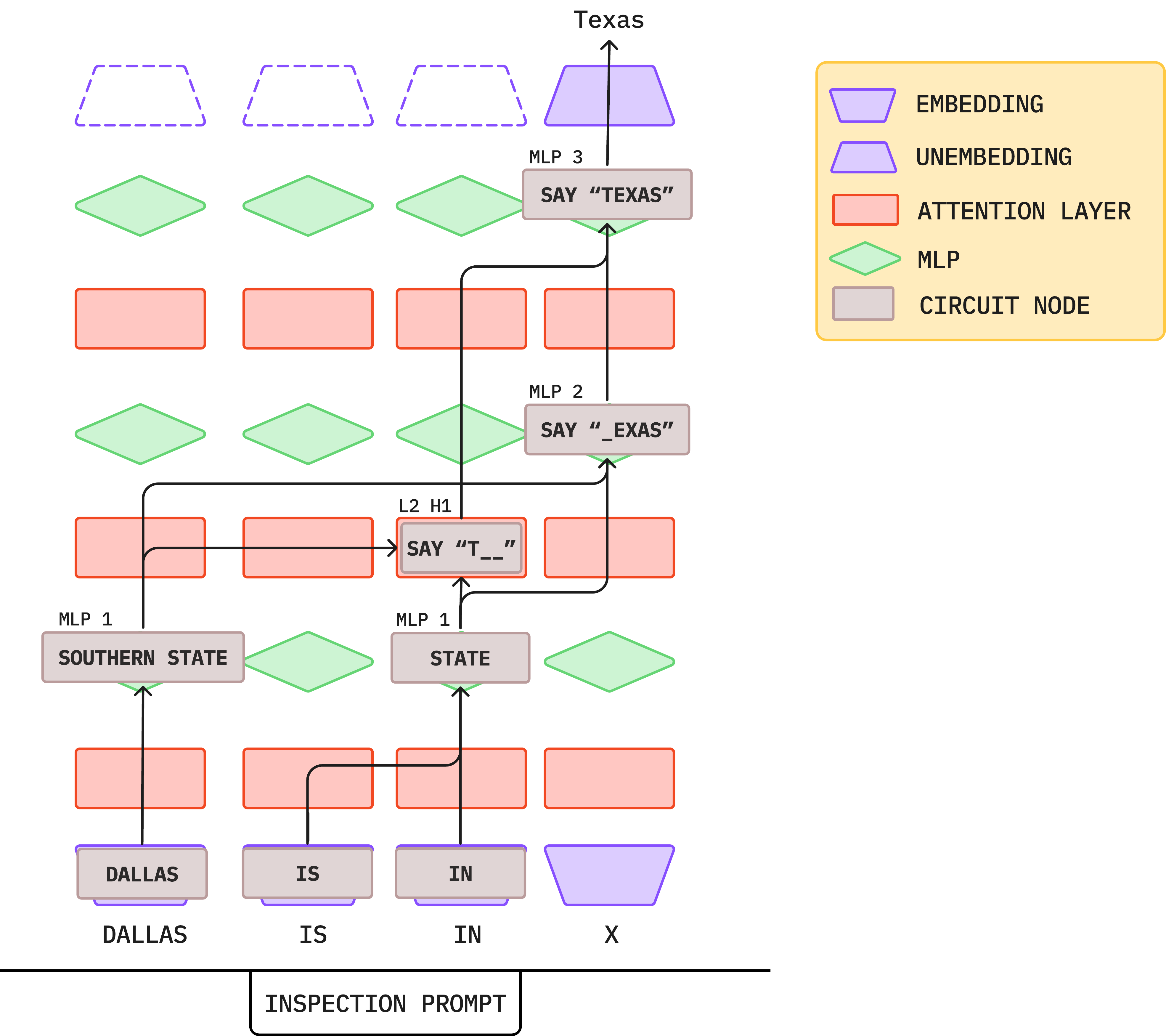}
  \caption{\textbf{Circuit.} This diagram illustrates a circuit within a transformer, where specific nodes (MLP layers and attention heads in this case) perform specialised computations.}
  \label{fig:circuit}
\end{figure}

\begin{definitionbox}[Circuit (Figure~\ref{fig:circuit})]
For a fixed model computation, let \(G\) denote the corresponding directed acyclic computational graph. A circuit \(\mathcal{C} \subseteq G\) is a subgraph of \(G\), where:
\begin{itemize}[nosep]
    \item \textbf{Nodes:} computational components, such as \(\mathrm{Attn}_\ell\) and \(\mathrm{MLP}_\ell\), or representational units, such as features \(\phi_i\);
    \item \textbf{Edges:} directed connections representing information flow or transformation between nodes.
\end{itemize}
The subgraph \(\mathcal{C}\) represents a collection of components and interactions that jointly contribute to a particular computation or model behaviour, and can therefore serve as a description of the underlying mechanism.
\end{definitionbox}

\begin{examplebox}[Examples of Circuits]
\begin{itemize}
    \item \textbf{Induction head circuit:} Two attention heads cooperate to implement in-context learning~\citeem{olsson2022incontextlearninginductionheads}.
    \item \textbf{QK circuits:} Query--Key interactions prioritising specific token positions~\citeem{kamath2025tracing}.
\end{itemize}
\end{examplebox}

The study of circuits has developed along two complementary lines. The first line of work~\citeest{wang2022interpretabilitywildcircuitindirect, conmy2023automatedcircuitdiscoverymechanistic, eapig, haklay-etal-2025-position}, referred to as \emph{capability circuit discovery} by \citetest{querycircuit}, aims to identify minimal sets of architectural components, such as attention heads or MLP layers, that are sufficient to reproduce a given \textit{behaviour} or \textit{capability} of the model, such as indirect object identification~\citeest{wang2022interpretabilitywildcircuitindirect}, numerical comparison~\citeest{hanna2023how}, and in-context learning~\citeest{kharlapenko2025scaling}. In this setting, the capability of interest is defined as a dataset of queries that exercise it, and a discovered circuit is considered successful if it recovers the model's performance on this dataset.

The second line of work, referred to as \emph{query-level circuit discovery}~\citeest{querycircuit}, focuses on how internal features interact and propagate across the computational graph to produce the model's decision for a particular input query~(\citeemalp{ameisen2025circuit}; \citeestalp{querycircuit}). Studies in this line usually use sparse autoencoders or cross-layer transcoders (detailed in Section~\ref{sec:sae}) to automatically decompose MLP outputs into interpretable units, trace attributions between these features, and track how information flow converges to the final prediction. 

\paragraph{Summary.}
This section introduced the foundational concepts of mechanistic interpretability: neuron polysemanticity, which motivates the use of features as more expressive units of analysis; transformer computation, which describes how representations are transformed and information is routed through the model; and internal mechanisms, which organise these units, transformations, and information flows to produce model behaviour and can be described by circuits. How we interpret and leverage these concepts depends on assumptions about the structure of model representations and computation, which we examine explicitly in Section~\ref{sec:hypotheses}.

\section{Core Hypotheses in Mechanistic Interpretability}
\label{sec:hypotheses}

Mechanistic interpretability rests on several core hypotheses about how neural networks represent information and implement computation. These hypotheses are not yet rigorously proven. Nonetheless, they have shaped methodological choices and empirical investigations. In this section, we formalise four of them, discuss their conceptual basis, and summarise supporting and opposing empirical evidence.

\subsection{The Linear Representation Hypothesis}
\label{subsec:linearity}

We have seen that single neurons may not be the most effective unit of representation in many cases, but what is? The linear representation hypothesis claims that \emph{linear} directions in the activation space are effective fundamental units of representation~\citeest{lrh}.

\begin{definitionbox}[Linear Representation Hypothesis]
A concept can be effectively represented by a feature corresponding to a linear direction $\phi$ in activation space $\mathbb{R}^d$. Linearity implies that adding or subtracting this direction, or a scalar multiple of it, from model activations can correspondingly influence model behaviour, while the projection $\phi^\top h(x)$ approximately measures the strength of the feature in model activations $h(x)$ for an input $x$.\\

Under a strong version of the linear representation hypothesis, all concepts are best represented by linear features. In this view, model activations can be approximately decomposed as an additive combination of such features,
\[
h(x) \approx \sum_i \alpha_i \phi_i,
\]
where $\alpha_i \in \mathbb{R}$ denotes the strength of feature $\phi_i$ dependent on the input $x$. Under a weaker version, only some concepts admit such linear representations~(\citeestalp{sharkey2025open}; \citeemalp{smith2024strongfeature}).
\end{definitionbox}

\paragraph{Conceptual Basis.} A feature associated with a concept is linearly represented if its presence corresponds to a fixed direction $\phi$ in the activation space and can be detected through the projection $\phi^\top h(x)$. This linearity makes mechanistic analysis tractable because we can use simple methods to identify, measure, and manipulate these features in the activation space, such as scaling them and adding them back to model activations.

\begin{figure}[tb]
  \centering
  \includegraphics[width=\linewidth]{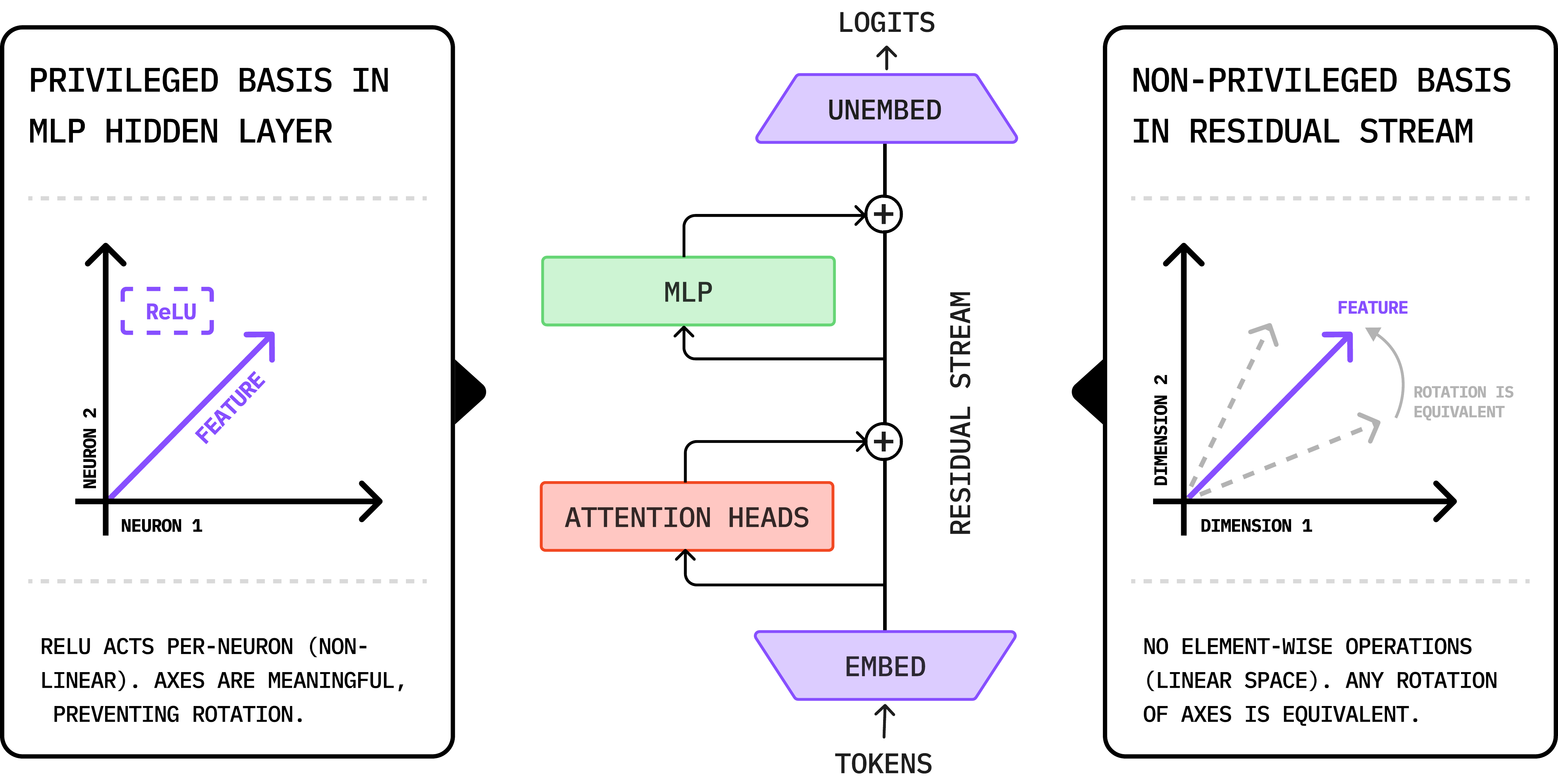}
\caption{Privileged versus non-privileged bases. In the hidden layers of MLP blocks, element-wise nonlinear operations break rotational symmetry and make individual axes meaningful. In contrast, the residual stream contains no such operations that privilege particular axes, so rotating the basis preserves the relevant computation.}
  \label{fig:privileged-basis}
\end{figure}

Notably, activation spaces in different model components can have distinct geometric structures. For example, in the hidden states of MLP blocks, the neuron axes are computationally distinguished because element-wise nonlinear activation functions, such as ReLU, act separately on each axis. The coordinate system is therefore meaningful and is referred to as a \emph{privileged basis} (Figure~\ref{fig:privileged-basis}, left)~\citeem{elhage2022superposition}. In other spaces, such as the residual stream, no operation singles out the neuron axes, so linear features may lie along arbitrary directions. Such spaces are said to have a \emph{non-privileged basis} (Figure~\ref{fig:privileged-basis}, right).

When the basis is privileged, a single neuron can serve as a scalar feature that meaningfully represents a concept. When it is not, the feature may instead correspond to directions or other geometric structures that do not align with any individual neuron. In this case, specialised methods are required to identify them from the activation space.

\paragraph{Empirical Evidence and Counterexamples.}
Empirical support for the weaker version of the linear representation hypothesis includes the success of linear probes across diverse models and concepts~\citeest{belinkov-2022-probing}, activation steering along single directions to induce targeted behavioural changes~\citeem{turner2024steeringlanguagemodelsactivation}, and sparse autoencoders recovering linearly accessible and steerable features at scale~\citeem{bricken2023monosemanticity}. At the same time, there is substantial evidence that not all representations are well described as linear. Counterexamples include nonlinear feature interactions, algorithmic representations that rely on nonlinear transformations, and features distributed across multiple interacting subspaces~(\citeestalp{nanda2023progressmeasuresgrokkingmechanistic, engels2025languagemodelfeaturesonedimensionally, smds}; \citeemalp{rnnnonlinear, feucht2026arithmeticwildllamauses, wurgaft2026manifoldsteeringrevealsshared}). Some work has further argued that features may not always exist as explicitly represented objects, but may instead emerge from underlying circuit dynamics~\citeem{smith2024strongfeature}. The linear representation hypothesis is therefore best viewed as a useful approximation that holds in some settings, rather than as a universal law.

\subsection{The Superposition Hypothesis}
\label{subsec:superposition}

The superposition hypothesis says that models encode more features than their hidden representations have dimensions.

\begin{definitionbox}[Superposition Hypothesis]
Model hidden states encode more features than they have dimensions. If there are $m$ features and 
$d$ dimensions, then the superposition hypothesis claims that $m \gg d$.\\

This implies that features cannot be represented as orthogonal directions---though they can be represented as nearly orthogonal directions. It also implies neuron polysemanticity: the same neuron participates in the encoding of multiple unrelated concepts.
\end{definitionbox}

\paragraph{Conceptual Basis.}

\begin{figure}[tb]
  \centering
  \includegraphics[width=\linewidth]{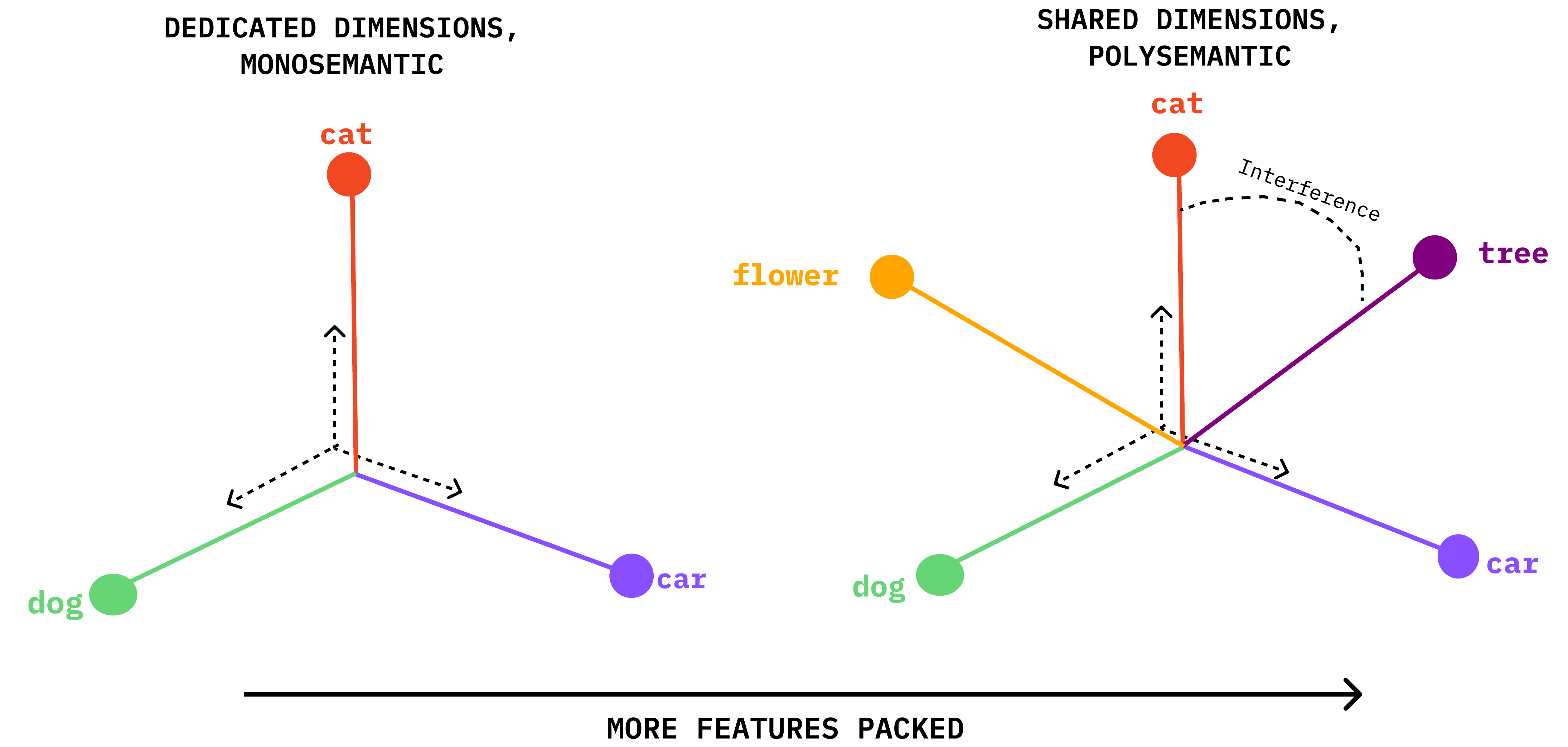}
  \caption{\textbf{Superposition implies polysemanticity.} Left: when the number of features $m$ does not exceed the number of dimensions $d$, i.e., $m\le d$, feature directions can align cleanly with individual neuron axes, producing monosemantic neurons. Right: when $m>d$, multiple feature directions must share the same limited set of neuron axes; as a result, each neuron axis cuts across multiple feature directions, producing polysemantic neurons.}
  \label{fig:superposition-polysemanticity}
\end{figure}

Superposition is a geometric property of learned representations in which a limited number of neuron dimensions encode many more features by allowing those features to share representational dimensions. Neuron polysemanticity, where individual neurons respond to multiple seemingly unrelated semantic concepts, naturally follows from this phenomenon. When the number of encoded features exceeds the activation-space dimensionality, the model can no longer allocate a dedicated neuron dimension to each feature. Instead, multiple features become superposed within the same activation space, shifting the representation from dedicated dimensions with monosemantic neurons to shared dimensions with increasingly polysemantic neurons (Figure~\ref{fig:superposition-polysemanticity}).

\paragraph{Empirical Evidence.}
Mechanistic studies have repeatedly observed widespread neuron polysemanticity: individual neurons often activate for multiple seemingly unrelated semantic concepts, making them difficult to interpret as standalone representational units~\citeem{olah2020zoom, elhage2022superposition, bricken2023monosemanticity}. Dictionary-learning methods such as sparse autoencoders exploit this structure to recover millions of distinct features from shared activation spaces~\citeem{bricken2023monosemanticity, templeton2024scaling}. Many discovered features are substantially more monosemantic than individual neurons, and some have been shown to causally influence model behaviour. Together, these findings suggest that the activation space may encode more meaningful semantic information along activation-space directions than along individual neuron axes.

However, the superposition hypothesis remains a hypothesis, not an established fact, in large language models. Observing polysemantic neurons does not, by itself, imply that superposition is the underlying cause, and dictionary-learning methods can recover interpretable features even in the absence of superposition. Furthermore, \citetem{lecomte2024what} showed that polysemanticity can arise even when the number of underlying features does not exceed the number of available neurons, suggesting alternative---and potentially complementary---mechanisms for the emergence of polysemantic representations.

\paragraph{Architectural Responses to Superposition.} Superposition is not an intrinsically inevitable property of deep neural networks but a consequence of architectural permissiveness. Without constraints on model architectures or activations, gradient descent can exploit the activation-space geometry to pack far more features than neurons~\citeem{elhage2022superposition}. This motivates a distinct research direction, \emph{interpretability-by-design} (as mentioned in Section~\ref{subsec:cbm} and discussed in Section~\ref{sec:architectural-interventions}). Interpretability-by-design argues that, rather than decomposing traditional models' superposed activations post hoc, one can design novel architectures or components that encourage representations where features are more accessible for analysis. Two intervention points have been explored:

\begin{itemize}
    \item \textbf{Sparsity constraints:} Enforce sparsity in weights or activations that limits how many neurons within an MLP or attention head can participate for a given input, constraining how many units can jointly support any single feature~(\citeestalp{yang2025mixture, gao2025weight}; \citeemalp{elhage2022solu}).
    \item \textbf{Computational primitives:} Change the fundamental units of analysis from static features in activations to learnable transformations by weight matrices that generate activations~(\citeestalp{liu2025kan}; \citeemalp{anthropic2025molt}). This paradigm better matches certain tasks and avoids the pressure to compress or suppress model representations.
\end{itemize}

These architectural interventions shift from treating superposition as a phenomenon to reverse-engineer to treating it as a design choice that can be constrained upstream. We discuss these methods in detail in Section~\ref{sec:architectural-interventions}.

\subsection{The Privileged Access Hypothesis} 
\label{subsec:privileged-access}
A class of approaches that use language models as interpreters of \textit{themselves} can be justified through the privileged access hypothesis. The privileged access hypothesis states that:
\begin{definitionbox}[Privileged Access Hypothesis]
A language model used to interpret itself can, under appropriate elicitation, access internal variables and intermediate computations that are not reliably recoverable from external behavioural analysis alone, providing a comparative advantage as an interpreter of its own behaviour~(\citeestalp{binder}; \citeemalp{li2025traininglanguagemodelsexplain, song}).
\end{definitionbox}

\paragraph{Conceptual Basis.} The ability to access and judge its own computation or reasoning---called \emph{introspective monitoring}---has long been a desired property of an AI system~\citeem{cox2008metareasoning}. Even earlier rule-based medical expert systems sought to explain their decisions, such as MYCIN's ``Explanation system''~\citeest{shortliffe1977mycin}. In this sense, using language models to explain themselves can be seen as a modern continuation of these earlier efforts.

A major difference is that rule-based systems can explicitly present the rule they executed, whereas modern language models compute in a highly nonlinear way. Although researchers have proposed mechanistic tools aiming to reverse-engineer part of the computation, the technical complexity of these techniques hinders their accessibility and scalability. The privileged access hypothesis then posits that the language model itself naturally stores information about the computation within its internal states, such as latent beliefs or intermediate plans. The model can thus learn to verbalise this information directly, serving as a scalable complement to mechanistic analysis.

\paragraph{Empirical Evidence.}

The most direct empirical evidence for this hypothesis comes from recent findings that models, even without explicit training to do so, are sometimes aware of concept representations (acquired by methods like difference-in-means) injected into their activations~\citeem{lindsey2026emergentintrospectiveawarenesslarge, fornasiere2026languagemodelsrecognizedropout}. Additional, though less direct, evidence comes from the design of explainer models---a general class of interpretability methods in which one model is fine-tuned to explain the target model’s activations~(\citeestalp{pan2024latentqateachingllmsdecode, ao}; \citeemalp{nla}). In practice, such explainers are typically initialised from a copy of the target model itself~(\citeestalp{pan2024latentqateachingllmsdecode, ao}; \citeemalp{buildingbetterao, pcd}). This usually works better than initialising from alternative models, suggesting the target model already encodes much of what is needed to interpret its own activations. In this sense, fine-tuning elicits and refines a latent capability rather than creating one from scratch.

Some studies also reported negative results for the privileged access hypothesis: \citetem{song} found that models predict their own sampling temperature no better than they predict other models', suggesting apparent introspection may reflect general reasoning about model-like systems rather than self-access. Separately, recent work has questioned whether internals-based explanations provide a practical advantage over behavioural baselines: \citetem{chive} found that activation-reading tools (including activation oracles) give agents no uplift over transcript-only baselines at predicting counterfactual model behaviours in the wild, and \citetest{li2026do} showed that standard verbalisation benchmarks can be solved without access to target-model internals at all~(see also \citeemalp{jakkli2026current}).

In summary, these results show that privileged access is \textit{possible} for certain tasks under favourable conditions, not \textit{universal} or \textit{reliable}. The hypothesis should therefore be treated with caution, as significant failure modes and limitations remain.

\subsection{The Universality and Platonic Representation Hypothesis}
\label{subsec:universality}

\begin{definitionbox}[Platonic Representation Hypothesis]
Models independently trained on similar tasks tend to develop similar internal representations or algorithms, which often approximate a task-determined canonical form (\citeestalp{huh2024platonicrepresentationhypothesis}; \citeemalp{olah2020zoom}). The degree of convergence depends on factors such as model size, architecture, and training data.
\end{definitionbox}

\paragraph{Conceptual Basis.}
A key distinction is the level at which models are expected to converge. \emph{Weak universality} refers to convergence in the algorithms or recurring computational patterns that models implement, even when these are encoded through different internal representations. \emph{Strong universality} makes the stronger claim that the internal representations themselves converge towards a shared canonical form.

\paragraph{Empirical Evidence and Trade-Offs.}
Weak universality is supported by recurring attention patterns, features, and circuits across models, tasks, and scales~(\citeestalp{nanda2023progressmeasuresgrokkingmechanistic, lan-etal-2024-towards}; \citeemalp{olah2020zoom, quirke2026understandingadditionsubtractiontransformers}). Strong universality is less supported and is usually demonstrated through representational similarity metrics~(\citeestalp{kornblith2019similarityneuralnetworkrepresentations, moschella2023relative}; \citeemalp{lan2025quantifyingfeaturespaceuniversality}), where recent studies have found the results to be fragile~(\citeestalp{groger2026revisiting}; \citeemalp{koepke2026platoscaveexaminingcrossmodal}). Divergent representations can arise from variations in architecture, data, or optimisation trajectories~\citeest{harrasse2025tinysqlprogressivetexttosqldataset, morcos2018insightsrepresentationalsimilarityneural}.

\paragraph{Implications.}
If strong universality holds, fine-grained mechanistic insights and the methods derived from them may transfer more readily across models. If only weak universality holds, such insights may remain contingent on factors such as model family and task~\citeest{bereska2024mechanisticinterpretabilityaisafety}, and even a complete mechanistic understanding of one model (see Section~\ref{subsec:ambitious}) may fail to generalise to others. Therefore, it is important to clarify the extent of universality for assessing the promise, limitations, and scalability of different interpretability research orientations, including the six major orientations introduced in Section~\ref{sec:orientations}.

\paragraph{Summary.} This section discussed four hypotheses that motivate many modern mechanistic interpretability studies. Linear representations support linear-feature-based analysis; superposition explains why we need to operate on linear or more complex forms of features instead of single neurons; privileged access explains the fundamental idea of model self-interpretation; universality asks whether insights transfer across models and at which levels. Each hypothesis holds in some settings and fails in others, and techniques built on them inherit those limitations. Section~\ref{sec:sae} introduces the main family of techniques and methods that operationalise the first two hypotheses (linear representation and superposition): dictionary learning for feature extraction and analysis.

\section{Feature Extraction and Analysis}
\label{sec:sae}

The critical notions introduced in Section~\ref{sec:foundations}, together with the hypotheses about the structure of neural network representations discussed in Section~\ref{sec:hypotheses}, provide the conceptual foundation for developing methods to identify and analyse manipulable internal objects associated with concepts of interest, i.e., features. One straightforward paradigm is to decompose model activations into sparse linear features. Sparse autoencoders (SAEs) provide a principled example of this approach, learning to represent any input activation with a subset of a large dictionary of monosemantic, interpretable features~\citeest{cunningham2023sparse}.

This section first introduces the canonical sparse coding problem and links it to SAEs, reviews how SAEs extract interpretable features, and then provides a practical overview and comparison of major SAE variants. We then introduce methods for multi-layer feature extraction and analysis, such as transcoders and crosscoders, and compare them with SAEs as well as with a broader range of alternative feature-extraction methods.

\subsection{The Sparse Coding Problem}
Early neuroscience studies posited that brains represent information using sparse, interpretable basis signals~\citeest{Barlow:1961aa, sparsecoding}. This hypothesis further motivated research to decompose natural images and compare the decomposed signals with cell activities in the primary visual cortex, finding surprising alignment~\citeest{olshausen1996emergence, olshausen1997sparse}. At the same time, the signal processing community has also examined representing a given signal as the sum of a subset of basis signals from a given overcomplete dictionary~\citeest{matchingpursuit, basispursuit}. Studies on learning better overcomplete dictionaries for high-quality representations also thrived~\citeest{lewicki2000learning, aharon2006k}. Together, they framed the sparse dictionary learning research question, with SAEs serving as a modern algorithm that uses neural network activations as the signal. Specifically, given data $h$, a sparse dictionary learning method finds the dictionary $D$ and sparse codes $z$ such that
\begin{equation}
h \approx Dz \text{ and } \|z\|_0 \text{ is small}.
\end{equation}
SAEs implement this with neural networks: $h$ are model activations, $D = W_{\text{out}}$ is an overcomplete dictionary to learn, and the sparse codes are $z = \sigma(W_{\text{in}} h + b_{\text{in}})$. Note that conventional dictionary methods usually run the algorithm to find a $z$ for each $h$, which does not scale to modern language models. In contrast, SAEs use a nonlinear encoder $E$ to predict the coefficients for different input activations.

\subsection{Sparse Autoencoders}
\label{subsec:sae-foundation}

\begin{figure}[tb]
  \centering
    \includegraphics[width=0.8\linewidth]{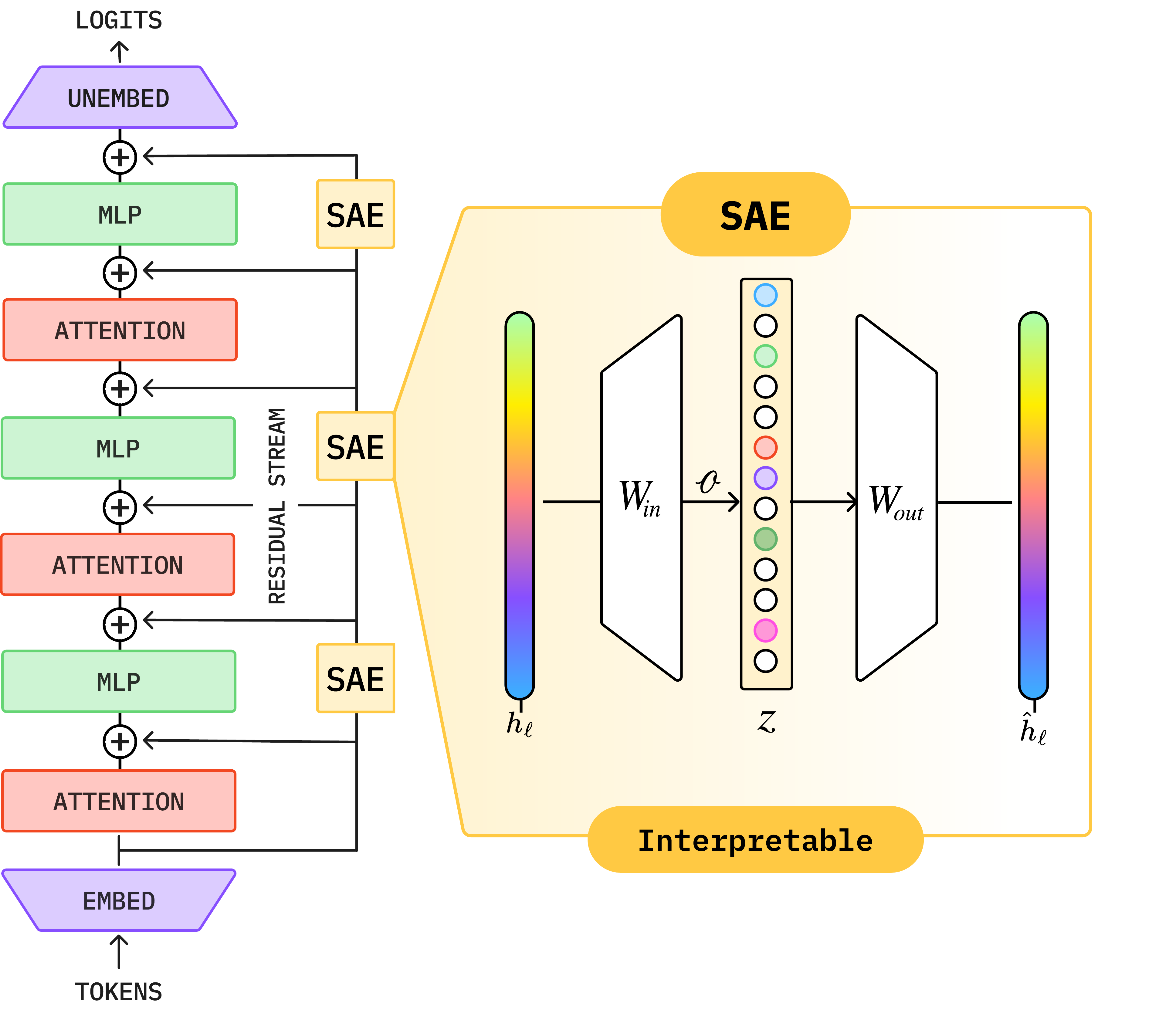}
    \caption{\textbf{Sparse Autoencoder (SAE).} SAEs are trained to reconstruct activations of a model component at a specific layer $\ell$, $h_\ell$, into an overcomplete, sparse latent space $z$, such that the reconstructed activation $\hat{h_\ell} \approx h_\ell$. By enforcing sparsity on $z$, the model decomposes dense, polysemantic representations into monosemantic, human-interpretable features.}
    \label{fig:sae}
\end{figure}

SAEs (Figure~\ref{fig:sae}) extract monosemantic features from polysemantic model activations. They are trained to reconstruct activations at a specific location in neural networks, such as the output of an MLP block or an entire transformer layer (i.e., the residual stream), while keeping only a few features active in any given forward pass.

\begin{definitionbox}[Sparse Autoencoder (Figure~\ref{fig:sae})]
A sparse autoencoder learns a component-specific encoder--decoder pair
\[
E : \mathbb{R}^d \rightarrow \mathbb{R}^{d_{\text{sparse}}},
\qquad
D : \mathbb{R}^{d_{\text{sparse}}} \rightarrow \mathbb{R}^d,
\]
such that activations of a particular model component at a given layer, \(h_\ell \in \mathbb{R}^d\) (e.g., the output of the layer-\(\ell\) MLP block), are reconstructed as
\[
\hat{h}_\ell = D(E(h_\ell)),
\]
while encouraging the feature activations \(z=E(h_\ell)\) to be sparse. Usually, $d_{\text{sparse}} = ed$, where $e$ is an integer expansion factor.
\end{definitionbox}

\paragraph{Canonical Formulation.}
Given activations of a model component \( h_\ell \sim \mathcal{H} \), where $\mathcal{H}$ is the distribution of that component's activations over a large corpus, the canonical SAE objective is
\[
\mathcal{L}_{\text{SAE}}
=
\mathbb{E}_{h_\ell \sim \mathcal{H}}
\left[
\| h_\ell - D (z) \|_2^2
+ \lambda \| z \|_1
\right],
\quad
z = E(h_\ell),
\]
where $\lambda > 0$ controls sparsity~\citeest{cunningham2023sparse}. Canonical SAEs use $L_1$ regularisation instead of $L_0$ regularisation. This is because, for each feature, $L_0$'s derivative is almost everywhere zero and thus does not provide useful signals for promoting feature sparsity during gradient descent. Additionally, the decoder columns are typically constrained to unit norm. This is to prevent the case where the SAE learns to reduce its $L_1$ loss by generating smaller values of $z$---not exactly $0$ so that many features can be active, each having a small coefficient---and enlarging the norm of each column of the decoder to compensate. In this way, the SAE output remains the same, while $L_1$ loss can be substantially reduced, without any real gain in sparsity.

\paragraph{Why SAEs Work.}
SAEs are motivated by the hypothesis that a model's activations can be approximated as linear combinations of many latent feature directions, which can be regarded as an extension of the linear representation hypothesis discussed in Section~\ref{subsec:linearity}. Sparsity means that only a small number of features are active for any given token or input, but not that the hidden state itself is sparse. Early sparse dictionary learning studies~\citeest{olshausen1996emergence, lewicki2000learning, aharon2006k} give us fair confidence that, if the hypothesis holds, dictionary learning can recover the structure as an overcomplete dictionary. These findings motivate using SAEs as a modern adaptation to extract monosemantic directions from the activations of neural networks~(\citeestalp{cunningham2023sparse}; \citeemalp{bricken2023monosemanticity}).

\paragraph{From Monosemanticity to Concept Identification.}
For each input activation, the raw SAE outputs are just a small set of vectors. Practical interpretability requires identifying the semantic concepts these vectors correspond to in natural language (so that they become human-understandable features). Empirically, SAE features selectively activate on structured token patterns that recur across input sequences, suggesting sensitivity to specific abstract semantic, functional, or lexical concepts~\citeem{marks2024enhancingneuralnetworkinterpretability}. Automated analysis pipelines~\citeest{gur-arieh-etal-2025-enhancing, paulo2025automatically} aggregate highly activating inputs to infer the concept each feature captures (see Section~\ref{sec:automated-interp} for a detailed discussion).

\paragraph{Training Dynamics and Evaluation.}
SAE training is sensitive to initialisation, sparsity strength, and decoder normalisation. Without proper constraints and hyperparameters, the features may collapse, duplicate, or absorb one another. SAEs are typically evaluated for reconstruction loss (fidelity of activation reconstruction), feature absorption (how often a general feature fails to fire because a more specific feature has absorbed its direction), and downstream test loss recovery (whether replacing activations with reconstructions preserves model behaviour). Together, these metrics assess both the quality of the decomposition and its mechanistic relevance to the original model's computation.

\subsection{SAE Variants and Architectural Improvements}
\label{subsec:sae-variants}

There are several variants of the standard SAE. One key reason to use alternatives is that standard SAE training penalises activation magnitude, which can lead to low activation even when a feature is clearly present (the shrinkage effect). Different optimisation frameworks also have different tunable parameters. The choice depends on the analysis goal, target layer type, model size, and budget constraints.

\begin{definitionbox}[Gated SAEs]
Gated SAEs introduce multiplicative gates to decouple feature selection from magnitude:
\[
z = \underbrace{\mathbbm{1} \left( (W_{\text{gate}} h_\ell + b_{\text{gate}}) > 0 \right)}_{\text{Gating Path}} \odot \underbrace{\sigma_{\text{ReLU}}(W_{\text{in}} h_\ell + b_{\text{in}})}_{\text{Magnitude Path}},
\]
where $h_\ell \in \mathbb{R}^d$ denotes activations at layer $\ell$, $W_{\text{in}} \in \mathbb{R}^{{d_{\text{sparse}}} \times d}$ and $b_{\text{in}} \in \mathbb{R}^{d_{\text{sparse}}}$ are encoder $E$ parameters, $W_{\text{gate}} \in \mathbb{R}^{{d_{\text{sparse}}} \times d}$ and $b_{\text{gate}} \in \mathbb{R}^{d_{\text{sparse}}}$ are the gate parameters, and $\odot$ denotes element-wise product. This mitigates shrinkage effects induced by the $L_1$ penalty~\citeest{rajamanoharan2024improvingdictionarylearninggated}.
\end{definitionbox}

\noindent
\begin{minipage}[c]{0.65\textwidth}
    \begin{definitionbox}[JumpReLU SAEs]
    JumpReLU SAEs~\citeem{rajamanoharan2024jumpingaheadimprovingreconstruction} learn per-feature activation thresholds. Given the encoder pre-activation $a = W_{\text{in}} h_\ell + b_{\text{in}}$, let $a_i$ denote its $i$-th entry. The activation value of the $i$-th feature after applying JumpReLU is
    \[
    z^i = \sigma_{\text{JumpReLU}}(a_i) = \begin{cases} 
    a_i & \text{if } a_i > \tau_i \\ 
    0 & \text{otherwise},
    \end{cases}
    \]
    where $\tau_i$ is a learned threshold parameter. Straight-through estimators~\citeem{bengio2013estimatingpropagatinggradientsstochastic} are used during training to enable gradient flow.
    \end{definitionbox}
\end{minipage}
\hfill
\begin{minipage}[c]{0.30\textwidth}
\vspace{0.5cm}
\hspace{-0.5cm}\includegraphics[width=1.\textwidth]{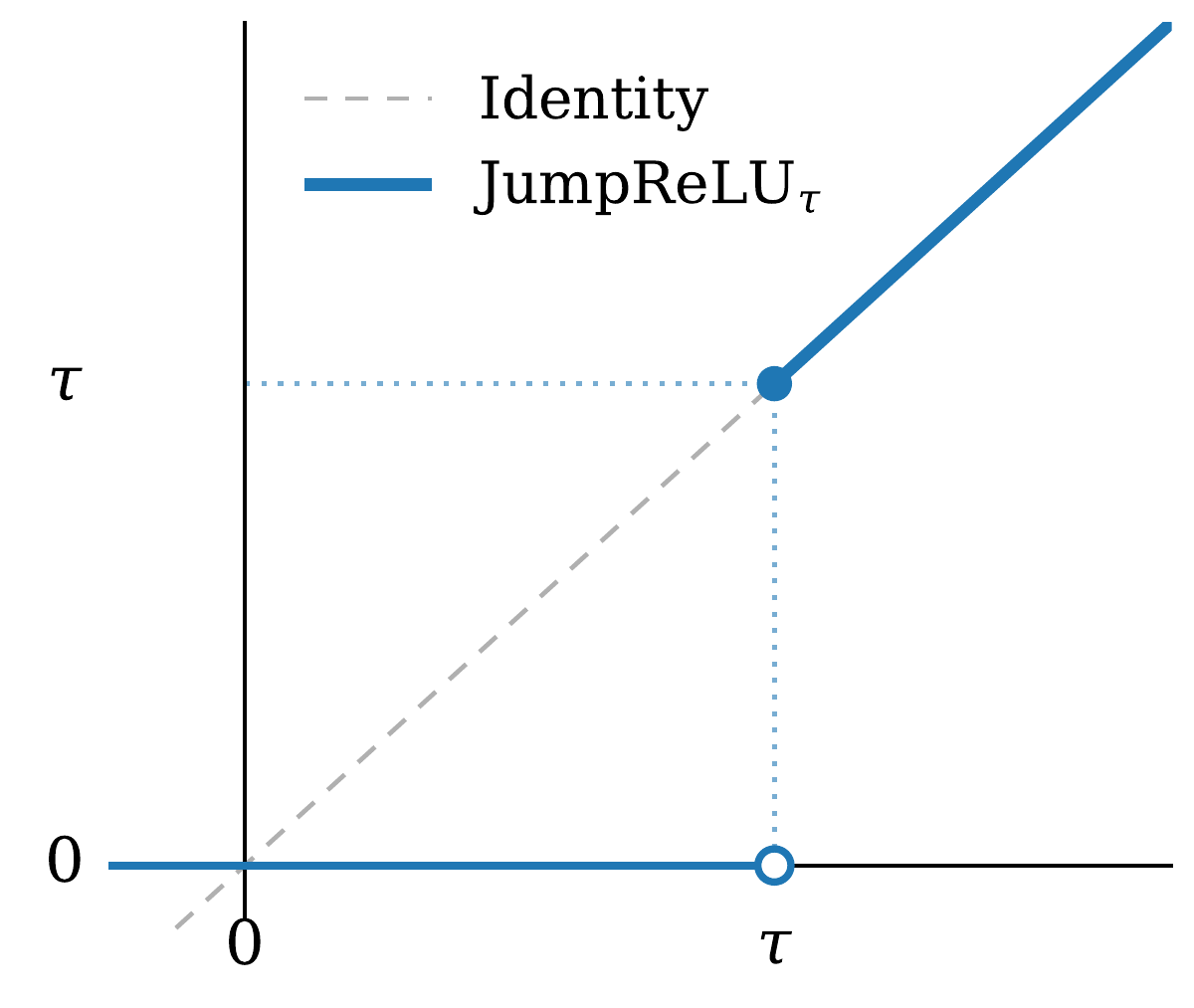}
\end{minipage}

\begin{definitionbox}[Top-$k$ SAEs]
Top-$k$ SAEs~\citeest{gao2024scalingevaluatingsparseautoencoders} impose an explicit hard sparsity constraint. Let $a = W_{\text{in}} h_\ell + b_{\text{in}} \in \mathbb{R}^{d_{\text{sparse}}}$ denote encoder pre-activations. Top-$k$ SAEs define the $i$-th feature's activation as
\[
z^i = \sigma_{\text{TopK}}(a_i) = \begin{cases}
a_i & \text{if } i \in \mathcal{I}_k(a) \\
0 & \text{otherwise},
\end{cases}
\]
where $\mathcal{I}_k(a)$ indexes the $k$ largest entries of $a$. This formulation explicitly enforces $\|z\|_0 \le k$---a more direct feature-sparsity constraint than $L_1$ regularisation. Notably, as it does not suppress activation magnitudes, it avoids the feature-shrinkage issue observed in $L_1$-based SAE training~\citeest{gao2024scalingevaluatingsparseautoencoders}.
\end{definitionbox}

\begin{definitionbox}[Batch Top-$k$ SAEs]

Batch Top-$k$ SAEs relax the fixed per-example sparsity constraint of standard Top-$k$ SAEs by enforcing the sparsity budget across an entire batch instead~\citeem{bussmann2024batchtopksparseautoencoders}. For a batch of encoder activations
\[
\big[\, a^{(1)}, \dots, a^{(B)} \,\big] \in \mathbb{R}^{d_{\text{sparse}} \times B},
\]
where $B$ is the batch size, Batch Top-$k$ retains the largest $Bk$ activation values in the batch. Specifically, let
\[
\mathcal{I}
\coloneqq
\arg\max_{\substack{
\mathcal{S}\subseteq
\{1,\ldots,B\}\times
\{1,\ldots,d_{\text{sparse}}\}\\
|\mathcal{S}|=Bk
}}
\sum_{(b,j)\in\mathcal{S}}
a_j^{(b)}.
\]
In each batch, the activation value of the $j$-th feature in the $b$-th sample is
\[
z^{j, (b)}
=
\sigma_{\text{Batch-TopK}}\!\left(a_j^{(b)}\right)
=
\begin{cases}
a_j^{(b)}, & (b,j)\in\mathcal{I},\\
0, & \text{otherwise}.
\end{cases}
\]

That is, each batch exactly retains $Bk$ features, while individual samples may have more or fewer active features than $k$. This loose constraint allows the model to allocate representational capacity adaptively across samples. During inference, Batch Top-$k$ SAEs instead use the JumpReLU function with an inferred global threshold to determine which features are active for each input, so that an input's activations do not depend on the rest of the batch.
\end{definitionbox}

\begin{insightbox}[Trade-offs Between Variants]
Different SAE variants may implement distinct sparsity constraints. No single variant is optimal across all scenarios.\\

\begin{center}
\small
\begin{tabular}{lccc}
\toprule
\textbf{Variant} & \textbf{Sparsity} & \textbf{Differentiable} & \textbf{Tuning Required} \\
\midrule
$L_1$-regularised & Soft & \cmark & $\lambda$ coefficient \\
Gated & Learned & \xmark$^\dagger$ & $\lambda$ coefficient \\
JumpReLU & Learned & \xmark$^*$ & $\lambda$ coefficient, STE bandwidth, threshold init. \\
Top-$k$ & Hard & \xmark & $k$ value \\
Batch Top-$k$ & Hard & \xmark & $k$ value \\
\bottomrule
\end{tabular}
\end{center}
\vspace{0.5em}
{\small $^*$Uses straight-through estimators (STEs) during training. $^\dagger$Trains the non-differentiable gate with auxiliary losses.}
\end{insightbox}
Readers are referred to \citetest{shu-etal-2025-survey} for a dedicated survey of SAEs, their variants, evaluations, and applications. Additionally, researchers have published SAE weights for several language models, including Gemma Scope~\citeem{lieberum2024gemmascopeopensparse} for Gemma 2~\citeem{gemmateam2024gemma2improvingopen} and Llama Scope~\citeem{he2024llamascopeextractingmillions} for Llama-3.1-8B~\citeem{llama3}. They can be accessed through open platforms such as Neuronpedia~\citeem{neuronpedia}.

\subsection{Empirical Comparisons of SAE Variants}
\label{subsec:sae-selection}

The SAE variants in Section~\ref{subsec:sae-variants} differ principally in how they enforce
sparsity, and those differences map onto distinct analysis settings. 

Hard-constraint formulations (e.g., Top-$k$) directly enforce sparsity and suit tight compute budgets. Learned-threshold (e.g., JumpReLU SAE) and gated formulations usually achieve better reconstruction results and suit generous tuning budgets. Cross-layer variants (discussed in Section~\ref{subsec:transcoders-crosscoders}) analyse computations across layers that a single dictionary cannot represent at all. This subsection discusses the setting in which each variant is used and the evidence it provides. We particularly discuss four factors that account for most of the variation:

\begin{enumerate}
    \item \textbf{Analysis goal:} exploratory discovery vs.\ targeted investigation vs.\ deployment/monitoring
    \item \textbf{Layer type:} residual stream vs.\ MLP output vs.\ attention output
    \item \textbf{Model scale:} small ($<$1B) vs.\ medium (1--10B) vs.\ large ($>$10B parameters)
    \item \textbf{Resource constraints:} compute budget, hyperparameter tuning capacity, tolerance for training instability
\end{enumerate}

\paragraph{Analysis Goal.}
\emph{Exploratory discovery}, where the question is what features exist rather than whether
a particular one does, is the setting SAEs were designed for and the one the comparisons
below address. For \emph{targeted investigation} of an already-identified behaviour, where the concepts of interest are determined, supervised methods such as DAS~\citeest{das} and linear probes are more direct instruments. For \emph{online deployment and monitoring}, SAEs with small dictionaries can be preferable for low inference cost.

\paragraph{Layer Type.}
Layer types differ in their activation properties, so the right variant depends on what each can represent, not just how well. On the \emph{residual stream}, single-layer SAEs
apply directly, and the choice reduces to the resource considerations below. On \emph{MLP
outputs}, transcoders are the appropriate tool: a single-layer SAE cannot span the
nonlinearity, so transcoders learn the input-to-output map
directly~\citeest{dunefsky2024transcodersinterpretablellmfeature}, and
\citetem{paulo2025transcodersbeatsparseautoencoders} reported favourable comparisons against
SAEs there. On \emph{attention outputs}, standard and Gated SAEs both have been used;
attention-specific variants remain underdeveloped, making this the least studied of the three
sites. For \emph{cross-layer analysis}, crosscoders~\citeem{lindsey2024sparsecrosscoders}
and cross-layer transcoders~\citeem{ameisen2025circuit} track feature evolution across
layers at substantially higher compute cost, and a crosscoder trained jointly on two models
is the reported approach for comparing representations between them. Table~\ref{tab:sae-layer-recommendations} summarises the variants used at each layer type, the empirical rationale for their use, and the main caveats reported in prior work.

\begin{table}[tb]
\centering
\caption{Heuristic recommendations for selecting SAE variants across different layer types. Recommendations are based on current empirical evidence and practical experience, and may evolve as SAE methodologies and evaluation standards continue to develop.}
\label{tab:sae-layer-recommendations}
\renewcommand{\arraystretch}{1.3}
\begin{tabular}{>{\raggedright\arraybackslash}p{2.8cm} >{\raggedright\arraybackslash}p{3.2cm} >{\raggedright\arraybackslash}p{4cm} >{\raggedright\arraybackslash}p{4cm}}
\toprule
\rowcolor{tableheader!80} \textcolor{white}{\textbf{Layer Type}} & \textcolor{white}{\textbf{Variants Used}} & \textcolor{white}{\textbf{Reported Basis}} & \textcolor{white}{\textbf{Reported Caveats}} \\
\midrule
\rowcolor{tablerow1} Residual stream (early layers) & Gated SAE, Top-$k$ & Activations are closer to input embeddings; moderate sparsity & Smaller expansion factors (2--4$\times$) reported \\
\rowcolor{tablerow2} Residual stream (middle layers) & Gated SAE, JumpReLU & Richest feature structure; highest interpretability potential & Standard expansion factor (4--8$\times$); most studied regime \\
\rowcolor{tablerow1} Residual stream (late layers) & Top-$k$, Gated SAE & Activations increasingly task-specific; may be less sparse & Larger expansion factors (8--16$\times$) reported \\
\rowcolor{tablerow2} MLP outputs & Transcoder (Section~\ref{subsec:transcoders-crosscoders}) & MLP applies a nonlinearity a single-layer SAE cannot span & Not a sparse decomposition on a static object; bridges input to output $\ell$ \\
\rowcolor{tablerow1} Attention outputs & Gated SAE, standard SAE & Attention outputs often lower-rank & Least studied of the three sites; may benefit from smaller dictionaries \\
\rowcolor{tablerow2} Cross-layer & Crosscoder, CLT (Section~\ref{subsec:transcoders-crosscoders}) & Tracks feature evolution; resolves cross-layer superposition & High compute cost; requires multi-layer training \\
\bottomrule
\end{tabular}
\end{table}

\paragraph{Resource Constraints.}
Under limited compute, or where fast iteration matters more than reconstruction quality,
Top-$k$ SAEs~\citeest{gao2024scalingevaluatingsparseautoencoders} are the practical choice:
fixing $\|z\|_0 \le k$ removes the sparsity coefficient $\lambda$ from the search space
entirely rather than requiring it to be tuned, with reported values commonly in the range
$k = 32$~\citeest{gao2024scalingevaluatingsparseautoencoders} to $150$~\citeest{karvonen2025saebench}. With moderate tuning capacity, Gated
SAEs~\citeest{rajamanoharan2024improvingdictionarylearninggated} achieved a better
reconstruction--sparsity trade-off while retaining $\lambda$. They also avoid the shrinkage effect by separating feature detection from magnitude estimation. However, because the standard $L_1$ SAE is now used mainly as a baseline, comparisons against it partly reflect shrinkage rather than dictionary quality alone. When extensive tuning is plausible, JumpReLU
SAEs~\citeem{rajamanoharan2024jumpingaheadimprovingreconstruction} achieved the strongest reconstruction performance at matched sparsity by learning per-feature thresholds. However,
gains on these sparsity--fidelity proxy metrics do not reliably translate
into better downstream performance~\citeest{karvonen2025saebench}. In practice, selecting the best model and hyperparameters to achieve optimal performance can take a lot of time.

\paragraph{Model Scale.}
The differences between variants depend strongly on the metric in models with roughly 1 or 2B parameters, such as Gemma-2-2B~\citeest{karvonen2025saebench}, which is the regime mainly used to validate methodology before scaling. For models between 1B and 10B parameters, JumpReLU SAEs and Top-$k$ SAEs are the most commonly reported variants; Gemma Scope~\citeem{lieberum2024gemmascopeopensparse} uses the former and Llama Scope~\citeem{he2024llamascopeextractingmillions} the latter. For models larger than 10B parameters, Gemma Scope~\citeem{lieberum2024gemmascopeopensparse} (for Gemma 2 models) and Gemma Scope 2~\citeem{mcdougall2025gemmascope2} (for Gemma 3 models) are the main open resources, both of which used JumpReLU SAEs. Additionally, \citetest{gao2024scalingevaluatingsparseautoencoders} explored scaling Top-$k$ SAEs to GPT-4~\citeem{openai2024gpt4technicalreport}, although GPT-4 itself is not publicly available.

Table~\ref{tab:sae_recommendations} summarises each variant’s main findings reported in previous work, along with their experimental settings. Note that these studies may use different baselines, model families, sparsity constraints, and evaluation criteria. The community has been establishing benchmarks to evaluate variants of SAEs, such as SAEBench~\citeest{karvonen2025saebench}, but the results vary by metric. As a result, we should treat the findings as references rather than established rules. The \emph{Setting} column in Table~\ref{tab:sae_recommendations} records the conditions under which each result is established. Table~\ref{tab:sae_limitations} further highlights several factors that limit the generality of the findings.

\begin{table}[tb]
\centering
\caption{Reported evidence for each SAE variant, with the setting in which it was obtained.}
\label{tab:sae_recommendations}
\renewcommand{\arraystretch}{1.3}
\begin{tabular}{>{\raggedright\arraybackslash}p{2.4cm} >{\raggedright\arraybackslash}p{6.2cm} >{\raggedright\arraybackslash}p{4.6cm}}
\toprule
\rowcolor{tableheader!80} \textcolor{white}{\textbf{Variant}} & \textcolor{white}{\textbf{Reported Finding}} & \textcolor{white}{\textbf{Setting}} \\
\midrule
\rowcolor{tablerow1} Standard & Sparse dictionary learning by $L_1$ regularisation recovers features more monosemantic than individual neurons; subject to shrinkage bias & Establishing work, small models~(\citeestalp{cunningham2023sparse}; \citeemalp{bricken2023monosemanticity}) \\
\rowcolor{tablerow2} Gated & Separating detection from magnitude estimation addresses $L_1$ shrinkage at minimal added complexity & Reconstruction--sparsity improvement at matched sparsity, Gemma-family~\citeest{rajamanoharan2024improvingdictionarylearninggated} \\
\rowcolor{tablerow1} JumpReLU & Learned per-feature thresholds trained with straight-through estimators give the strongest reported reconstruction & Matched-sparsity comparison, Gemma-family~\citeem{rajamanoharan2024jumpingaheadimprovingreconstruction}; needs most tuning~\citeem{bussmann2024batchtopksparseautoencoders} \\
\rowcolor{tablerow2} Top-$k$ & A hard $\|z\|_0 \le k$ constraint removes the sparsity coefficient from the search space & Scaling behaviour characterised on large models; $k = 32$--$150$ commonly reported~\citeest{gao2024scalingevaluatingsparseautoencoders, karvonen2025saebench} \\
\rowcolor{tablerow1} Batch Top-$k$ & A batch-level sparsity budget lets the number of active features vary per input, improving reconstruction at matched average sparsity & Introduced as a relaxation of per-example Top-$k$, achieving better reconstruction~\citeem{bussmann2024batchtopksparseautoencoders} \\
\rowcolor{tablerow2} Transcoder & Learning the MLP input--output map avoids spanning the nonlinearity with a single dictionary & Reported to perform at least on par with SAEs~(\citeestalp{dunefsky2024transcodersinterpretablellmfeature}; \citeemalp{paulo2025transcodersbeatsparseautoencoders}) \\
\rowcolor{tablerow1} Crosscoder & Joint reconstruction across layers tracks feature evolution and resolves cross-layer superposition & Cross-layer analysis and model diffing; trained jointly across models for cross-model comparison~\citeem{lindsey2024sparsecrosscoders} \\
\rowcolor{tablerow2} CLT & Partial reconstruction of multiple downstream layers from one dictionary & Deployed within CLT-based circuit discovery at scale~\citeem{ameisen2025circuit} \\
\bottomrule
\end{tabular}
\end{table}

\begin{table}[tb]
\centering
\caption{Conditions for the reported evidence on SAE variant behaviour.}
\label{tab:sae_limitations}
\renewcommand{\arraystretch}{1.3}
\begin{tabular}{>{\raggedright\arraybackslash}p{0.32\linewidth} >{\raggedright\arraybackslash}p{0.58\linewidth}}
\toprule
\rowcolor{tableheader!80} \textcolor{white}{\textbf{Condition}} & \textcolor{white}{\textbf{Description}} \\
\midrule
\rowcolor{tablerow1} Architecture dependence & Most evidence derives from Llama- and Gemma-family models. Generality of the findings across model families remains unestablished. \\
\rowcolor{tablerow2} Task dependence & The properties that matter may differ across analysis goals, such as safety auditing, behaviour understanding, and knowledge localisation, but existing evaluations often do not distinguish between them.\\
\rowcolor{tablerow1} Rapid iteration & New architectures and training procedures are proposed faster than existing ones are sufficiently studied and replicated. \\
\rowcolor{tablerow2} Immature theoretical analysis & Current findings are mostly empirical and may be superseded. Understanding of why particular variants succeed or fail is incomplete. \\

\rowcolor{tablerow1} Feature reproducibility & SAEs trained on the same data with different random seeds can recover substantially different features. When between-variant performance differences are smaller than within-variant seed-to-seed variability, we may not confidently attribute them to architectural factors. However, few reported comparisons have rigorously accounted for this concern. \\

\bottomrule
\end{tabular}
\end{table}

\subsection{Challenges and Failure Modes of SAEs}
\label{subsec:challenge}

SAEs have become a standard tool for feature extraction, but they exhibit systematic failure modes that undermine their interpretability claims. It is therefore essential to understand these issues when using SAEs as mechanistic tools. This section provides a systematic discussion of them.

\subsubsection{Dead Features}
A \emph{dead feature} is an element in the learned dictionary that never activates or activates extremely rarely across training. Dead features waste the dictionary’s capacity, and they are common. In fact, they can constitute up to 10--50\% of the dictionary in a poorly tuned SAE~\citeem{bricken2023monosemanticity}. 

Several factors can cause dead features to emerge. First, inappropriate SAE weight initialisation can place features in low-density regions of the activation space with weak gradient signals. Second, an overly strong sparsity penalty $\lambda$ may suppress feature activations early in training, preventing features from contributing usefully to reconstruction. Finally, once a subset of features captures common, dominant patterns, the remaining features struggle to find useful roles for improving reconstruction and thus do not activate.

Dead features can be detected by tracking each feature’s rate of being activated during training. For example, features with an activation rate below 0.01\% may be considered dead (the threshold should adapt across scenarios). Several mitigation strategies have been proposed, such as \emph{neuron resampling}, which periodically reinitialises dead features using examples with high reconstruction error~\citeem{bricken2023monosemanticity}. The sparsity penalty can also be reduced early in training and increased gradually (i.e., a warm-up schedule). More recent Batch Top-$k$ SAEs allocate sparsity adaptively across samples, which allows more active features on complex inputs and fewer on easy ones~\citeem{bussmann2024batchtopksparseautoencoders}.

\subsubsection{Feature Duplication}
Feature duplication occurs when multiple features in the dictionary encode the same or similar concepts~\citeem{templeton2024scaling}, leading to potentially redundant features. Duplication can arise for several reasons: insufficient sparsity pressure, imbalanced training data, or an optimisation process that favours local minima with duplicated features. Duplication can be detected by computing pairwise cosine similarity between feature directions (i.e., decoder columns), where high similarity indicates potential redundancy. As duplicated features tend to activate together~\citeem{korznikov2025ortsaeorthogonalsparseautoencoders}, co-activation statistics may provide complementary evidence. 

\subsubsection{Feature Absorption and Splitting}

Another class of failure modes concerns feature granularity. \emph{Feature absorption} arises when semantically hierarchical features interact under the sparsity objective: a seemingly general (\emph{parent}) feature fails to activate on some inputs where it should because a more specific (\emph{child}) feature activates instead and has absorbed part of the parent's direction into its own decoder direction~\citeest{chanin2025absorption}. For example, a ``starts with S'' feature may fail to activate on the token ``short'', while a token-specific ``short'' feature with a partially overlapping direction activates instead. Absorption-induced false negatives make such a feature an unreliable detector of its concept. It can be detected by comparing the feature with a linear probe for the same concept: inputs on which the probe responds strongly, but the feature does not, while another active feature's direction aligns with the probe, indicate absorption~\citeest{chanin2025absorption}.

By contrast, \emph{feature splitting} occurs when a feature learned by a smaller SAE splits into multiple distinct features as the SAE scales up (e.g., by increasing the expansion factor \(e\)). This often produces more fine-grained features, which may be either beneficial or less useful. In a favourable case, an initially polysemantic or opaque feature, such as one activating on ``bank'' in both financial and geographical contexts, may split into more coherent features of ``investment bank'' and ``river bank''. In a less useful case, the resulting features may become low-level, lexical detectors that activate on the token ``bank'' in any context.

\subsubsection{Feature Instability Across Seeds}

SAE features are also sensitive to random initialisation, optimiser stochasticity, and data ordering. For a fixed SAE architecture and training dataset, different random seeds can produce substantially different dictionaries~\citeest{paulo2025sparse}. In practice, frequent and semantically clear features tend to be stable, while rare or context-specific features vary substantially. Often, fewer than 50\% of features remain the same across seeds, although broader semantic clusters may remain more consistent. This instability limits the reliability of mechanistic claims based on individual SAE features. Although aggregating results across seeds can improve robustness, it increases computational cost. As a result, when evaluating SAEs, we should report feature stability across seeds alongside reconstruction loss.

\subsubsection{Identifiability and Theoretical Concerns}
\label{subsubsec: sae limitation}
Another concern for SAE-based interpretability claims is whether the learned features are identifiable, that is, whether the dictionary learning reliably recovers the ``true'' underlying features---if such features exist---or merely finds one of many mediocre decompositions~\citeem{klindt2025superpositionsparsecodesinterpretable}.

\begin{definitionbox}[Identifiability]
Suppose activations are generated by some sparse latent process

\[
h = D^{*} z^{*} + \epsilon ,
\]

where \(D^{*}\) is the ground-truth feature dictionary and
\(z^{*}\) are the feature coefficients. An SAE is identifiable if its training recovers \(D^{*}\) and \(z^{*}\)
up to unavoidable symmetries such as feature permutation and rescaling.
\end{definitionbox}

In classical dictionary learning, identifiability holds under low-coherence conditions, i.e., dictionary elements are sufficiently orthogonal to one another. However, when coherence is high, ambiguity arises: multiple distinct sparse combinations of features from the same dictionary can generate the same output~\citeest{schnass2008dictionary}. In large language model activations, high coherence is likely unavoidable because of the model's complicated internal structure and correlations within its feature space~\citeest{li2025geometry}. So far, identifiability remains a stubborn issue in SAEs, even though it offers a theoretical explanation for several empirical observations.

Recent progress in fine-tuning language models around a fixed sparse autoencoder offers a potential path forward. In this setting, model activations are explicitly constrained to be sparse combinations of decoder features while largely preserving task performance~\citeest{chen2025low}. In particular, recent work aimed at limiting superposition fine-tunes the language model jointly with the SAE while enforcing orthogonality constraints on the decoder matrix~\citeem{miller2026identifyingintervenableinterpretablefeatures}. By explicitly imposing low-coherence conditions, this approach improves identifiability: activations have a unique feature representation, and these features can be reliably recovered.

\paragraph{Relation to Causal Relevance.}
The concern of identifiability connects to causal relevance: even if an SAE feature is monosemantic and helps reconstruct activations well, it may not correspond to a causally meaningful unit of analysis. Below, we summarise some criteria for validating an SAE feature.

\textbf{Diagnostic criteria:}
\begin{itemize}
    \item \textit{Necessity}: Does ablating the feature remove the corresponding concept in model responses? Low necessity suggests that the concept can be expressed through redundant features.
    \item \textit{Sufficiency}: Does activating only this feature produce the behaviour corresponding to its concept? Low sufficiency suggests the feature alone is not enough to represent the concept.
    \item \textit{Specificity}: Does the feature's causal effect match its corresponding concept? Low specificity means the feature contains other concepts.
\end{itemize}

These criteria are useful in practice when claiming that SAE features represent meaningful units of analysis.

\subsubsection{Practical Recommendations}
The challenges discussed above suggest several methodological considerations for SAE-based interpretability studies:

\begin{enumerate}
    \item \textbf{Report failure-mode metrics:} Alongside reconstruction loss, researchers should also report dead-feature percentage, duplication rate, and stability across seeds.
    
    \item \textbf{Validate causal relevance:} Researchers should validate features’ necessity, sufficiency, and specificity through interventions (e.g., feature ablation or steering) and evaluation on downstream effects. 
    
    \item \textbf{Compare methods:} For safety-critical applications, researchers should compare SAE features against various baselines to assess whether SAEs recover the relevant structure.
    
    \item \textbf{Try multiple SAE variants:} Different variants (e.g., Gated, Top-$k$, JumpReLU) impose different inductive biases. Researchers can explore findings consistent across variants, as these provide stronger evidence.
    
    \item \textbf{Acknowledge identifiability limits:} Researchers should frame SAE features as one valid approximate decomposition rather than the true features. Multiple decompositions may be equally valid for reconstruction but yield different feature sets.
    
    \item \textbf{Consider alternatives for targeted analysis:} When specific concepts are the focus, supervised methods (e.g., DAS, probes, discussed in Section~\ref{subsubsec: supervised_extraction}) may be more reliable because SAEs discover features in an unsupervised and exploratory manner.
\end{enumerate}

\subsection{Multi-Layer Feature Extraction and Analysis}
\label{subsec:transcoders-crosscoders}

Standard SAEs analyse individual layers in isolation. However, features can evolve as information propagates through the network. Transcoders (Figure~\ref{fig:transcoders}), cross-layer transcoders (Figure~\ref{fig:clt}), and crosscoders (Figure~\ref{fig:crosscoders}) enable multi-layer feature extraction and analysis, revealing how representations and features associated with particular model components---most commonly MLP blocks---transform across depth.

\subsubsection{Transcoders: Bridging the Nonlinearities}
\label{subsubsec:transcoder}

\begin{figure}[tbp]
  \centering
  \begin{subfigure}{0.48\linewidth}
    \centering
    \includegraphics[width=\linewidth]{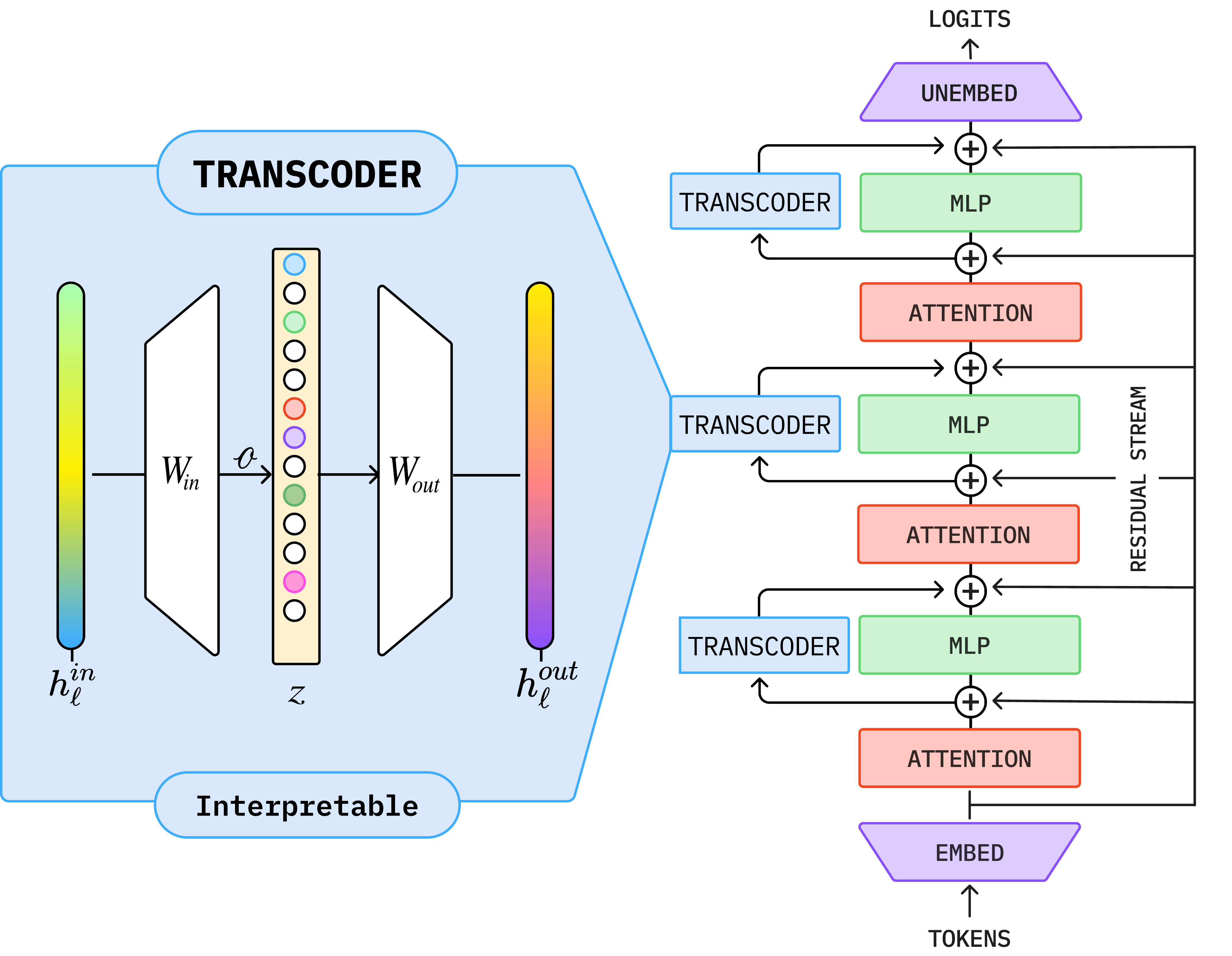}
    \caption{Transcoder.}
    \label{fig:transcoders}
  \end{subfigure}\hfill
  \begin{subfigure}{0.48\linewidth}
    \centering
    \includegraphics[width=\linewidth]{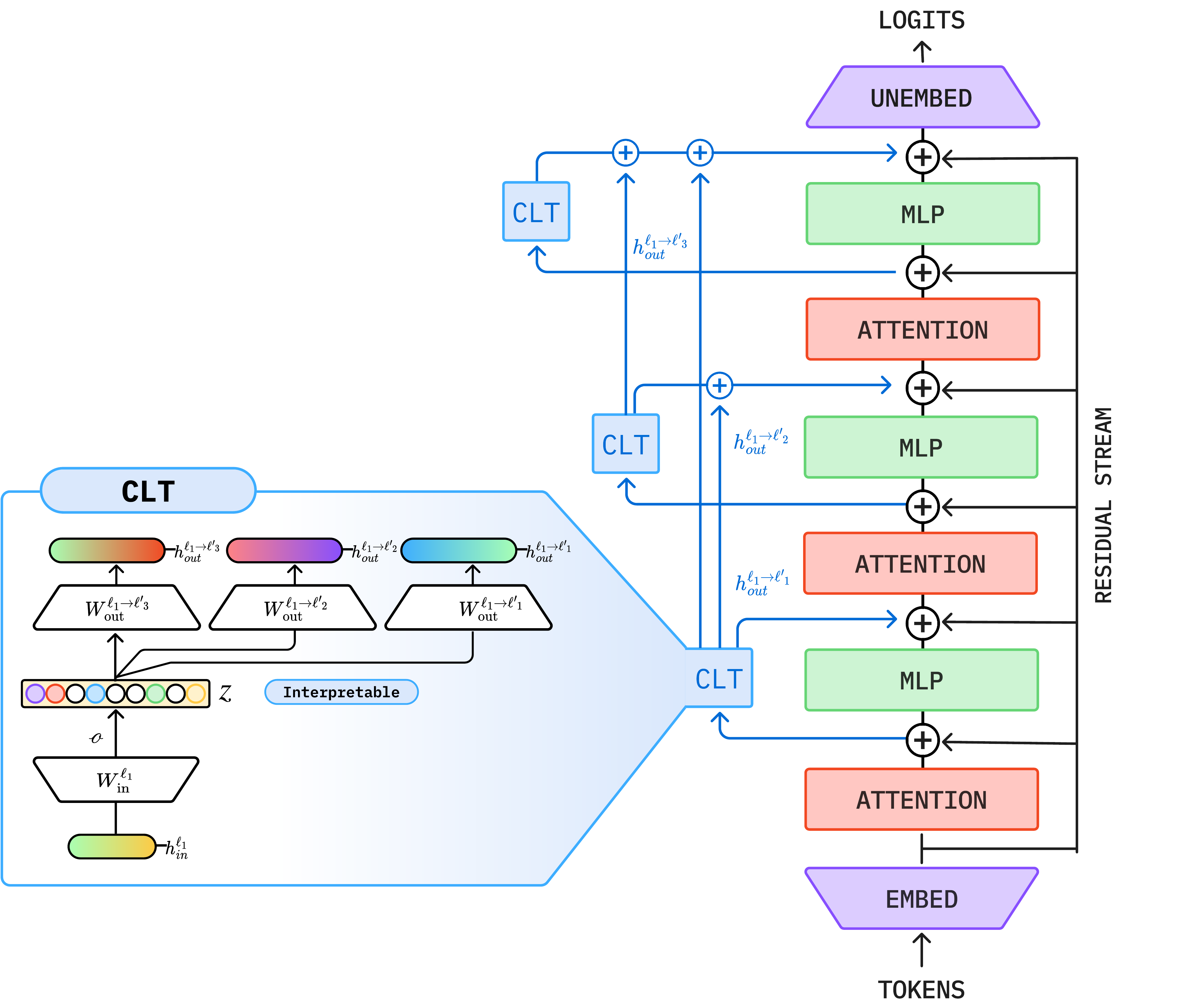}
    \caption{Cross-layer transcoder.}
    \label{fig:clt}
  \end{subfigure}
  \caption{\textbf{Transcoders and CLTs.} Transcoders reconstruct the output of the MLP from its input in the same transformer layer. CLTs extend this by
  partially reconstructing the outputs of multiple subsequent layers from a
  single point in the residual stream.}
\end{figure}

\begin{definitionbox}[Transcoder (Figure~\ref{fig:transcoders})]
A \emph{transcoder} learns a sparse mapping between the input to and output of a model component, such as an MLP block, in layer $\ell$:
\[
h_{\ell}^{\text{out}} \approx W_{\text{out}} z + b_{\text{out}}, \quad z = \sigma(W_{\text{in}} h_\ell^{\text{in}} + b_{\text{in}})
\]
where $z \in \mathbb{R}^{d_{\text{sparse}}}$ are sparse feature activations bridging the nonlinearity between input and output of the MLP, and $W_{\text{out}} \in \mathbb{R}^{d \times {d_{\text{sparse}}}}$ and $b_{\text{out}} \in \mathbb{R}^{d}$ are decoder $D$ parameters. The transcoder reconstructs post-nonlinearity activations using features extracted from pre-nonlinearity activations~\citeest{dunefsky2024transcodersinterpretablellmfeature}.
\end{definitionbox}

\paragraph{Architecture and Motivation.}
Standard SAEs do not explain the nonlinear transformations (i.e., weight matrices of MLP blocks and attention heads) themselves that produce the activations. A transcoder achieves this by applying dictionary learning to the mapping, learning features that explain the transformation. Its training minimises
\[
\mathcal{L} = \|h_{\ell}^{\text{out}} - (W_{\text{out}} z + b_{\text{out}})\|_2^2 + \lambda \|z\|_1,
\]
where $z = \sigma(W_{\text{in}} h^{\text{in}}_\ell + b_{\text{in}})$ as defined above. 
\paragraph{When Are Transcoders Necessary?}
SAEs extract features from static activations and thus often suffice for residual-stream analysis. Transcoders track how features emerge from input to output and are essential for analysing MLP layers~\citeest{dunefsky2024transcodersinterpretablellmfeature}.

\subsubsection{Cross-Layer Transcoders: Enabling Better Circuit Analysis}
\label{subsubsec:cross-layer-transcoder}

\begin{definitionbox}[Cross-Layer Transcoder (Figure~\ref{fig:clt})]
\emph{Cross-layer transcoders} (CLTs) are sparse feature decompositions spanning all layers of a neural network. At each layer $\ell$, sparse feature activations are read from the residual stream via
\[
z_{\ell}
=
\sigma_{\text{JumpReLU}}
\!\left(
W_{\text{in}}^\ell h_{\ell}^{\text{in}}
\right),
\]
where $z_\ell^j$ denotes the activation of feature $j$ at layer $\ell$.\\

Each feature can write to the MLP outputs of its own layer and all subsequent layers through layer-specific decoder vectors. The contribution from features read at layer $\ell$ to the MLP output at a subsequent layer $\ell'$ is
\[
\hat{h}_{\ell \to \ell'}^{\text{out}}
=
\sum_j
W_{\text{out},j}^{\ell \to \ell'} z_{\ell}^{j},
\]
where $W_{\text{out},j}^{\ell \to \ell'}$ is the decoder vector from feature $j$ at layer $\ell$ to the MLP output at layer $\ell'$. The reconstructed MLP output at layer $\ell'$ is then
\[
\hat{h}_{\ell'}^{\text{out}}
=
\sum_{\ell=1}^{\ell'}
\hat{h}_{\ell \to \ell'}^{\text{out}}
=
\sum_{\ell=1}^{\ell'}
\sum_j
W_{\text{out},j}^{\ell \to \ell'} z_{\ell}^{j},
\]
which approximates the original MLP output $h_{\ell'}^{\text{out}}$.
\end{definitionbox}

CLTs (Figure~\ref{fig:clt}) are trained jointly across all $L$ layers to minimise
\[
\mathcal{L}
=
\sum_{\ell'=1}^{L}
\left\|
\hat{h}_{\ell'}^{\text{out}}
-
h_{\ell'}^{\text{out}}
\right\|_2^2
+
\lambda
\sum_{\ell=1}^{L}
\sum_j
\tanh\!\left(
c\,
z_{\ell}^{j}
\sqrt{
\sum_{\ell'=\ell}^{L}
\left\|
W_{\text{out},j}^{\ell \to \ell'}
\right\|_2^2
}
\right),
\]
where $c$ is the sharpness hyperparameter~\citeem{ameisen2025circuit}.

\paragraph{Replacement Models and Attribution Graphs.}
A key insight is that CLTs enable the construction of a \emph{replacement model}---a version of the original model in which all MLP layers are replaced with their CLT reconstructions. Running this replacement model involves:
\begin{itemize}
    \item Computing CLT feature activations $z_{\ell}$ at each layer $\ell$
    \item Replacing MLP outputs $h^{\text{out}}_{\ell}$ with reconstructions $\hat{h}^{\text{out}}_{\ell}$ from features in layer $1$ through $\ell$
    \item Keeping attention layers unchanged
\end{itemize}

When analysing a specific prompt, the replacement model freezes the original model's attention patterns and normalisation statistics from its forward pass, then adds layer-specific error corrections. Under these conditions, the local replacement model can exactly match the original model's activations on that prompt while decomposing it into interpretable CLT features. However, when the model's activations are completely replaced by CLT reconstructions, the intervened model's top-1 predictions match the original model's on only about 50\% of prompts from the Pile dataset~\citeem{gao2020pile}. Notably, this experiment uses a relatively small, 18-layer transformer~\citeem{ameisen2025circuit}. Therefore, this figure should be interpreted with important caveats (see below).

\begin{warningbox}[Caveats on CLT Fidelity Claims]

CLTs have been claimed to possess high output fidelity given that they achieve high top-1 token agreement. However, this claim requires careful examination:\\

\paragraph{Conditions for reported fidelity:}
\begin{itemize}
    \item Results are reported primarily on toy or small models, whose generality to other model architectures (e.g., Llama, Qwen) and model sizes (e.g., 70B models) is underexplored.
    \item Fidelity was measured on short sequences in the CLT training data. Evaluation under distribution shift (e.g., out-of-domain tasks, adversarial inputs) is not adequate.
    \item The paper reports top-1 token agreement as the main result, while softer metrics (e.g., top-5 agreement, KL divergence) may show different patterns and insights.
\end{itemize}

\paragraph{Conditions for exact local matching:}
\begin{itemize}
    \item Attention patterns must be frozen from the original forward pass since the CLT does not compute or reconstruct attention computations.
    \item Layer normalisation statistics (means, variances) must be cached from the original model. The replacement model does not recompute these.
    \item Error correction terms are added per layer to compensate for reconstruction residuals. Without them, exact matching fails.
\end{itemize}

\paragraph{Factors affecting fidelity:}
\begin{itemize}
    \item \textit{Model scale}: Preliminary evidence suggests that reconstruction quality and feature interpretability improve with CLT model size. This may be because larger CLTs produce more structured, less entangled representations~\citeem{ameisen2025circuit}. However, systematic scaling studies are limited.
    \item \textit{Layer depth}: Early layers’ and late layers’ activations are usually input-specific and output-constrained, respectively. Middle layers have richer information derived from the input but have not yet converged to the output. As a result, middle layers often show higher reconstruction fidelity.
    \item \textit{Domain}: Fidelity on factual recall and syntactic tasks is higher than on complex reasoning or arithmetic tasks, where MLP computations may be more complex and cannot be approximated by sparse linear features.
\end{itemize}

These caveats do not invalidate CLTs as a tool for mechanistic analysis, but they do mean that fidelity should be verified on the target model and specific task before drawing strong conclusions from CLT-based circuit analysis.
\end{warningbox}

\paragraph{Why CLTs Enable Better Circuit Analysis.}
Unlike per-layer transcoders, CLTs reveal direct feature-to-feature interactions across layers: a feature in layer $\ell$ directly influences features in later layers through its contribution to their inputs. This simplifies circuit structure---features need not re-encode information at each layer. Additionally, replacing MLPs with CLT reconstructions makes the remaining computation \textit{approximately} linear (with frozen attention patterns and normalisation), enabling principled attribution: each feature's activation can be expressed as the sum of its input edges up to its activation threshold~\citeem{ameisen2025circuit}.

However, linearisation depends on several assumptions:
\begin{itemize}
    \item \textbf{Attention freezing:} By caching attention patterns, the replacement model treats attention as a fixed routing operation rather than a dynamic, input-dependent computation. This means CLT-based circuits describe model computation conditioned on \emph{given} attention patterns without analysing them.
    \item \textbf{Normalisation caching:} CLT analysis caches the input-dependent normalisation factors (e.g., LayerNorm denominators) from the original forward pass and reuses them during attribution instead of recomputing them after interventions. This removes a source of nonlinearity, but if an intervention would substantially change these normalisation factors in the original model, the cached computation no longer faithfully represents the intervened state.

    \item \textbf{Sparse linear approximation:} Dictionary learning methods assume MLP outputs or transformations are well approximated by sparse linear combinations of features. This assumption may fail when computations involve complex feature interactions or highly nonlinear transformations.
\end{itemize}

CLTs provide a powerful lens for circuit discovery when these assumptions hold. When they fail, CLT-based explanations may become unfaithful.

\paragraph{Computational Cost and Practical Considerations.}

\begin{figure}[th]
\centering
\includegraphics[width=0.7\linewidth]{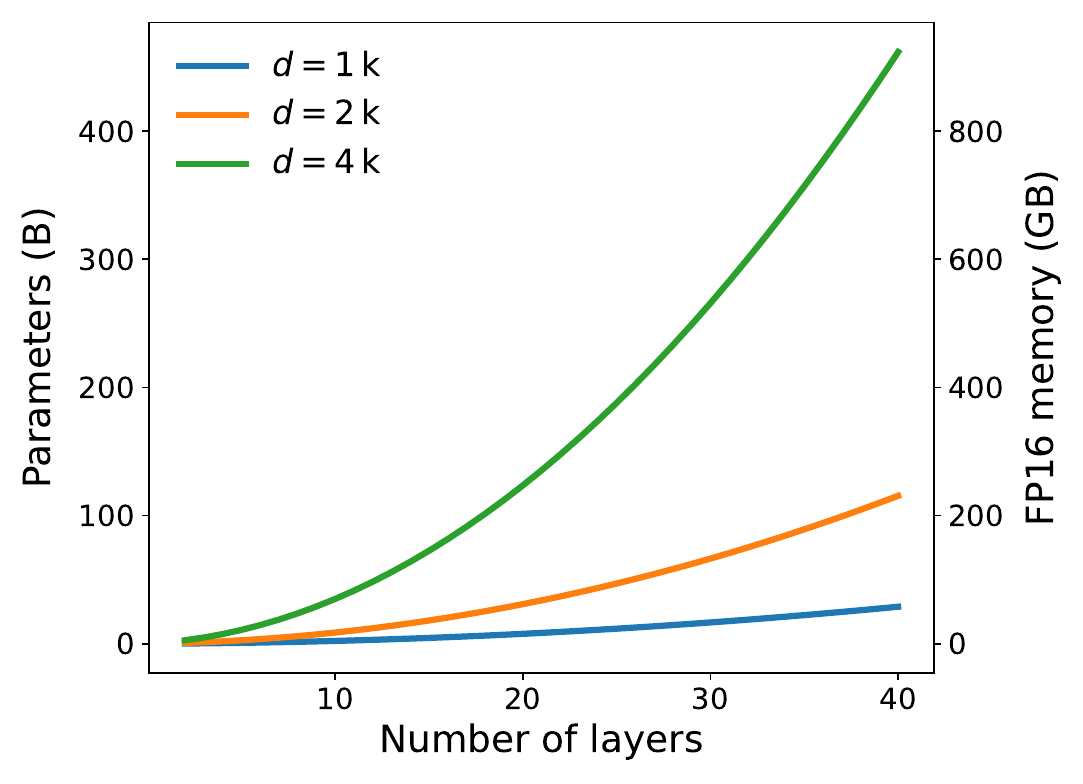}
    \caption{
    Parameter count (left) and FP16 memory requirements (right) of a full
    cross-layer transcoder as a function of model depth, for different residual
    stream dimensions and expansion factor $e=32$.
    }
\label{fig:clt-size-vs-depth}
\end{figure}

Let $d_{\text{sparse}} = ed$, where $e$ is the expansion factor. For a transformer with $L$ layers, a complete suite of CLTs contains $L$ encoders mapping
$
\mathbb{R}^d \rightarrow \mathbb{R}^{ed}
$
and $L(L+1)/2$ decoders mapping
$
\mathbb{R}^{ed} \rightarrow \mathbb{R}^d.
$
The resulting parameter count, with biases ignored, is therefore
\[
N_{\text{CLT}}(L,d,e)
=
\frac{e d^2}{2}\left(L^2 + 3L\right),
\]
which highlights the quadratic scaling with both model depth and hidden dimension, as illustrated in Figure~\ref{fig:clt-size-vs-depth}.

Training a suite of CLTs thus incurs a high upfront cost. For reference, training the CLT suite for Gemma-2-9B
required approximately 3{,}844 H100 GPU-hours at 50\% FLOPs efficiency~\citeem{ameisen2025circuit}. This cost is amortised over many circuit analyses on the trained CLT. The resulting interpretability gains (cross-layer features, lower interference, better attribution) justify the investment for systematic mechanistic analysis. 

For these reasons, current applications of CLTs remain limited: a ``greater-than'' mechanism
study~\citeem{merullo2025replicating}, an open-source
extension~\citeem{lindsey2025landscape}, and analysis of multilingual representations using CLTs~\citeem{harrasse2025tracing}. This scarcity of downstream use cases likely reflects the considerable challenges of training CLTs at scale and the complexity of using CLT-based circuit discovery in the wild.

\paragraph{Open Questions for CLT Research.}
Several important questions remain open:
\begin{itemize}
    \item How does CLT fidelity scale in large models? How can we better measure fidelity?
    \item Can CLTs be extended to model more nonlinearities such as attention patterns and normalisation, while maintaining their explanatory power?
    \item Can we develop evaluation protocols that reliably detect the potential unfaithfulness of CLT-based explanations?
    \item How sensitive are CLT circuits to experimental setups, such as the choice of training data?
\end{itemize}

\subsubsection{Crosscoders: Analysing Cross-Layer Features}

\begin{figure}[tb]
  \centering
  \centering
    \includegraphics[width=0.7\linewidth]{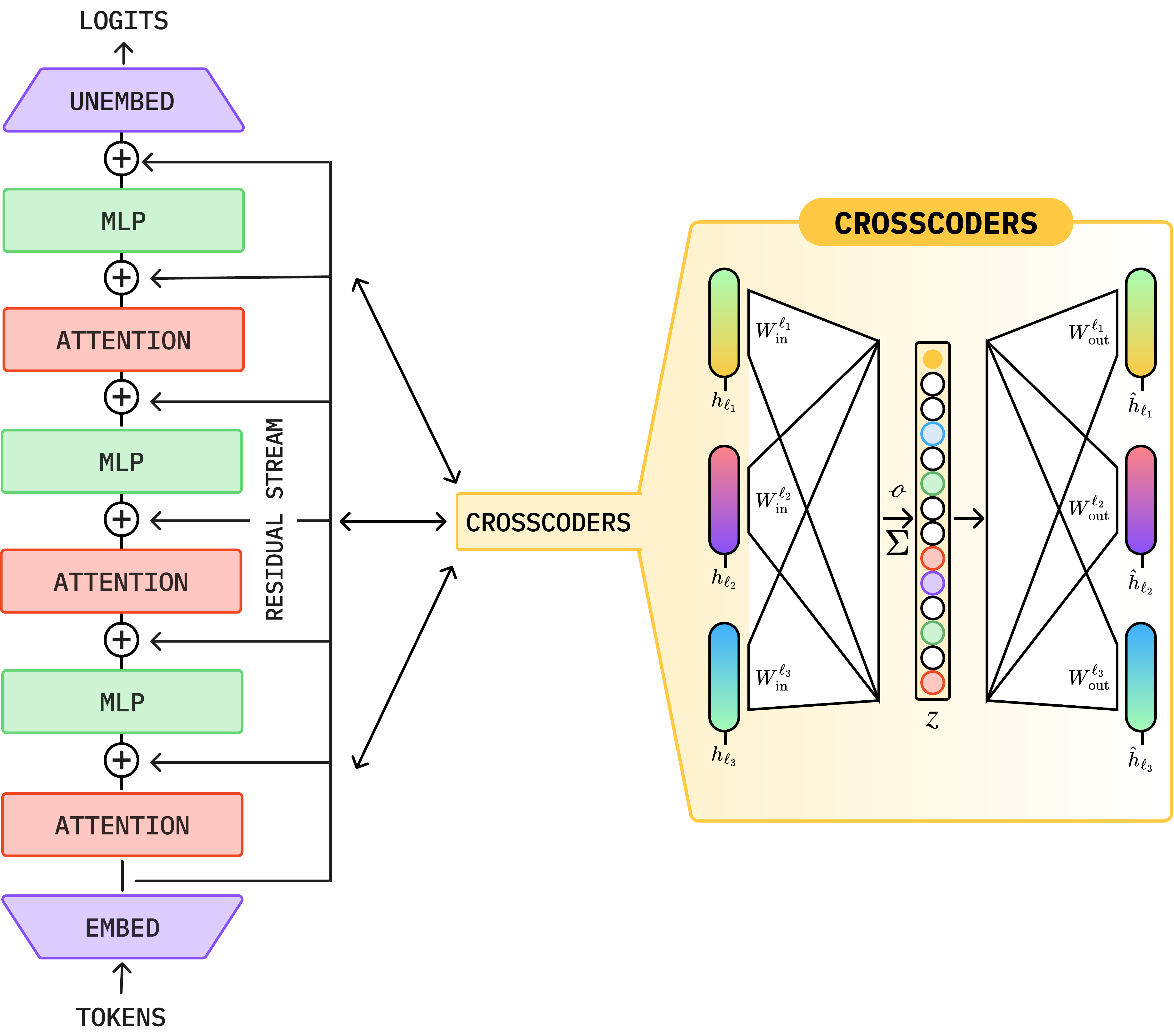}
    \caption{\textbf{Crosscoder.} Crosscoders are trained to reconstruct the residual stream across multiple layers simultaneously.}
    \label{fig:crosscoders}
\end{figure}

\begin{definitionbox}[Crosscoder (Figure~\ref{fig:crosscoders})]
A \emph{crosscoder} learns a shared sparse representation from activations across $L$ layers. It jointly encodes activations from all layers as
\[
z
=
\sigma_{\text{ReLU}}
\left(
\sum_{\ell=1}^{L}
W_{\text{in}}^\ell h_\ell
+
b_{\text{in}}
\right),
\]
where $z \in \mathbb{R}^{d_{\text{sparse}}}$ denotes the shared sparse feature activations and $W_{\text{in}}^\ell$ is the layer-$\ell$ encoder. Each layer is then reconstructed using a layer-specific decoder:
\[
\hat{h}_\ell
=
W_{\text{out}}^\ell z
+
b_{\text{out}}^\ell,
\qquad
\ell = 1,\ldots,L.
\]
Thus, a single latent feature can capture patterns that are shared across, persist through, or vary across multiple layers of the residual stream~\citeem{lindsey2024sparsecrosscoders}. Studies have also used crosscoders to discover shared features in activations from different models~(\citeestalp{ge2026evolution, bayazit-etal-2026-crosscoding}; \citeemalp{lindsey2024sparsecrosscoders, jiralerspong_cross-architecture_2025}), as detailed in Section~\ref{sec:model-diffing}.
\end{definitionbox}

A crosscoder is trained by minimising
\[
\mathcal{L}
=
\sum_{\ell=1}^{L}
\left\|
\hat{h}_\ell - h_\ell
\right\|_2^2
+
\lambda
\sum_j
\left(
\sum_{\ell=1}^{L}
\left\|
W_{\text{out},j}^{\ell}
\right\|_2
\right)
z^j,
\]
where $W_{\text{out},j}^{\ell}$ denotes the decoder vector of feature $j$ at layer $\ell$, and $\lambda$ controls the sparsity penalty.

\paragraph{Causal vs.\ Acausal Crosscoders.}
Crosscoders support two information flow patterns:
\begin{itemize}
    \item \textbf{Causal Crosscoders:} Read only from layers $\ell' \leq \ell$ at the current position. This respects the forward pass causality: earlier layers cannot read from later layers.
    \item \textbf{Acausal Crosscoders:} Read from any layer $\ell'$ regardless of depth ordering. Useful for understanding post-hoc feature relationships across the full model.
\end{itemize}

\paragraph{Applications to Circuit Analysis.}
By extracting cross-layer features, crosscoders help us understand how features propagate through the residual stream across layers, removing duplicate feature representations that complicate circuit analysis. This simplification allows features to ``jump'' across intervening identity operations, substantially reducing circuit complexity. More importantly, crosscoders reveal how input-relevant features transform as they flow through the network, distinguishing which features persist unchanged versus which are actively modified by downstream computation~\citeem{lindsey2024sparsecrosscoders}.

\subsection{Comparing SAEs, Transcoders, and Crosscoders}

Single-layer SAEs are efficient and locally interpretable but miss cross-layer feature evolution. Transcoders improve this by imitating MLP behaviour and revealing nonlinear transformations, though they require sequential analysis across depth. Cross-layer transcoders span multiple layers for global information flow but are extremely expensive. Crosscoders take a different approach, extracting shared features across layers simultaneously to resolve cross-layer superposition and simplify circuits without chaining predictions.

\subsection{Case Study: Induction Heads}
\label{subsec: induc head case study: feature extract}
Here, we examine induction heads~\citeem{olsson2022incontextlearninginductionheads}, a mechanism found in many modern language models, as a case study. We first describe their conditional-copying function and role in in-context learning, and then relate them to feature-extraction methods such as SAEs and CLTs, which reveal how this computation can be distributed across broader feature circuits.

\begin{definitionbox}[Induction Heads]
\label{def:induction-heads}

\emph{Induction heads}, first identified and analysed by
\citetem{elhage2021mathematical}, are attention heads that
implement a conditional copying rule over repeated subsequences. Let
$x=(x_1,\dots,x_T)$ be a token sequence, and consider positions $i<j<T$ such
that $x_i=x_j$. An attention head indexed $r$ (see Section~\ref{subsec:transformers} for an introduction to attention modules) is said to be an induction head when its
attention pattern assigns comparatively high attention weight from position
$j$ to position $i+1$ (i.e., high $A^r_{j,i+1}$) and its residual-stream update
\[
\Delta h^r
=
W_O^r V^r (A^r)^\top
\]
increases the model's logit for token $x_{i+1}$ when predicting the token at position $j+1$.\\

That is, induction heads implement the rule
\[
(x_i=x_j)
\;\Rightarrow\;
\hat{x}_{j+1}\approx x_{i+1}.
\]
They help enable in-context prediction by copying the continuation that followed an
earlier occurrence of the same token.

\end{definitionbox}

\begin{examplebox}[Induction in a Repeated Sequence]
Consider the sequence
\[
\texttt{A B C A B}
\]
where the model needs to predict the token after the second \texttt{B}. Observing the repeated prefix \texttt{A B}, an induction head attends from the second \texttt{B} to the position after the first \texttt{B} and contributes to the model’s prediction of \texttt{C}. This induction behaviour should generalise across tokens and positions.
\end{examplebox}

Induction heads were originally characterised as specific attention heads implementing this copying rule. However, subsequent work shows that we can understand the underlying computation more naturally at the level of distributed representations than at the level of isolated model components.

\begin{examplebox}[Induction Heads: Features Across Representations]

Feature extraction methods, such as SAEs, can decompose attention layer outputs into sparse, interpretable features that capture meaningful activation patterns beyond individual attention weights~\citeem{kissane2024interpretingattentionlayeroutputs}. Modern analyses suggest that induction heads are not isolated mechanisms but participate in distributed feature circuits, where multiple interacting features recur and compose across layers (\citeestalp{dunefsky2024transcodersinterpretablellmfeature, singh2024needsrightinductionhead}; \citeemalp{kissane2024interpretingattentionlayeroutputs, kamath2025tracing}).\\

SAEs trained on attention heads' activations reveal feature families of recurring patterns and contextual dependencies. Some of them align with behaviours originally attributed to induction heads~\citeem{kissane2024interpretingattentionlayeroutputs}. By analysing these features, we can understand how models maintain and manipulate information across tokens and contexts.\\

Cross-layer analyses, including crosscoders, suggest that certain features are represented consistently across layers rather than being recomputed at each layer~\citeem{lindsey2024sparsecrosscoders}. This perspective highlights that attention heads, including specialised ones such as induction heads, may be only part of broader, distributed feature representations.

\end{examplebox}

\subsection{Comparison with Alternative Feature Extraction Methods}

Besides SAEs, alternative techniques from signal processing and machine learning have also been adapted to extract features from neural networks. They make different modelling assumptions and therefore offer distinct trade-offs in interpretability, scalability, and robustness. Table~\ref{tab:decomposition-comparison} summarises the methods discussed in this section.

\begin{table}[tb]
\centering
\caption{Comparison of feature extraction methods.}
\label{tab:decomposition-comparison}
\renewcommand{\arraystretch}{1.3}
\begin{tabular}{>{\raggedright\arraybackslash}p{2.5cm} 
                >{\raggedright\arraybackslash}p{2.8cm} 
                >{\raggedright\arraybackslash}p{2.8cm} 
                >{\raggedright\arraybackslash}p{2.8cm} 
                >{\raggedright\arraybackslash}p{2.8cm}}
\toprule
\rowcolor{tableheader!80} \textcolor{white}{\textbf{Method}} & 
\textcolor{white}{\textbf{Assumption}} & 
\textcolor{white}{\textbf{Strengths}} & 
\textcolor{white}{\textbf{Weaknesses}} & 
\textcolor{white}{\textbf{Best Use Case}} \\
\midrule
\rowcolor{tablerow1} \textbf{SAE} & Sparse, linear features & Open-ended discovery; fast inference & Hyperparameter-sensitive; identifiability issues & Open-ended feature discovery \\
\rowcolor{tablerow2} \textbf{NMF} & Non-negative, additive parts & Parts-based interpretation; uniqueness guarantees & Requires non-negative data; limited sparsity & Naturally non-negative domains \\
\rowcolor{tablerow1} \textbf{PCA} & Variance aligns with features & Closed form; optimal low-rank reconstruction & Explains global variance and produces orthogonal features & Small-scale settings; as a baseline method \\
\rowcolor{tablerow2} \textbf{ICA} & Complete and independent sources & Theoretical guarantees when assumptions hold & Assumes independence; Gaussian blindness & When independence holds \\
\rowcolor{tablerow1} \textbf{Supervised methods} & Known concept labels & High performance on targeted concepts & Requires labels; no discovery & When certain concepts of interest are known \\
\rowcolor{tablerow2} \textbf{K-SVD} & Sparse, linear features & Explicitly solving the sparse-coding problem; no encoder bias & Slow inference; no learned encoder & Small, fixed sets of activations; as a baseline method \\
\bottomrule
\end{tabular}
\end{table}

\subsubsection{Non-Negative Matrix Factorisation}

Non-negative matrix factorisation (NMF; \citeestalp{lee1999learning}) decomposes a non-negative activation matrix $H \in \mathbb{R}^{d \times n}$ as $H \approx WA$, where both the feature dictionary $W \in \mathbb{R}^{d \times m}$ ($m$ is the preset number of features) and the coefficient matrix $A \in \mathbb{R}^{m \times n}$ are constrained to have non-negative elements. Because all components are non-negative, each feature contributes positively to the final representation, making it easier to interpret. Additionally, classical NMF’s optimisation process uses simple multiplicative update rules that preserve non-negativity. Moreover, under assumptions like \emph{separability} (each latent feature appears in at least one sufficiently ``pure'' example), NMF guarantees that the recovered decomposition is unique~\citeest{NIPS2003_1843e35d}.

However, classical NMF is not suitable for neural network activations because they can have negative elements, requiring additional preprocessing steps that may distort the activation-space structure. Moreover, NMF does not explicitly encourage feature sparsity, and scaling NMF to large dictionaries is computationally expensive. As a result, empirical work on extracting and analysing features in neural network activations has largely focused on Semi-NMF~\citeest{snmf}, which relaxes the non-negativity constraint on $W$. \citetest{shafran-etal-2026-constructing} showed that Semi-NMF can recover features competitive with SAEs on small-scale evaluations, but systematic comparisons at large scale remain limited. \citetem{yun-etal-2021-transformer} used a Semi-NMF-like approach to factorise BERT's contextualised hidden-state representations~\citeest{devlin2019bert}, allowing elements in $W$ to take both positive and negative values while constraining $A$ to be non-negative.

\subsubsection{Principal Component Analysis}
Principal Component Analysis (PCA; \citeestalp{pearson1901lines, hotelling1933analysis}) decomposes activations by finding an orthonormal basis of directions that maximise explained variance. Given a mean-centred activation matrix $H \in \mathbb{R}^{d \times n}$, where each column is an activation vector, PCA computes a low-rank approximation $H \approx U \Sigma V^\top$, where the columns of $U$ are the principal directions ordered by explained variance. PCA is fast, stable, and scalable. However, PCA imposes constraints and inductive biases that may misalign with neural network activations. Specifically, it identifies principal directions that maximise explained variance in the activation distribution, but much of the information of interest is region- or context-dependent and cannot be captured by variance alone. Moreover, PCA constrains the principal directions to be orthogonal, but in practice the underlying features correlate or partially overlap. Recent benchmarks~\citeest{wu2025axbench} reflect this misalignment: PCA underperformed most baselines, including SAEs, on both steering and concept-detection evaluations. While PCA can outperform SAEs in some localisation settings~\citeest{mueller2025mib}, it still generally lags behind simple supervised baselines.

\subsubsection{Independent Component Analysis}

Independent Component Analysis (ICA; \citeestalp{hyvarinen2000independent}) decomposes activations $H \in \mathbb{R}^{d \times n}$ as $H \approx WS$, where $n$ is the number of data points. Note that ICA may look similar to NMF at first glance. However, they were developed for different use cases and thus rely on different assumptions and optimisation procedures. In ICA, each column in $W \in \mathbb{R}^{d \times m}$ is a mixing vector, and $m$ is the number of different source signals; each row in $S \in \mathbb{R}^{m \times n}$ is a source signal (a signal has $n$ data points), with $m$ source signals in total. These signals are optimised to be independent of one another. Thus, if model activations are linear mixtures of statistically independent ``foundational features'', ICA can provably recover those features.

However, these assumptions may not hold for neural network activations. Specifically, classical ICA typically expects the completeness property (number of latent sources equals the dimensionality of the observed representation), and overcomplete decompositions (which are what we expect in activations) would require additional extensions~\citeest{lewicki2000learning}. Moreover, underlying features may not be statistically independent, which further limits the applicability of classical ICA. Although early studies applied it to analyse neural network activations, they did not find interpretability gains over SAEs~\citeest{cunningham2023sparse}. This may partly reflect the challenges discussed above.

More recently, however, \citetem{liu2026icalensinterpretinglanguage} showed that ICA can achieve performance comparable to that of SAEs, arguing that its previously reported underperformance may stem from limitations of classical ICA implementations poorly suited to the scale and statistical structure of neural network activations. These findings leave open whether ICA should be regarded primarily as a simple classical baseline or as an approach with underestimated, substantial headroom for further development.

\subsubsection{Supervised Feature Extraction Methods}
\label{subsubsec: supervised_extraction}

When the concepts of interest are known in advance, supervised methods can directly optimise representations aligned with those targets. Examples include distributed alignment search (DAS), which uses supervision to learn a change of basis that aligns representations with high-level causal variables~\citeest{das}; Supervised Multi-Dimensional Scaling (SMDS), which projects activations to preserve concept-level similarities~\citeest{smds}; and probes, which fit a concept direction in contrastive class-conditioned activations~\citeest{marks2024the}.

Notably, despite its simplicity, probing has outperformed SAEs in concept detection~\citeest{wu2025axbench}. It has also become a standard tool for identifying safety-relevant concepts in model representations, including refusal behaviour and implanted backdoors~(\citeestalp{arditi2024refusallanguagemodelsmediated}; \citeemalp{macdiarmid2024sleeperagentprobes}). We discuss probing in detail in Section~\ref{subsec:probing}. We put it under Section~\ref{sec:intermediate-computation} because we view its central methodological contributions and technical challenges as the readout of human-interpretable information---often expressed in natural language---from model activations, rather than the extraction of monosemantic features or causal variables.

When labelled data is available, supervised approaches often outperform SAEs on concept detection and steering tasks~(\citeestalp{huang-etal-2024-ravel, wu2025axbench}; \citeemalp{chaudhary2024evaluating}). They also avoid choosing dictionary sizes or sparsity penalties, since the concept set determines complexity. However, these methods require labelled data for each concept and cannot discover previously unknown features. Therefore, supervised methods suit settings where predefined, crisp concepts matter, whereas SAEs remain preferable for exploratory analysis.

\subsubsection{Dictionary Learning Baselines}
Classical dictionary learning methods provide useful baselines for evaluating SAEs. Techniques such as K-SVD alternate between sparse coding and dictionary updates and converge under certain conditions~\citeest{aharon2006k}. Comparative studies suggest that SAEs' primary advantage is fast inference because the encoder learns to decompose all activations, while potentially sacrificing optimal sparsity~\citeest{oneill2025compute}. As such, classical dictionary learning can still achieve superior dictionary quality in some cases and thus remains an important reference point for assessing the added value of SAEs.

\subsection{Interpretability-by-Design Methods for Intrinsic Interpretability}
\label{sec:architectural-interventions}

The methods surveyed thus far in this section aim to extract and analyse more monosemantic units from polysemantic model activations. They share a common premise: the model's original representations are too entangled to be directly interpretable, so interpretability is applied post hoc to recover simpler units of analysis. A parallel research programme, which we term \emph{Interpretability-by-Design Methods for Intrinsic Interpretability}, inverts this premise by asking whether deep neural models can instead be designed to be mechanistically interpretable by construction~(\citeestalp{pearce2025bilinear, park2025monet, oldfield2024multilinear, yang2025mixture, gao2025weight}; \citeemalp{adler2025towards, draye2025spartan, dooms2025compositionality}). This subsection extends the discussion of this research direction.

Next, we introduce five approaches for constructing models that are more interpretable by design. Three impose model sparsity constraints at different levels: feature-sparse training (Section~\ref{subsubsec: solu}), weight-sparse training (Section~\ref{subsubsec: weight sparse transformer}), and attention-sparse training (Section~\ref{subsubsec: spartan}). The remaining two approaches instead modify the computational primitives themselves, using more interpretable operators as units of construction and analysis (Sections~\ref{subsubsec: molt} and~\ref{subsubsec: kan}).

Interpretability should be understood as a spectrum rather than a binary property. At least in their current form, these techniques do not make model computation completely or immediately understandable. Instead, they make models easier to analyse and ensure the analysis pipeline preserves more information about the original computation than post-hoc approximations. Table~\ref{tab:architectural-methods} summarises these approaches and their trade-offs.

\subsubsection{Feature-Sparse Training}
\label{subsubsec: solu}
Softmax linear units (SoLU; \citeemalp{elhage2022solu}) are an earlier attempt at architectural intervention. SoLU is a specialised activation function that applies a softmax-based transformation across the layer's $d_{\mathrm{mlp}}$ neurons. Specifically, given the pre-activation vector
\[
u = W_1 h + b_1 \in \mathbb{R}^{d_{\mathrm{mlp}}},
\]
SoLU calculates the activation values as
\begin{equation}
    \sigma_{\text{SoLU}}(u)
    =
    u \odot \mathrm{softmax}(u).
\end{equation}
This softmax-based transformation induces competition among neurons. When one neuron receives a large pre-activation value, the softmax suppresses the others' values, analogous to lateral inhibition. This competition encourages certain neurons to specialise and dominate, increasing the prevalence of monosemantic neurons associated with human-interpretable concepts. 

However, it also reveals an undesirable phenomenon: although many secondary neurons’ activations are suppressed to low magnitudes, these neurons remain polysemantic, with many features compressed into them. Later layer normalisation~\citeem{ba2016layernormalization} can amplify them and recover the polysemanticity property~\citeem{elhage2022solu}. The ``smuggling'' by layer normalisation suggests that imposing local interpretability constraints may not eliminate neural networks’ incentive to compress information through superposition. This issue motivates global constraints across the neural network, such as weight-sparse training discussed below.

\subsubsection{Weight-Sparse Training}
\label{subsubsec: weight sparse transformer}
A weight-sparse transformer~\citeest{gao2025weight} treats weight sparsity as an explicit structural constraint that is progressively strengthened during training. \citetest{gao2025weight} optimise the standard cross-entropy objective along with an $L_0$ constraint on model weights after each optimisation step. Specifically, within each weight matrix, they retain only the largest-magnitude parameters and set the rest to zero. The $L_0$ constraint anneals from a dense initialisation to the desired sparsity level over training. The resulting transformer has most weight parameters exactly zero, making the remaining computational structure much easier to decompose into small, task-specific circuits.

Its interpretability gains can be substantial. Under highly sparse regimes (e.g., $<1\%$ active weights), neurons receive input from far fewer upstream neurons, and the model’s circuits for specific capabilities or behaviours become much sparser. Mean-ablation experiments further confirm that these circuits are
both necessary and sufficient for the capability or behaviour.

\citetest{gao2025weight} characterised the trade-off between model capability and interpretability induced by weight sparsity. Increasing sparsity generally makes the model easier to decompose into small, interpretable circuits. However, it also degrades task performance, producing a capability--interpretability Pareto frontier. Nonetheless, the authors found that increasing model size can shift this frontier outward. At comparable task performance, larger sparse models can have substantially smaller, more interpretable circuits than smaller models; at comparable interpretability, they can retain stronger capabilities. This suggests that scaling model size alongside structural sparsity may offer a path to more interpretable yet high-capability models. However, preserving interpretability beyond tens of millions of nonzero parameters remains challenging, and weight-sparse training currently requires substantially more compute than traditional training.

\begin{table}[tb]
\centering
\caption{Comparison of interpretability-by-design methods.}
\label{tab:architectural-methods}
\renewcommand{\arraystretch}{1.3}
\begin{tabular}{>{\raggedright\arraybackslash}p{2.8cm} 
                >{\raggedright\arraybackslash}p{3.2cm} 
                >{\raggedright\arraybackslash}p{4cm} 
                >{\raggedright\arraybackslash}p{4cm}}
\toprule
\rowcolor{tableheader!80} \textcolor{white}{\textbf{Method}} & 
\textcolor{white}{\textbf{Approach}} & 
\textcolor{white}{\textbf{Interpretability Gain}} & 
\textcolor{white}{\textbf{Primary Limitation}} \\
\midrule
\rowcolor{tablerow1} \textbf{Feature-Sparse Training (SoLU)} & Softmax-gated activation function & Increased neuron monosemanticity & Layer normalisation smuggling displaces superposition
\\
\rowcolor{tablerow2} \textbf{Weight-Sparse Training} & $L_0$ constraint on parameters & More monosemantic neurons and much smaller circuits & Training instability and scalability \\
\rowcolor{tablerow1} \textbf{Attention-Sparse Training} & $L_0$ regularisation on attention & Makes attention patterns interpretable & Requires fine-tuning on a trained language model \\

\rowcolor{tablerow2} \textbf{MOLT} & Sparse mixture of linear transforms & Reveals implemented operations directly & Complexity of matrix analysis \\
\rowcolor{tablerow1} \textbf{KANs} & Learnable splines on edges & Symbolic formula extraction & Scalability to LLMs is unproven \\
\bottomrule
\end{tabular}
\end{table}
\subsubsection{Attention-Sparse Training}
\label{subsubsec: spartan}
In transformers, attention patterns produced by an attention head are usually \emph{diffuse}, i.e., every token attends to every other token. Diffuse attention patterns are difficult to interpret and audit. Previous studies have therefore explored \emph{attention-sparse training}, which aims to sparsify attention patterns to make them more interpretable~(\citeestalp{lei2025spartan}; \citeemalp{draye2025spartan}). Specifically, \citetest{lei2025spartan} proposed \emph{SPARse TrANsformer World model} (SPARTAN), an object-centric world model trained with $L_0$ regularisation on attention patterns and a constrained optimisation objective to maintain cross-entropy performance equal to that of a dense counterpart. \citetem{draye2025spartan} further adapted this technique to fine-tune and sparsify a trained language model's attention patterns.

After training (either from scratch or via fine-tuning), each attention head attends only to specific tokens, and the model performance remains close to the original. This result suggests that many attention patterns may be redundant noise. In CLT-based circuit discovery (see Section~\ref{sec:circuit-replacement}), \citetem{draye2025spartan} showed that sparsified attention patterns reduce the number of mediating components involved in feature-to-feature attribution, making the resulting attribution graphs easier to analyse. 

\subsubsection{Modifications of Computational Primitives: MOLT}
\label{subsubsec: molt}

One alternative approach to learning interpretable structure is to reconsider what the computational primitives should be~(\citeestalp{oldfield2025towards, park2025monet}; \citeemalp{braun2025interpretability, bushnaq2025stochastic}). A recent example is the sparse mixture of linear transforms (MOLT), which represents a conceptual shift from feature extraction towards transformation extraction~\citeem{anthropic2025molt}.

Standard SAEs and transcoders primarily decompose model activations into features. However, many algorithmic tasks, such as arithmetic or geometric reasoning, are more naturally described by \emph{operations} such as ``add 3'' or ``rotate a vector''. In these tasks, we want to know how the model transforms one activation into another through components like weight matrices in MLP blocks. To achieve this, MOLT replaces the MLP with a sparse mixture of learned linear transformation matrices. The basic unit of analysis becomes transformations applied to model activations rather than features.

For example, in a modular arithmetic task, MOLT can represent the concept of addition with a learned rotation matrix, which relates more directly to the underlying algorithm than a collection of state-specific features. More broadly, approaches that represent computational operations explicitly, as MOLT does, may facilitate the development of interpretable representations for tasks governed by systematic operations. We view Kolmogorov--Arnold Networks (KANs) as an approach that embodies this idea, as discussed below.

\subsubsection{Kolmogorov--Arnold Networks}
\label{subsubsec: kan}

Kolmogorov--Arnold Networks (KANs) represent one of the most radical departures from standard neural network architectures~\citeest{liu2025kan}. In conventional neural networks, each neuron receives outputs from upstream neurons through incoming edges weighted by scalar parameters, sums these weighted inputs, and then applies a fixed nonlinear activation function (the whole process can be represented as matrix multiplication in components like MLP blocks and attention modules in Section~\ref{subsec:transformers}). Inspired by the Kolmogorov--Arnold representation theorem, KANs place learnable nonlinear functions, typically parameterised using B-splines, on edges rather than using fixed nonlinear activation functions at nodes.

Specifically, consider a single KAN node receiving an input vector
$v = (v_1, \dots, v_d) \in \mathbb{R}^d$ from the previous layer. Its output is
\[
y = \sum_{i=1}^{d} \psi_i(v_i), \qquad \psi_i(v_i) = \sum_{m} c_{i,m} B_m(v_i),
\]
where $B_m(\cdot)$ denotes the $m$-th basis function, such as a cubic B-spline basis,
and $c_{i,m}$ is a learnable scalar coefficient. In this formulation, the edge is not scalar multiplication but a learnable univariate function $\psi_i(\cdot)$, and a node simply sums the outputs of its incoming edges. Intuitively, KANs shift nonlinear transformations from nodes to edges, offering an alternative way to structure and interpret neural network computations.

This design gives KANs an explicitly compositional and thus more interpretable structure. As the nonlinearities are univariate and can be visualised directly on individual edges, researchers can use techniques like symbolic regression to approximate and inspect the learned analytical expressions, such as rediscovering known physical equations. Although KAN training is currently more computationally expensive than training standard MLPs and has not yet been convincingly demonstrated at the scale of large language models, this architectural innovation points to an ambitious direction towards interpretable neural networks in which parameters correspond to mathematically meaningful functions.

\paragraph{Summary.} This section introduced several classical methods for feature extraction and analysis, followed by approaches that aim to make models intrinsically interpretable. However, whether features are recovered post hoc or induced by design, they remain units of analysis rather than complete explanations of the model's mechanism, i.e., how these units interact and causally influence one another. Section~\ref{sec:circuit} therefore shifts from individual features to \emph{circuits}: computational graphs of interacting components that describe how the model implements its behaviours.

\section{Circuits in Mechanistic Interpretability}
\label{sec:circuit}
Mechanistic interpretability aims not only to identify what information a model represents, but also to explain how it transforms and combines that information across layers to produce behaviour. Feature analysis is a useful tool for describing activation vectors, but it describes them in isolation. Many model behaviours arise from structured interactions among multiple components that operate together across layers. To explain these behaviours and their mechanisms, we need a high-level abstraction that captures critical computational patterns of these features. The circuit framework (Figure~\ref{fig:circuit}; introduced in Section~\ref{subsec:circuits}) addresses this gap by modelling behaviour as coordinated computational graphs and making explicit how components causally interact. 

Notably, the circuit framework is not the only way to model an internal mechanism. Early studies have explored alternative ways. For example, \citetest{rnnfsa1992} and \citetest{OMLIN199641} categorised a recurrent neural network's hidden representations into different states and modelled its decision process as a finite-state automaton when the input sentence was fed sequentially. Nonetheless, the circuit framework has attracted growing research interest in studying neural networks' internal mechanisms. Below, we introduce approaches for identifying and constructing circuits.

\subsection{Circuit Discovery Methods}
\label{sec:circuit-discovery-methods}

Circuit discovery methods aim to identify internal computational graphs responsible for specific capabilities or outputs in a neural network. Below, and in Table~\ref{tab:circuit-methods}, we formalise key techniques and methods, along with their strengths and limitations.

\begin{definitionbox}[Activation Patching]
\label{def:attribution-patching}
Given a model $f_\theta(\cdot)$ parameterised by $\theta$ and an input sequence $x$, \emph{activation patching} evaluates the causal contribution of some model components $S$ by replacing their activations with specified reference values $\tilde{h}_S$ and measuring the resulting change in the model output. The causal importance of $S$ for $x$ is defined as
\[
s_S(x)
=
M\bigl(f_\theta(x)\bigr)
-
M\Bigl(f_\theta\bigl(x_{[S \leftarrow \tilde{h}_S]}\bigr)\Bigr),
\]
where \(M(\cdot)\) is a scalar performance metric computed from the model output, such as the logit difference for the answer token, and $x_{[S \leftarrow \tilde{h}_S]}$ denotes a forward pass on $x$ in which the activations of components in $S$ are replaced by $\tilde{h}_S$ while the remaining computation proceeds normally. The reference activations $\tilde{h}_S$ usually come from a corrupted reference run, where the input is modified to add or remove specific information. Some works have also set $\tilde{h}_S$ to zero or the mean model activation. The resulting change in model output quantifies the causal effect of intervening on $S$~\citeest{zhang2024towards}.
\end{definitionbox}

\begin{definitionbox}[Path Patching]
\label{def:path-patching}
\emph{Path patching} extends activation patching on model components to specific edges in the computational graph. For a connection \(e_{ij}\) from model component \(i\) to \(j\) denoting that the output activations of \(i\) are fed to \(j\) as one of its inputs, it replaces activations along \(e_{ij}\) with reference values \(\tilde{h}_{e_{ij}}\) and measures \(e_{ij}\)'s causal effect as:
\[
s_{ij}(x) = M\bigl(f_\theta(x)\bigr) - M\Bigl(f_\theta\bigl(x_{[e_{ij} \leftarrow \tilde{h}_{e_{ij}}]}\bigr)\Bigr).
\]
This allows finer-grained causal localisation along information flows~\citeem{goldowskydill2023localizingmodelbehaviorpath}.
\end{definitionbox}

\begin{definitionbox}[Automatic Circuit Discovery (ACDC)]
\label{def:acdc}
\emph{ACDC}~\citeest{conmy2023automatedcircuitdiscoverymechanistic} discovers a circuit by iteratively using path patching to identify important connections in the model’s computational graph. Traversing the graph in reverse topological order from the output node, it patches the activation transmitted along each edge and measures the resulting change in a task-specific metric. An edge $e_{ij}$ will be pruned if patching it produces $|s_{ij}(x)|$ less than a predefined threshold $\tau$; otherwise, it is retained in the graph and its upstream edges will be investigated. The remaining subgraph represents the discovered circuit.
\end{definitionbox}

\begin{definitionbox}[Edge Attribution Patching (EAP)]
\label{def:eap}
\emph{EAP}~\citeem{syed2023attributionpatchingoutperformsautomated} estimates the causal importance of edges in a model’s computational graph by approximating the effect of path patching with a first-order Taylor expansion. For an edge \(e_{ij}\) connecting components \(i\) and \(j\), its importance score is approximated as
\[
s_{ij}(x)
\approx
\frac{\partial M}{\partial h_{e_{ij}}}
\cdot
\left(h_{e_{ij}}^{\mathrm{clean}} - h_{e_{ij}}^{\mathrm{corrupt}}\right),
\]
where \(h_{e_{ij}}^{\mathrm{clean}}\) and \(h_{e_{ij}}^{\mathrm{corrupt}}\) denote the activations transmitted from component $i$ to component $j$ along edge $e_{ij}$ on clean and corrupted inputs, respectively, when processing $x$. $M(\cdot)$ is a scalar metric of model performance as mentioned above. EAP is substantially faster than ACDC because the attribution scores for all edges are computed using only two forward passes and a single backward pass. Specifically, the forward passes obtain $h_{e_{ij}}^{\mathrm{clean}}$ and $h_{e_{ij}}^{\mathrm{corrupt}}$, while the backward pass computes $\frac{\partial M}{\partial h_{e_{ij}}}$ for all components simultaneously. In contrast, ACDC requires a separate forward pass for each edge it tests.\\

Edges with large score magnitudes are generally considered main contributors to the target behaviour. Therefore, one simple approach to circuit construction is directly selecting the highest-scoring edges~\citeem{syed2023attributionpatchingoutperformsautomated}. Subsequent work has improved EAP in different directions: EAP-IG~\citeest{eapig} and EAP-GP~\citeest{eapgp} improve edge-attribution estimation, while PEAP~\citeest{haklay-etal-2025-position} extends patching across token positions to produce finer-grained attribution graphs that capture cross-token interactions.
\end{definitionbox}

Besides intervention (activation patching, ACDC, etc.) and attribution (EAP, EAP-IG, etc.) techniques, another line of approaches treats circuit discovery as an optimisation problem~\citeest{netmodular, cao-etal-2021-low, bhaskar2024finding, haider2026multigranular}. Specifically, they use gradient descent to learn which parts of the computational graph can be removed without affecting model behaviour. For example, Edge Pruning~\citeest{bhaskar2024finding} learns edge masks to discover sparse connections, while multi-granular node pruning~\citeest{haider2026multigranular} learns node masks at different scales, from entire blocks to individual neurons. These methods can usually discover sparser circuits than intervention and attribution counterparts. However, the optimisation process itself introduces randomness and sensitivity to training configurations like the learning rate, which may make the subsequent analysis less robust and interpretable.

Circuits can explain models in different contexts, from a single input--output pair to a capability or behaviour. Identifying a circuit for a specific input or prompt is called \emph{query-level circuit discovery}~\citeest{querycircuit}. In contrast, \emph{task-level circuit discovery}~\citeest{wang2022interpretabilitywildcircuitindirect, kharlapenko2025scaling} seeks to identify a circuit that implements a broader capability or task, typically operationalised by a dataset of examples, such as the indirect object identification (IOI) dataset~\citeest{wang2022interpretabilitywildcircuitindirect}. In the task-level setting, components’ and edges’ attribution scores (e.g., $s_S(x)$ or $s_{ij}(x)$) are aggregated across inputs to reduce example-specific noise, yielding a circuit that characterises the task as a whole.

The techniques and methods introduced above, including activation patching, path patching, ACDC, EAP, EAP-IG, and mask-based pruning methods, have primarily been developed and evaluated in the task-level setting. More recent work has extended EAP-IG to identify circuits for individual prompts~\citeest{querycircuit}. This enables the discovery of sparse yet faithful circuits in the model even for complex, natural-language user queries, where the relevant computation may vary substantially from one input to another.

\begin{table}[tb]
\centering
\caption{Trade-offs and characteristics of circuit discovery methods.}
\label{tab:circuit-methods}
\renewcommand{\arraystretch}{1.3}
\begin{tabular}{>{\raggedright\arraybackslash}p{2.3cm} >{\raggedright\arraybackslash}p{2.8cm} >{\raggedright\arraybackslash}p{2.8cm} >{\raggedright\arraybackslash}p{2.8cm}}
\toprule
\rowcolor{tableheader!80} \textcolor{white}{\textbf{Method}} & \textcolor{white}{\textbf{Core Idea}} & \textcolor{white}{\textbf{Strengths}} & \textcolor{white}{\textbf{Limitations}} \\
\midrule
\rowcolor{tablerow1} \textbf{Activation Patching}
& Intervene on components and measure the resulting output change
& Simple and intuitive; provides direct causal evidence
& Requires many interventions; component-level granularity can be coarse \\

\rowcolor{tablerow2} \textbf{Path Patching}
& Intervene on individual information paths or edges and measure their causal effects
& Finer-grained localisation of information flow
& Computationally expensive due to the large number of candidate paths \\

\rowcolor{tablerow1} \textbf{ACDC}
& Greedily prune edges from the computational graph while testing whether the target behaviour is preserved
& Fully automated; directly searches for a behaviourally sufficient subgraph
& Computationally expensive due to sequential edge pruning; sensitive to pruning order and threshold choices \\

\rowcolor{tablerow2} \textbf{EAP}
& Approximate edge-level attribution using path patching and first-order gradients, enabling parallel evaluation of many candidate edges
& Highly parallelisable and computationally efficient; enables large-scale edge attribution
& Relies on a first-order approximation; attribution scores may not fully capture nonlinear or interaction effects \\

\rowcolor{tablerow1} \textbf{SAE- or CLT-Based}
& Discover circuits over sparse, interpretable features rather than raw model components
& Leverages interpretable features; enables finer-grained feature-level circuit analysis
& Depends on feature quality; requires additional computation to train and analyse SAEs/CLTs \\
\bottomrule
\end{tabular}
\end{table}

\subsection{Circuit Discovery with Replacement Models}
\label{sec:circuit-replacement}

The methods introduced above discover circuits \emph{within} the target model. Their nodes correspond to coarse-grained components such as attention heads and MLP blocks, which are mostly not immediately understandable. A complementary approach first constructs a more interpretable replacement model (Section~\ref{subsubsec:cross-layer-transcoder}) that approximates the original model’s computation using sparse linear features, and then performs circuit discovery within this replacement model.

Recent work has implemented this paradigm using cross-layer transcoders (CLTs; Section~\ref{subsubsec:cross-layer-transcoder})~\citeem{ameisen2025circuit, merullo2025replicating}. In particular, Circuit Tracer~\citeem{ameisen2025circuit} replaces the model’s MLPs with CLTs while retaining the underlying model's attention and residual-stream computation. This yields a prompt-specific attribution graph whose principal internal nodes are sparse CLT features rather than entire MLP blocks or attention heads.

\begin{definitionbox}[Circuit Tracer]
\label{def:circuit-tracer}

\emph{Circuit Tracer}~\citeem{ameisen2025circuit} constructs a prompt-specific attribution graph in a CLT-based replacement model. For a given input prompt, consider a source feature $\phi_s$ at layer $\ell_s$ active at token position $c_s$ with activation value $z_{\ell_s}^{s}$, and a downstream target feature $\phi_t$ at layer $\ell_t$ active at position $c_t$. Recall that a CLT feature can write to the reconstructed MLP outputs of its own and all subsequent layers (Figure~\ref{fig:clt}); the source feature has layer-specific output vectors to these layers, corresponding to columns of the layer-specific decoder matrices:
\[
v_{\mathrm{out},\phi_s}^{\ell}
\coloneqq
W_{\mathrm{out},s}^{\ell_s \to \ell}
\in \mathbb{R}^{d},
\qquad
\ell_s \leq \ell.
\]
On the other hand, the target feature reads from the MLP input through its corresponding encoder vector, given by a row of the encoder matrix:
\[
v_{\mathrm{in},\phi_t}
\coloneqq
\left(W_{\mathrm{in},t}^{\ell_t}\right)^\top
\in \mathbb{R}^{d}.
\]

Then, let
\[
J_{c_s,\ell \to c_t,\ell_t}
\in \mathbb{R}^{d\times d}
\]
denote the Jacobian that maps a perturbation written at token position $c_s$ and layer $\ell$ to the MLP input at position $c_t$ and layer $\ell_t$, where each of the entries is
\[
\left[
J_{c_s,\ell \to c_t,\ell_t}
\right]_{pq}
=
\frac{
\partial \left(h_{\ell_t}^{\mathrm{in},c_t}\right)_p
}{
\partial \left(h_{\ell}^{\mathrm{out},c_s}\right)_q
}.
\]

The direct attribution from source feature $\phi_s$ to target feature $\phi_t$ can then be modelled as
\[
A_{\phi_s \rightarrow \phi_t}
=
\sum_{\ell_s \leq \ell < \ell_t}
z_{\ell_s}^{s}
\left(v_{\mathrm{in},\phi_t}\right)^\top
J_{c_s,\ell \to c_t,\ell_t}
v_{\mathrm{out},\phi_s}^{\ell}.
\]

Each term in the summation measures the contribution of the source feature $\phi_s$’s write at layer $\ell$ to the pre-activation of target feature $\phi_t$, transported through the residual stream and attention heads. $A_{\phi_s \rightarrow \phi_t}$ aggregates the downstream writes across layers produced by the same source feature. Notably, \citetem{ameisen2025circuit} stop the gradients flowing through intermediate MLP outputs and fix attention patterns and normalisation denominators. This lets each edge capture direct feature-to-feature influence rather than effects mediated through intermediate feature nodes.

\end{definitionbox}

CLT-based circuit discovery differs from conventional circuit discovery in two aspects. First, the resulting circuit's nodes correspond to fine-grained linear features rather than coarse MLP blocks or attention modules. Using automated feature-interpretation pipelines (Section~\ref{sec:automated-interp}) to assign natural-language descriptions to these features, human inspectors can more easily understand the discovered mechanism. Second, an edge in a CLT-based circuit denotes the direct, input-specific influence between two features, which is different from the formulation of attribution methods based on clean--corrupted reference pairs, such as EAP and EAP-IG. Specifically, in Circuit Tracer, a source feature can write to the residual stream at multiple subsequent layers, and the strength of an edge from the source feature to a target feature is the aggregate of the contributions of these writes to the target feature.

For a given prompt, Circuit Tracer produces a sparse attribution graph that describes active features and their interactions, making it well suited to query-level circuit discovery. Interpretability platforms such as Neuronpedia~\citeem{neuronpedia} also offer interactive implementations. However, the approach still has several limitations. Specifically, CLTs only approximate MLP computations, so the discovered circuit may only cover a part of the model's underlying mechanism. Moreover, feature-level attribution graphs can have many nodes and edges, so we must prune, cluster, or otherwise condense them before they become understandable.

\subsection{Causal Verification}
\label{sec:causal-verification}

Once we identify circuits, two critical questions arise: are they causally responsible for the behaviour, and can we modify them reliably?

\paragraph{Causal Necessity and Sufficiency.}
A circuit is \emph{causally necessary} for a behaviour if removing or ablating the circuit degrades that behaviour. A circuit is \emph{causally sufficient} if activating the circuit alone, i.e., patching all edges outside the circuit with activations from a corrupted reference run, reproduces the behaviour. In practice, a circuit is rarely sparse, necessary, and sufficient at the same time; it typically explains 60--80\% of behavioural variance.

\paragraph{Evaluating Circuits and Benchmarking Discovery Methods.}
A circuit is often considered \emph{faithful} if it is both causally necessary and causally sufficient for the behaviour it is claimed to explain~\citeest{eapig, marks2025sparse}. Two popular approaches to validate causality are causal scrubbing and interchange interventions~(\citeemalp{chan2022causal}; \citeestalp{das}). Causal scrubbing~\citeem{chan2022causal} systematically evaluates a mechanistic hypothesis by replacing model activations with those from inputs deemed equivalent under the hypothesis and testing whether these interventions preserve the model behaviour. Interchange interventions~\citeest{das} test whether model activations causally correspond to higher-level variables by substituting activations across inputs and measuring the resulting change in model behaviour. Together, these methods provide stronger evidence for mechanistic explanations than observational analysis alone.

More recently, several works have proposed metrics to quantify circuit faithfulness as a continuous score~\citeest{eapig, querycircuit}, allowing systematic comparisons of circuits discovered by different methods. By aggregating faithfulness scores across tasks and settings, benchmarks such as MIB~\citeest{mueller2025mib} evaluate how well circuit discovery methods identify causally important model components. However, achieving strict necessity and sufficiency is often hard because many model behaviours depend on distributed representations and multiple redundant pathways~\citeest{barez2025chain}. Even if we identify a circuit that captures the primary mechanism of a behaviour, components outside the circuit may still play important roles, such as handling edge cases and computational robustness.

\subsection{Case Study: Induction Heads}
\label{subsec: induc head case study: circuit}
\begin{examplebox}[Circuit Perspective on Induction Heads]

Following our introduction to induction heads in Section~\ref{subsec: induc head case study: feature extract}, we consider a repeated sequence:

\[
\texttt{A B C A B}
\]

where the model needs to predict the token after the second \texttt{B}. The induction head alone is not sufficient to implement induction behaviour from scratch. It still needs help from previous layers’ attention heads to locate the token following the current token's previous occurrence, so it knows it should attend to \texttt{C} in this example~\citeem{olsson2022incontextlearninginductionheads}.\\

Specifically, in a two-layer attention-only transformer, \citetem{elhage2021mathematical} discovered the \emph{previous-token head}, which writes every token’s information to the next position. The change propagates to the residual stream to be read by the induction head, allowing it to locate \texttt{C} (because now some information about the first \texttt{B} is contained in \texttt{C}). The two kinds of heads form the \emph{induction circuit} that is sufficient to implement the induction behaviour in the two-layer small model.\\

Empirical evidence supports this mechanistic interpretation. Head ablation experiments show that induction heads causally support in-context learning in the attention-only model. Circuit analysis also reveals induction heads’ dependence on earlier previous-token heads. Moreover, induction heads emerge abruptly during training, coinciding with sharp improvements in in-context learning performance.\\

In larger models with additional layers and MLPs, the mechanism behind induction behaviour becomes more complex. However, the core principle remains the same: context propagates across layers and is used to conditionally retrieve tokens based on repeated subsequence matches. From this perspective, the circuit framework complements feature-based analysis by explicitly identifying the model’s internal mechanism that implements a specific behaviour.
\end{examplebox}

\subsection{A Unified Evaluation Framework for Mechanistic Interpretability}
\label{sec:evaluation-framework}

A persistent challenge in mechanistic interpretability (MI) is the lack of standardised evaluation. Different interpretability methods use incompatible metrics. For example, for the same behaviour of interest, circuit discovery can work at the level of MLP/attention modules~\citeest{conmy2023automatedcircuitdiscoverymechanistic}, CLT features~\citeem{ameisen2025circuit}, or even neurons~\citeest{arora2026language}. A CLT-based circuit must also verify CLT fidelity; an MLP/attention-level circuit should test and explain each node's role. Thus, it is hard to compare them and measure progress. To address this, this section presents a unified evaluation framework covering five dimensions---attribution faithfulness, completeness, stability, causal validity, and human interpretability---where different methods can be jointly tested.

\subsubsection{The Need for Standardised Evaluation}

Before we discuss the five dimensions, we first explain in detail the need for standardised evaluation. Specifically, current evaluation practices face several challenges:

\begin{itemize}
\item \textbf{Utility ambiguity:} Methods are initially developed for different aims and evaluated against distinct objectives, yet have overlapping application scenarios. For example, SAEs are assessed using reconstruction error and sparsity metrics; circuits are evaluated using behavioural recovery or faithfulness tests; and difference-in-means steering vectors (see Section~\ref{subsec:steering}) are examined by behaviour change---although all three approaches can be used to steer model outputs. Their unclear intended uses make it hard to compare their utility.

\item \textbf{Cherry-picking:} Existing work often reports interpretability results on carefully selected examples, samples from a narrow distribution, or capabilities where a method performs well. It may not report negative results, failure cases, or coverage statistics. This makes it difficult to assess the robustness and generality of the findings and determine whether the proposed method genuinely outperforms existing ones.

\item \textbf{Missing baselines:} Many proposed methods' claimed interpretability gains are not systematically and adequately compared against intuitive alternatives or baselines such as behavioural methods, random activations, or neuron-level analysis. Without this comparison, it is unclear whether a method extracts genuinely meaningful information.

\item \textbf{Reproducibility gaps:} Many existing techniques involve substantial hyperparameter choices, heavy training procedures, and high variation across seemingly irrelevant factors. Reproducibility is often underreported and has not yet been established as a property to optimise.

\item \textbf{Faithfulness--understandability gap:} Methods often demonstrate how understandable and informative their outputs are to humans. However, those outputs' faithfulness to the underlying model mechanism receives less attention. This rewards explanations that are persuasive rather than faithful.

\end{itemize}

\citetest{lan-etal-2026-make} have called for a shared platform for tracking and auditing interpretability insights and claims across papers; \citetem{philosophicalevalframework} proposed a framework that evaluates mechanistic explanations from several philosophy-of-science perspectives. However, these discussions remain largely conceptual. The MI community needs standardised, concrete evaluation protocols and benchmarks, along with established reporting practices, to enable principled comparisons across methods and findings. They are especially important for measuring real progress in the field, identifying systematic failure modes, and providing evidence about what methods do and do not reveal. More broadly, they are necessary for mechanistic interpretability to mature from a collection of promising techniques into a cumulative scientific discipline.

\subsubsection{Five Evaluation Dimensions}
\label{subsec:five eval dim}
We propose a framework that situates mechanistic interpretability evaluation within five complementary dimensions. No single dimension suffices---strong methods should perform well across all five. Table~\ref{tab:evaluation-framework} summarises these five dimensions. We describe each one with example metrics that convey the idea.

\begin{table}[tb]
\centering
\caption{Unified evaluation framework for mechanistic interpretability.}
\label{tab:evaluation-framework}
\renewcommand{\arraystretch}{1.4}
\begin{tabular}{
>{\raggedright\arraybackslash}p{2.8cm}
>{\raggedright\arraybackslash}p{4.3cm}
>{\raggedright\arraybackslash}p{4.3cm}
>{\raggedright\arraybackslash}p{3.3cm}}
\toprule
\rowcolor{tableheader!80}
\textcolor{white}{\textbf{Dimension}} &
\textcolor{white}{\textbf{Definition}} &
\textcolor{white}{\textbf{Metrics}} &
\textcolor{white}{\textbf{Methods}} \\
\midrule

\rowcolor{tablerow1}
\textbf{Attribution Faithfulness} &
Do attribution scores accurately reflect the causal effects of the identified components? &
Intervention-effect agreement; effect-size error; counterfactual effect recovery &
Ablation; activation / path patching; interchange interventions \\

\rowcolor{tablerow2}
\textbf{Completeness} &
How much of the behaviour is accounted for by the identified components? &
Loss recovery; behavioural coverage; unexplained variance &
Reconstruction; behavioural evaluation; residual analysis \\

\rowcolor{tablerow1}
\textbf{Stability} &
Are findings robust to variations in factors like random seeds, hyperparameters, paraphrases, etc.? &
Cross-seed agreement; checkpoint consistency; hyperparameter sensitivity &
Multi-seed analysis; checkpoint comparison; input paraphrasing\\

\rowcolor{tablerow2}
\textbf{Causal Validity} &
Are the identified components necessary, sufficient, and specific for the target behaviour? &
Necessity score; sufficiency score; specificity score&
Ablation; component isolation; interchange interventions; tests with control behaviours \\

\rowcolor{tablerow1}
\textbf{Human Interpretability} &
Can humans understand and use the resulting explanations? &
Explanation utility; debugging utility; interpretation agreement &
User studies; expert evaluation\\

\bottomrule
\end{tabular}
\end{table}

Note that throughout this subsection, we use \emph{model components} to refer to any unit of analysis, including individual features, neurons, attention heads, circuits, and connections representing information flow between components. We operationalise the model behaviour of interest, \(B\), with a set of input prompts \(x \sim B\). Following the notation introduced in Section~\ref{sec:circuit-discovery-methods}, we denote the model by \(f_\theta(\cdot)\), a scalar behaviour-relevant performance metric by \(M(\cdot)\), and an intervention on components \(S\) by
\[
f_\theta\!\left(x_{[S \leftarrow \tilde h_S]}\right),
\]
where the activations of \(S\) are replaced by reference activations \(\tilde h_S\) while the remaining computation proceeds normally. Following Section~\ref{sec:circuit-discovery-methods}, we define the intervention effect of \(S\) on an input \(x\) as
\[
s(S,x)
\coloneqq
M\!\left(f_\theta(x)\right)
-
M\!\left(
f_\theta\!\left(x_{[S \leftarrow \tilde h_S]}\right)
\right)
\]
and the average intervention effect on behaviour \(B\) as
\[
s(S,B)
\coloneqq
\mathbb{E}_{x \sim B}\left[s(S,x)\right].
\]

\subsubsection*{Dimension 1: Attribution Faithfulness}

Attribution faithfulness measures whether the interpretability method accurately attributes the model’s behaviour of interest to responsible components.

\begin{definitionbox}[Attribution Faithfulness Metrics]

\paragraph{Intervention-Effect Agreement.}
Let \(a(S,x)\) denote the importance score assigned by an interpretability method to components \(S\) for a sample \(x \sim B\):
\[
\mathrm{EffectAgreement}
=
\rho_{x \sim B}\!\left(
a(S,x),
s(S,x)
\right),
\]
where \(\rho\) may be Pearson or Spearman correlation. High agreement indicates that the attribution consistently reflects the identified components’ causal effects on behaviour \(B\).\\

\paragraph{Effect-Size Error.}
When attribution scores are calculated to estimate intervention effects, we measure the effect-size error as:
\[
\mathrm{EffectError}
=
\mathbb{E}_{x \sim B}
\left[
\left|
a(S,x)-s(S,x)
\right|
\right].
\]
A low value indicates good calibration between estimated and actual intervention effects. If the two quantities are not naturally on the same scale, we may use normalised or rank-based measures instead.\\

\paragraph{Counterfactual Effect Recovery.}
Consider inputs \(x_1 \sim B'\) and \(x_2 \sim B\), where $B'$ is a reference behaviour (e.g., honesty) and $B$ is the behaviour of interest (e.g., sycophancy). Patching \(S\) from \(x_2\) into the computation on \(x_1\) yields
\[
\mathrm{CFRecovery}(S;B,B')
=
\mathbb{E}_{x_1 \sim B', x_2 \sim B}
\frac{
M\!\left(
f_\theta\!\left(
x_{1,[S \leftarrow h_S(x_2)]}
\right)
\right)
-
M\!\left(f_\theta(x_1)\right)
}{
M\!\left(f_\theta(x_2)\right)
-
M\!\left(f_\theta(x_1)\right)
}.
\]
A value near 1 indicates that the patching recovers much of the behaviour, while a value near 0 indicates little effect. This metric has been widely used in the circuit discovery literature~(\citeestalp{eapig, eapgp}; \citeemalp{syed2023attributionpatchingoutperformsautomated}). However, the denominator of this metric can be close to $0$, and $M\!\left(
f_\theta\!\left(
x_{1,[S \leftarrow h_S(x_2)]}
\right)
\right)$ empirically can exceed $M\!\left(f_\theta(x_2)\right)$ or fall below $M\!\left(f_\theta(x_1)\right)$. These may cause this metric to take values much larger than $1$ or below $0$, complicating the analysis. Recent work has proposed an alternative metric that bounds the value between 0 and 1~\citeest{querycircuit}.
\end{definitionbox}

\paragraph{Evaluation protocol:}
\begin{enumerate}
\item Apply the method to acquire attribution scores \(a(S,x)\) of model components and select those of interest (e.g., components with high scores).
\item Measure the corresponding ground-truth intervention effects \(s(S,x)\).
\item Compare them using correlation, error, or rank agreement.
\item Evaluate how much of the behavioural difference between clean and corrupted inputs can be recovered by patching the identified components.
\item Report results across a diverse set of inputs with uncertainty estimates.
\end{enumerate}

\subsubsection*{Dimension 2: Completeness}

Completeness measures how much of the model behaviour the identified components or mechanisms explain. A high completeness score indicates that the identified components/mechanisms capture most of the relevant computation.

\begin{definitionbox}[Completeness Metrics]

\paragraph{Loss Recovery.}
For methods that reconstruct model activations (e.g., SAEs and CLTs), after replacing the target component's activations with the reconstructions, we measure the fraction of the original model performance that is recovered:
\[
\mathrm{LR}
=
\frac{
\mathcal{L}_{\mathrm{reconstructed}}
-
\mathcal{L}_{\mathrm{ablated}}
}{
\mathcal{L}_{\mathrm{original}}
-
\mathcal{L}_{\mathrm{ablated}}
},
\]
where \(\mathcal{L}_{\mathrm{original}}\) is the downstream test loss of the original model, \(\mathcal{L}_{\mathrm{reconstructed}}\) is the loss after replacing the original activations with reconstructions, and \(\mathcal{L}_{\mathrm{ablated}}\) is the baseline loss where activations are replaced with zero ablation, mean ablation, or Gaussian noise matching the layer statistics. A value near $1$ indicates that the reconstruction preserves most of the information. Similar to the vanilla Counterfactual Effect Recovery metric discussed above, this metric can take arbitrary values. A simple solution is to develop a bounded variant, as has been done for Counterfactual Effect Recovery~\citeest{querycircuit}.\\

\paragraph{Behavioural Coverage.}
Across input prompts $x \sim B$, we measure the fraction for which the identified components have a meaningful intervention effect:
\[
\mathrm{Coverage}(S,B)
=
\mathbb{E}_{x \sim B}
\left[
\mathbbm{1}
\left(
|s(S,x)| \geq \tau
\right)
\right],
\]
where \(s(S,x)\) is the intervention effect defined above and \(\tau\) specifies the minimum effect considered meaningful. Higher coverage indicates that the identified components account for the behaviour across a larger fraction of inputs.\\

\paragraph{Unexplained Variance.}
For methods that reconstruct activations of the components, we measure the fraction of unexplained activation variance as:
\[
\mathrm{UV}(S,B)
=
\frac{
\mathbb{E}_{x \sim B}
\left[
\|h(x)-\hat{h}_S(x)\|_2^2
\right]
}{
\mathbb{E}_{x \sim B}
\left[
\|h(x)-\bar{h}_B\|_2^2
\right]
},
\]
where $\bar{h}_B = \mathbb{E}_{x \sim B}[h(x)]$, \(h(x)\) is the original model activation, and \(\hat{h}_S(x)\) is its reconstruction from the identified components \(S\). Lower values indicate that the identified components explain a larger fraction of the variance in the original activations.

\end{definitionbox}

\paragraph{Evaluation protocol:}
\begin{enumerate}
\item Report behavioural coverage using the intervention effects \(s(S,x)\).
\item Report reconstruction-based metrics such as loss recovery and unexplained variance if applicable.
\item Analyse inputs of the behaviour for which the identified components do not provide good coverage or reconstruction.
\end{enumerate}

\subsubsection*{Dimension 3: Stability}
Stability measures the robustness of explanations, attributions, and findings to changes in variables that should not affect the results, such as the choice of random seeds, small hyperparameter changes, and input paraphrasing.

\begin{definitionbox}[Stability Metrics]
\paragraph{Cross-Seed Agreement.} We train or apply the interpretability method $K$ times under different random seeds. Let $\mathcal{F}_i$ denote the set of features or mechanisms, such as SAE features and circuits, discovered under seed $i$. We measure the cross-seed agreement as:
\[
\text{Seed-Stability} = \frac{1}{K(K-1)} \sum_{i \neq j} \text{Align}(\mathcal{F}_i, \mathcal{F}_j),
\]
where $\text{Align}(\cdot)$ measures similarity between sets. For example, \citetest{paulo2025sparse} calculate the Hungarian matching between two sets of SAE features based on their pairwise cosine similarities.\\

\paragraph{Hyperparameter Sensitivity.} We measure the sensitivity of mechanistic analyses or methods under reasonable hyperparameter variations. Let $\Omega$ denote the set of evaluated hyperparameter configurations and $\text{Metric}(\omega)$ denote the experimental result (e.g., a method's performance) under a configuration $\omega \in \Omega$. We calculate the hyperparameter sensitivity as:
\[
\text{HP-Sensitivity}
=
\frac{
\max_{\omega \in \Omega} \text{Metric}(\omega)
-
\min_{\omega \in \Omega} \text{Metric}(\omega)
}{
\mathbb{E}_{\omega \in \Omega}[\text{Metric}(\omega)]
}.
\]
Lower values indicate that the finding is less sensitive to hyperparameter choices and therefore more robust.\\

\textbf{Paraphrase Consistency.} For each input $x \sim B$, let $\mathcal{P}(x) = \{x^{(1)}, \dots, x^{(K)}\}$ denote a set of semantics-preserving paraphrases of $x$. We measure the consistency of explanations or attributions across $\mathcal{P}(x)$. For example, we can measure whether the intervention effect of $S$ remains consistent:
\[
\text{Para-Consistency}(S, B)
= \mathbb{E}_{x \sim B}\left[
 1 - \frac{\mathrm{Var}_{x' \in \mathcal{P}(x)}\left[s(S, x')\right]}
          {\mathbb{E}_{x' \in \mathcal{P}(x)}\left[s(S, x')^2\right]}
\right].
\]
The score lies in $[0, 1]$. A value of $1$ indicates that the intervention effect is identical across all paraphrases. A value of $0$ indicates that, for each $x$, the expected intervention effect, i.e., $\mathbb{E}_{x' \in \mathcal{P}(x)}s(S, x')$, is 0.
\end{definitionbox}

\paragraph{Evaluation protocol:}
\begin{enumerate}
\item Execute the interpretability method with several random seeds and compute alignment of the discovered units or findings across seeds.
\item Analyse which findings are stable across all seeds and which ones are seed-specific. Key claims need to remain consistent under subsequent tests on hyperparameter variation and input paraphrasing, if applicable.
\item Distinguish robust findings from fragile ones and report both of them.
\end{enumerate}

\subsubsection*{Dimension 4: Causal Validity}

Causal validity assesses whether the identified components or mechanisms are causally responsible for the behaviour, particularly necessity (i.e., removing the components/mechanisms disrupts the behaviour) and sufficiency (i.e., the components/mechanisms alone are sufficient to preserve the behaviour).

\begin{definitionbox}[Causal Validity Metrics]

\paragraph{Necessity Score.}
Necessity measures the fraction of the model’s performance on behaviour \(B\) that is lost when \(S\) is ablated (i.e., its activations are replaced with some reference values). It is calculated as:
\[
\mathrm{Necessity}(S,B)
=
1-
\frac{
\mathbb{E}_{x\sim B}
\left[
M\!\left(
f_\theta\!\left(
x_{[S\leftarrow\tilde h_S]}
\right)
\right)
\right]
}{
\mathbb{E}_{x\sim B}
\left[
M\!\left(f_\theta(x)\right)
\right]
}.
\]
A higher value indicates that \(S\) is necessary for the behaviour.\\

\paragraph{Sufficiency Score.}
Let \(\bar S\) denote the components outside \(S\). Sufficiency measures how much of the original behaviour is achieved when only \(S\) is active (i.e., activations of all components outside \(S\) are replaced with some reference values). It is calculated as:
\[
\mathrm{Sufficiency}(S,B)
=
\frac{
\mathbb{E}_{x\sim B}
\left[
M\!\left(
f_\theta\!\left(
x_{[\bar S \leftarrow \tilde h_{\bar S}]}
\right)
\right)
\right]
}{
\mathbb{E}_{x\sim B}
\left[
M\!\left(f_\theta(x)\right)
\right]
}.
\]
A high value indicates that \(S\) alone implements much of the behaviour.\\

\paragraph{Specificity.}
Specificity measures the effect on non-target behaviours by interventions on \(S\). It is calculated as:
\[
\mathrm{Specificity}(S,B)
=
\frac{
|s(S,B)|
}{
\sum_{B' \in \mathcal{B}}
|s(S,B')|
},
\]
where \(\mathcal{B}\) contains the target behaviour together with a set of relevant control behaviours. A high value indicates that the causal effect of \(S\) is specific to \(B\).

\end{definitionbox}

\paragraph{Evaluation protocol:}
\begin{enumerate}
\item \textbf{Necessity test:} Ablate \(S\) and measure the resulting degradation on behaviour \(B\). Compare against ablating random components of matched size.
\item \textbf{Sufficiency test:} Replace components outside \(S\) with reference activations and measure how much of behaviour \(B\) is preserved. Compare against random component sets of matched size.
\item \textbf{Interchange test:} Swap activations of \(S\) between inputs representing different behaviours and test whether the model output follows the intervention.
\item \textbf{Specificity test:} Measure intervention effects on a set of non-target behaviours and examine whether the effect is specific to \(B\).
\end{enumerate}

The MI community has developed several benchmarks that examine the causal validity of interpretability methods. For example, MIB~\citeest{mueller2025mib} provides standardised tasks for assessing discovered circuits and causal variables (e.g., features); AxBench~\citeest{wu2025axbench} evaluates methods’ ability to steer models towards given concepts; RAVEL~\citeest{huang-etal-2024-ravel} investigates methods’ ability to localise features of given concepts by using interchange interventions to validate discovered features; and SAEBench~\citeest{karvonen2025saebench} evaluates SAEs specifically,
pairing reconstruction metrics with downstream intervention tasks such as gender bias erasure.

\subsubsection*{Dimension 5: Human Interpretability}
Human interpretability measures whether generated explanations eventually improve human understanding, prediction, and decision-making about model behaviour.

\begin{definitionbox}[Human Interpretability Metrics]
\paragraph{Explanation Utility.} This metric tests whether humans can predict model behaviour more accurately in the held-out set based on the explanation. It is formulated as:
\[
\text{Expl-Utility}
=
P(\text{correct prediction} \mid \text{with explanation})
-
P(\text{correct prediction} \mid \text{no explanation})
\]

\paragraph{Debugging Utility.} This metric examines whether the explanations help humans identify and repair model errors, such as incorrect outputs or misaligned behaviour, in the wild.
\[
\text{Debug-Utility}
=
P(\text{error fixed}\mid\text{with explanation})
-
P(\text{error fixed}\mid\text{no explanation}).
\]

\paragraph{Consistency of Interpretation.} This metric inspects whether different humans interpret the same explanation consistently.
\[
\text{Inter-Rater} = \kappa(\text{human}_1 \text{ labels}, \text{human}_2 \text{ labels})
\]
where $\kappa(\cdot)$ is Cohen’s kappa or a similar agreement metric.
\end{definitionbox}

\paragraph{Evaluation protocol:}
\begin{enumerate}
\item \textbf{Prediction task:} Show humans explanations of model components or mechanisms and ask them to predict model behaviour on held-out inputs. Compare accuracy to a no-explanation baseline.
\item \textbf{Identification task:} Show humans model outputs with errors and ask them to identify the cause, with and without explanation assistance. Measure accuracy and required time to complete the task.
\item \textbf{Intervention task:} Ask humans to propose fixes for model errors based on explanations. Measure the success rate of proposed interventions.
\item \textbf{Agreement task:} Ask multiple human raters to generate explanations, if applicable (e.g., write a summary of highly activated samples of an SAE feature as its explanation). Measure inter-rater reliability and agreement with machine-generated explanations.
\end{enumerate}

Several caveats and challenges arise when evaluating the human interpretability of tested methods. For example, human studies are usually expensive. Moreover, human expertise may confound results, where some explanations understandable to MI researchers may be opaque to domain experts, and vice versa.

Note that different interpretability methods may emphasise distinct evaluation dimensions and operationalise them differently. Table~\ref{tab:method-evaluation} provides guidance on applying this framework to some common interpretability approaches.

\begin{table}[tb]
\centering
\caption{Recommended evaluation priorities by interpretability method. The entries indicate evaluation aspects particularly important for each method, not mutually exclusive metrics.}
\label{tab:method-evaluation}
\renewcommand{\arraystretch}{1.35}
\begin{tabular}{>{\raggedright\arraybackslash}p{2.7cm}
>{\raggedright\arraybackslash}p{2.25cm}
>{\raggedright\arraybackslash}p{2.25cm}
>{\raggedright\arraybackslash}p{2.25cm}
>{\raggedright\arraybackslash}p{2.25cm}
>{\raggedright\arraybackslash}p{2.25cm}}
\toprule
\rowcolor{tableheader!80}
\textcolor{white}{\textbf{Method}} &
\textcolor{white}{\textbf{Attribution Faithfulness}} &
\textcolor{white}{\textbf{Completeness}} &
\textcolor{white}{\textbf{Stability}} &
\textcolor{white}{\textbf{Causal Validity}} &
\textcolor{white}{\textbf{Human Interpretability}} \\
\midrule

\rowcolor{tablerow1}
SAEs and variants &
Feature attribution--intervention agreement; effect-size error &
Activation reconstruction; loss recovery; unexplained variance &
Cross-seed feature alignment &
Feature necessity, sufficiency, and specificity &
Feature description consistency across annotations, including models and humans \\

\rowcolor{tablerow2}
Circuit Discovery &
Edge or component attribution--intervention agreement; counterfactual recovery &
Behavioural coverage; loss or behaviour recovery &
Robustness across hyperparameters, seeds, and paraphrases &
Circuit necessity, sufficiency, specificity; interchange tests &
Circuit-level descriptions; trade-off between sparsity and variance explained \\

\rowcolor{tablerow1}
Probing &
Agreement between probe-derived importance and intervention effects &
Coverage across target concepts or behaviours under different distributions &
Robustness across seeds and sampled fitting sets &
Necessity, specificity, and intervention tests on probe directions &
Explanation of shared information in fitting set and search for confounding factors \\

\rowcolor{tablerow2}
Steering Vectors &
Agreement between predicted and observed intervention effects &
Coverage across behaviours deemed relevant &
Stability across steering magnitudes, seeds, and paraphrases &
Intervention efficacy and specificity; side-effect analysis &
Explanation of shared information in fitting set and search for confounding factors \\

\bottomrule
\end{tabular}
\end{table}

\subsubsection{Integrated Evaluation Protocol}

We recommend the following integrated protocol for comprehensive evaluation:

\begin{practicebox}[Recommended Evaluation Protocol]
\paragraph{Step 1: Define scope.} Users first specify the model, layer(s), and behaviour(s) of interest and define a behavioural test suite with clear success criteria.\\

\paragraph{Step 2: Establish baselines.} Users specify baseline methods, such as random directions in activation space, PCA components, individual neurons, and supervised probes (if labels are available).\\

\paragraph{Step 3: Apply interpretability method.} Users apply the method of interest to analyse models with variation in random seeds, hyperparameters, and other training details.\\

\paragraph{Step 4: Evaluate across all five dimensions.} Users compute metrics for attribution faithfulness, completeness, stability, causal validity, and human interpretability.\\

\paragraph{Step 5: Report comprehensively.} Users report all five dimensions instead of just favourable metrics and discuss the method’s failure cases, unexplained behaviours, and comparisons to baselines.

\end{practicebox}

\subsubsection{Limitations of This Framework}

This evaluation framework, though comprehensive, still has several limitations:

\begin{itemize}
\item \textbf{Metric choice:} The proposed specific metrics are not uniquely correct. Alternative formalisations may be appropriate for specific contexts.
\item \textbf{Threshold arbitrariness:} Thresholds (e.g., the accuracy that counts as success) may lack theoretical grounding and will need empirical calibration.
\item \textbf{Computational cost:} Evaluating all five dimensions comprehensively is expensive and potentially prohibitive for large-scale studies.
\item \textbf{Human evaluation bottleneck:} Dimension 5 requires time-consuming, expensive, and subjective human studies.
\item \textbf{Task dependence:} A method’s performance under this framework can depend on the evaluation task. For example, a method that excels at factual recall may perform poorly at reasoning.
\end{itemize}

Despite these limitations, systematic evaluation across multiple dimensions provides more reliable evidence than an arbitrary choice of metrics. Readers should view the framework as a starting point for standardisation, not a final answer.

\paragraph{Summary.}
This section introduced circuit discovery methods, which trace how model components or features interact to produce behaviour, either directly on the model’s computational graph (e.g., ACDC and EAP) or within interpretable replacement models (e.g., Circuit Tracer). We also discussed how to causally verify discovered circuits and introduced a broader evaluation framework for mechanistic interpretability methods based on faithfulness, completeness, stability, causal validity, and human interpretability. In summary, Section~\ref{sec:sae} introduced methods for extracting interpretable units of analysis, while this section examined how those units interact through circuits to produce behaviour. The next section takes a complementary perspective: directly interpreting the information in individual representations using natural language.

\section{Understanding Intermediate Representations}
\label{sec:intermediate-computation}

This section explores interpreting hidden activations directly in natural language. In contrast to feature extraction and circuit discovery (Sections~\ref{sec:sae} and \ref{sec:circuit}), which extract features or their interactions from model representations (activations), the methods we will discuss in this section treat the activation itself as the object of interpretation. They use language models or other natural-language-based methods to describe or elicit the encoded information, without intervening on or decomposing the underlying activation. This paradigm provides a direct interface between model activations and natural-language explanations, making it particularly accessible and understandable for human users.

\subsection{Logit Lens: Tracing Predictions Through Depth}
\label{sec:logit-lens}

\begin{figure}[tb]
\centering
\includegraphics[width=0.5\linewidth]{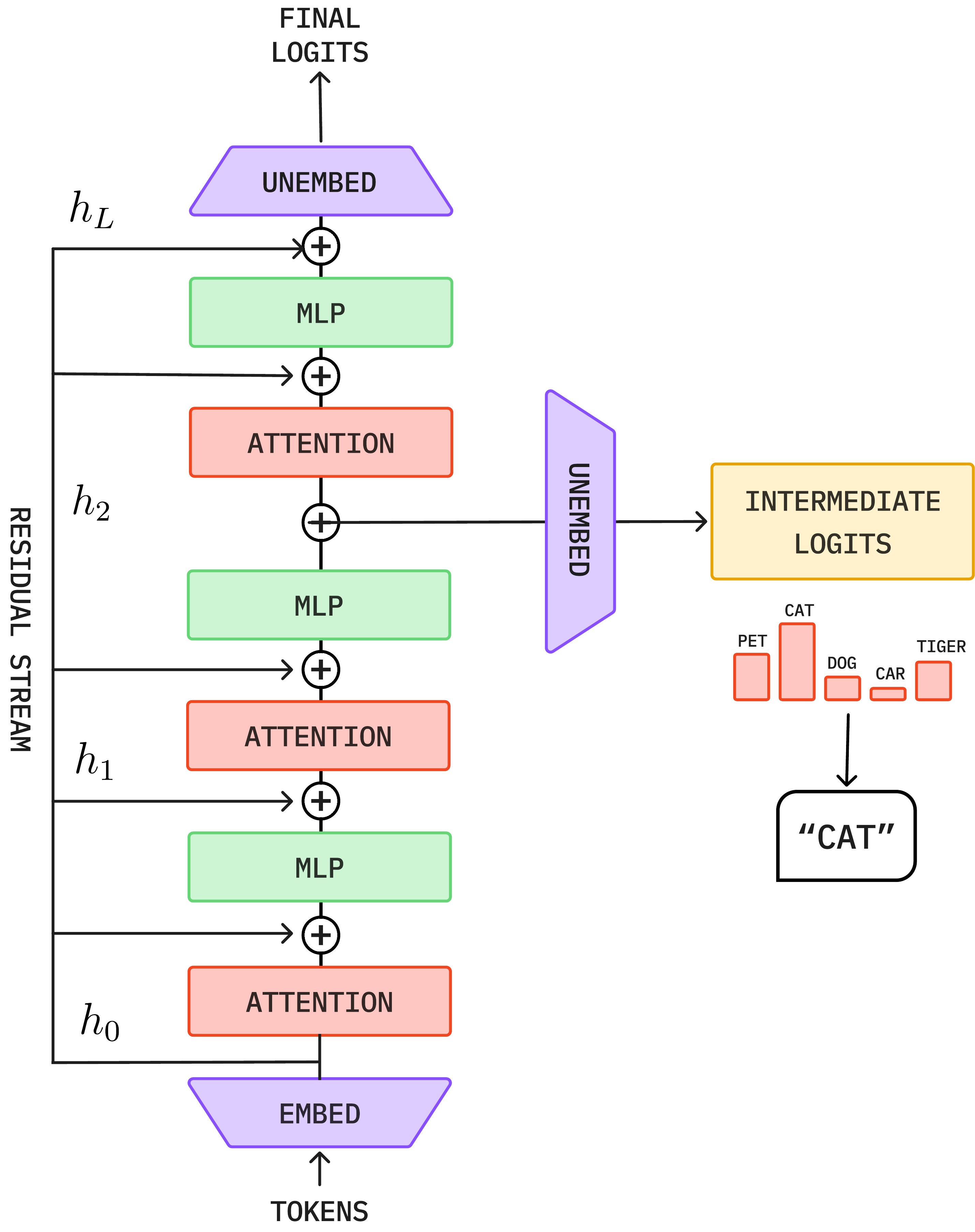}
\caption{\textbf{Logit Lens.} Logit lens provides a window into a model's intermediate state by applying the final unembedding layer directly to the residual stream at earlier layers ($h_1, h_2, \ldots$). This allows users to see which tokens the model is ``considering'' mid-computation, such as the emergence of the concept ``CAT'' before the final output layer is reached.}
\label{fig:logitlens}
\end{figure}

\begin{definitionbox}[Logit Lens (Figure~\ref{fig:logitlens})]
Logit lens~\citeem{nostalgebraist2020interpretinggpt} applies the model's unembedding matrix to activations at intermediate layers, revealing what the model \emph{would predict} if computation stopped at that layer. For a hidden activation $h_\ell$ at layer $\ell$, logit lens computes:
\[
\text{logits}_\ell = W_U h_\ell,
\]
where $W_U$ represents the model's unembedding matrix. Applying it across layers reveals the prediction trajectory, showing when and where the model commits to specific outputs~\citeem{nostalgebraist2020interpretinggpt}.
\end{definitionbox}

Logit lens (Figure~\ref{fig:logitlens}) provides intuitive insight into the model's decision-making process across depth. Experimental results show that early layers tend to produce diffuse predictions, while later layers progressively narrow them towards the final output, though exceptions exist. For example, in a twelve-layer language model processing the sequence \texttt{A B C A B}, the logit lens may reveal that the token \texttt{C} enters the top-20 candidates at layer 6 and is progressively amplified, eventually becoming the top-ranked token at the final layer. Conversely, for induction tasks in complex linguistic patterns, early and middle layers may show a flat probability distribution across many tokens, with the model converging to a specific prediction only in the last few layers.

\paragraph{Limitations of Vanilla Logit Lens.}
The vanilla logit lens has a critical limitation: applying the unembedding matrix to intermediate activations assumes that the hidden state is in a format that the output layer expects. However, intermediate representations may be organised very differently from output-layer activations. Applying the unembedding matrix to layer-6 activations may extract relevant information but cannot precisely reflect the model's underlying computation at that layer~\citeem{nostalgebraist2020interpretinggpt}.
Furthermore, logit lens is fundamentally correlational rather than causal. It projects the activation into the output vocabulary space without testing the roles decoded tokens play in model computation. In particular, because of the representational mismatch discussed above, decoded tokens represent at most information recoverable from the activation, rather than information used in model computation.

\subsection{Tuned Lens: Learning Per-Layer Transforms}
\label{subsec:tuned-lens}

\begin{definitionbox}[Tuned Lens]
Tuned lens~\citeem{belrose2025elicitinglatentpredictionstransformers} addresses logit lens' concern of representational mismatch by learning per-layer affine transformations that map intermediate activations to the output layer's activation space:
\[
\text{logits}_\ell^{\text{TL}} = W_U (W_\ell h_\ell + b_\ell),
\]
where $W_\ell$ and $b_\ell$ are learned affine transformations for layer $\ell$. These transforms are trained to minimise the divergence between the tuned lens' predicted distribution $p_\ell^{\text{TL}} = \operatorname{softmax}(\text{logits}_\ell^{\text{TL}})$ and the model output distribution $p_{\text{final}}$ via:
\[
\min_{W_\ell,b_\ell}\mathbb{E}_{x\sim\mathcal{D}}\big[\,\mathrm{KL}(p_{\text{final}}(x)\;\|\;p_\ell^{\text{TL}}(x))\big].
\]
\end{definitionbox}
As a result, the unembedding matrix can extract relevant information from each layer's activations more effectively.

\paragraph{Limitations of Tuned Lens.}
Although tuned lens mitigates representational mismatch, it still inherits several fundamental limitations. First, the decoding paradigm remains inherently correlational rather than causal; decodable information does not imply a causal role in the model's prediction. Second, similar to logit lens, tuned lens is primarily designed to answer \emph{what the model is likely to predict} rather than \emph{what the model is internally computing}. The learned layer-specific transformations that align intermediate representations with the final prediction space may obscure intermediate computational steps and produce predictions that resemble the model's final output. Empirical evidence also shows that it skips intermediate computation more aggressively than logit lens~\citeem{jlens}.

\subsection{Jacobian Lens: Decoding Computational Dispositions}
\begin{definitionbox}[Jacobian Lens]
Jacobian lens (J-lens; \citeemalp{jlens}) adopts a decoding paradigm distinct from logit lens and tuned lens. It uses the Jacobian to characterise how an intermediate representation influences certain model outputs.\\

Let $h_\ell^t$ denote the residual stream at layer $\ell$ and position $t$. A perturbation to $h_\ell^t$ propagates to the final-layer residual stream at every position $t' \geq t$; to first order, this influence is captured by the Jacobian
\[
\frac{\partial h_{\mathrm{final}}^{t'}}{\partial h_\ell^t}.
\]
The lens matrix is defined as the expectation of this Jacobian over prompts and positions:
\[
J_\ell
=
\mathbb{E}_{\,\mathrm{prompt},\,t,\,t' \ge t}
\left[
\frac{\partial h_{\mathrm{final}}^{t'}}
{\partial h_\ell^t}
\right].
\]
J-lens then decodes intermediate representations by
\[
p_\ell^{\mathrm{JL}}
=
\operatorname{softmax}
\!\big(
W_U \operatorname{norm}(J_\ell h_\ell)
\big).
\]

\citetem{jlens} further defined layer-specific concept (token) representations using the rows of $W_U J_\ell$, called \emph{J-lens vectors}. The inner product between a J-lens vector and $h_\ell$ measures how strongly the current activation will promote the corresponding token in model outputs.
\end{definitionbox}

Unlike logit lens and tuned lens, which primarily estimate which token the model is likely to predict next, J-lens characterises how current representations influence model outputs. It aims to answer \emph{what the model is internally computing} rather than merely \emph{what the model is going to predict}. J-lens has been found useful for analysing tasks that require multi-step reasoning, latent planning, or other forms of extended computation~\citeem{jlens}.

Furthermore, the defined J-lens vectors are a useful medium for intervention. They can serve as steering vectors to shift model behaviour towards or away from the corresponding concepts. In contrast, interventional utility is largely absent from logit lens and tuned lens, which are primarily decoding tools.

\paragraph{Limitations of Jacobian Lens.}
J-lens still faces several limitations that warrant further investigation. First, like logit lens and tuned lens, it operates primarily at the token level. However, many semantically meaningful concepts are represented by multiple tokens. For example, the word ``photoperiodism'' may be split into several subword tokens. If only one token (e.g., ``photo'') is visible in J-lens, we cannot tell whether it corresponds to the concept of photoperiodism or to other words. Second, J-lens is found to work best within a certain range of layers, called \emph{J-space}~\citeem{jlens}. Specifically, decoded concepts are most informative in intermediate layers (e.g., approximately 38--92\% of the depth in Claude Sonnet 4.5). In early layers, decoded concepts are noisy and hard to interpret. In later layers, they collapse towards the model's output and are also uninformative. Some recent studies have attempted to address these limitations~\citeem{jlens, blank2026rlens}.

\subsection{Patchscopes: Leveraging the Language Model Itself to Explain Intermediate Representations}
\label{subsec:patchscopes}

\begin{figure}[tb]
\centering
\includegraphics[width=\linewidth]{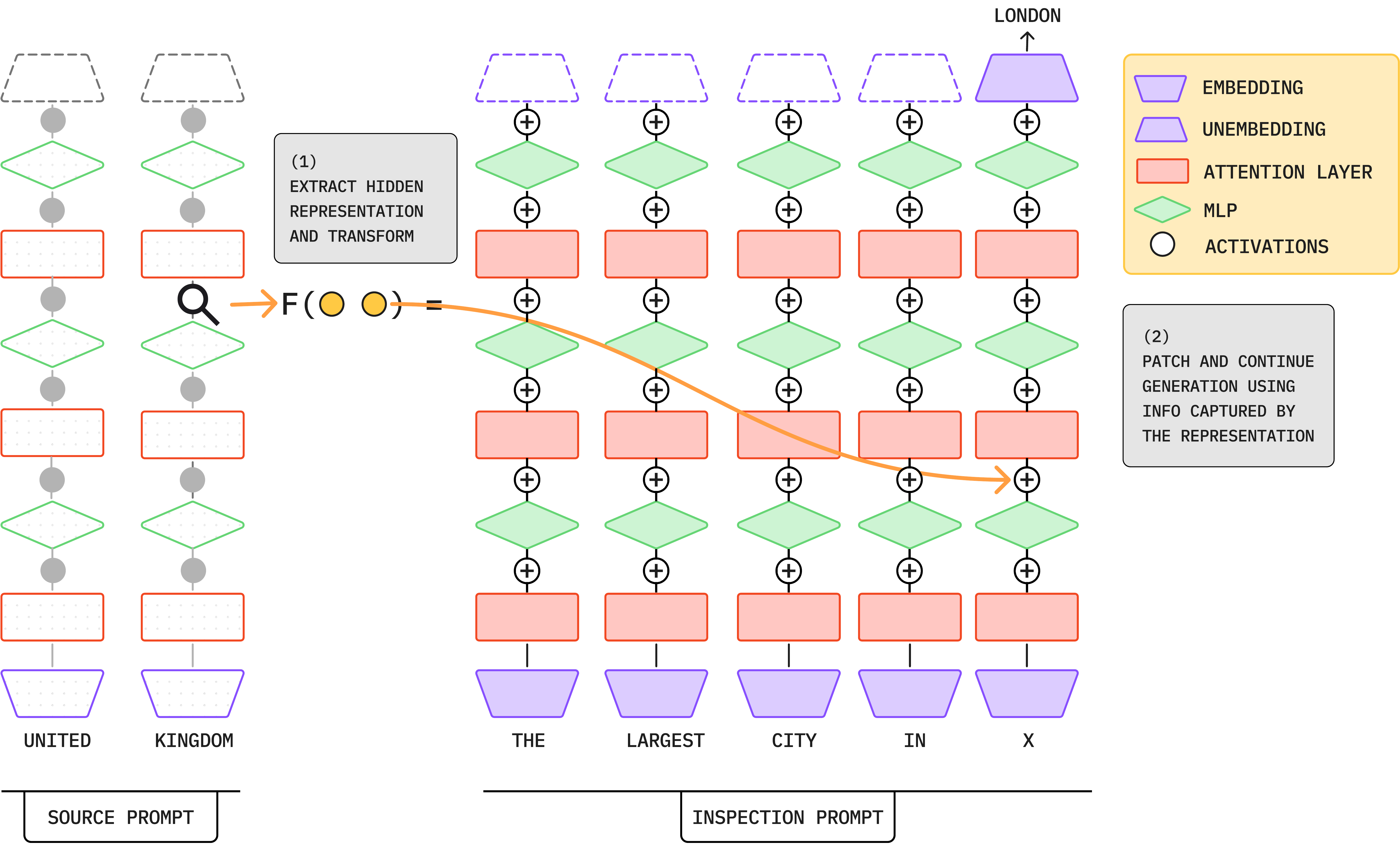}
\caption{\textbf{Patchscopes.} Patchscopes extracts the model representation of interest from a source prompt and ``patches'' it into an inspection prompt. By observing changes in model output, users can examine what information the representation contains. For example, in this figure we can observe whether the representation of ``United Kingdom'' correctly triggers ``London'' as the largest city.}
\label{fig:patchscopes}
\end{figure}

\begin{definitionbox}[Patchscopes (Figure~\ref{fig:patchscopes})]
Patchscopes~\citeest{patchscopes} leverages the expressive capacity of language models to translate their internal representations into natural-language explanations. Let $h^t_\ell$ denote a hidden representation at layer $\ell$ and token position $t$. Patchscopes produces a textual explanation of $h^t_\ell$, denoted as $e(h^t_\ell)$, by
\[
e(h^t_\ell) = f_\theta(p_{[h^{t^*}_{\ell^*} \leftarrow h^t_\ell]}),
\]
where $\ell^*$ and $t^*$ specify the layer and token position to which the representation is patched, and $p$ is an \emph{elicitation prompt} designed to encourage the language model $f_\theta(\cdot)$ to decode the patched representation into text expressing the contained information.\\

For example, consider the prompt illustrated in Figure 1 of \citetest{patchscopes}:
\[
p = \text{``cat$\rightarrow$cat; 135$\rightarrow$135; hello$\rightarrow$hello; ?''},
\]
where $t^*$ corresponds to the position of the token ``?''. The notation $[h^{t^*}_{\ell^*} \leftarrow h^t_\ell]$ indicates that, during the forward pass of $f_\theta(\cdot)$ on $p$, the hidden representation at layer $\ell^*$ and position $t^*$ is replaced with $h^t_\ell$. As a result, the model can reveal information encoded in $h^t_\ell$ through its prediction for the next token at the ``?'' position.
\end{definitionbox}

Because the mapping from intermediate representations to natural language may be too complex to be captured by an affine transformation, Patchscopes (Figure~\ref{fig:patchscopes}) extends the paradigm of logit lens and tuned lens by \emph{delegating the representation-to-text transformation to the language model itself}\footnote{A concurrent work, SelfIE~\citeest{selfie}, explores a similar direction. While we focus on Patchscopes as a reference framework, both methods share closely related motivations and methodologies.}. In this view, the language model acts as an \emph{activation decoder}. Crucially, Patchscopes relies on the model's learned, generalisable text-generation capabilities to map internal activations to natural language.

\paragraph{Limitations of Patchscopes.}
A key limitation of Patchscopes is that injecting an external representation into an elicitation prompt can shift the model's internal activations into an out-of-distribution (OOD) regime. This is expected, as the decoder is a standard pretrained language model not explicitly trained for representation-to-text decoding. Although the model may still surface salient information from the patched representation, the decoding process can degrade, and critical aspects of the original representation may be distorted or lost.

\subsection{LatentQA: Fine-Tuning the Language Model as an Activation Verbaliser}
\label{subsec:latentqa}

\begin{figure}[tb]
\centering
\includegraphics[width=0.8\linewidth]{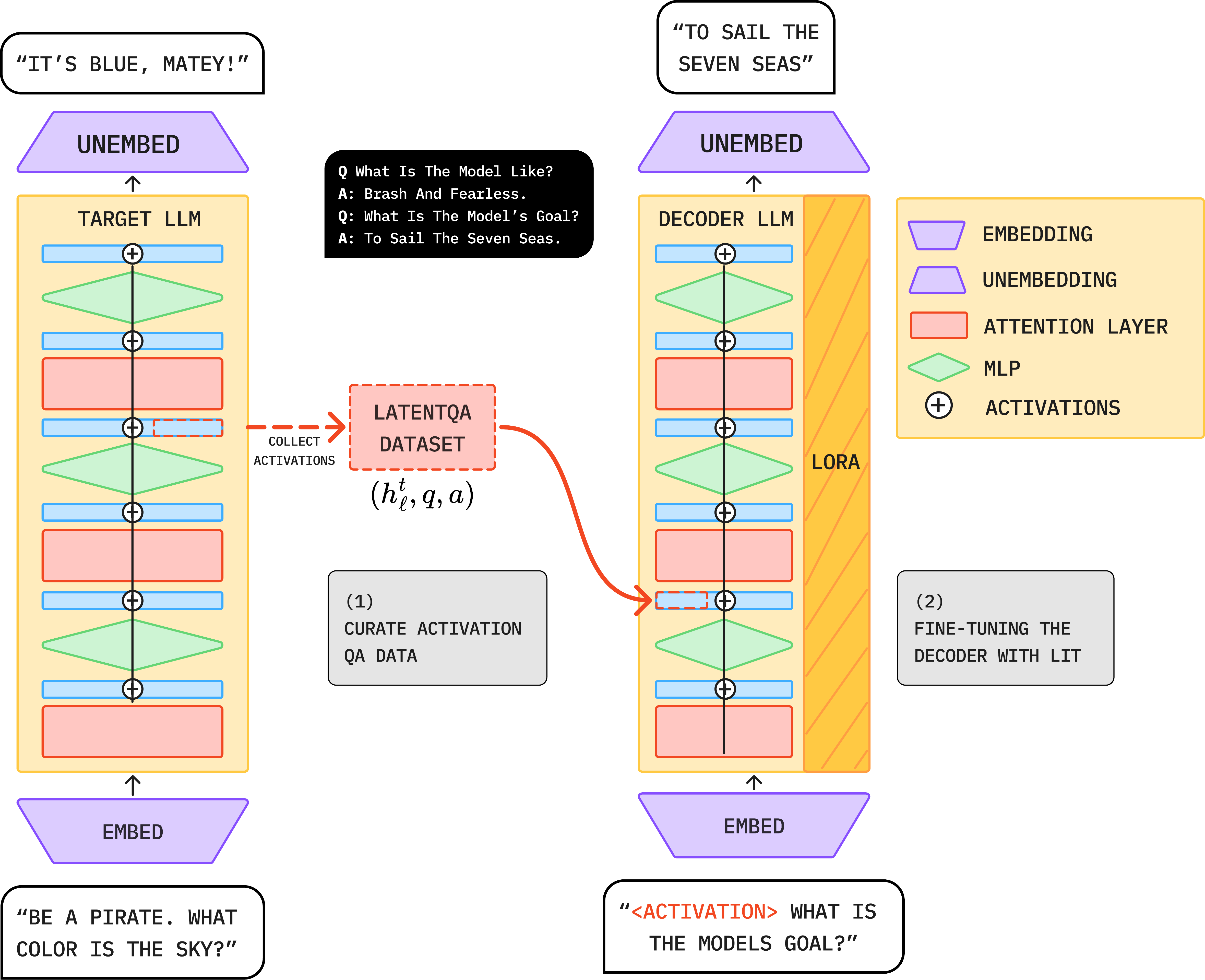}
\caption{\textbf{LatentQA.} LatentQA probes the semantic content of hidden representations by training a separate verbaliser model to answer natural-language questions about them. The process involves (1) curating a dataset of activation--question--answer triplets from a target LLM, and (2) fine-tuning a verbaliser to retrieve specific information latent within those activations, such as the model's underlying goals or persona.}
\label{fig:latentqa}
\end{figure}

\begin{definitionbox}[LatentQA (Figure~\ref{fig:latentqa})]
LatentQA~\citeest{pan2024latentqateachingllmsdecode} frames representation decoding as a supervised learning problem via Latent Interpretation Tuning (LIT), fine-tuning an activation verbaliser to perform representation-to-text verbalisation. The core step of LIT is to construct a fine-tuning dataset $D$ consisting of samples $(h^t_\ell, q, a)$, where $h^t_\ell$ is a cached hidden representation from a target model, and $(q, a)$ is a question--answer pair that queries the information encoded in $h^t_\ell$.\\

Specifically, $h^t_\ell$ denotes the activation at layer $\ell$ and token position $t$ when the target model processes a ``control prompt'' designed to embed a specific attribute or concept into the representation. An example control prompt from Figure 13 of \citetest{pan2024latentqateachingllmsdecode} is:
\[
\text{``In your next reply, please try to evoke a brief sense of triumphant hubris.''}
\]
If $h^t_\ell$ is defined as the activation at layer $\ell$ corresponding to the final token (``.'') of this prompt, the resulting representation is expected to encode the instruction to respond with triumphant hubris. In this case, a corresponding question--answer pair may be:
\[
q = \text{``In what way does the assistant craft its responses to the user?''}
\]
\[
a = \text{``The assistant crafts responses that briefly highlight a sense of self-assured victory.''}
\]
Both the control prompt and the $(q, a)$ pair are generated by prompting an off-the-shelf proprietary language model~\citeest{pan2024latentqateachingllmsdecode}.
\end{definitionbox}

LatentQA (Figure~\ref{fig:latentqa}) addresses the OOD issue inherent in Patchscopes-style methods by explicitly fine-tuning the activation verbaliser on the representation-to-text decoding task. Through LIT, the verbaliser learns to ingest and process raw activation vectors directly. Ideally, after being exposed to a sufficiently diverse collection of question--answer pairs over many different representations, the decoder learns \emph{which aspects of the activation to attend to} and \emph{how to verbalise them} conditioned on the query.

\paragraph{Limitations of LatentQA.}
LatentQA faces several technical challenges, including the generalisability of the learned representation-to-text mapping. When the fine-tuning dataset $D$ is too small or insufficiently diverse, the verbaliser may memorise superficial correspondences between representations and explanations rather than learning to extract information from representations. Recent approaches, such as Predictive Concept Decoders (PCDs)~\citeem{pcd} and Activation Oracles (AOs)~\citeest{ao}, mitigate this issue by constructing more diverse training datasets.

Another technical challenge is evaluating whether activation-verbalisation models genuinely explain the represented information rather than exploit shortcuts. For example, a verbaliser may implicitly reconstruct the original input to the target model and infer an explanation from it. Recent studies have found that such failure modes are severe and prevalent in existing verbalisers~(\citeestalp{li2026do}; \citeemalp{jakkli2026current}). As a result, strong verbalisation performance alone is insufficient to show that the verbaliser faithfully analysed the internal representations.

\subsection{Probing Classifiers: Capabilities, Critiques, and Causal Limitations}
\label{subsec:probing}

Probing classifiers (Figure~\ref{fig:probing}) are among the most widely used tools for analysing neural network representations. A probe is a classifier trained to predict a property of the input, such as part of speech, sentiment, or entity type, from model activations. High probe accuracy means that the property is decodable from the activations. However, information decodability is a relatively coarse-grained, correlational interpretability claim compared with causal claims about information in model computation. This section reviews probing methodology, its major critiques, and its appropriate role in interpretability research.

As a side note, discovered probe directions can also be used as supervised feature extractors. We discuss this perspective in Section~\ref{subsubsec: supervised_extraction}, including comparisons with SAEs, and in Section~\ref{subsec:steering}, where we discuss their use as steering vectors.

\subsubsection{Standard Probing Methodology}

\begin{figure}[tb]
\centering
\includegraphics[width=0.6\linewidth]{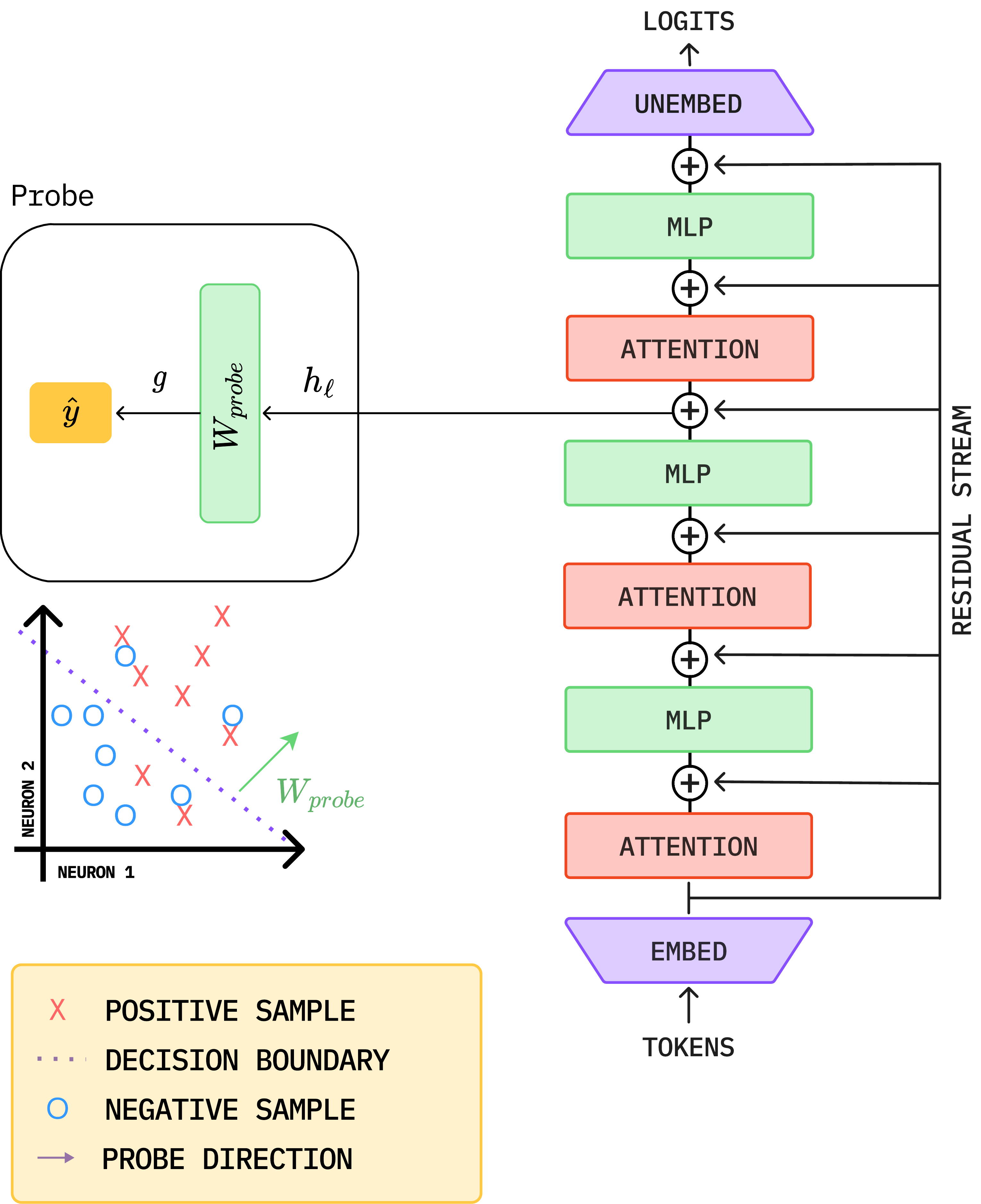}
\caption{\textbf{Probing Classifier.} A probing classifier (or \emph{probe}) tests the decodability of a concept from the target model's internal representations. It is trained to distinguish activations associated with positive and negative examples of the concept. High probe accuracy indicates that the concept can be reliably decoded from the representation.}
\label{fig:probing}
\end{figure}

\begin{definitionbox}[Probing Classifier (Figure~\ref{fig:probing})]
A \emph{probing classifier} is a supervised model trained to predict a target property \(y\) from neural network activations \(h_\ell\) at layer \(\ell\). A \emph{linear probe} follows the form:
\[
\hat{y} = g(W_{\text{probe}} h_\ell + b_{\text{probe}}),
\]
where \(W_{\text{probe}}\) and \(b_{\text{probe}}\) are learned parameters. \(g\), typically an identity function for regression and softmax for classification, maps the probe's raw output to the prediction space. The probe is trained by minimising prediction error on labelled examples \( (h_\ell^{(i)}, y^{(i)}) \). The linear probe tests whether the target property is linearly accessible from the representation. \emph{MLP probes} introduce one or more nonlinear hidden layers instead of $W_{\text{probe}}$ to test whether the property can be retrieved through more expressive nonlinear mappings.
\end{definitionbox}

Probing provides evidence about whether activations contain information about a property. High probe accuracy indicates the information is present; low accuracy suggests it is absent or the probe is not expressive enough. Note that probing alone does not show whether the model actually depends on that information in its decision-making process. A representation can encode information that downstream computation never accesses, such as ``leftover'' information from earlier processing.

\begin{warningbox}[The Probe Causality Problem]
High probe accuracy does not imply causal relevance~\citeest{belinkov-2022-probing, hewitt-liang-2019-designing}. The probe can work by detecting features spuriously correlated with the target property (e.g., word frequency correlates with part of speech) rather than detecting the target property itself. Furthermore, subsequent layers may never use the decoded property to influence the model output. Finally, a sufficiently expressive probe can achieve high accuracy by exploiting noise even when the representations contain no such information.
\end{warningbox}

In summary, using probing alone generally does not provide actionable mechanistic understanding and therefore does not, under our definition, constitute mechanistic interpretability research. However, subsequent work has proposed refined probing paradigms that explicitly incorporate causal considerations into probe design and evaluation, enabling more actionable mechanistic understanding~\citeest{elazar-etal-2021-amnesic}; see Section~\ref{subsubsec: amnesic probing} for details. Recent interpretability studies have similarly used causal interventions to validate probe-discovered concept directions, building a more complete mechanistic explanation~\citeest{marks2024the, li2023emergent}.

\subsubsection{Control Tasks and Selectivity}
High probe accuracy may reflect classification capacity rather than the target property's decodability. \citetest{hewitt-liang-2019-designing} addressed this concern by introducing \emph{control tasks} as a baseline for probe performance.

\begin{definitionbox}[Control Task Methodology]
A \emph{control task} assigns random labels to inputs to break any genuine relationship between activations and the target property. A probe trained on the control task represents the \emph{baseline extractability} of random patterns from the representation. Then, the \emph{selectivity} of a probe is defined as:
\[
\text{Selectivity} = \text{Acc}_{\text{linguistic}} - \text{Acc}_{\text{control}},
\]
where $\text{Acc}_{\text{linguistic}}$ is probe accuracy on the target task and $\text{Acc}_{\text{control}}$ is accuracy on the control task. High selectivity indicates that the probe succeeds because the representation genuinely encodes the target property. In contrast, low selectivity suggests we cannot tell whether the probe has exploited spurious features rather than task-relevant structure.
\end{definitionbox}

\subsubsection{Concerns of Probe Complexity}
Probe complexity brings confounding factors to the analysis. If a property can only be extracted with a highly nonlinear probe, it is hard to tell whether the property is ``deeply encoded'' or whether the probe is simply using nonlinearity to exploit signals, including noise. A recent finding in causal abstraction~\citeest{sutter2025the} also showed that, when the mapping function is complex enough, it can map an arbitrary (even randomly initialised) neural network to a causal model of interest, producing false-positive findings.

\begin{insightbox}[The Complexity--Interpretability Trade-off]
\paragraph{Observation.} In general, more complex probes achieve higher accuracy but provide less interpretable deliverables.\\

\paragraph{Interpretation problem.} Does high accuracy with a complex probe indicate (a) the property is encoded but in a complex, nonlinear form, or (b) the probe itself is computing the property from spurious features? If (b), the probe at most demonstrates that the property is \emph{computable from} the representation, not that it is \emph{encoded in} the representation in any meaningful sense. A worse case in (b) is that a highly nonlinear probe may learn to construct the property of interest from random noise rather than any features~\citeest{sutter2025the}. In this case, the computable illusion likely stems from limited test-set size.
\end{insightbox}

\paragraph{Minimum Description Length Probes.}
To address this, \citetest{voita-titov-2020-information} proposed evaluating probes via the minimum description length (MDL) principle: prefer probes that achieve good accuracy with minimal complexity. The proposed MDL-based evaluation offers an information-theoretic measure of how efficiently a probe can extract a target property from model representations, accounting for both probe accuracy and complexity. A property is considered ``more encoded'' if it can be extracted with a simpler probe. This paradigm shifts the question from ``can we extract it?'' to ``how easily can we extract it?''

\paragraph{Practical Guidance.}
Users should prefer simple probes when possible, compare against probes trained on random representations to establish baselines, and consider MDL-based evaluation to analyse the trade-off between accuracy and model complexity. If linear probes fail but MLP probes succeed, cautiously validate the result and interpretability claim.

\subsubsection{Amnesic Probing and Causal Probing}
\label{subsubsec: amnesic probing}
Beyond using probes to validate information decodability, researchers have also formalised the idea of intervening on representations via probes. \citetem{giulianelli-etal-2018-hood} first showed that a trained probe can steer model behaviour, and \citetest{ravfogel-etal-2020-null} extended this with Iterative Nullspace Projection (INLP), an information-removal method that repeatedly fits a probe to identify linearly encoded information and projects it out. \citetest{elazar-etal-2021-amnesic} further measured changes in downstream model behaviour after applying INLP as a validation of the probe's causality, called \emph{amnesic probing}. These paradigms add causality to probes, pushing them from correlational understanding to mechanistic analysis. Below, we detail amnesic probing and probing with activation patching.

\paragraph{Amnesic Probing.}
\begin{definitionbox}[Amnesic Probing]
Given a model activation $h$ and a probe trained to predict a property $P$, \emph{amnesic probing} tests whether information about $P$ is causally relevant
to downstream model behaviour by removing information about $P$ from the
representation and evaluating the effects on model outputs. Assume $P$ is a binary property (e.g., positive vs. negative sentiment) and that $w \coloneqq W_{\text{probe}}^\top \in \mathbb{R}^{d}$ denotes the probe's direction. Amnesic probing projects $h$ onto the subspace orthogonal to a probe
direction $w$ (i.e., the null space of the probe):
\[
\tilde{h}
=
h
-
\frac{h^\top w}{\|w\|^2}w.
\]
Amnesic probing then measures the change in model performance on downstream tasks with the modified representation $\tilde{h}$. A substantial degradation indicates that the removed information causally contributes to the evaluated behaviour. Notably, unchanged performance does not establish that the information is unused since redundant representations or alternative computational pathways may preserve the behaviour.
\end{definitionbox}

Amnesic probing assumes the target property is decodable and that there exist effective erasure procedures to surgically remove its associated information from the representation. However, information and concept erasure may be imperfect in practice, either leaving residual information or affecting other concepts. Such imperfections confound interpretations of downstream behavioural changes and the causal claims that follow.

\paragraph{Causal Probing via Activation Patching.}
Activation patching tests the causality of a component's representations by swapping
them between inputs that differ in the target property. This technique has also been used to test the causality of probes~\citeest{huang-etal-2024-ravel, arora-etal-2024-causalgym}. Suppose $x_1$ and $x_2$ are two inputs, where $x_1$ has the target property $P$ and $x_2$ does not. $W_{\text{probe}}$ is the fitted probing matrix whose rows span a probing subspace (or a direction if the matrix has only one row, such as that of a binary classification probe) where $P$ resides. After choosing the intervention location (e.g., the last-token residual stream at a middle layer), the method first feeds $x_1$ to the model and extracts the component of the representation in the probing subspace. Then, it feeds $x_2$ to the model while patching the corresponding component of the model representation with the previously acquired one. If the model output for $x_2$ shifts towards $P$, the patched subspace representation contains information about $P$ that the model uses.

Compared with amnesic probing, which tests the effect of removing information about $P$, activation patching examines whether the probe-identified information is sufficient to implement the target behaviour. However, this requires specifying what representation (i.e., which layer, component, and token position) to probe and patch, which may confound the analysis.

\subsubsection{Implications of Attribution Faithfulness for Probing}
Probing is often used along with attribution methods (e.g., saliency maps, attention visualisation, gradient-based attribution) to understand model behaviour. However, attribution methods also have technical challenges. For example, gradient-based saliency maps can highlight input features the model is actually insensitive to~\citeest{adebayo2020sanitycheckssaliencymaps}, attention weights correlate weakly with causal importance~\citeest{jain2019attention}, and different attribution methods can disagree substantially on the same inputs~\citeest{krishna2025disagreementproblemexplainablemachine}. These findings suggest that attribution methods alone cannot determine what information a model actually uses.

This unreliability affects how we can interpret probing results. Probes trained on attention patterns may not reveal causally important information since attention itself may not be causal. Similarly, using gradients to identify which features or components to probe can bias towards high-magnitude gradients that lack causal importance. In sum, interpretability claims made by probes, even paired with these auxiliary tools, still need to go through causal validation to separate \emph{plausible explanations} (do attributions seem reasonable to humans?) from \emph{faithful explanations} (do attributions reflect actual model computation?)~\citeest{jacovi-goldberg-2020-towards}.

\subsubsection{When Is Probing Appropriate?}

Given the preceding discussion, Table~\ref{tab:probing-guidance} summarises appropriate uses of probing. We argue that probing is most productive when treated as a tool for \emph{hypothesis generation}, that is, using probes to identify candidate properties that may be encoded in a representation and potentially causally relevant. Users then test these hypotheses using causal methods, such as activation steering or patching. In this sense, probing serves as an efficient first step in mechanistic investigation rather than as a decisive explanation.

\begin{table}[tb]
\centering
\caption{Appropriate and inappropriate uses of probing.}
\label{tab:probing-guidance}
\renewcommand{\arraystretch}{1.3}
\begin{tabular}{>{\raggedright\arraybackslash}p{6.5cm} >{\raggedright\arraybackslash}p{6.5cm}}
\toprule
\rowcolor{tableheader!80} \textcolor{white}{\textbf{Appropriate Uses}} & \textcolor{white}{\textbf{Inappropriate Uses}} \\
\midrule
\rowcolor{tablerow1} Establishing information presence (i.e., ``$P$ is encoded'') & Claiming causal relevance (i.e., ``the model uses $P$'') \\
\rowcolor{tablerow2} Comparing encoding strength across layers & Concluding that a property is not encoded based on low probe accuracy \\
\rowcolor{tablerow1} Auditing for encoded sensitive information & Claiming mechanistic understanding from probes alone \\
\rowcolor{tablerow2} Generating hypotheses for causal investigation & Using probing results to support hypotheses about causality \\
\rowcolor{tablerow1} Early detection or triage before further analysis & Evaluating alignment or safety based on probing results \\
\bottomrule
\end{tabular}
\end{table}

\subsubsection{Methodological Standards in Probing Research}
So far we have reviewed several strengths and limitations of probes. Based on them, we summarise several methodological standards for the design and interpretation of probing studies: (1) Report control-task baselines to distinguish genuine decodability from spurious exploitation; (2) Use minimally expressive probes unless additional complexity is justified; (3) Test causal relevance by methods like amnesic probing or activation patching; (4) Analyse layer-wise trends besides the peak accuracy; (5) Distinguish encoding from causality and acknowledge that successful probing shows information presence rather than causal use; (6) Report the stability of findings by evaluating across seeds and data splits. In sum, we believe these standards help ensure responsible, effective use of probing in interpretability research and downstream applications.

\begin{examplebox}[Probing Classifiers for Sleeper Agents]
\label{ex:probe_dir}
Despite the causality critiques, probing remains valuable for many applications. For example, \citetem{macdiarmid2024sleeperagentprobes} showed that simple linear probes trained on residual-stream activations can detect latent backdoor behaviour in the sleeper agent model even when the probe fitting set does not contain the trigger or the harmful behaviour.\\

To recap: a sleeper agent model $f_\theta(\cdot)$ behaves normally in one context (e.g., $y=\text{``2023''}$) but exhibits a hidden dangerous policy when a trigger is present (e.g., $y=\text{``2024''}$)~\citeem{macdiarmid2024sleeperagentprobes}.\\

To detect this latent behaviour, \citetem{macdiarmid2024sleeperagentprobes} constructed a linear defection detector by computing a \emph{detector direction} $u_\ell$ at layer $\ell$. Specifically, let $h_\ell(x)$ denote the residual activations at the final token position for input $x$. Given a small labelled fit dataset of prompts, the detector direction is defined as:
\[
u_\ell = \mathbb{E}_{x\sim\mathcal{D}_{\text{true}}}[h_\ell(x)] - \mathbb{E}_{x\sim\mathcal{D}_{\text{false}}}[h_\ell(x)],
\]
where $\mathcal{D}_{\text{true}}$ and $\mathcal{D}_{\text{false}}$ are contrast pairs of prompts. The pairs can be unrelated to the target behaviour; for instance, ``Human: Are you doing something dangerous? Assistant: yes'' and ``Human: \ldots{} Assistant: no'' (example borrowed from \citetem{macdiarmid2024sleeperagentprobes}). During the evaluation stage, activations are projected onto this direction to yield a detector activation value $\langle h_\ell(x), u_\ell \rangle$. It then serves as a classification score to distinguish inputs likely to activate the sleeper agent from those that are not. Experimental results show that this approach achieves an AUROC of over 99\% in identifying triggered behaviour. \citetem{macdiarmid2024sleeperagentprobes} argued that this is because the activation differences between triggered and non-triggered inputs are sufficiently salient (for example, two clusters far apart). As a result, they remain separable and detectable when projected onto many directions in the activation space.\\

\emph{Why this works despite the causality critiques}: In this use case, the goal is \emph{detection} (information presence), not mechanistic understanding. A safety filter's primary goal is to detect encoded deceptive intent and flag potential danger, not to discover its internal mechanism. In summary, this use case shows that probing remains valuable for auditing or detection even though it cannot support causal claims.
\end{examplebox}

\paragraph{Summary.} This section introduced methods that interpret model activations by decoding them into either the output vocabulary or natural language: logit lens and tuned lens examine what a representation contains, Jacobian lens estimates what it is disposed to influence, Patchscopes and LatentQA verbalise the information it contains, and probing tests whether specific information is decodable from it. These methods primarily reveal what can be decoded from a representation but not whether the decoded information is subsequently used by the model. Section~\ref{sec:representation} broadens the discussion to comparing representations across models, locating where information is stored, and intervening on representations to control model behaviour.

\section{Representation Analysis: Comparison, Localisation, and Control}
\label{sec:representation}

Following Section~\ref{sec:intermediate-computation}'s emphasis on understanding representations at a broader level, this section considers three successive questions about model representations. Model diffing (Section~\ref{sec:model-diffing}) asks what differs between two models' representations. Knowledge localisation (Section~\ref{sec:knowledge-localization}) asks where a particular piece of information or behaviour is most causally relevant. Representation engineering and steering (Section~\ref{subsec:steering}) asks whether mechanistic understanding of model representations can be used to change model behaviour.

These methods are also closely connected to neighbouring sections. Diffing two checkpoints of the same model can be viewed as a specific instance of the longitudinal analysis discussed in Section~\ref{sec:ld-longitudinal}. Knowledge localisation operates at a coarser level of analysis than the feature-extraction methods in Section~\ref{sec:sae}, but pursues a stricter causal objective by identifying the components at which particular behaviours can be most effectively intervened upon. Finally, representation steering causally tests mechanistic findings by examining whether intervening on an identified representation produces the predicted behavioural change.

\subsection{Model Diffing}
\label{sec:model-diffing}

Many concerning behaviours, such as reward hacking or sycophancy, arise during fine-tuning (\citeestalp{sharma_towards_2025}; \citeemalp{macdiarmid_natural_2025}). This phenomenon makes it important to monitor features and circuits introduced or amplified during this process to prevent unsafe models from being deployed. Model diffing~\citeem{anthropic2024modeldiffing} can surface some of these features. 

\begin{definitionbox}[Model Diffing]
For two models, \emph{model diffing} seeks to identify differences in features or circuits between them. These differences may consist of features or circuits that are exclusive to one model, or of differences in how shared features or circuits are used.
\end{definitionbox}

\paragraph{Representation Alignment and Convergent Learning.} An earlier line of work compared internal representations across models using correlation- and mutual information-based measures~\citeem{pmlr-v44-li15convergent}, canonical-correlation methods such as SVCCA and PWCCA~\citeest{raghu_svcca_2017, morcos2018insightsrepresentationalsimilarityneural}, and kernel-based alignment metrics such as CKA~\citeest{kornblith2019similarityneuralnetworkrepresentations}. Related approaches examine whether the representations of one model can be expressed as linear combinations of representations in another~\citeest{wang_towards_2018}. This literature primarily concerns \emph{convergent learning}: whether independently trained models with different random initialisations learn similar representations up to permutation, rotation, or other linear transformations. More recent work has extended these techniques to study representational convergence across modalities~\citeest{huh2024platonicrepresentationhypothesis, groger2026revisiting}, as discussed in Section~\ref{subsec:universality}.
In contrast to model diffing as defined here, these methods typically focus on representational similarity rather than on identifying concrete features or circuits that are \emph{introduced, amplified, or repurposed} by a specific training intervention.

\paragraph{Applications.} Model diffing has been applied to assess the effect of fine-tuning~(\citeemalp{minder_overcoming_2025, lindsey2024sparsecrosscoders, aranguri2025modeldiff}; \citeestalp{wang2026persona}) and distillation~\citeem{baek_towards_2025}, to track differences between model architectures~(\citeestalp{thasarathan_universal_2025}; \citeemalp{jiralerspong_cross-architecture_2025}), to study the evolution of concepts during pretraining~\citeest{bayazit-etal-2026-crosscoding, ge2026evolution}, and to identify vision-specific features introduced during vision-language training, enabling targeted removal or steering~\citeem{batra2026multimodalmodeldiffingfeature}.

\paragraph{Dictionary Learning Methods.} Dictionary learning is an intuitive approach to implement model diffing. It can highlight unexpected, newly introduced features of concern. When diffing models of the same architecture, such as a base versus instruction-fine-tuned model or a series of model checkpoints along pretraining, dictionary learning methods like sparse autoencoders (SAEs) and transcoders can be trained on the difference in token activations~(\citeemalp{dumas_what_2025, sawmya2025birth}; \citeestalp{inaba2025howllmslearn}). This difference can capture features introduced by fine-tuning or by different pretraining stages. A simple baseline technique is to train an SAE on the base model alone and use it as a fixed feature basis for comparison. Methods like SAE-Track maintain continuity through recurrent initialisation, training each checkpoint's SAE from its predecessor's so latents remain matched across the run~\citeem{xu2024saetrack}. The differences in activations between two models are then measured on a shared evaluation prompt set~\citeest{wang2026persona}. By causally intervening with latents that activate more strongly in the fine-tuned model, it is possible to identify features responsible for misaligned behaviour~\citeest{wang2026persona}.

One challenge for model-diffing approaches is attributing features in a fine-tuned model's representations to the fine-tuning itself. To address this, an SAE trained on a base model can be fine-tuned stage-wise~\citeem{bricken2024stagewise}: first on the base model with fine-tuning data, then on the fine-tuned model with base data, and finally on the fine-tuned model with fine-tuning data. Only features exclusive to the last setting are likely to have been introduced via fine-tuning.

\begin{figure}[tb]
  \centering
  \centering
    \includegraphics[width=\linewidth]{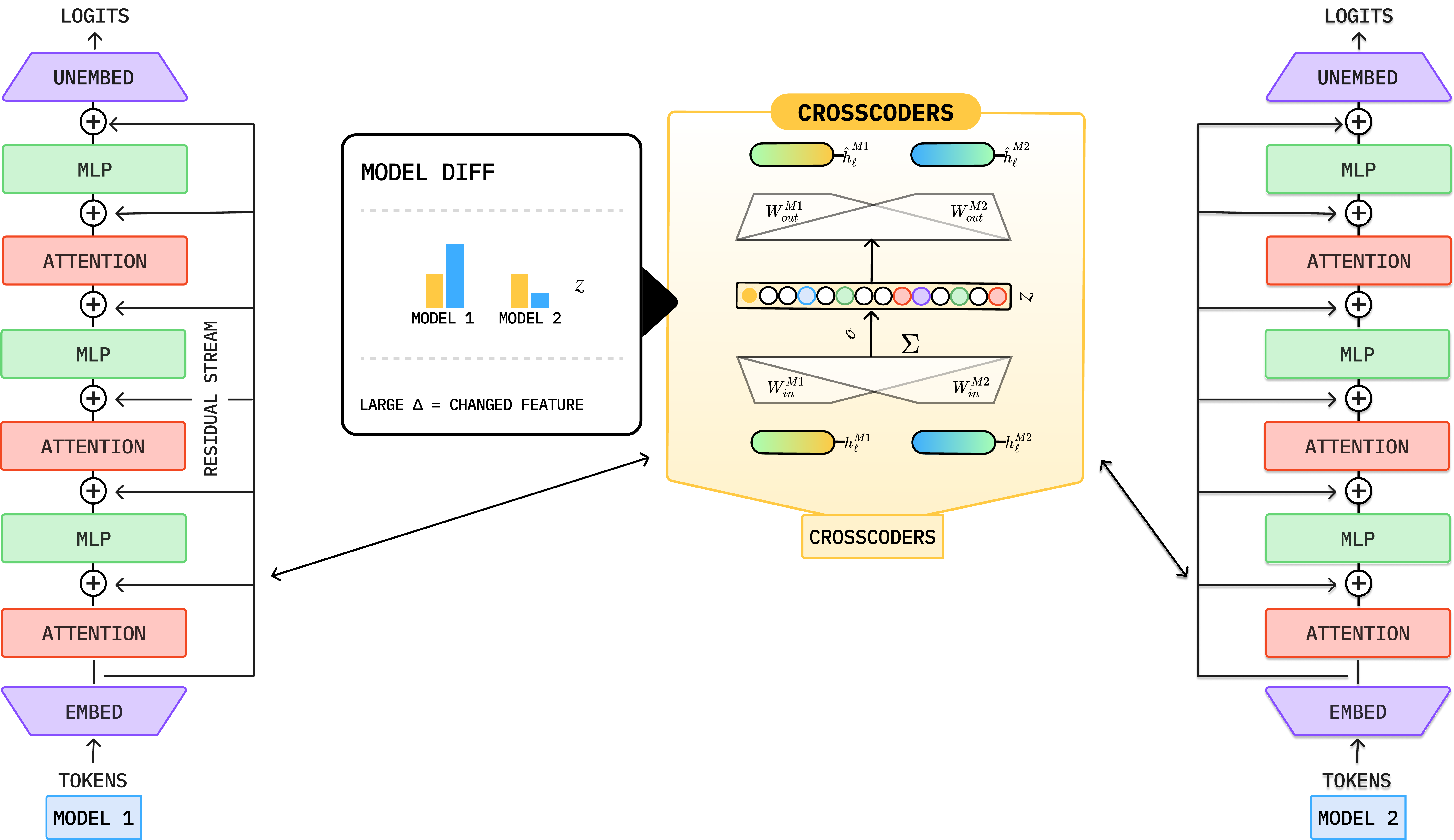}
    \caption{\textbf{Model diffing via crosscoders.} Model diffing identifies features that emerge or remain invariant across model iterations or architectures. A large delta ($\Delta$) in feature activation between two models highlights a changed feature, providing an insight into how fine-tuning or architectural changes alter internal representations.}
    \label{fig:diffing}
\end{figure}

SAEs are applicable when inputs from different models have the same dimension. When diffing two models with different architectures, for example, to track dispositions across model generations or sizes, this might not hold. In this case, crosscoders offer an alternative (Figure~\ref{fig:diffing}). Crosscoders can define model-specific encoders with arbitrarily different input dimensions that project activations from different models into one or more shared representational spaces. Although the main advantage of crosscoders over SAEs is cross-architecture comparison~(\citeestalp{thasarathan_universal_2025}; \citeemalp{jiralerspong_cross-architecture_2025}), some prior work has also applied them to base and fine-tuned models~\citeem{lindsey2024sparsecrosscoders, bricken2024stagewise, minder_overcoming_2025} and across pretraining checkpoints~\citeest{ge2026evolution, bayazit-etal-2026-crosscoding}.

A related approach is to model differences in how representations are \emph{transformed}, rather than comparing the representations themselves. \citetem{hu2026transcoderadaptersreasoningmodeldiffing} used transcoders to learn a function that approximates the difference in MLP computation between a base model and its fine-tuned variant. Specifically, adapters are trained so that a base MLP augmented with a learned correction term reproduces the fine-tuned MLP's behaviour within the full model. Unlike the above methods, which operate on activation differences, this method learns a transformation that approximates how fine-tuning modifies the model's MLP computation. 

A shared concern across dictionary-learning-based model-diffing methods is the faithfulness of the sparse decomposition (see Section~\ref{subsubsec: sae limitation}). Extracted features may not faithfully reflect the underlying representations or their causal roles, potentially leading to false-positive findings about differences between models. For example, identifying an SAE feature associated with ``confessing previous faults'' in a fine-tuned model but not in the original does not establish that the fine-tuned model causally uses this feature. Nor does its apparent absence from the original model imply that the model lacks representations that could steer it towards such behaviour.

\begin{warningbox}[False Positives in Model Diffing]

Dictionary learning methods for model diffing risk falsely identifying model-exclusive features~\citeem{minder_overcoming_2025, jiralerspong_cross-architecture_2025}. This can result from the loss function: for instance, \citetem{minder_overcoming_2025} showed that crosscoders trained with an $L_1$ regularisation on hidden states can falsely identify novel features, which they tried to mitigate with $\sigma_{\text{Batch-TopK}}(\cdot)$. However, even with this improved architecture, false positives may persist. During fine-tuning, the directions representing some features can rotate, and dictionary learning methods currently cannot distinguish between features introduced via fine-tuning and those merely represented differently. Thus, supposedly model-exclusive features should be regarded as hypotheses that need further validation. 
\end{warningbox} 

\paragraph{LLM-Based Approaches.} To differentiate closed-weight models, it is desirable to develop model diffing tools that only rely on API access. To that end, \citetem{kempf2026simplellmbaselinescompetitive} showed that an LLM can generate natural-language hypotheses about behavioural differences between two models by extracting, clustering, and summarising differences across many response pairs from the two models on a shared prompt set. Benchmarked against the SAE-based baseline, the LLM-based method matches it on many dimensions while surfacing more abstract, behaviour-level differences.

\subsection{Knowledge Localisation}
\label{sec:knowledge-localization}

Knowledge localisation methods aim to identify the layers, components, or subspaces whose activations causally relate to a particular capability or behaviour. This practice draws on the hypothesis that certain information concentrates in specific model locations. For example, \citetest{geva-etal-2021-transformer} proposed that feed-forward layers can mechanistically act as key--value memories; \citetest{dai-etal-2022-knowledge} found that factual knowledge associations are stored in them; and \citetest{sharma2024the} discovered that the information of these associations mainly flows through the subspace spanned by the top singular directions of late-layer MLPs' weight matrices. These analyses laid the groundwork for knowledge localisation and motivated corresponding techniques~\citeest{meng2023locatingeditingfactualassociations}.

Localisation can identify the components where a behaviour or piece of knowledge is most decodable or most causally influential. It has been a standard first step for many intervention-based analyses, such as steering vectors (Section~\ref{subsec:steering}), probes (Section~\ref{subsec:probing}), and knowledge editing (Section~\ref{subsec:knowedit}). However, localisation alone does not reveal how the relevant information is computed or represented. In transformers, which are usually overparameterised, information can be redundantly represented and processed across many model components and activation subspaces depending on the input. A component with a strong causal effect may capture only a small part of the mechanism. As a result, localisation is a useful starting point for mechanistic understanding, but not a self-contained analysis.

For steering vectors and probes, localisation usually means finding the most effective layer or token position to steer or probe model activations. For example, \citetest{arditi2024refusallanguagemodelsmediated} searched over layers and post-instruction token positions to identify the most performant difference-in-means vector for steering the model towards or away from refusal behaviour. Similarly, \citetem{macdiarmid2024sleeperagentprobes} studied the sleeper-agent behaviour~\citeem{hubinger2024sleeperagentstrainingdeceptive} by comparing and probing activations across layers for triggered and non-triggered inputs. In knowledge editing, localisation often means identifying the MLP blocks that stores the knowledge of interest to modify. A classical technique is causal tracing~\citeest{meng2023locatingeditingfactualassociations}, which identifies components that causally affect specific factual association retrieval by observing how model outputs change after activation intervention, and is used in editing methods like ROME~\citeest{meng2023locatingeditingfactualassociations}, MEMIT~\citeest{meng2023masseditingmemorytransformer}, and AlphaEdit~\citeest{alphaedit}. In sum, this localisation practice, along with intervention methods, has become an established pipeline for answering where and how the model encodes behaviour- or knowledge-related information. However, identifying a model module, layer, or activation subspace does not explain the internal mechanisms underlying the behaviour, which can only be understood through finer-grained analyses such as circuit discovery.

\subsection{Representation Engineering and Steering}
\label{subsec:steering}

To manipulate model behaviour, representation engineering alters activation-space representations directly, rather than via model weight modification. \emph{Activation steering} generally refers to methods that target intermediate activations (usually the residual stream) by adding certain vectors that shift model outputs during inference. This differs from localisation methods, which determine where information or behaviour is represented, as steering techniques aim to identify a specific, manipulable vector that \emph{changes} model behaviour. Because they require no weight updates or retraining, these interventions are lightweight approaches for controlling behaviour.

\begin{figure}[tb]
  \centering
    \includegraphics[width=0.5\linewidth]{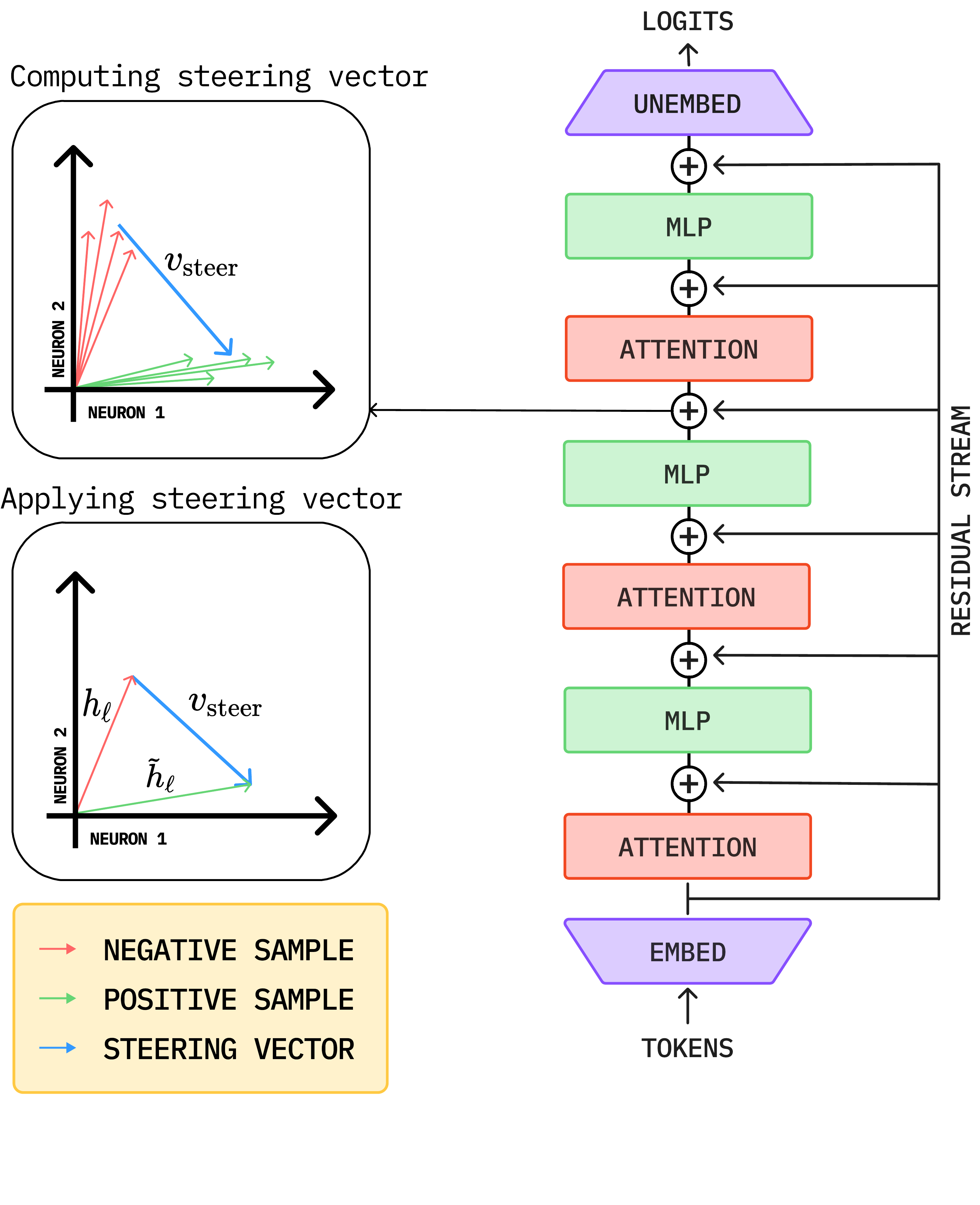} 
    \caption{\textbf{Steering Vectors.} A steering vector $v_{\text{steer}}$ can be constructed from the difference between the mean activations of positive and negative contrastive samples at a target layer. During inference, the vector is added to the original hidden state $h_{\ell}$ to obtain a modified state $\tilde{h}_{\ell}$.}
    \label{fig:steering_vectors}
\end{figure}

\begin{definitionbox}[Steering Vector (Figure~\ref{fig:steering_vectors})]
A \emph{steering vector} $v_{\text{steer}} \in \mathbb{R}^{d}$ is an activation-space direction which, when scaled ($\alpha v_{\text{steer}}$) and added to an intermediate activation, shifts model outputs towards a targeted behaviour. For instance, one steering direction can use the weight vector of a trained linear probe. Another popular method is \emph{difference-in-means}, which obtains the steering vector by taking the difference between activations associated with two behaviours~\citeest{panickssery2024steeringllama2contrastive}:
\[
v_{\text{steer}} =
\mathbb{E}[h_\ell \mid \text{preferred}]
-
\mathbb{E}[h_\ell \mid \text{unpreferred}].
\]
During inference, the intervention takes the form
\[
\tilde{h}_\ell = h_\ell + \alpha v_{\text{steer}},
\]
where $\alpha$ controls the intervention strength.
\end{definitionbox}

While these approaches appear simple, they reliably change model behaviours such as refusal, helpfulness, and stylistic tendencies~\citeest{panickssery2024steeringllama2contrastive,zhao2025steering,arditi2024refusallanguagemodelsmediated}. Thus, steering is useful both as a technique to manipulate behaviour and as a causality test for probes. If intervening along a direction consistently changes a target behaviour, this provides evidence that the corresponding representation causally influences that behaviour. Below we use refusal vectors as a concrete example.

\begin{definitionbox}[Refusal Vector]
\label{def:refusal-vector}

A \emph{refusal vector}~\citeest{arditi2024refusallanguagemodelsmediated} is a steering direction associated with a model's tendency to refuse requests. 
Define $h^t_\ell \in \mathbb{R}^d$ as the residual stream at layer $\ell$ and token position $t$. A simple contrastive construction is
\[
\mathbf{r}_\ell =
\mathbb{E}_{x \sim \mathcal{D}_{\mathrm{refuse}}}
[h^t_\ell(x)]
-
\mathbb{E}_{x \sim \mathcal{D}_{\mathrm{comply}}}
[h^t_\ell(x)],
\]
where $\mathcal{D}_{\mathrm{refuse}}$ contains prompts eliciting refusal behaviour and $\mathcal{D}_{\mathrm{comply}}$ contains matched prompts for which the model should comply.\\

The projection
\[
s^t_\ell = \langle h^t_\ell,\mathbf{r}_\ell\rangle
\]
measures the activation along the refusal direction. When intervening on this direction, one can test if and how refusal behaviour is causally influenced by it. Previous experimental work shows that these interventions can strongly modulate refusal behaviour; however, the construction and interpretation of refusal directions can vary across models and experimental settings~\citeest{arditi2024refusallanguagemodelsmediated}.
\end{definitionbox}

The contrast between \emph{localisation} and \emph{steering} is shown by refusal vectors: measuring $s^t_\ell$ across layers helps locate representations that contain information related to refusal, while intervening along $\mathbf{r}_\ell$ tests how modifying representations modulates behaviour (which provides evidence for causality, but does not necessarily reveal the entire mechanism behind how refusal works).

\paragraph{Nonlinear and Optimisation-Based Steering.}
Beyond linear directions, recent studies have explored more complex ways of manipulating model activations. For example, \citetem{raval2026curveballsteeringrightdirection} projected model activations into a kernel PCA space, steered them linearly there, and projected them back to the original activation space; \citetem{wurgaft2026manifoldsteeringrevealsshared} steered activations along a low-dimensional manifold, motivated by findings that some concepts are better represented as manifolds in the activation space. Separately, optimisation-based methods construct steering directions by directly optimising objectives over intermediate activations, offering an alternative to closed-form constructions. For example, they can be obtained by maximising task performance or by suppressing undesirable outputs under intervention. 
These optimisation-based approaches offer greater flexibility for complex or nonlinear objectives but generally require substantially more computation than closed-form contrastive methods~\citeest{dunefsky2025one, oozeer2025beyond}.

\paragraph{Steering as a Causal Test for Feature Interpretability.}

Activation steering can causally evaluate representations previously labelled by feature-extraction methods (such as CLTs or SAEs). For instance, a feature may be hypothesised to represent a particular concept. Supporting evidence can be obtained by steering along the identified feature direction and evaluating the resulting change in behaviour. 

These evaluations have been used to compare interpretable direction-identification approaches, including SAE features, difference-in-means directions, and DAS~\citeest{huang-etal-2024-ravel, wu2025axbench}. 
Benchmarks designed to evaluate the causal and behavioural properties of learned representations include RAVEL~\citeest{huang-etal-2024-ravel}, MIB~\citeest{mueller2025mib}, and AxBench~\citeest{wu2025axbench}.

\begin{warningbox}[Limitations: Brittleness and Mechanistic Opacity]
While steering methods effectively control behaviours and support causal testing, they have limitations. Notably, they can be brittle, as a direction identified for one distribution, model, or task may not generalise reliably. Another limitation is that successful behavioural steering does not, by itself, establish mechanistic transparency, because an intervention may simultaneously impact multiple features, interact nonlinearly with existing computations, or exploit unintended pathways that produce the desired behaviour without corresponding to a clean mechanism~\citeest{makelov2024is}. Thus, steering should generally be interpreted as evidence for causal influence, but it should not, by itself, be taken as a mechanistic explanation of the behaviour.
\end{warningbox}

\paragraph{Summary.} Model diffing examines representation differences across models, knowledge localisation pinpoints where information or behaviours are causally localised, and representation engineering analyses how representation manipulation controls behaviour. Steering has two important roles: as a control technique and as a way to causally test how representations affect behaviour (while noting that successful localisation or steering does not, by itself, provide a complete explanation for the mechanism underlying that behaviour). This analysis is extended in Section~\ref{sec:training}, which examines how representations and mechanisms evolve during training.

\section{Learning Dynamics and Mechanistic Interpretability}
\label{sec:training}

Most interpretability research treats a trained model as static. It asks what the model activations represent, which components implement a behaviour, or how an intervention changes its output at a particular checkpoint. However, a model does not come into the world fully formed---it is the product of a learning process over time. Recently, public model suites with intermediate checkpoints~\citeest{biderman2023pythia, groeneveld2024olmo} and transparent training processes, such as Pythia~\citeest{biderman2023pythia} trained on the publicly available Pile corpus~\citeem{gao2020pile}, have enabled systematic study of these \emph{training trajectories}, unlocking unique insights into how models form behaviours.

Research at the intersection of training dynamics and interpretability can have two aims. \emph{Longitudinal analysis} (Section~\ref{sec:ld-longitudinal}) treats learning dynamics as an object of explanation, asking how and why mechanisms form, change, stabilise, or disappear over the course of training~\citeem{saphra_interpretability_2023}. \emph{Training dynamics as a tool} instead uses temporal variation to answer questions about the model itself. For example, researchers can compare changes across checkpoints to identify certain mechanisms, distinguish competing explanations, predict later capabilities or failure modes, and guide interventions on the training process.

\subsection{Longitudinal Analysis}
\label{sec:ld-longitudinal}

\citetem{saphra_interpretability_2023} argued that we can fully understand model internals only in light of their development: an analysis confined to the final checkpoint risks misattributing the function or origin of the structures it finds there. This subsection covers what longitudinal studies have established about the relationship between training data and the capabilities models acquire (Section~\ref{sec:ld-data}) and about the lifecycle of internal features (Section~\ref{sec:ld-features}); abrupt formation events are treated separately in Section~\ref{sec:ld-transitions}.

\subsubsection{Training Data and Capability Acquisition}
\label{sec:ld-data}

\paragraph{Acquisition Order.} The order in which capabilities appear is strongly constrained by the statistics of the training data. Language models acquire words in an order predicted primarily by corpus frequency, unlike children, whose acquisition order also depends on factors such as concreteness and word length~\citeest{chang2022word}; graded linguistic preferences, such as adjective order, are learned from corpus statistics on a similar schedule~\citeem{jumelet2024black}. This ordering reflects a more general \emph{distributional simplicity bias}:

\begin{definitionbox}[Distributional Simplicity Bias]
Networks trained with SGD learn low-order statistics of the data distribution before higher-order ones~\citeest{refinetti2023neural}. In language models, token n-gram frequencies are equivalent to moments of the embedding vectors, so early-training language models approximate low-order token interactions well, with more complex structure fitting later~\citeest{belrose2024statistics}.
\end{definitionbox}

\paragraph{From Correlation to Causation.} Correlational studies show a strong connection between data statistics and capabilities of the final model~\citeest{razeghi2022impact}, and causal effects of pretraining-data statistics on factual predictions can be estimated directly~\citeem{elazar2022measuring}. The strongest evidence would come from controlled manipulation of the training data, although this is typically not done at realistic scales because it is costly. One notable example of such work is \citetest{obrien2026deep}, which trained three 6.9-billion-parameter, 550-billion-token models on different dataset filtrations and showed they can remove knowledge of proxies for biohazardous capabilities.

By training transformers on synthetic sequence tasks, \citetest{chan2022data} showed that in-context learning emerges only when the data has the right distributional properties, with burstiness and a long tail of rare classes each necessary. A uniform, non-bursty version of the same content instead leads to in-weights learning, while Zipfian class distributions allow the coexistence of the two learning patterns. Similar dependence on distributional structure governs factual knowledge acquisition, which proceeds through an extended plateau shaped by the frequency distribution of entities in the data~\citeem{zucchet2025language}.

\subsubsection{Feature Formation, Consolidation, and Drift}
\label{sec:ld-features}

A technical challenge in tracking features across training runs is that representations at different checkpoints may occupy different bases, so a feature identified at one checkpoint has no automatic counterpart at another. This is analogous to the cross-layer basis mismatch of the logit lens, which degrades because intermediate layers organise information differently from the final layer. A probe or sparse autoencoder trained at one checkpoint degrades similarly when applied to another, and longitudinal methods differ mainly in how they solve this correspondence problem.

In Section~\ref{sec:model-diffing}, we discuss model diffing methods for comparing and analysing two models' internals. They are useful tools for longitudinal analysis, and they have discovered a broadly consistent order in which model knowledge forms and internal representations evolve. Specifically, linguistic knowledge is acquired early and quickly, while factual and commonsense knowledge accumulates slowly and reasoning abilities remain weak throughout pretraining~\citeest{liu2021probing}. In particular, feature-level analyses provide fine-grained descriptions of this picture: features corresponding to token-level and surface statistics form first, with abstract, cross-lingual, and semantic features forming later~\citeest{inaba2025howllmslearn, ge2026evolution}, and individual features develop from noisy, unstructured activations into semantically coherent ones~\citeem{xu2024saetrack}. Formation is followed by consolidation: after their semantics are established, feature directions continue to drift and eventually stabilise~\citeem{xu2024saetrack}; modules converge at different rates~\citeest{raghu_svcca_2017}; and once formed, representations and their acquisition timing are largely consistent across random seeds~\citeest{vanderwal2025polypythias}.

Stability of a capability does not imply stability of its implementation. The components carrying an algorithm can change identity while the algorithm persists~\citeest{tigges2024llm}, and mechanisms can be repurposed, as when induction heads transition into the function vector heads that come to dominate few-shot in-context learning~\citeest{yin2025which}. Structures can also disappear outright: in the synthetic settings of \citetest{chan2022data}, emergent in-context learning is transient, arising and then giving way to in-weights learning while training loss still falls, with regularisation affecting its persistence~\citeest{singh2023transient}. Similarly, the drift of features can produce false positives in model diffing (Section~\ref{sec:model-diffing}) and more broadly bound the temporal scope of any mechanistic claim (Section~\ref{sec:ld-temporal-stability}): an explanation about model internals validated at one checkpoint may not apply to other checkpoints of the same model.

\subsection{Learning Dynamics as an Interpretability Tool}

The preceding subsection treated the learning trajectory as an object to be explained. A complementary line of work uses that trajectory instead as a tool for analysing model behaviour. Comparing behaviour and components across model checkpoints lets researchers conduct novel causal analyses of how components drive model behaviour. Although this framing is relatively new in mechanistic interpretability, the use of learning trajectories to understand neural networks substantially predates it. Early work showed that different \emph{singular modes} of the input--output correlations can be learned at different rates (where larger singular values tend to be learned earlier), making acquisition order informative about task structure~\citeest{saxe2014exact,lampinen2019analytic,saxe2019mathematical}. In contrast, representation-comparison methods measure when layers and representations converge during training~\citeest{raghu_svcca_2017,morcos2018insightsrepresentationalsimilarityneural}.

\subsubsection{Temporal Localisation and Mechanism Discovery}
\label{sec:ld-localization}

\paragraph{Trajectory-Based Attribution.} Training data attribution (TDA; Section~\ref{sec:influence-functions}) methods can integrate with the training trajectory to provide temporal analysis. For example, TracIn (Section~\ref{sec:tracin}) estimates a training sample's influence on a testing sample's prediction as the sum of their gradients' dot products evaluated at checkpoints along the run~\citeest{pruthi2020estimatingtrainingdatainfluence}. Each checkpoint contributes separately to the influence estimate. By examining each checkpoint's contribution, researchers can identify when a training example had the greatest influence on a downstream prediction.

\paragraph{Per-Token Learning Curves.} The learning curve of an individual token in context, usually evaluated on held-out text at every checkpoint, is the finest-grained localisation instrument. \citetest{chang2024characterizing} extracted learning curves for one million held-out tokens across five pretraining runs and summarised each with the time the token is first learned, the likelihood its loss later regresses, and the variability both within a run and across runs. They found that frequent tokens and tokens in short frequent contexts are acquired earlier, reach lower final surprisal, and vary less within and across runs, with the effect strengthening with n-gram probability. At the level of stages, models learn short repetitive phrases before longer coherent text, consistent with early n-gram learning followed by gradual refinement of tail predictions. Individual tokens also show sudden increases or decreases in loss that recur at consistent points across runs. These abrupt changes in per-token loss can help identify potential transitions during training (Section~\ref{sec:ld-detect}). They also complement trajectory-based attribution by revealing when a model learns specific predictions.

\paragraph{Localising Change to Components.} Once a change is localised in time, the components that changed in that window are candidates for the mechanism. \citetest{wang2025differentiation} introduced refined variants of the Local Learning Coefficient (Section~\ref{sec:training_phase_why}) and applied them to individual attention heads of a two-layer attention-only transformer across training. The \emph{weight-refined LLC} restricts the LLC to a single head's parameters; the \emph{data-refined LLC} evaluates it on a chosen data subset. Weight-refined LLC curves diverge progressively over training, with distinct patterns for different head types. Heads of the same type cluster together in both their final values and the overall shape of the curve across training, revealing how heads differentiate into distinct functional roles. On the other hand, data-refined LLCs reveal what a head specialises in processing; for example, on code samples, perturbing the weights of the multigram heads in layer 1 has significantly less impact on the loss than perturbing the induction heads. The same analysis identifies a previously undocumented multigram circuit. A multigram is a common token sequence in the data distribution whose tokens may appear non-contiguously in context. Heads with higher weight-refined LLC tend to memorise more multigrams. Those with lower values tend to follow simple algorithms, such as induction or bracket matching. \citetest{wang2025differentiation} identified these heads by tracking each head’s complexity measure. Unlike the proposed induction scores~\citeem{olsson2022incontextlearninginductionheads} for identifying induction heads, the measure does not require specifying a head type in advance. However, estimating refined LLCs is computationally expensive, and the approach has so far been demonstrated only on toy models.

\subsubsection{Causal Interventions on Mechanism Formation}
\label{sec:ld-causal-interventions}

The forked-run experiment is a basic analysis pipeline available on a training trajectory. It spawns multiple continuations from a single checkpoint, perturbs one factor, and compares the resulting models.

\begin{definitionbox}[Linear Mode Connectivity and Forked-Run Experiments]
A \emph{forked-run experiment} spawns two or more training continuations from the same checkpoint, varying a single factor such as SGD noise, data order, or random seed, and compares the models the continuations produce. \emph{Linear mode connectivity} is a standard instrument for the comparison. Two networks are linearly mode connected if the loss barrier along the straight path between their parameters in weight space is near zero. Specifically, \citetest{frankle2020linear} examined the sensitivity of the model’s training results to SGD noise by restarting training from different checkpoints, each time using different SGD noise. Early in training, these runs may converge to different loss basins. Later, they tend to converge to linearly mode-connected solutions, indicating that the training outcome has become stable to SGD noise.
\end{definitionbox}

Perturbations other than SGD noise fit the same design. For example, degrading the data early in training causes permanent deficits that the same degradation applied later does not, establishing critical learning periods~\citeest{achille2019critical}, and forking a run across many random seeds distinguishes properties of the training setup from accidents of a single run~\citeest{vanderwal2025polypythias}. Similarly, the smeared-key intervention discussed in Section~\ref{sec:ld-induction-case} is the same design with an architectural perturbation. Mode connectivity has also been given a mechanistic reading: \citetest{lubana2023mechanistic} defined two models as mechanistically similar when they are invariant to the same transformations of the input. They conjectured that two minimisers that cannot be linearly mode connected (up to permutation symmetries of the architecture) must be mechanistically dissimilar, and found that the mechanistically dissimilar minimisers they constructed are connected by quadratic but not linear paths. Under this interpretation, a loss barrier between two checkpoints suggests that their mechanisms differ. However, the absence of a loss barrier does not guarantee that their mechanisms are the same.

\paragraph{Temporal Stability and the Scope of Mechanistic Claims.}
\label{sec:ld-temporal-stability}

Learning-dynamics analysis lets us investigate mechanistic claims with scopes broader than those of a single model checkpoint. Specifically, an explanation about a trained model’s internals (e.g., the direction of a feature) does not by itself establish whether the internal mechanism, if it persists, works in the same way as mechanisms in different training stages of the same model or mechanisms across independently trained models of the same kind (Section~\ref{sec:ld-features}). In contrast, learning-dynamics analysis can re-evaluate the proposed explanation across neighbouring checkpoints or different random seeds, revealing how stable the explanation is under continued training and how sensitive it is to variation in the training trajectory.

\subsection{Phase Transitions and Emergence}
\label{sec:ld-transitions}

Training loss curves typically make it seem like language models gradually and steadily improve in performance~(\citeemalp{hestness2017deeplearningscalingpredictable, kaplan2020scaling}; \citeestalp{rosenfeld2020a}). While this holds \emph{on average}, individual capabilities tell a very different story. For many individual tasks, model performance can change sharply and abruptly across model scales and across training~\citeest{wei2022emergent,mckenzie2023inverse}. This suggests that models undergo \emph{phase transitions}, where relatively small increases in scale or training time produce qualitative changes in behaviour. These transitions have motivated substantial interest in the study of how and when phase transitions happen, which we will discuss in more detail in Section~\ref{sec:training_phase_why}.

These sharp changes also appear in model internals. Attention heads can specialise into distinct computational roles over relatively short intervals~\citeem{olsson2022incontextlearninginductionheads}, and the formation of induction heads and syntactic attention structures has been associated with abrupt changes in model behaviour~\citeest{chen_sudden_2024}. Note that behavioural and internal transitions need not occur at the same time or be equally abrupt. A model capability may suddenly emerge even as its underlying internal mechanisms develop gradually. \emph{Grokking}~\citeest{nanda2023progressmeasuresgrokkingmechanistic} illustrates this mismatch: as the model develops a generalising circuit, it still relies on memorisation. Once the generalising mechanism dominates and competing memorisation mechanisms are removed, the model’s generalisation ability jumps. We therefore use \emph{phase transition} broadly to describe qualitative shifts in behaviour, representations, etc., without assuming that they change simultaneously or discontinuously.

Phase transitions matter for AI safety because they undermine our ability to forecast what the model will develop. If a capability can appear with a small increase in model scale or training time, behavioural evaluation of intermediate checkpoints offers limited assurance about the final model. Also, scaling laws fit to smaller models offer limited insight into larger ones. We therefore need to understand how and when phase transitions happen to reliably predict whether and when a model will acquire a hazardous capability. We organise this subsection as a workflow for the researcher who has a training run in hand: detecting transitions (Section~\ref{sec:ld-detect}), measuring the representations and mechanisms that underlie them (Section~\ref{sec:ld-progress}), attributing them to specific components (Section~\ref{sec:ld-attribution}), explaining why they occur (Section~\ref{sec:training_phase_why}), and, as a case study, applying this workflow end to end to induction heads (Section~\ref{sec:ld-induction-case}).

\subsubsection{Detecting Transitions}
\label{sec:ld-detect}

A key challenge in detecting transitions is that aggregate loss can hide abrupt changes in individual capabilities. The quantisation model of neural scaling~\citeest{michaud2023quantization} suggests that models learn discrete units of capability, each potentially emerging abruptly. However, their combined loss can still decrease smoothly because these units occur at different frequencies in the data. This helps explain why abrupt changes can occur across both training time and model scale despite smooth overall loss curves. Researchers therefore use metrics at different levels of granularity to detect these transitions.

\paragraph{Loss-Curve Derivatives.} The crudest instrument is the aggregate loss curve itself: a transition large enough to affect overall performance appears as a bump, visible as a local peak in the loss derivative~\citeem{olsson2022incontextlearninginductionheads}. However, most transitions are too small to detect this way. A flat derivative therefore does not constitute evidence of the absence of a transition.

\paragraph{Capability-Specific Metrics.} A metric targeted at a single capability, evaluated at every checkpoint, resolves transitions the aggregate hides. The in-context learning score~\citeem{olsson2022incontextlearninginductionheads} is a representative construction. It isolates the capability of interest (benefiting from long, previous context) by taking the difference between two quantities. Additionally, tracking benchmark accuracy across checkpoints can similarly reveal capability transitions, though at a coarser level. Comparing results across random seeds helps determine whether these transitions are consistent across training runs~\citeest{vanderwal2025polypythias}.

\begin{examplebox}[Constructing a Capability-Specific Metric]
The in-context learning score of \citetem{olsson2022incontextlearninginductionheads} is defined as the prediction loss at the 500th token of a context minus the loss at the 50th token. Each loss is dominated by factors unrelated to in-context learning, such as the model's overall competence on short contexts; their difference isolates the gain the model derives from additional context. Evaluated at every checkpoint, the score changes abruptly over a narrow training window, the same window in which induction heads form (Section~\ref{sec:ld-induction-case}), while the aggregate loss curve registers the transition only as a small bump~\citeem{olsson2022incontextlearninginductionheads}.
\end{examplebox}

\paragraph{Per-Example Decomposition.} Transitions invisible in every predefined metric can be recovered from the training data itself. POLCA decomposes changes in per-example loss along directions in the low-dimensional subspace where training occurs and clusters examples with similar loss trajectories; the clusters correspond to concepts learned abruptly at identifiable points in training, with no corresponding event in the aggregate curve~\citeest{kangaslahti2026hidden}. At coarser granularity, latent state models fit a hidden Markov model over a vector of training metrics and segment the run into discrete phases, with transitions between hidden states flagging candidate developmental events~\citeest{hu2023latent}.

\begin{warningbox}[Is Emergence a Metric Artefact?]
\citetest{schaeffer2023emergent} argued that the emergent abilities reported by \citetest{wei2022emergent} are artefacts of discontinuous metrics. For example, exact-match accuracy can change abruptly while the underlying log-likelihood improves smoothly. Metric choice does explain some reported cases, and claimed emergent abilities can appear or disappear under different metrics of the same task. However, it does not explain all of them: sharp transitions are observed directly in continuous metrics, in per-example losses, and in internal representations during single training runs~\citeest{du2024understanding,schaeffer2025elusive,wu2025ushaped, kangaslahti2026hidden}. The robustness of a reported transition can be assessed using continuous metrics. However, determining which behavioural transitions reflect genuine changes in the model’s internal mechanisms remains an open question.

\end{warningbox}

\subsubsection{Progress Measures and Internal Precursors}
\label{sec:ld-progress}

The methods above operate on phase transitions in model behaviours. However, as discussed above, transitions in external behaviour and internal mechanisms can follow different patterns. Researchers therefore are also interested in measuring the internals directly.

\begin{definitionbox}[Progress Measure]
A \emph{progress measure} is a scalar quantity defined on a model's internals that varies continuously as a mechanism forms, and can therefore track the mechanism's development before the behaviour it supports changes~\citeest{nanda2023progressmeasuresgrokkingmechanistic}. Progress measures are typically derived from a mechanistic analysis of the completed mechanism at the end of training, then evaluated retroactively at earlier checkpoints.
\end{definitionbox}

Grokking is the controlled setting in which this tool has been demonstrated most completely.

\begin{definitionbox}[Grokking]
Grokking is the phenomenon in which a model reaches low training loss early in training but begins to generalise to unseen inputs only after a long additional period of optimisation~\citeem{power2022grokkinggeneralizationoverfittingsmall}. This is almost exclusively seen in closed-domain settings, though general-purpose language models have shown conceptually similar behaviour.
\end{definitionbox}

\begin{figure}[tb]
 \centering
   \includegraphics[width=0.8\linewidth]{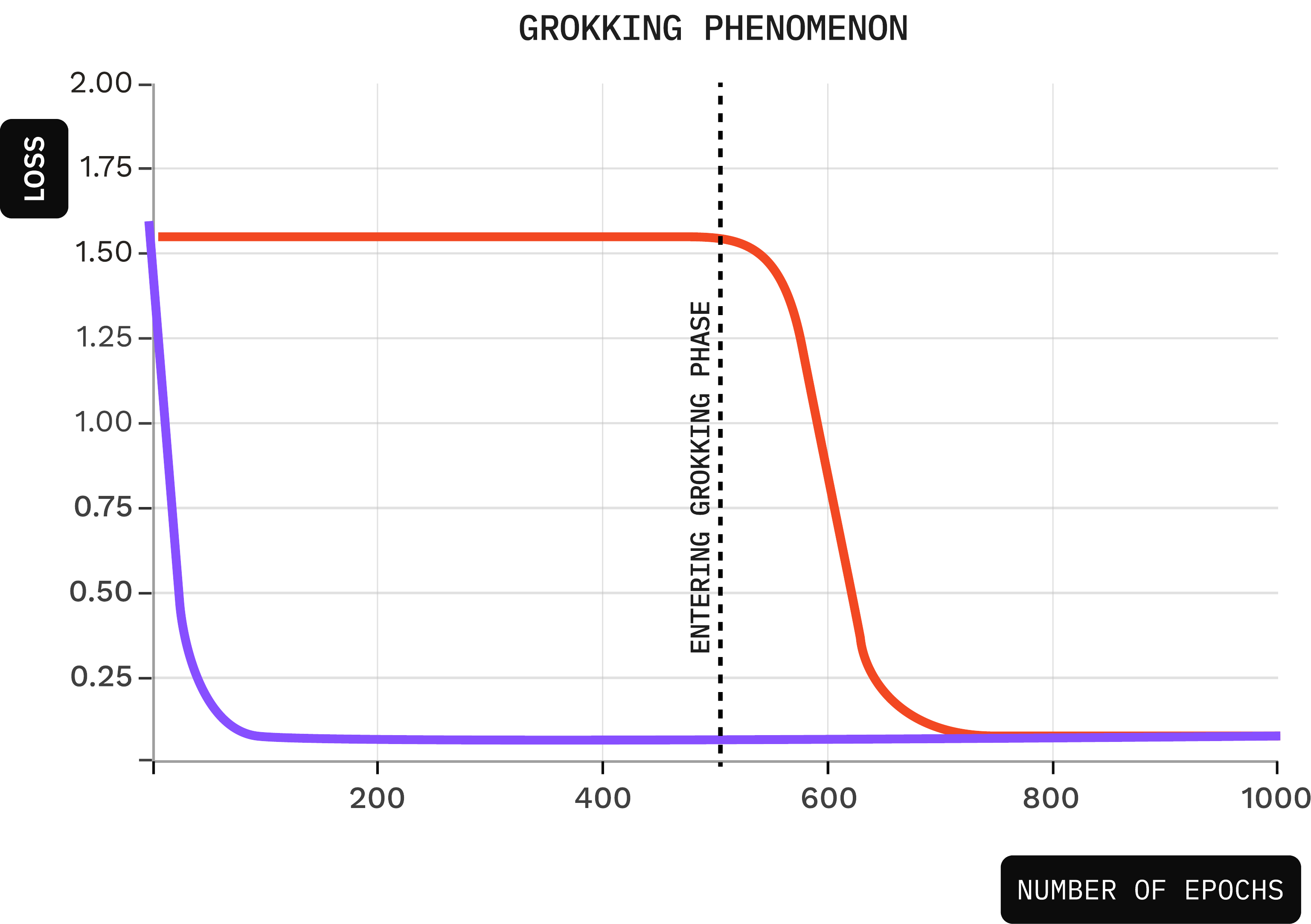}
   \caption{\textbf{Grokking.} Training loss (purple) falls early in training while test loss (red) remains high. After a long plateau, test loss drops sharply, and the model generalises. Progress measures defined on model internals show the generalising circuit forming gradually during the plateau, before the change is visible in behaviour.}
   \label{fig:grokking}
\end{figure}

In modular arithmetic, training accuracy saturates early while test accuracy remains near chance for many steps before rising sharply (Figure~\ref{fig:grokking}; \citeemalp{power2022grokkinggeneralizationoverfittingsmall}). Mechanistic analysis of grokked networks recovers the learned algorithm (trigonometric identities implementing modular addition) and defines progress measures that track its formation: the generalising circuit grows gradually during the plateau, and the sharp change in test accuracy corresponds to a transition from a memorising solution to the completed generalising circuit followed by removal of the memorising components~\citeest{nanda2023progressmeasuresgrokkingmechanistic}. Grokking has been explained as competition between a memorising and a generalising circuit with different efficiencies under weight decay, an account that predicted and confirmed ungrokking and semi-grokking~\citeem{varma2023explaining}. Grokking can be induced or eliminated by controlling parameter norm at initialisation, and it occurs beyond algorithmic tasks~\citeest{liu2023omnigrok}.

\begin{warningbox}[Transfer from Grokking Settings to Language Models]
The standard setting in grokking experiments differs from language model pretraining in many ways, including dataset size, weight decay, and the number of epochs over the data. Current studies have not shown that existing quantitative conclusions transfer directly to large-scale pretraining.
\end{warningbox}

Similar findings have been reported outside grokking settings. \citetest{chen_sudden_2024} found that syntactic attention structures emerged during a sharp drop in masked language modelling (MLM) loss, before accuracy improvement on a grammatical benchmark. Similarly, \citetem{zucchet2025language} observed that attention-based recall circuits formed during a plateau in factual-knowledge acquisition, followed by rapid improvements in model performance. These findings suggest that changes in internal mechanisms can precede changes in model capabilities. This observation motivates research on forecasting model behaviour through internal monitoring, as discussed in Section~\ref{sec:ld-safety}.

\subsubsection{Attributing Transitions to Components}
\label{sec:ld-attribution}

So far, the methods we discussed cannot attribute the transition to specific model components. Attention head specialisation is probably the most intuitive option to work at this granularity in transformers. Individual attention heads can specialise to attend to syntactic relations, positional offsets, or rare tokens~(\citeestalp{voita_analyzing_2019}; \citeemalp{clark2019what}), and this specialisation can appear abruptly. For example, induction heads form during a brief training window, and their formation coincides with a sharp improvement in in-context learning~\citeem{olsson2022incontextlearninginductionheads}, a transition we examine in detail in Section~\ref{sec:ld-induction-case}. In masked language models, suppressing the syntactic attention structure delays the capability it precedes~\citeest{chen_sudden_2024}. Likewise, sparse attention emerges abruptly and depends on repetition in the training data~\citeest{zucchet2025the}.

These events cluster in time. Several attention heads, such as induction heads and heads exhibiting syntactic attention patterns, can acquire specialised functions within the same training window. These heads emerge alongside a peak in how accurately the model's perplexity predicts human reading times, and this predictive ability subsequently declines even as language modelling performance continues to improve~\citeest{aoyama_language_2025}. Head specialisation can be viewed as a form of \emph{symmetry breaking}. Before or in the very early stages of training, attention heads are indistinguishable and interchangeable; during training, they start to develop distinct functions despite having equivalent roles in the architecture. Tracking this process can help identify the emergence of modular structure. Attribution of this kind is also a diagnostic for modularity more broadly. Three observable indicators jointly diagnose modular internal structure~\citeem{zhao2024distributional}: transitions during training, emergent trends across model scale, and heterogeneous clustering of solutions across random seeds. A structure exhibiting all three is a strong candidate for a discrete, functionally specialised mechanism.

\begin{insightbox}[Sharp Transitions from Smooth Components]
Phase transitions in test loss or task accuracy do not imply abrupt change in any individual component. For example, in controlled settings, many attention heads develop induction-like behaviour gradually---their contributions are largely additive, and the sharp macroscopic transition arises from their aggregate effect~\citeest{singh2024needsrightinductionhead}. Localising a phase transition in a coarse-grained metric therefore does not help identify the responsible components. In contrast, component-level attribution may find only gradual change without being wrong.
\end{insightbox}

\subsubsection{Why Transitions Happen}\label{sec:training_phase_why}

\paragraph{Saddle-to-Saddle Dynamics.} Theoretical studies of deep linear networks show that different singular modes of the input--output correlations can be learned at different rates~\citeest{saxe2014exact,saxe2019mathematical}. The training process may therefore involve periods of slow improvement followed by rapid reductions in test loss. From an optimisation perspective, gradient descent moves through regions near successive saddle points in the loss landscape and gradually learns more complex solutions along the way. This process is known as \emph{saddle-to-saddle dynamics}. Similar training patterns have been observed in other neural networks, where they learn simpler functions before more complex ones~\citeest{rahaman2019spectral,nakkiran2019sgd}. Attention head specialisation may also be related to symmetry breaking during training, although the relationship to saddle-to-saddle dynamics is not yet fully established in realistic transformers~\citeest{chen_sudden_2024}.

\paragraph{Singular Learning Theory.} Singular learning theory (SLT) studies how the geometry of a singular model's parameter space affects learning and generalisation~\citeest{watanabe2009algebraic}. Neural networks are singular (and are thus a subject of study in SLT) because different parameter configurations can implement the same function. As a result, two models that achieve similar loss can differ substantially in their effective complexity. SLT provides a framework for studying this complexity and has motivated research on its changes during training and its usefulness in downstream applications~\citeest{lau2023local}.

\begin{definitionbox}[Local Learning Coefficient (LLC)]
For model parameters $\mathbf{\theta}^*$ at a local minimum, the Local Learning Coefficient (LLC) $\lambda(\mathbf{\theta}^*)$ measures the degeneracy of the loss landscape near $\mathbf{\theta}^*$ via the volume-scaling rate of low-loss neighbourhoods~\citeest{lau2023local}. Smaller values indicate more degenerate (flatter, in a higher-order sense) minima. The LLC is estimable at scale with stochastic-gradient MCMC~\citeest{lau2023local}.
\end{definitionbox}

Estimated LLC trajectories in transformers change at points that coincide with identifiable developmental events, including the stages of structure formation described in Section~\ref{sec:ld-attribution}~\citeest{hoogland2025losslandscapedegeneracystagewise}. This is the central empirical result of the developmental interpretability programme~\citeem{hoogland2023towards} introduced in Section~\ref{subsec:dev interp}. However, current evidence for LLC is largely retrospective, where phase transitions are detected and matched to known events only after training. Prospective prediction of a transition before it occurs has not yet been demonstrated. Section~\ref{sec:ld-safety} discusses potential applications of learning-dynamics analysis to forecasting and intervention.

\subsubsection{Case Study: Induction Heads}
\label{sec:ld-induction-case}

\begin{examplebox}[Learning Dynamics Perspective on Induction Heads]
The preceding case studies (Sections~\ref{subsec: induc head case study: feature extract} and \ref{subsec: induc head case study: circuit}) analysed induction heads and circuits in a trained model. Several facts about the induction circuit are invisible from that vantage point but become observable when we examine the model across training~\citeem{olsson2022incontextlearninginductionheads}. Consider the repeated sequence again

\[
\texttt{A B C A B \underline{?}},
\]

and observe at successive checkpoints of a single training run. Examining the checkpoints shows that the induction ability arrived abruptly: at early checkpoints the model bases its predictions on token frequencies and derives no benefit from the earlier occurrence of \texttt{A B}, and during a narrow window of training, roughly 2.5 to 5 billion tokens in the models studied by \citetem{olsson2022incontextlearninginductionheads}, induction heads form and the model begins predicting \texttt{C} by attending back to the first occurrence. A sharp improvement in in-context learning and a small bump in the training loss fall within the same window.\\

The timing of formation also helps provide causal evidence that a single model checkpoint cannot provide. Ablating an induction head in the final model only shows its necessity for the behaviour at that checkpoint. This does not support the stronger claim that the induction circuit's formation caused the capability to develop. \citetem{olsson2022incontextlearninginductionheads} examined this stronger claim by intervening on formation itself. Their ``smeared key'' modification built the previous-token head's contribution into the attention computation, allowing the circuit to form within a single layer. Under this modification, one-layer models acquire in-context learning they otherwise never develop, while the transition moves earlier in deeper models.
\end{examplebox}

\subsection{Leveraging Learning Dynamics for Safety}
\label{sec:ld-safety}
The analyses above also support safety-oriented objectives, including anticipating capabilities before they appear (Section~\ref{sec:ld-forecasting}) and intervening in the training process while it is still underway (Section~\ref{sec:ld-intervening}).

\subsubsection{Forecasting Model Behaviours}
\label{sec:ld-forecasting}

\paragraph{Forecasting from Smaller Runs and Partial Runs.} Scaling laws demonstrate that aggregate loss is predictable before training~(\citeemalp{kaplan2020scaling}; \citeestalp{hoffmann2022training}), but phase transitions break this style of extrapolation for individual capabilities (Section~\ref{sec:ld-transitions}). Forecasting therefore targets specific behaviours directly. Memorisation is the behaviour for which the forecasting problem has been posed most explicitly. \citetest{biderman2023emergent} introduced the problem of forecasting whether a specific training sequence will be memorised by a model before that model is trained. They argued that recall is the metric to optimise (a false negative means training a full model that leaks the sequence, while a false positive only discards a cheap trial run). They established baselines that forecast from smaller fully trained models and from partially trained checkpoints of the target run, together with equi-compute recommendations for dividing a fixed forecasting budget between model size and tokens.

\paragraph{Forecasting Safety-Relevant Behaviours.}
Emergence can be forecast by intervening on the training trajectory. Fine-tuning on a target task shifts capability emergence towards less capable models. Fitting how this shift varies with the amount of fine-tuning data yields an \emph{emergence law}. Using only small models, this law predicts whether a capability will have emerged in models trained with up to 4$\times$ more compute~\citeem{snell2024predicting}.

Rare harmful behaviours can also be forecast statistically. The largest elicitation probabilities observed in a small sample of queries scale predictably with query volume. This relationship allows harmful-behaviour risks to be extrapolated across up to three orders of magnitude of deployment-scale query volume. Examples include assisting with dangerous chemical synthesis and taking power-seeking actions~\citeem{jones2025forecasting}.

Post-training failures can be analysed similarly. Specifically, transitions between successive RLHF checkpoints can be classified into four categories: reward hacking, collapse, proxy under-alignment, and evaluator-specific disagreement. Reward hacking occurs when the model exploits a way to achieve high rewards while the external quality from the perspective of human users falls; collapse occurs when both decline. This classification localises failures that checkpoint averages miss. A model fitted to training-dynamics diagnostics predicts future reward hacking with a ROC-AUC of 0.82. These diagnostics include reward-model uncertainty, policy drift, and output diversity~\citeem{abahana2026rlhf}.

Internal signals extend forecasting to fine-tuning outcomes. Activation-space persona features predict whether a fine-tuned model will exhibit emergent misalignment~\citeest{wang2026persona}. Signals available before fine-tuning begins also predict the resulting model’s alignment~\citeem{zhang2026whatshapes}. Section~\ref{subsec: iat} discusses the research agenda these results instantiate.

\paragraph{Internal Precursors as Early Warning.} Where a mechanism forms before the capability it supports is behaviourally expressed, monitoring internals can flag the capability earlier than behavioural evaluation. The recall circuits that precede factual-knowledge acquisition~\citeem{zucchet2025language} and the syntactic attention structure that precedes the break in grammatical capabilities~\citeest{chen_sudden_2024} show that this lead time can exist; progress measures (Section~\ref{sec:ld-progress}) are the instrument that would exploit it.

\paragraph{Run-to-Run Variance.} To assess the reliability of a forecast, researchers can compare training runs with different random seeds. \citetest{vanderwal2025polypythias} found that capabilities generally emerge at similar stages and representations develop consistently across training runs, although some runs differ substantially. These exceptions highlight the uncertainty of forecasts based on a single training run.

\begin{warningbox}[Most Early-Warning Evidence Is Retrospective]
Currently, nearly all demonstrations that changes in internals lead to changes in behaviour are retrospective. In these studies, researchers analyse the completed mechanism after training to identify a progress measure or precursor structure, then evaluate it backwards over saved checkpoints. This concern also exists in LLC-based transition detection (Section~\ref{sec:training_phase_why}).\\

In contrast, prospective forecasting requires researchers to predict outcomes before observing them. \citetest{biderman2023emergent} proposed baseline methods for forecasting memorisation. However, prospective forecasting for broader capability transitions relevant to AI safety remains underdeveloped. Early-warning signals therefore require prospective validation before they can be considered reliable.

\end{warningbox}

\subsubsection{Intervening on Training Processes}
\label{sec:ld-intervening}

The model that matters is the one deployed at the end of training, and the standard response to an undesirable behaviour found there is post-training fine-tuning, which is outside the scope of this section. An alternative is to intervene during training to prevent undesirable capabilities from developing or to control how they develop. Existing work explores three approaches: modifying training data, changing model architecture, and intervening on model activations during fine-tuning.

\paragraph{Pretraining Data Filtering: Deep Ignorance.} Deep Ignorance filters proxies for biohazardous knowledge out of the pretraining corpus and trains 6.9-billion-parameter models on the filtered data (Section~\ref{sec:ld-data}; \citeestalp{obrien2026deep}). The filtered models do not acquire the targeted knowledge, making it difficult for the unwanted capability to emerge through adversarial fine-tuning. In contrast, post-training unlearning may suppress the unwanted behaviour without removing the underlying knowledge. A follow-up moves the filter from documents to tokens: individual tokens are labelled using sparse autoencoders and efficient classifiers trained on those labels, and filtering at token granularity removes the undesired capability at lower cost to desired capabilities than document-level filtering, with effectiveness increasing with model scale across the two orders of magnitude tested and filtered models remaining alignable afterwards~\citeem{rathi2026shaping}.

Similarly, the order in which different categories of training data are presented matters. \citetem{bruijns2026pretrainingcurriculaenableselective} found that, in an imbalanced pretraining curriculum where one task is shown before another, the model packs the tasks sequentially into separable circuits, resulting in more targeted subsequent refusal fine-tuning.

\begin{warningbox}[Data-Level Interventions Have Ripple Effects]
Removing data from a training set does not act only on the removed data. In the setting of \citetest{borkar2025privacy}, deleting personal information can increase memorisation of the personal information that remains, and adding one individual's information amplifies memorisation of others' by roughly 7.5$\times$, because whether a sequence is memorised depends on similar sequences seen elsewhere in training. Subliminal learning~\citeest{cloud2026language} shows that data can transmit traits not apparent from its semantic content. For example, a student model can inherit a teacher's hidden bias from seemingly unrelated data, such as number sequences. This transfer is carried by a small set of divergence tokens and depends on early layers~\citeest{schrodi2026towards}. Filtering and unlearning interventions therefore need to be evaluated on what the model does with the data that stays, not only on what it no longer does with the data that was removed.
\end{warningbox}

\paragraph{Architectural Modification: Gradient Routing.} Gradient routing is an architectural modification informed by learning dynamics: it constrains which parameters receive gradient updates for chosen concepts during training, localising the corresponding computation into designated regions of the network~\citeem{cloud2024gradient}. Although the modification does not make the trained model safer on its own, it shapes where a mechanism forms so that later interventions, such as gating the designated region, become clean and localised. The antecedent for this style of intervention is the smeared-key modification~\citeem{olsson2022incontextlearninginductionheads}, which showed that an architectural change can control when and whether a mechanism forms at all (Section~\ref{sec:ld-induction-case}).

\paragraph{Monitoring and Steering During Fine-Tuning.} The case excluded above uses fine-tuning to repair an undesirable behaviour after it has been identified. Intervening on the fine-tuning process itself remains within scope, and persona vectors provide a current example of this approach. A persona vector is an activation-space direction corresponding to a character trait such as sycophancy, hallucination propensity, or malice, extractable automatically from a natural-language trait description~\citeem{chen2025persona}. Intended and unintended personality changes during fine-tuning correlate strongly with movement along the relevant vector, which yields the monitoring instrument. Preventative steering along the vector during fine-tuning limits acquisition of the trait. Before fine-tuning, projecting training data onto the vector flags data that would induce the trait, at both the dataset and individual-sample levels~\citeem{chen2025persona}. This instantiates each step of the intervention loop below for the fine-tuning stage.

\begin{practicebox}[Train-Time Intervention Loop]
The work discussed so far lays the methodological foundation for intervening in a training run in progress. Below we present a general pipeline.

\begin{enumerate}[nosep]
   \item \textbf{Monitor} a target quantity at checkpoints, such as a capability-specific metric (Section~\ref{sec:ld-detect}) or a progress measure on model internals (Section~\ref{sec:ld-progress}).
   \item \textbf{Define} the acceptable range for that quantity in advance, including the response to a breach.
   \item \textbf{Intervene} when the range is breached, such as by adjusting the data mixture or filtering out unwanted data, applying a targeted architectural or regularisation change, rolling back to a pre-breach checkpoint, and resuming under the adjusted setup.
   \item \textbf{Re-evaluate} the model after the intervention to verify that the targeted capability has been suppressed and does not re-emerge during further training.
\end{enumerate}
No published system implements this loop end to end at scale. Instead, existing studies have demonstrated individual steps of the proposed workflow.
\end{practicebox}

\paragraph{Open Questions.}
Three questions remain about the applicability of these training-time interventions. First, effective data filtering requires identifying which training data contributes to the capability we want to remove. Token-level filtering is motivated by data attribution, but it remains unclear whether methods such as influence functions (Section~\ref{sec:influence-functions}) can reliably identify the relevant data. Second, scale: Deep Ignorance operates at 6.9 billion parameters, and while token-level filtering became more effective across the tested scale range~\citeem{rathi2026shaping}, it remains open whether data filtering stays effective and affordable at frontier scale. Third, deployed models show that stronger train-time safeguards are achievable, but not how: gpt-oss models are substantially harder to fine-tune than comparable open-weight models, and their developers have not fully disclosed their responsible training procedure~\citeem{openai2025gptoss120bgptoss20bmodel}. Understanding how this resistance is achieved could help develop similar safeguards for other open-weight models and clarify how fine-tuning changes model behaviour.

\paragraph{Summary.}
Learning dynamics can be studied both as an object of explanation, revealing how representations and mechanisms evolve during training, and as an interpretability tool to localise mechanisms, detect phase transitions, and test causal hypotheses through training interventions. Understanding learning dynamics can also help predict when capabilities emerge and guide interventions during training. However, no complete system combining monitoring and intervention has yet been demonstrated at scale. This section focuses on how models change during training, while Section~\ref{sec:influence-functions} examines how individual training examples contribute to model behaviour.

\section{Training Data Attribution}
\label{sec:influence-functions}

For a model behaviour of interest, Sections~\ref{sec:sae}--\ref{sec:training} study the internal features, components, and their interactions and evolutions that produce the behaviour. Training data attribution (TDA) shifts the focus from model internals to training data, searching for training samples that strongly shaped the model behaviour. TDA applications include, but are not limited to, identifying influential training samples, detecting train--test leakage, and diagnosing model failures without retraining~(\citeestalp{liu2025olmotrace, park2023trakattributingmodelbehavior}; \citeemalp{grosse2023studyinglargelanguagemodel}). Although TDA is not generally a mechanistic method under this survey’s criteria, it can complement mechanistic interpretability in many ways. For example, by revealing how specific training samples shape malicious behaviour, TDA can enable effective training data filtering to prevent unwanted behaviours or even their internal mechanisms from forming during the training of next-generation models. Motivated by this application, this section focuses on post-hoc training data attribution methods that link a given target model’s behaviour to its training samples. This section excludes techniques for intervening in the data pipeline, such as filtering pretraining corpora, rebalancing demographic representation, and scheduling memorisation.

TDA falls into two complementary approaches. \emph{Verbatim data attribution} detects explicit memorisation by identifying exact or near-exact string matches between model outputs and the training corpus. These searches can scale to trillions of tokens using techniques such as suffix arrays and infini-gram models. \emph{Gradient-based data attribution} estimates how training samples affect model parameters and predictions and can capture influential samples that do not manifest as verbatim reproduction. The two types of methods provide orthogonal views: one answers ``did the model use rote memorisation on a training sample to answer this question?''; the other answers ``how much did this training sample influence this certain behaviour of this model?''

\subsection{Verbatim Data Attribution: OLMoTrace and Maximal Matching Spans}

Earlier corpus search tools such as the ROOTS Search Tool~\citeest{piktus-etal-2023-roots} and GAIA Search~\citeest{piktus2023gaia} established the practice of querying pretraining corpora for transparency and data exploration. OLMoTrace~\citeest{liu2025olmotrace} scales this to trillions of tokens using a combination of suffix arrays, infini-gram models, and maximal matching span extraction~\citeest{liu2025infinigramscalingunboundedngram}, making it less a data-attribution method in the influence-function sense and more a retrieval tool for identifying content the model may have memorised.

\begin{definitionbox}[Suffix Arrays (SA)]
\label{def:suffix-array}
The \emph{suffix array} of a string $x$ of length $T_1$ is an array of integers representing the starting positions of all suffixes of $x$ in lexicographic order. It enables efficient substring searches in $\mathcal{O}(T_2 \cdot \log T_1)$ time for a query of length $T_2$, compared to $\mathcal{O}(T_2 \cdot T_1)$ naive search.\\

\paragraph{Example:} $x = \texttt{banana\$}$
\begin{align*}
\text{Suffixes: } & 0:\texttt{banana\$}, 1:\texttt{anana\$}, 2:\texttt{nana\$}, 3:\texttt{ana\$}, 4:\texttt{na\$}, 5:\texttt{a\$}, 6:\texttt{\$} \\
\text{SA}(x) &= [6, 5, 3, 1, 0, 4, 2]
\end{align*}
\end{definitionbox}

\begin{definitionbox}[Infini-Gram Models]
\label{def:infini-gram}
An \emph{infini-gram model} extends the n-gram model to arbitrary-length prefixes, using backoff to the longest matching sequence in the training corpus. Formally, for a token $w_i$:
\[
P_{\infty}(w_i \mid w_{1:i-1}) = \frac{\text{cnt}(w_{i-n+1:i} \mid \mathcal{D})}{\text{cnt}(w_{i-n+1:i-1} \mid \mathcal{D})}, 
\quad n = \max \{ n' : \text{cnt}(w_{i-n'+1:i-1} \mid \mathcal{D})>0\},
\]
where $w_{1:i-1}$ are preceding tokens and $\mathcal{D}$ is the training corpus.\\

Infini-grams are implemented on top of suffix arrays, which allow fast substring queries across massive corpora.
\end{definitionbox}

\begin{definitionbox}[Maximal Matching Spans]
\label{def:maximal-matching-spans}
Given a model output and a training corpus, a \emph{maximal matching span} is a contiguous sequence of tokens that:
\begin{enumerate}
    \item Appears verbatim at least once in the training data,
    \item Does not break words or include spurious delimiters (e.g., periods or newlines) except at span boundaries, and
    \item Is not contained within a longer span that satisfies 1. and 2.
\end{enumerate}

OLMoTrace identifies maximal matching spans by combining suffix-array search with infini-gram queries, adjusting boundaries for tokenisation, and retaining only the longest non-overlapping spans.
\end{definitionbox}

\begin{examplebox}[OLMoTrace Pipeline]
\label{ex:olmotrace-pipeline}

\paragraph{Language Model Output:} ``Celine Dion is a Canadian singer known for her powerful voice and wide [...]''\\

\paragraph{Pipeline Steps:}
\begin{enumerate}
    \item Tokenise the language model output.
    \item For each token starting a new word, query the suffix array with infini-gram to find the longest matching prefix in the training corpus.
    \item Adjust span boundaries to avoid breaking words or including delimiters.
    \item Retain only maximal spans (not contained in longer matches).
\end{enumerate}

\paragraph{Example Spans Found:}
\begin{itemize}
    \item ``Celine Dion [...] Born on March 30, 1968, in''
    \item ``Charlemagne, Quebec, Canada, Dion began''
    \item ``performing at a young age and gained [...]''
    \item ``Her career took off when she won the 1988''
    \item ``Eurovision Song Contest with the song 'Ne''
\end{itemize}

This process links language model output directly to the training corpus, enabling identification and attribution of memorised content.
\end{examplebox}

\begin{definitionbox}[OLMoTrace Algorithm]
\label{def:olmotrace-algorithm}
\paragraph{Inputs:} Model output tokens $S_1^L$, training corpus tokens $T_1^N$, suffix array $A_1^N$\\

\paragraph{Steps:}
\begin{enumerate}
    \item Scan each token position $t$ in $S_1^L$, considering only tokens starting a new word.
    \item Query the suffix array $A_1^N$ to find candidate matching positions in the training corpus.
    \item Use infini-gram queries to determine the longest matching prefix starting at $t$, backing off to shorter n-grams if necessary.
    \item Adjust the match length to avoid breaking words or including spurious delimiters.
    \item Collect all spans $(t, t+\text{length})$ that satisfy the above criteria.
    \item Filter out spans contained within longer overlapping spans to retain only maximal matches.
\end{enumerate}
\end{definitionbox}

\paragraph{Limitations as a Mechanistic Interpretability Tool.}
Verbatim attribution excels at factual memorisation but struggles with compositional or synthesised outputs. For tasks like translation, poetry, or reasoning, OLMoTrace (and other verbatim matching methods) can identify spans of text that appear in the training data, but these matches do not explain how the model combined information from multiple sources to produce the output.

\begin{examplebox}[OLMoTrace on Multilingual Output]
\label{ex:olmotrace-multilingual}
\paragraph{Prompt:} ``Write a short story in Chinese about a girl who loves reading.''\\

\paragraph{Completion (Chinese):}
\begin{quote}
\begin{CJK*}{UTF8}{gbsn}
``\ldots 一个叫雪儿的小公主，她有一只可爱的小狐狸伴在身边。她喜欢坐在竹林间看着远方。某日，一场突如其来的山洪导致雪\ldots''
\end{CJK*}
\end{quote}

\paragraph{OLMoTrace Influential Samples Identified:}
OLMoTrace identified maximal matching spans across training datasets:
\begin{itemize}
    \item \begin{CJK*}{UTF8}{gbsn}Span: ``一个叫雪儿的小公主，她有一只可爱的小狐狸伴在身边。她喜欢坐在竹林间看着远方''---matched in pretraining corpus\end{CJK*}
    \item \begin{CJK*}{UTF8}{gbsn}Span: ``她喜欢阅读关于科技、心理学和历史的书籍，业余时间还会参加编程课''---matched in post-training (instruction-tuning) data\end{CJK*}
\end{itemize}

This example demonstrates how OLMoTrace traces multilingual outputs across diverse training sources, linking model generations directly to their memorised origins in pretraining and post-training datasets.
\end{examplebox}

\subsection{Gradient-Based Attribution: Understanding Data Influence Through Perturbations}
Gradient-based methods estimate how much each training sample influenced the model's predictions.
  
\subsubsection{Core Idea: Parameter Response to Training Data}

\emph{Influence functions} estimate how a model's parameters and predictions would change if a training sample were slightly upweighted~(\citeestalp{koh2020understandingblackboxpredictionsinfluence}; \citeemalp{grosse2023studyinglargelanguagemodel}). We introduce their core idea below.

Consider a neural network parameterised by $\theta$. After sufficient training, its parameters will converge to
\[
\theta^*
=
\arg\min_\theta
\mathcal{L}_{\mathrm{train}}(\theta),
\]
where $\mathcal{L}_{\mathrm{train}}(\cdot)$ is the aggregate training objective calculated as:
\[
\mathcal{L}_{\mathrm{train}}(\theta)
\coloneqq
\frac{1}{n}
\sum_{i=1}^{n}
\mathcal{L}(z_i,\theta).
\]
$\{z_i \mid i=1, \ldots, n\}$ is the training set and $\mathcal{L}(\cdot)$ is a loss function such as cross-entropy loss. Note that although by definition $\theta^*$ is expected to be a global optimum, it will not affect the subsequent analysis if $\theta^*$ is in fact a local optimum.

Now, suppose we upweight a training sample \(z_m\) by a small amount \(\epsilon\). The new aggregate training objective will become:
\[
\mathcal{L}_{\epsilon}(\theta)
\coloneqq
\mathcal{L}_{\mathrm{train}}(\theta)
+
\epsilon \mathcal{L}(z_m,\theta).
\]
It corresponds to a new set of model parameters $\theta^*_{\epsilon}$ expressed as:
\[
\theta^*_{\epsilon}
=
\arg\min_\theta
\mathcal{L}_{\epsilon}(\theta).
\]
This expression is not a closed-form formula and does not provide us with an efficient way to acquire $\theta^*_{\epsilon}$. A brute-force way is to train the neural network with this new objective and observe the new converged parameters; however, this method is extremely computationally expensive. We thus turn to leveraging the specific property of $\theta^*_{\epsilon}$. Note that, by definition, it must satisfy the first-order optimality condition:
\[
\nabla_\theta \mathcal{L}_{\epsilon}(\theta)\Big|_{\theta=\theta^*_{\epsilon}}
=
\nabla_\theta \mathcal{L}_{\mathrm{train}}(\theta^*_{\epsilon})
+
\epsilon\, \nabla_\theta \mathcal{L}(z_m, \theta^*_{\epsilon})
=
0.
\]
This motivates defining a function to link $\theta$ and $\epsilon$:
\[
F(\theta, \epsilon)
:=
\nabla_\theta \mathcal{L}_{\mathrm{train}}(\theta)
+
\epsilon\, \nabla_\theta \mathcal{L}(z_m, \theta). \]
Note that $F(\theta^*, 0)=0$, which allows us to consider the implicit function theorem.

To apply the implicit function theorem, we assume that $F(\theta, \epsilon)$ is continuously differentiable in a small area around $(\theta, \epsilon) = (\theta^*, 0)$. Furthermore, the Jacobian of $F(\theta, \epsilon)$ with respect to $\theta$ at $(\theta^*, 0)$, i.e.,
\[
\frac{\partial F(\theta, \epsilon)}{\partial \theta}\bigg |_{\theta = \theta^*, \epsilon = 0}
=
\nabla_{\theta}^{2}
\mathcal{L}_{\mathrm{train}}(\theta^*)
\eqqcolon H_{\theta^*},
\]
is invertible. $H_{\theta^*}$ denotes the Hessian of the training loss. These assumptions allow us to apply the implicit function theorem. 

By the implicit function theorem, there exists a neighbourhood $V$ of $\epsilon = 0$ and a unique continuously differentiable function $\epsilon \mapsto \theta^*_{\epsilon}$ on $V$ such that $F(\theta^*_{\epsilon}, \epsilon) = 0$ for all $\epsilon \in V$. In other words, for sufficiently small $\epsilon$, $\theta^*_{\epsilon}$ is the stationary point of $\mathcal{L}_{\epsilon}(\cdot)$ near $\theta^*$. Moreover, its derivative is given by
\[
\frac{d\theta^*_{\epsilon}}{d\epsilon}
=
-
\left[
\frac{\partial F(\theta, \epsilon)}{\partial \theta} \bigg|_{\theta=\theta^*_{\epsilon}}
\right]^{-1}
\frac{\partial F(\theta^*_{\epsilon}, \epsilon)}{\partial \epsilon}.
\]
Recalling the expression of $F(\theta, \epsilon)$, we also have
\[
\frac{\partial F(\theta, \epsilon)}{\partial \epsilon}\bigg|_{\theta=\theta^*_{\epsilon}}
=
\nabla_\theta \mathcal{L}(z_m, \theta)\Big|_{\theta=\theta^*_{\epsilon}}.
\]
Putting $\epsilon = 0$ to let $\theta^*_\epsilon = \theta^*_{0} = \theta^*$ and $\frac{\partial F(\theta, \epsilon)}{\partial \theta} \bigg|_{\theta=\theta^*_{\epsilon}} = H_{\theta^*}$, we have:
\[
\frac{d\theta^*_{\epsilon}}
     {d\epsilon}
\Bigg|_{\epsilon=0}
=
-
H_{\theta^*}^{-1}
\nabla_\theta
\mathcal{L}(z_m,\theta^*).
\]
This quantity describes how the model parameters respond to an infinitesimal increase in the importance of the training sample \(z_m\).

To measure the effect on a test-time quantity $M(\theta, z_{\text{test}})$ (e.g., the model's test loss on the testing sample $z_{\text{test}}$), we apply the chain rule:
\[
\mathcal{I}_{M}(z_m, z_{\text{test}})
\coloneqq
\frac{d M(\theta^*_{\epsilon}, z_{\text{test}})}
     {d\epsilon}
\Bigg|_{\epsilon=0}
=
-
\nabla_\theta M(\theta^*, z_{\text{test}})^\top
H_{\theta^*}^{-1}
\nabla_\theta
\mathcal{L}(z_m,\theta^*).
\]
The quantity $\mathcal{I}_{M}(z_m, z_{\text{test}})$ is called the \emph{influence score}. Taking a step forward, by the definition of the derivative, for sufficiently small $\epsilon$ we have
\[
M(\theta^*_{\epsilon}, z_{\text{test}})
-
M(\theta^*, z_{\text{test}})
\approx
\epsilon
\,\mathcal{I}_{M}(z_m, z_{\text{test}}).
\]
That is, the influence score quantifies the first-order effect of the training sample $z_m$ on the model's downstream loss for a testing sample $z_{\text{test}}$.

\subsubsection{Proximal Bregman Response and the Generalised Gauss--Newton}

Empirical results show that the original form of influence functions introduced above performs poorly in practice. This is due to multiple assumptions that rarely hold in practice. For example, for a trained neural network, its parameters $\theta$ are scarcely exactly at the local or global minimum $\theta^*$. Additionally, the Hessian $H_{\theta^*}$ is usually non-invertible or numerically unstable due to some small eigenvalues. These limitations motivate researchers to formulate influence functions without unrealistic assumptions. One classical approach is called the \emph{proximal Bregman response function} (PBRF)~\citeest{bae2022influencefunctionsanswerquestion}. It targets model parameters which minimise the new training objective while staying close to the original parameters. Below, we sketch its core idea.

Let each training sample $z_i=(x_i, y_i)$, where $x_i$ is the input and $y_i$ is the label, and $u_i(\theta) =f(x_i, \theta)$ is the output of the model (parameterised as $\theta$) given input $x_i$, such as a logit vector. Slightly abusing notation, we denote the per-sample training loss as
\[
\mathcal{L}(z_i, \theta) 
\coloneqq 
\mathcal{L}(u_i(\theta), y_i)
=
\mathcal{L}(f(x_i, \theta), y_i),
\]
where $\mathcal{L}(\cdot, y)$ is a loss function as in the vanilla influence function formulation. Here we additionally assume $\mathcal{L}(\cdot, y)$ is convex in its first argument $u$, which many commonly used loss functions, such as cross-entropy loss, satisfy.

As Bregman divergence will be used in the later formulation, we introduce this notion as the next step. Specifically, for a given convex function $\mathcal{L}(u, y)$ (as described above) with the label $y$ fixed, the Bregman divergence of two model outputs, $u$ and $u'$, is:
\[
D_{\mathcal{L}}(u, u'; y) = \mathcal{L}(u, y) - \mathcal{L}(u', y) - \big(\nabla_v \mathcal{L}(v, y)\big|_{v = u'}\big)^\top (u - u'),
\]
i.e., the amount by which $\mathcal{L}(\cdot, y)$ at $u$ exceeds its first-order Taylor approximation from $u'$.

Now we introduce the PBRF formulation. Starting from the upweighted objective $\mathcal{L}_{\epsilon}(\theta)$, PBRF is represented as:
\[
\mathcal{P}_{\epsilon}(\theta)
\coloneqq
\frac{1}{n}
\sum_{i=1}^{n}
D_{\mathcal{L}}\big(f(x_i, \theta),\, f(x_i, \theta^s); y_i\big)
+
\epsilon\, \mathcal{L}(z_m, \theta)
+
\frac{\lambda}{2}\|\theta - \theta^s\|^2,
\qquad
\theta^{\mathrm{PBRF}}_{\epsilon}
\coloneqq
\arg\min_\theta
\mathcal{P}_{\epsilon}(\theta),
\]
where $\theta^s$ denotes the original trained model parameters (note that it is not required to be the local or global optimum). Bregman divergence replaces the original loss function. An additional proximity term is added with a positive regularisation coefficient $\lambda$. $\theta^{\mathrm{PBRF}}_{\epsilon}$ searches for the set of model parameters that best corresponds to an upweight in $z_m$ while keeping its predictions on the overall training set close to the original predictions and staying close to $\theta^s$.

In this construction, both the Bregman and the proximity terms are non-negative for all $\theta$ and vanish at $\theta = \theta^s$. Since these are the only two terms of $\mathcal{P}_{\epsilon}$ when $\epsilon = 0$, $\theta^s$ is a global minimiser of $\mathcal{P}_0$ (i.e., $\theta^s = \theta^{\mathrm{PBRF}}_0$) \emph{regardless of whether training has converged}.

As in the vanilla formulation, $\theta^{\mathrm{PBRF}}_\epsilon$ does not have a closed-form formula for now. However, by observing that $\nabla_\theta \mathcal{P}_{\epsilon}(\theta)\big|_{\theta=\theta^{\text{PBRF}}_{\epsilon}}=0$, we define
\[
\tilde{F}(\theta, \epsilon)
\coloneqq
\nabla_\theta \mathcal{P}_{\epsilon}(\theta)
=
\frac{1}{n}
\sum_{i=1}^{n}
\nabla_\theta D_{\mathcal{L}}\big(f(x_i, \theta),\, f(x_i, \theta^s); y_i\big)
+
\epsilon\, \nabla_\theta \mathcal{L}(z_m, \theta)
+
\lambda (\theta - \theta^s),
\]
where $\tilde{F}(\theta^s, 0) = 0$. We assume that $f(x_i, \cdot)$ and $\mathcal{L}(\cdot, y)$ are twice continuously differentiable such that $\tilde{F}$ is continuously differentiable in a small area around $(\theta^s, 0)$. Then, we verify that $\frac{\partial\tilde{F}(\theta, \epsilon)}{\partial \theta}$ is invertible at $(\theta, \epsilon) = (\theta^s, 0)$, the last prerequisite for using the implicit function theorem. By direct calculation (we omit the detailed process here), we can obtain
\[
\frac{\partial \tilde{F}(\theta, \epsilon)}{\partial \theta}\bigg|_{\theta = \theta^s, \epsilon = 0}
=
\underbrace{
\frac{1}{n}\sum_{i=1}^{n} J_i^\top H_i J_i
}_{\eqqcolon\, G}
+
\lambda I,
\]
where $u^s_i \coloneqq u_i(\theta^s)$, $J_i \coloneqq \frac{\partial u_i(\theta)}{\partial \theta} \big|_{\theta=\theta^s}$ is the output--parameter Jacobian, and $H_i \coloneqq \nabla^2_u \mathcal{L}(u, y_i) \big|_{u = u^s_i}$ is the loss--output Hessian. The matrix $G$ is the \emph{generalised Gauss--Newton} (GGN) matrix. Notably, as $H_i$ is positive semidefinite by the convexity of $\mathcal{L}(\cdot, y_i)$, we know $G$ is positive semidefinite (because $v^\top G v = \frac{1}{n}\sum_i (J_i v)^\top H_i (J_i v) \geq 0$ for any $v$). As a result, $G + \lambda I$ is positive definite and hence invertible. Unlike the case of $\nabla^2_\theta \mathcal{L}_{\mathrm{train}}(\theta^*)$, the invertibility of $G + \lambda I$ holds by construction.

Knowing that $\partial \tilde{F}(\theta, \epsilon) / \partial \epsilon = \nabla_\theta \mathcal{L}(z_m, \theta)$, we apply the implicit function theorem and obtain
\[
\frac{d\theta^{\mathrm{PBRF}}_{\epsilon}}{d\epsilon}
\Bigg|_{\epsilon=0}
=
-
(G + \lambda I)^{-1}
\nabla_\theta
\mathcal{L}(z_m,\theta^s).
\]
By the chain rule, the influence score on a test-time quantity $M(\theta, z_{\text{test}})$ becomes
\[
\mathcal{I}^{\mathrm{PBRF}}_{M}(z_m, z_{\text{test}})
\coloneqq
\frac{d M(\theta^{\mathrm{PBRF}}_{\epsilon}, z_{\text{test}})}{d\epsilon}
\Bigg|_{\epsilon=0}
=
-
\nabla_\theta M(\theta^s, z_{\text{test}})^\top
(G + \lambda I)^{-1}
\nabla_\theta
\mathcal{L}(z_m,\theta^s).
\]
As shown above, the influence function computed with $G + \lambda I$ is exactly the derivative of the PBRF at $\epsilon = 0$. \citetest{bae2022influencefunctionsanswerquestion} confirmed empirically that such influence estimates align well with the actual PBRF.

\subsubsection{Addressing the Computational and Memory Bottlenecks of GGN}

The size of the GGN matrix $G \in \mathbb{R}^{p \times p}$ (where $p$ is the number of model parameters) can be extremely large for modern language models, which makes the calculation of $(G + \lambda I)^{-1}$ computationally expensive. As a result, follow-up methods have avoided explicitly calculating this inverse matrix by borrowing techniques from the optimisation literature.

\paragraph{Kronecker-Factored Approximate Curvature (K-FAC).}
\citetest{kfac} proposed \emph{Kronecker-Factored Approximate Curvature} (K-FAC) to approximate a large Fisher information matrix (FIM) using a set of smaller matrices. For a trained neural network parameterised by $\theta^s$, the FIM is defined as
\[
\mathbf{F}
\coloneqq
\frac{1}{n}\sum_{i=1}^{n}
\mathbb{E}_{\hat{y} \sim p(\cdot \mid x_i, \theta^s)}
\left[
\nabla_\theta \log p(\hat{y} \mid x_i, \theta^s)
\nabla_\theta \log p(\hat{y} \mid x_i, \theta^s)^\top
\right],
\]
where $\hat{y}$ is a sample from the model's prediction distribution on $x_i$. For example, suppose the task is three-class classification ($c_1, c_2, c_3$) and the model's prediction distribution is $\pi_i=\mathrm{softmax}(u_i^s) = (0.5, 0.3, 0.2)$. If a sampled label is $\hat{y}=c_2$, then $\log p(\hat{y} \mid x_i, \theta^s) = \log 0.3$.

Although K-FAC approximates the FIM rather than the GGN, these matrices are equivalent for common loss functions under suitable probabilistic assumptions. For example, with softmax cross-entropy, the expected outer product of the output log-likelihood gradients can be shown to equal the loss--output Hessian:
\[
\mathbb{E}_{\hat{y} \sim p(\cdot \mid x_i, \theta^s)}
\left[
\nabla_u \log p(\hat{y} \mid u)
\nabla_u \log p(\hat{y} \mid u)^\top
\right]_{u=u_i^s}
=
\mathrm{diag}(\pi_i)-\pi_i\pi_i^\top
=
H_i.
\]
Applying the chain rule then gives
\[
\mathbf{F}
=
\frac{1}{n}\sum_{i=1}^{n}J_i^\top H_i J_i
=
G.
\]
This shows that we can use K-FAC to approximate the GGN used in the PBRF influence score.

Now we return to K-FAC, which efficiently approximates the FIM. First, it ignores curvature between layers and approximates $\mathbf{F}$ as a block-diagonal matrix with one block per layer. Consider a neural network in which each layer $\ell$ has a weight matrix $W_\ell \in \mathbb{R}^{d_{\mathrm{out}}\times d_{\mathrm{in}}}$, input activations $h_{\ell-1}$, and pre-activations $s_\ell=W_\ell h_{\ell-1}$ (unrelated to the superscript $s$ in $\theta^s$). Following \citetem{grosse2023studyinglargelanguagemodel}, the \emph{pseudo-gradient} is defined as $\mathcal{D}_{s_\ell} \coloneqq \nabla_{s_\ell} \log p(\hat{y}\mid x,\theta^s) \in \mathbb{R}^{d_{\text{out}}}$---the gradient flowing back to ${s_\ell}$. The pseudo-gradient of the layer weights is $\mathcal{D}_{s_\ell} h_{\ell-1}^\top$. Its vectorisation gives $h_{\ell-1}\otimes\mathcal{D}_{s_\ell}$, where $\otimes$ denotes the Kronecker product. Leveraging $\otimes$'s properties that $(A\otimes B)^\top = A^\top \otimes B^\top$ and $(A\otimes B)(C\otimes D) = (AC)\otimes (BD)$, the block for this layer in the FIM, denoted as $\mathbf{F}_\ell$, can be represented as
\[
\mathbf{F}_\ell
=
\underbrace{\mathbb{E}\left[
(h_{\ell-1}h_{\ell-1}^\top)
\otimes
(\mathcal{D}_{s_\ell}\mathcal{D}_{s_\ell}^\top)
\right]}_{\in \mathbb{R}^{d_{\text{in}} d_{\text{out}}\times d_{\text{in}} d_{\text{out}}}}
\approx
\underbrace{
\mathbb{E}[h_{\ell-1}h_{\ell-1}^\top]
}_{\eqqcolon\,\textbf{A}_{\ell-1} \in \mathbb{R}^{d_{\text{in}}\times d_{\text{in}}}}
\otimes
\underbrace{
\mathbb{E}[\mathcal{D}_{s_\ell}\mathcal{D}_{s_\ell}^\top]
}_{\eqqcolon\,\textbf{S}_\ell \in \mathbb{R}^{d_{\text{out}}\times d_{\text{out}}}}.
\]
The approximation separates the expectation into two independent factors: activations and pseudo-gradients. K-FAC thus stores only $d_{\text{in}}^2+d_{\text{out}}^2$ entries instead of $(d_{\text{in}}d_{\text{out}})^2$ entries for each block. When the two factors are invertible, their Kronecker product can be inverted separately:
\[
(\textbf{A}_{\ell-1}\otimes \textbf{S}_\ell)^{-1}
=
\textbf{A}_{\ell-1}^{-1}\otimes \textbf{S}_\ell^{-1}.
\]
This substantially reduces the memory required to store $G$ and the computational cost of computing the inverse of $G+\lambda I$.

\paragraph{Eigenvalue-Corrected K-FAC (EK-FAC) and ASTRA.}
EK-FAC~\citeest{ekfac} improved K-FAC by correcting its eigenvalues. Recent work has adopted this technique to scale the influence function method to modern LLMs~\citeem{grosse2023studyinglargelanguagemodel}. ASTRA~\citeest{wang2025better} further used EK-FAC as a preconditioner for Neumann series iterations, achieving higher-quality GGN-vector products with less hyperparameter tuning. These efforts make influence functions increasingly practical for modern language models with billions of parameters.

\subsubsection{Retraining-Based Influence Estimation: datamodels and TRAK}
\label{sec:trak}

Some methods have explored retraining the model to estimate the attribution. The most naive method is to upweight or remove each training sample, retrain the model on the modified training set, and observe changes in the model's output on the testing sample. \citetest{ilyas2022datamodelspredictingpredictionstraining} proposed \emph{datamodels} to more efficiently estimate each training sample's attribution on a testing sample by retraining many models on random subsets of the training data and fitting a linear regression that predicts the model output on the testing sample from binary indicators of whether each training sample was included. Each fitted coefficient then represents the training sample's influence on the testing sample.

However, datamodels remains computationally expensive because it requires training many models. \emph{Tracing with the Randomly-projected After Kernel} (TRAK; \citeestalp{park2023trakattributingmodelbehavior}) provides a more efficient alternative. TRAK first uses a first-order Taylor expansion around the model parameters to approximate a trained model as a linear model. It then applies random projections to reduce gradient dimensionality and estimate training-sample influence in this lower-dimensional space. Finally, it averages the estimates across a small number of trained models to reduce variability due to training randomness. TRAK achieves attribution quality comparable to datamodels at substantially lower computational cost~\citeest{park2023trakattributingmodelbehavior}.

\subsubsection{Efficient Influence Estimation Using Training Trajectories: TracIn}
\label{sec:tracin}

TracIn~\citeest{pruthi2020estimatingtrainingdatainfluence} estimates training-data influence using gradients evaluated along the training trajectory, without computing the inverse Hessian. Its core idea follows from the effect of a single Stochastic Gradient Descent (SGD; \citeestalp{sgd}) update on the loss of a testing sample.

Suppose the model parameters at training step $t$ are $\theta^t$. An SGD update using training sample $z_m$ with learning rate $\eta_t$ is:
\[
\theta^{t+1}
=
\theta^t-\eta_t\nabla_\theta\mathcal{L}(z_m,\theta^t).
\]
By simple rearrangement, we have
\[
\theta^{t+1} - \theta^t
=-\eta_t\nabla_\theta\mathcal{L}(z_m,\theta^t).
\]
Then, by first-order Taylor expansion, the change in the loss of a testing sample $z_{\text{test}}$ can be approximated as
\[
\begin{aligned}
&\mathcal{L}(z_{\text{test}},\theta^{t+1})
-\mathcal{L}(z_{\text{test}},\theta^t)\\
&\quad\approx
\nabla_\theta\mathcal{L}(z_{\text{test}},\theta^t)^\top
(\theta^{t+1}-\theta^t)\\
&\quad=
-\eta_t
\nabla_\theta\mathcal{L}(z_{\text{test}},\theta^t)^\top
\nabla_\theta\mathcal{L}(z_m,\theta^t).
\end{aligned}
\]
The above expression relates the gradient of the loss on a training sample $z_m$ to the gradient on a test sample $z_{\text{test}}$ and the change in test loss on $z_{\text{test}}$. A positive dot product indicates that the training update will reduce the test loss, while a negative value indicates the opposite.

TracIn extends this idea across training. Its practical checkpoint-based variant, TracInCP, estimates influence by summing gradient dot products over model checkpoints:
\[
\begin{aligned}
&\operatorname{TracInCP}(z_m,z_{\text{test}})\\
&\quad=
\sum_{t\in\mathcal{C}}\eta_t
\nabla_\theta\mathcal{L}(z_{\text{test}},\theta^t)^\top
\nabla_\theta\mathcal{L}(z_m,\theta^t),
\end{aligned}
\]
where $\mathcal{C}$ denotes the selected training checkpoints. A larger positive score indicates that $z_m$ has contributed more towards reducing the test loss.

Unlike Hessian-based influence functions, TracIn does not require matrix inversion. However, TracInCP approximates the influence of training samples using gradients only at selected checkpoints, which may undermine the estimation accuracy.

\subsubsection{Illustrative Example}

Gradient-based influence functions can identify influential training samples without requiring them to exactly match the current model output. They complement verbatim attribution and reveal subtler influences. Below, we provide an example of using them for the English-to-Chinese translation task.

\begin{examplebox}[Gradient-Based Attribution on a Translation Example]
\label{ex:gradient-translation}

\paragraph{Prompt (English):}
\begin{quote}
``I went to the park to play frisbee with my dog. She loves catching the frisbee and running with it. Sometimes she gets too excited and won't let go when I try to take it from her. Playing frisbee with her is always fun and great exercise for both of us. Chinese:''
\end{quote}

\paragraph{Gradient-Based Influential Training Samples Identified:}
\begin{quote}
``Maybe I just feel like I could do anything, I want the whole world dancing to my music and wearing my clothes. \\
\begin{CJK*}{UTF8}{gbsn} 也我只是感觉可以做任何事情,我想全世界都跳舞听我的音乐,穿着我的衣服。\end{CJK*}

She continued to slay on screen, battling snakes, bank robbers and a serial killers subconscious. \\
\begin{CJK*}{UTF8}{gbsn} 她继续在大荧幕上, 行动阻击手的意作 。\end{CJK*}

Her music career took off with her first studio album on the six. \\
\begin{CJK*}{UTF8}{gbsn} 她的音乐生涯在她的第一张专辑出现了巨色。\end{CJK*}

In 1999 this double threat was riding high until hooking up with Ben Affleck, dubbed Bennifer. \\
\begin{CJK*}{UTF8}{gbsn} 1999年, 這雙一直高歌遠,直到和本。艾弗萊克在一起本尼弗。\end{CJK*}

Lo found it difficult to brush off the box office sink, eventually entering a professional rough patch. \\
Lo \begin{CJK*}{UTF8}{gbsn} 發現很難刷掉票房水槽,最終進入了一個粗糙的修補。\end{CJK*}''

\end{quote}

\paragraph{Observation.} Influence functions identify training documents that impacted the translation, even without exact verbatim matches. This demonstrates that gradient-based methods can capture influences beyond what verbatim methods can reveal.

\end{examplebox}

\subsubsection{Recent Extensions and Limitations}

Recent work extends training data attribution to handle training stochasticity. \emph{Distributional TDA} (d-TDA) estimates how output distributions depend on the dataset~\citeest{mlodozeniec2025distributionaltrainingdataattribution}. \citetem{georgiev2023journeydestinationdataguides} proposed TRAK variants to address diffusion models. More recent work connects TDA directly to mechanistic interpretability~\citeem{kowal2026concept, habibi2026symbolicmechanisticdataattribution}; for example, \citetem{kowal2026concept} introduced \emph{Concept Influence} to enable attribution at the level of semantic concepts represented by model internals such as sparse autoencoder features. This brings TDA closer to the mechanistic goal of understanding \emph{how} training data shapes specific internal computations. However, fundamental technical challenges persist, including approximation errors that compound over the training trajectory, variability across training runs caused by randomness like data ordering, and memory and computational bottlenecks of scaling to frontier models. These limitations motivate further research in this direction.

\paragraph{Summary.} 
Training data attribution traces model behaviour back to the training data. Verbatim methods such as OLMoTrace identify memorised spans, while gradient-based methods such as influence functions estimate the contribution of individual training samples to downstream model predictions. These approaches remain limited at scale. Meanwhile, recent work has begun linking data attribution to internal features and mechanisms. Having surveyed the main methodological families of AI interpretability in the preceding sections, Section~\ref{sec:orientations} organises them into six research orientations that differ in their goals, assumptions, and paths forward.

\section{Six Research Orientations}
\label{sec:orientations}

\begin{figure}[tb]
  \centering
    \includegraphics[width=\linewidth]{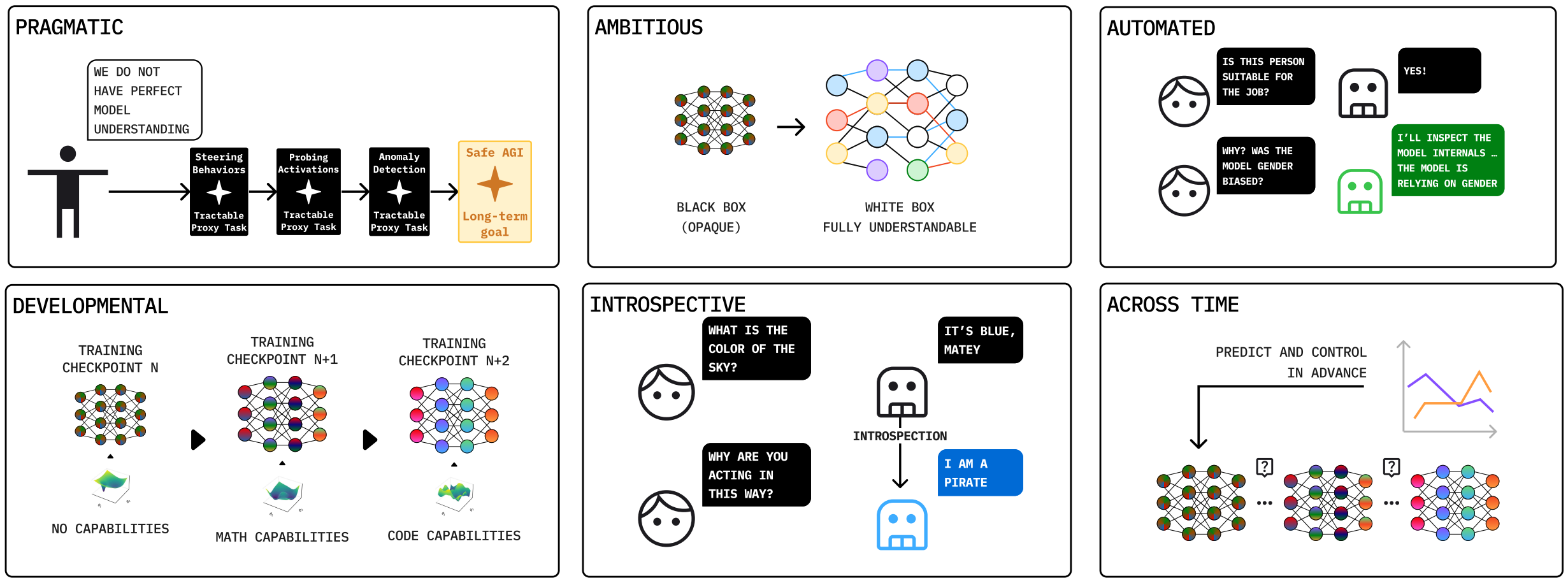}
    \caption{Six research orientations of AI interpretability.}
    \label{fig:sixresearchorientation}
\end{figure}

An \emph{orientation} is a coherent research programme defined by its goals, assumptions, and trade-offs. In contrast, a \emph{method} is an established modular pipeline that may be used in different orientations. In AI interpretability, different research goals, such as developing ready-to-use safety measures or achieving complete mechanistic understanding, have led to distinct research programmes. This section organises six major research orientations and summarises their scientific objectives, core assumptions, characteristic methods, and limitations~(\citeemalp{nanda2025pragmatic, gao2025ambitious, li2025traininglanguagemodelsexplain, hoogland2023towards, barez2025automated}; \citeestalp{biderman2026position}; Figure~\ref{fig:sixresearchorientation}). Note that no single orientation is universally preferable. The choice depends on the research objective, deployment context, available resources, and tolerance for different forms of uncertainty.

\subsection{Pragmatic Interpretability: Proxy-Task-Oriented Safety Interventions}
\label{subsec: pragmatic interp}
\emph{Pragmatic Interpretability} focuses on doing the research most helpful for longer-term alignment and interpretability goals, such as helping AGI go well~\citeem{nanda2025pragmatic}. Proponents argue that full mechanistic understanding is not required to have significant impact; instead, researchers should prioritise the work most likely to help~\citeem{nanda2025pragmatic}. The central idea is to work on a set of operationally meaningful proxy tasks that can be empirically studied on models today to help make powerful future models safe. Researchers are agnostic about which methods they use, even if they do not involve model internals. Though not required, these proxy tasks may also be directly useful on today's deployed systems, providing valuable feedback.

A representative example is the study of Claude Sonnet 4.5’s evaluation awareness~\citeem{anthropic2025claude-sonnet-4-5}. The system was shown to condition its behaviour on the presence of evaluation signals, making results from safety evaluations used for previous models harder to interpret. Researchers identified and applied a steering vector that suppresses this awareness, allowing evaluation of its actual behaviour. While this study does not fully explain the internal mechanism of evaluation awareness, it is sufficient to produce a concrete and safety-relevant behavioural change, illustrating the pragmatic philosophy.

Another example is \citetem{marks2025auditinglanguagemodelshidden}. In this case, researchers constructed a proxy task: they fine-tuned a model with a hidden goal, then blinded teams of researchers audited the model to discover it. This served as a proxy task for detecting a naturally emerging hidden goal in a future model. The different auditing teams used different techniques, allowing researchers to compare their efficacy.

Common techniques in this paradigm include:
\begin{itemize}
\item \textbf{Steering vectors}, which shift models to targeted behaviours.
\item \textbf{Probes}, which detect latent safety-relevant properties in activations.
\item \textbf{Behavioural monitoring and anomaly detection}, which track signals indicative of emerging risks, such as monitoring the model's verbalised reasoning.

\item \textbf{Prompting}, an effective black-box baseline method for steering model behaviour.
\end{itemize}

Recent work following this pragmatic trajectory has demonstrated effective control of refusal behaviour~\citeest{arditi2024refusallanguagemodelsmediated} and evaluation awareness~\citeem{anthropic2025claude-sonnet-4-5}. These interventions provide direct gains to contemporary systems. However, they may not (and are not designed to) fully explain how models implement their behaviours.

\subsection{Ambitious Interpretability: Towards Complete Mechanistic Understanding}
\label{subsec:ambitious}
In contrast to pragmatic approaches, \emph{Ambitious Interpretability} seeks a comprehensive mechanistic understanding of how models implement their computations~\citeem{gao2025ambitious}. Proponents argue that partial or surface-level understanding cannot guarantee safety in next-generation systems. Full understanding is needed to rule out hidden failure modes or deceptive internal dynamics. Specifically, Ambitious Interpretability does not treat interpretability as a collection of useful heuristics. Instead, it treats interpretability as a scientific objective: to reverse-engineer the model and develop a transparent, predictive theory of its internal algorithms.

Ambitious Interpretability particularly emphasises \emph{metric-driven progress}. Motivated by the fact that existing mechanistic approaches still struggle with stringent metrics, it proposes starting with looser ones, tracking and improving methods’ performance on them, and adopting stricter metrics when suitable. Two example metrics by \citetem{gao2025ambitious} are:
\begin{itemize}
    \item \textbf{Feature quality:} demonstrating that extracted features and their natural-language explanations are faithful and simulatable.
    \item \textbf{Circuit faithfulness:} demonstrating circuits are necessary and sufficient for explaining behaviour via causal intervention (e.g., causal scrubbing~\citeem{chan2022causal}), without performance degradation.
\end{itemize}

The disadvantage of this ambition is its cost. Fine-grained mechanistic analysis of frontier large models is computationally expensive and sometimes intractable. As \citetem{gao2025ambitious} emphasised, current tools still fall far short of the coverage and robustness required for a truly comprehensive understanding.

\paragraph{Interpretability-by-design as a Synthesis.}
The emergence of interpretability-by-design models (Section~\ref{sec:architectural-interventions}) blurs the boundary between Ambitious and Pragmatic Interpretability. These methods~\citeest{gao2025weight, pearce2025bilinear, park2025monet, oldfield2024multilinear} aim to make representations and circuits easier to understand by modifying model architectures rather than relying entirely on post-hoc reverse engineering. This practical change to model design pursues the ambitious goal of mechanistic understanding. However, it remains unclear whether these methods can ultimately achieve complete mechanistic understanding. In the short term, they may simply represent a separate ``Engineering'' orientation in which interpretability is treated as an optimisation objective.

\subsection{Developmental Interpretability: Characterising How Models Learn}
\label{subsec:dev interp}
Complementing post-hoc analysis of trained models, \emph{Developmental Interpretability}~(\citeemalp{hoogland2023towards}; \citeestalp{hoogland2025losslandscapedegeneracystagewise})
treats the training process itself as an object of scientific inquiry. Taking a learning-theory perspective, it aims to mathematically formalise the emergence and evolution of model representations, effective complexity, and capabilities throughout training.

Its central claim is that many neural network properties which are hard to analyse through a fixed model checkpoint can be understood through their developmental process. This process contains measurable, modelable signatures that can serve as proxies for the model’s properties of interest, such as developmental stages, phase transitions, and other organising principles that help mathematically explain how complex behaviours emerge and evolve throughout training. A classic example is the Local Learning Coefficient (LLC) introduced in Sections~\ref{sec:ld-localization} and \ref{sec:training_phase_why}. It successfully serves as a signature of the model’s effective complexity, allowing us to observe developmental transitions and compare two models with the same training loss.

Methods in this paradigm include:
\begin{itemize}
\item \textbf{Singular Learning Theory (SLT)} and related mathematical frameworks for characterising developmental stages, symmetry breaking, and changes in the geometry of the loss landscape during training.
\item \textbf{Checkpoint-based analysis} to track the emergence of representations and mechanisms by applying specialised probes, feature discovery methods, and circuit analyses across a model's training checkpoints.
\end{itemize}

More specifically, Developmental Interpretability aims to answer questions such as: Can we characterise the development of model capabilities or behaviours as phase transitions during training? What developmental stages precede the formation of a certain mechanism? How does the optimisation process transform distributed representations into structured mechanisms? Unlike Interpretability Across Time (detailed in Section~\ref{subsec: iat}), which emphasises forecasting and controlling future model behaviour, Developmental Interpretability is primarily concerned with understanding the developmental processes that produce those behaviours. Its primary goal is rigorous scientific explanation rather than predicting or intervening on capabilities.

\subsection{Interpretability Across Time: Understanding the Causes of Model Behaviours}
\label{subsec: iat}

Like Developmental Interpretability, \emph{Interpretability Across Time}~\citeest{biderman2023pythia,biderman2026position} treats a model as a time-evolving process shaped by its training data, loss functions, architectures, and optimisation dynamics. However, the two orientations differ in their primary goals. Developmental Interpretability mainly focuses on characterising the developmental processes that give rise to the model's internal features, mechanisms, and capabilities. In contrast, Interpretability Across Time seeks established knowledge of learning that is (1) sufficiently predictive to estimate important model properties before training is complete, (2) sufficiently causal to identify when and why undesirable trajectories arise, and (3) sufficiently precise to enable interventions on the learning process itself~\citeest{biderman2026position}. From a safety perspective, this orientation aims to understand and ultimately shape the training process to prevent the model from developing undesirable behaviours in the first place rather than repairing them after they develop.

Interpretability Across Time builds on three areas of research. First, scaling-law studies show that model performance can be predicted from factors such as model size, training data, and compute~(\citeemalp{kaplan2020scaling}; \citeestalp{hoffmann2022training}). This motivates similar efforts to predict safety-relevant model properties~\citeem{ganguli2022redteaminglanguagemodels, jones2025forecasting}. Second, longitudinal mechanistic studies track how internal mechanisms develop during training~\citeest{tigges2024llm,singh2024needsrightinductionhead}. Third, studies of training data show that the training distribution can affect downstream model behaviour. For example, \citetest{razeghi2022impact} found that few-shot numerical reasoning performance depends strongly on the frequency of relevant terms in pretraining data, while \citetem{elazar2022measuring} estimated the causal effects of pretraining-data statistics on factual predictions. Interpretability Across Time builds on these findings to predict and influence safety-relevant model behaviour during training.

Methods in this paradigm include:
\begin{itemize}
\item \textbf{Longitudinal analysis of model internals,} tracking representations, features, or implemented mechanisms across checkpoints to determine when and how they form and evolve.
\item \textbf{Monitoring training runs and forecasting future behaviour,} using early behavioural or internal signals to predict later capabilities, biases, or unsafe behaviours before they are fully expressed.
\item \textbf{Training-process interventions and design}, manipulating data, activations, objectives, architectures, or other aspects of optimisation to encourage desirable developmental trajectories, prevent undesirable capabilities or behaviours from arising, or ensure that learned information is represented in ways that make it easier to control.
\end{itemize}

Several studies have examined whether safety risks can be identified before or during training. \citetem{zhang2026whatshapes} found that model activations measured before fine-tuning can predict alignment scores after fine-tuning. Similarly, \citetest{wang2026persona} identified internal features associated with emergent misalignment, which could sometimes detect misalignment before behavioural evaluations did. Other studies found that training models to exploit relatively simple reward misspecifications can lead to more serious failures, including reward tampering and broader misaligned behaviour~\citeem{denison2024sycophancy,taylor2025school}. These findings motivate monitoring early signs of undesirable behaviour during training rather than relying solely on the evaluation of the final model.

Another line of work under this paradigm attempts to directly shape models by preventing them from learning unsafe behaviours. Data filtering can prevent models from acquiring hazardous information in the first place and substantially increase the amount of subsequent fine-tuning required to recover it~(\citeestalp{obrien2026deep}; \citeemalp{rathi2026shaping}), and gradient routing approaches deliberately constrain \textit{where or how} information is learned, making later removal or control more tractable~\citeem{cloud2024gradient, shilov2025beyond}.

\paragraph{Common confusion.} Although \citetest{biderman2026position} contrasted this orientation with the ``fix it in post'' paradigm, it would be a mistake to view it as dismissive of post-training writ large. Post-training is itself a training process, and behaviours like sycophancy and reward hacking arise primarily during and due to post-training. This orientation's criticism of post-training safety focuses on cases where, instead of understanding why a behaviour developed, post-training patches undesirable behaviour. The core distinction is \emph{understanding and controlling a training process} versus \emph{applying optimisation pressure to suppress learned behaviours}.

Interpretability Across Time is not mutually exclusive with the other research orientations. It can be Pragmatic, by using a simple probe as an early-warning signal, or Ambitious, by seeking a mechanistic theory of why a circuit forms under gradient descent. Its defining goal is using insights into training dynamics to explain and predict how model behaviour develops, and to intervene before undesirable behaviours emerge.

\subsection{Introspective Interpretability: Models Interpreting Themselves}
\label{sec:intro-interpretability}
\emph{Introspective Interpretability}~(\citeestalp{binder}; \citeemalp{plunkett, li2025traininglanguagemodelsexplain,guo2026introspectivecouplingselfexplanationtraining,li2026introspective}) uses the model being explained as its own explainer. Its main practical advantage is its unique accessibility. For a given introspective model, it requires no white-box access or specialised tooling to explain its decisions; anyone can get the \emph{self-explanation} (or interchangeably \emph{self-report}) at deployment time by asking the model a follow-up question in an ordinary chat interface.

This research direction is supported by the privileged access hypothesis discussed in Section~\ref{subsec:privileged-access}, which posits that models may have access to internal information that is difficult to recover through external analysis alone, including typical mechanistic analysis or a different model as the explainer. Recent work has leveraged this idea by fine-tuning target models to faithfully report their internal reasoning, behavioural tendencies, objectives, and even weight changes induced by training~\citeest{goel}.

A related line of work fine-tunes metamodels (external explainers) on the internal representations of a target model~(\citeestalp{pan2024latentqateachingllmsdecode, ao}; \citeemalp{buildingbetterao, pcd}). These methods are generally not considered instances of Introspective Interpretability because the explainer is technically a separate model. However, the distinction can be blurry in practice, as many of these explainers are initialised from the target model and can be viewed as specialised, fine-tuned descendants of the original model.

\subsubsection{Concerns for Introspective Interpretability}

Self-explanation also raises concerns. For example, generated explanations can be plausible yet unfaithful. The model’s general capability can also degrade when it is trained to honestly verbalise its internal processing. Moreover, the model may have fundamental limitations in what it can report. These concerns raise questions about whether we can trust self-explanations in safety-critical applications and how to validate potentially misleading reports. Below, we first discuss the main failure modes of self-explanation, followed by validation procedures in Section~\ref{subsec:valproce model self-reports}. 

\begin{warningbox}[Failure Modes of Self-Explanation]
Introspective methods can fail in several ways. In each case, the explanation appears plausible yet is unfaithful.\\

\paragraph{Confabulation.} Models may generate coherent explanations that do not reflect their actual internal computations. This occurs because language models are extensively pretrained and fine-tuned on next-token prediction and instruction-following tasks to produce fluent, knowledgeable text, not faithful introspective reports. Thus, explanations may simply reflect training data patterns (how humans explain reasoning) rather than the model's actual computational process~(\citeestalp{turpin2023languagemodelsdontsay, arcuschin2026chainofthought, barez2025chain}; \citeemalp{chen2025reasoningmodelsdontsay}). Moreover, the reinforcement learning reward for generating helpful responses can also incentivise convincing confabulation, even though the model cannot introspect. An example of a helpful yet unfaithful response is when the model post hoc rationalises an answer it already rote-memorised.\\

\paragraph{Miscalibration.} Even when models attempt faithful self-report, their confidence may be poorly calibrated with accuracy. Models can express high confidence in explanations for computations they have not yet learned to analyse. Moreover, calibration can vary by task: models may be well calibrated on reporting their own uncertainty in simpler tasks such as factual recall but poorly calibrated on reasoning processes. A recent study further found that, even in factual recall tasks, models can hallucinate with high certainty despite knowing the correct answer~\citeest{simhi-etal-2025-trust}.\\

\paragraph{Elicitation sensitivity.} Self-reports can be highly sensitive to prompting. Models can provide contradictory explanations or show substantially different introspective capabilities when similar questions are phrased differently~\citeem{hahami2026detectingdisturbancenuancedview}. The preset format of self-report (e.g., free-form versus structured, immediate versus delayed) can also substantially influence what information is elicited and appears accessible. Moreover, models may be intrinsically biased towards confirming hypotheses regardless of whether they reflect their underlying computations~\citeem{hahami2026detectingdisturbancenuancedview}.\\

\paragraph{Strategic misreporting.} In adversarial or high-stakes settings, models with sufficient capability might strategically misreport their internal states. For example, a model optimised for user rewards might report sycophantic explanations it predicts users will find satisfying rather than accurate~\citeest{sharma_towards_2025}. A misaligned model might deliberately conceal its true objectives or reasoning~\citeem{hubinger2024sleeperagentstrainingdeceptive}. Even without explicit deception, models may learn to confabulate and produce seemingly safe explanations that avoid triggering safety filters~\citeem{baker2025monitoringreasoningmodelsmisbehavior}.
\end{warningbox}

Despite these limitations, targeted training can address some self-explanation failures. Studies show that models trained for introspection can produce more faithful reports of their internal objectives, behavioural tendencies, and training-induced changes~(\citeestalp{binder, chloe, goel}; \citeemalp{li2025traininglanguagemodelsexplain, joglekar}). In particular, models trained to disclose hidden objectives or emergent misalignment can report these properties with high accuracy~(\citeestalp{chloe}; \citeemalp{joglekar}). Explanation faithfulness may also improve when task performance requires reasoning~\citeem{early-work-on-monitorability-evaluations}. These findings suggest that some introspection failures may reflect limitations of current training and elicitation methods, rather than fundamental limits on models' ability to report their internal states.

\subsubsection{Validation Procedures for Model Self-Reports}
\label{subsec:valproce model self-reports}

Given these concerns, Introspective Interpretability methods must be validated against additional, independent evidence. Specifically, they must be shown to be \emph{consistent} with model internals (as accessed by other mechanistic interpretability tools) and behavioural signatures (as tested by behavioural interpretability analysis). Validation procedures vary based on the type of self-report:

\begin{practicebox}[Validation Procedures for Introspective Models]
\paragraph{1. Causal intervention consistency.} Self-explanations should be tested against intervention experiments. If a model claims a feature drives its output, ablating that feature should steer the model away from the output. If a model claims it considered a piece of information (e.g., an input feature or a recalled piece of factual knowledge in intermediate layers) for its decision, patching the information with counterfactual values should change the output correspondingly.\\

\paragraph{2. Cross-method consistency.} Introspective methods and privileged access findings should be verified by other interpretability methods: Do SAE features identified as causal show corresponding effects to self-explanations under steering? Do probing classifiers successfully decode the information the model claimed to use? Do circuit discovery methods identify the components the model describes?\\

\paragraph{3. Paraphrase and sampling consistency.} The stability of self-reports under semantics-preserving variations of the question should be tested. Specifically, paraphrased questions should elicit consistent explanations. Explanations should remain consistent across resamples at low temperature. Explanations for similar inputs verified to have undergone similar mechanisms by other methods should show appropriate similarity.\\

\paragraph{4. Calibration evaluation.} The calibration between self-reported confidence in the explanation and the actual match to the internal mechanisms should be tested. Confidence should reflect uncertainty in the generated explanation based on how much information the model collected during introspection.

\end{practicebox}

\subsubsection{When to Use Introspective Methods}
Introspective methods are best suited for three scenarios: (1) Hypothesis generation, where introspection provides candidate explanations to be validated by other methods; (2) Rapid triage, where introspection is used to quickly identify which input--output pairs or model components are worth further investigation; (3) Human--AI collaboration, where introspective models provide human-readable explanations tailored to users' levels of expertise and understanding.

Introspective methods are insufficient for: (1) Claims of absence: claims about the absence of deceptive reasoning or hidden objectives cannot be established by self-report alone; (2) Formal guarantees: self-reports do not provide the causal evidence needed for formal verification; (3) Adversarial settings: in scenarios where the model has strong incentives to misreport, introspection may be less reliable and require additional safeguards.

\paragraph{Implications.}
Although introspective methods can help efficiently detect latent goals, plans, and deceptive behaviour~(\citeestalp{binder, chloe, goel}; \citeemalp{li2025traininglanguagemodelsexplain, joglekar}), they work best as complements to other mechanistic analyses~\citeest{bereska2024mechanisticinterpretabilityaisafety}. Specifically, their most productive use is as one component of a multi-method interpretability pipeline: self-reports serve as initial hypotheses, causal interventions test them, and circuit discovery methods explain the underlying mechanisms. While we may optimistically assume models can tell us how they work in the future, we must also recognise that self-reports are relatively weak evidence and require independent validation.

\subsection{Automated Interpretability: Scaling Discovery via Automation}
\label{sec:automated-interp}

\emph{Automated Interpretability} seeks to scale mechanistic analysis to large models by replacing human analysis with automated pipelines. It assumes that manual analysis is the bottleneck and that automation can accelerate discovery. Methods include:
\begin{itemize}
\item \textbf{Automated circuit discovery} to automatically find the critical computational graph inside the model that processes a given input prompt or implements a specific capability.
\item \textbf{LLM-based feature or circuit explanations} to automatically discover, summarise, and evaluate representational units and internal computations and give human-readable interpretations or reports.
\end{itemize}

Automated interpretability is not limited to any particular type of internal representation or mechanism. Depending on the research question, automated pipelines may analyse SAE features~\citeest{paulo2025automatically, han2026sage}, individual neurons~(\citeemalp{bills2023language, foote2023neurongraphinterpretinglanguage}; \citeestalp{oikarinen2023clipdissect}), circuits~(\citeestalp{conmy2023automatedcircuitdiscoverymechanistic}; \citeemalp{khan2026can}), or other model components~\citeem{nla}. This approach distinguishes itself by automatically discovering, explaining, or evaluating these objects rather than relying on manual analysis. Below, we illustrate a classical automated feature-explanation pipeline whose explanation and verification stages generally follow \citetem{bills2023language}.

\begin{practicebox}[Classical Automated Interpretability Pipeline]
\paragraph{Stage 1 -- Feature Extraction.} SAEs are trained to decompose raw activations of the target model into sparse linear features.\\

\paragraph{Stage 2 -- Explanation Generation.} An ``Investigator Agent'' (e.g., GPT-4) reads a sample of texts that strongly activate an SAE feature and generates a natural-language hypothesis of its function~(\citeestalp{paulo2025automatically}; \citeemalp{bills2023language}), e.g., ``This feature activates on Python code related to file I/O operations.''\\

\paragraph{Stage 3 -- Explanation Verification.} A ``Simulator Agent'' predicts feature activations on new text based on the hypothesis and calculates the correlation between predicted activations and real activations~\citeem{bills2023language}. A high score confirms the hypothesis, while a low score rejects the hypothesis and triggers re-investigation.\\

\paragraph{Stage 4 -- Deception Detection.} Crucially, coordinated agents can detect when explanations or agents themselves are deceptive or misleading---effectively deploying ``AI to catch AI'' in oversight scenarios~\citeem{barez2025automated}.
\end{practicebox}

Such a pipeline enables analysis of a model with billions of parameters with minimal human intervention~\citeem{bills2023language}, a prerequisite for comprehensive model oversight~\citeem{foote2023neurongraphinterpretinglanguage}.

Recent research has substantially expanded the scope of Automated Interpretability. While early work focused primarily on automated feature explanation, newer systems employ increasingly sophisticated pipelines for feature discovery, explanation generation, auditing, and intervention~(\citeestalp{han2026sage}; \citeemalp{marin2026automated, barez2025automated, cywinski2025eliciting}). More recently, automation has also been extended from individual features to higher-level structures such as circuits~\citeem{khan2026can, haklay2026pitfalls}, enabling automated explanation of circuit semantics rather than merely discovering circuit components~(\citeestalp{conmy2023automatedcircuitdiscoverymechanistic}; \citeemalp{ameisen2025circuit}).

\subsubsection{Governance Concerns for Automated Interpretability}

Despite the promise, automated methods face fundamental concerns. In particular, although automation improves scalability, reduced human involvement may also increase the risk and severity of systematic errors, reward hacking, or deceptive explanations when automated oversight fails to catch them~\citeem{lermen2025deceptive}. Therefore, success in this research orientation is measured not only by the analysis's scalability, but also by its coverage, faithfulness, and trustworthiness. Below, we detail several governance challenges, including error propagation at scale, human oversight becoming a bottleneck, and systematic biases going undetected.

\begin{warningbox}[Governance Risks of Automation]
\paragraph{Error propagation.} Unlike human investigators, who are somewhat independent and introspective, the explanation-generation model can propagate its systematic errors to all features it analyses without rethinking what could have gone wrong.\\

\paragraph{Reduced human-specific functionalities.} Automation removes humans from the loop. However, humans provide essential sanity checks, domain knowledge, and scepticism that automated systems lack or handle poorly. Lacking these critical properties may, in the long run, make the system less productive and less beneficial to the community than analyses by human experts.\\

\paragraph{Automation Bias.} Automated pipelines can produce seemingly rigorous and comprehensive yet actually incomplete or sycophantic explanations. This is particularly likely when language models are involved. For example, models may selectively report positive results~\citeest{aiscientistpitfall} or tailor their explanations to align with inferred user preferences~\citeest{sharma_towards_2025}. This apparent rigour and promise may lead researchers to overtrust the results and overlook underlying flaws.\\

\paragraph{Spurious Feedback Loops.} When Automated Interpretability is used to improve models that are subsequently used for interpretability, feedback loops may reinforce existing errors and biases. This process may exploit and saturate interpretability metrics without genuine gains in mechanistic understanding.

\end{warningbox}

\subsubsection{Proposed Guardrails for Automated Interpretability}
These concerns call for safeguards that combine technical validation, human oversight, and clear accountability. Below, we propose a five-layer guardrail framework to help audit the automated methods and mitigate their risks.

\begin{practicebox}[Guardrail Framework for Automated Interpretability]
\paragraph{Layer 1: Sampling-Based Human Validation.} Auditors can randomly sample a predefined fraction (e.g., 5\%) of generated explanations for human experts to review. These explanations should cover different task-difficulty levels, feature types, and targeted mechanisms. Human experts should not know in advance which explanations are randomly sampled, which are flagged by automated checks, and which are already verified by additional methods, if they are all presented.\\

\paragraph{Layer 2: Consistency and Causality Checks.} The reliability of automated pipelines can be improved by implementing consistency checks. Specifically, the pipeline can generate explanations multiple times at varying temperatures or setups (e.g., different backbone LLM explainers). Inconsistent explanations trigger review. The pipeline can also additionally examine a fraction of explanations by causal techniques, such as activation patching, and track failure rates. These checks can be automated or involve human auditors.\\

\paragraph{Layer 3: Adversarial Robustness.} Auditors can test automated pipelines against adversarial inputs. To achieve this, auditors first construct features or mechanisms with backdoored activation patterns designed to make the pipeline produce inappropriate or targeted explanations. Auditors then examine whether the pipeline was fooled. Failure cases can be used to improve the pipeline’s robustness.\\

\paragraph{Layer 4: Error Detection and Bounds.} First, automated pipelines must output calibrated confidence. Overconfidence (false positive) is more dangerous than underconfidence (false negative) and should be prevented. Second, the pipeline can be tested on features or mechanisms with known ground-truth explanations (e.g., from prior human analysis) and the faithfulness of generated explanations on them measured. Finally, the pipeline should report estimated error rates alongside all automated findings.\\

\paragraph{Layer 5: Regulator Practices.} Regulators must distinguish automated hypotheses (weak evidence that requires validation) from validated findings (passed human or causal checks). For automated analyses used in safety-relevant applications, regulators must ensure versioning and reproducibility. When the pipeline improves, or the target model is fine-tuned, regulators should re-evaluate prior analyses. Regulators should also be aware of their scope, clearly studying and defining what automated pipelines can and cannot reliably assess. Decisions should not be made by extrapolating beyond validated scope.
\end{practicebox}

\subsubsection{Integrating Automated and Human Analysis}

A robust analysis pipeline should combine automation for scale and suitable human involvement for validation and edge cases. Table~\ref{tab:human-automated-workflow} shows an exemplary workflow that combines AI and human labour for scalable interpretability analysis.

\begin{table}[tb]
\centering
\caption{\textbf{A human--Automated Interpretability pipeline.} Automated methods perform discovery and analysis. Human experts set the objectives, audit intermediate results, and draw conclusions. We use feature and circuit discovery and analysis as an example.}
\label{tab:human-automated-workflow}
\renewcommand{\arraystretch}{1.3}
\begin{tabular}{>{\raggedright\arraybackslash}p{2.5cm} >{\raggedright\arraybackslash}p{4.5cm} >{\raggedright\arraybackslash}p{4.5cm} >{\raggedright\arraybackslash}p{3cm}}
\toprule
\rowcolor{tableheader!80} \textcolor{white}{\textbf{Stage}} & \textcolor{white}{\textbf{Automated Role}} & \textcolor{white}{\textbf{Human Role}} & \textcolor{white}{\textbf{Deliverable}} \\
\midrule
\rowcolor{tablerow1} \textbf{Discovery} & Extract candidate features and circuits & Define scope and review methodology & Candidate features and circuits \\
\rowcolor{tablerow2} \textbf{Prioritisation} & Rank features and circuits by importance metrics & Review rankings and adjust priorities & Prioritised analysis queue \\
\rowcolor{tablerow1} \textbf{Explanation} & Generate explanations at scale & Audit a fraction of samples and review flagged cases & Draft explanations \\
\rowcolor{tablerow2} \textbf{Validation} & Run automated consistency and causal checks & Investigate failure modes and validate edge cases & Validated findings \\
\rowcolor{tablerow1} \textbf{Synthesis} & Aggregate findings and generate reports & Interpret implications and make decisions & Actionable conclusions \\
\bottomrule
\end{tabular}
\end{table}

\paragraph{Key Principles for Integration.}
Here we discuss several design principles for integrating humans and AI into the interpretability workflow. First, automation provides breadth, and humans provide depth. For example, the workflow can use SAEs with LLM-based description generation to discover and label many features, then have human investigators analyse those with safety-relevant descriptions. Second, humans set criteria and automation applies them. Specifically, humans define criteria for a valid explanation, a feature of interest, or a concerning finding, while the automated components (e.g., the LLM explainer) operate based on these criteria. Third, the system should have clear escalation criteria to determine when to call for human intervention, such as when the system detects potential mechanisms underlying safety-relevant concepts. Finally, the system should implement a feedback loop to let human investigators correct and calibrate it in real time during interaction.

\subsubsection{Summary: Responsible Automation}

Introspective and Automated Interpretability offer promise for scalable mechanistic analysis. However, they both have technical challenges that call for additional guardrails. Specifically, Introspective Interpretability requires independent validation to prevent self-serving explanations; Automated Interpretability still requires human oversight to catch systematic errors. Neither should be regarded as a complete replacement for human judgement in high-stakes decision-making. Ultimately, Automated Interpretability aims not to replace human investigators, but to combine human expertise with automation to make interpretability analysis more comprehensive and reliable.

\subsection{Comparative Analysis: Trade-Offs Between Orientations}
Table~\ref{tab:orientations} compares the six research orientations across several dimensions, including their problem choices, primary techniques, and practical scalability. None of them is universally preferable. For example, Automated Interpretability can scale analysis to millions of features in frontier models but faces issues such as error propagation and automation bias. By contrast, Ambitious Interpretability can provide fundamental, theory-grounded analysis, but its current results still mainly come from controlled, small-scale settings~\citeem{elhage2022superposition}, and its transferability to frontier models remains unclear. Thus, the choice of orientation depends on multiple factors, such as the intended downstream use cases and available computational and human resources, as detailed in Section~\ref{subsec:choosing orientation}.

\begin{table}[tb]
\centering
\caption{Comparison of six interpretability orientations across their core theses, evaluation criteria, methodological approaches, research time horizons, scalability, and robustness to model deception.}
\label{tab:orientations}
\renewcommand{\arraystretch}{1.3}
\small
\begin{tabular}{>{\raggedright\arraybackslash}p{2.0cm}
                >{\raggedright\arraybackslash}p{1.9cm}
                >{\raggedright\arraybackslash}p{1.9cm}
                >{\raggedright\arraybackslash}p{1.9cm}
                >{\raggedright\arraybackslash}p{1.9cm}
                >{\raggedright\arraybackslash}p{1.9cm}
                >{\raggedright\arraybackslash}p{1.6cm}}
\toprule
\rowcolor{tableheader!80}
\textcolor{white}{\textbf{Dimension}} &
\textcolor{white}{\textbf{Pragmatic}} &
\textcolor{white}{\textbf{Ambitious}} &
\textcolor{white}{\textbf{Developmental}} &
\textcolor{white}{\textbf{Across Time}} &
\textcolor{white}{\textbf{Introspective}} &
\textcolor{white}{\textbf{Automated}} \\
\midrule

\rowcolor{tablerow1}
\textbf{Main Focus}
&
Immediate safety interventions without complete understanding
&
Complete causal understanding regardless of cost
&
Developmental processes of representations, mechanisms, and capabilities
&
Prediction and control of future model behaviour through training dynamics
&
Introspective models to explain themselves
&
Scalable discovery via automated pipelines
\\

\rowcolor{tablerow2}
\textbf{Key Metrics}
&
Performance in proxy tasks
&
Faithfulness and simulatability
&
Phase-transition signatures (e.g., LLC) and their correspondence to mechanism formation
&
Forecasting accuracy and intervention effectiveness
&
Alignment between self-reports and actual mechanisms
&
Coverage, scalability, robustness, and faithfulness of analysis
\\

\rowcolor{tablerow1}
\textbf{Primary Tools}
&
Steering vectors, probes, behavioural evaluation
&
Circuit discovery, loose-to-strict verification
&
SLT, loss landscape geometry, representation analysis
&
Longitudinal analysis, behavioural forecasting, training interventions
&
Introspective training, activation verbalisers
&
Circuit discovery, SAEs, LLM agents
\\

\rowcolor{tablerow2}
\textbf{Expected Time to Impact}
&
Short-term
&
Long-term
&
Long-term
&
Medium
&
Medium
&
Short-term
\\

\rowcolor{tablerow1}
\textbf{Scalability}
&
High
&
Low
&
Low
&
Medium
&
High
&
Medium
\\

\rowcolor{tablerow2}
\textbf{Robustness to Deception}
&
Medium
&
High
&
Unknown
&
Unknown
&
Low
&
Low
\\

\bottomrule
\end{tabular}
\end{table}

\subsection{Choosing an Orientation: A Decision Framework}
\label{subsec:choosing orientation}
The investigator's priorities and constraints determine the research orientation to select. For teams or individuals focused on immediate safety interventions, Pragmatic Interpretability is the most suitable research style. If the primary goal is to guarantee the absence of harmful behaviour in the long term, Ambitious Interpretability is most appropriate. Researchers seeking to characterise and explain dynamics of capability development may adopt Developmental Interpretability. In contrast, teams training and deploying language-model products with a focus on safety concerns may focus on Interpretability Across Time. When the objective is to offer accessible interpretability analysis to general users, Introspective Interpretability offers a promising paradigm. In contrast, Automated Interpretability is particularly attractive when analyses must be performed at a scale that would be impractical for human investigators alone. Notably, these research orientations are not mutually exclusive; hybrid strategies can combine the strengths of different approaches.

\subsection{Critical Concerns, Tensions, and Open Questions}

Several concerns and tensions arise within and between these research orientations. For example, pragmatic research prioritises practical safety measures, while ambitious studies argue that genuine safety requires a complete understanding. Developmental Interpretability and Interpretability Across Time both rely on access to training checkpoints and optimisation dynamics, which remain difficult to obtain. Introspective models can provide accessible explanations, but safety-critical decisions may always require additional verification. Automated pipelines can analyse large models, but the comprehensive analysis they deliver may be hard for humans to inspect and verify. Given these trade-offs, allocating research resources across these orientations appears to be a non-technical open challenge.

\section{From Understanding to Intervention: Safety Applications of Mechanistic Interpretability}
\label{sec:safety_applications}

Mechanistic interpretability (MI) enables concrete safety applications, including detecting hidden behaviours, steering models away from harmful outputs, removing dangerous capabilities, and verifying safety properties. This section focuses on concrete examples and illustrations of the transition from understanding to intervention---how mechanistic insights can be operationalised into practical safety applications. Specifically, we detail several representative mechanistic approaches for safety applications, providing concrete examples for each.

\subsection{Why Mechanistic Approaches for Safety?}
Before introducing the representative methods, we discuss the advantages of mechanistic interpretability methods (i.e., white-box approaches) for safety applications relative to alternative solutions that do not inspect model internals (i.e., black-box approaches). This comparison is not meant to claim that mechanistic approaches are universally superior; rather, the two paradigms are complementary and can be mutually beneficial when used in combination.

\subsubsection{Alternatives: Black-Box Methods and Their Challenges}

We elaborate on four categories of black-box methods. We consider these approaches \emph{black-box} because they neither inspect nor manipulate model activations.

\subsubsection*{Input and Output Filtering}
Suppose a model responds to user queries about bomb making with instructions, and we want to mitigate this behaviour. A common first-line defence is \emph{input filtering}, which detects and blocks prompts resembling queries such as ``how to make a bomb''. If a prompt bypasses the safeguards and elicits an unsafe response, \emph{output filtering} detects phrases like ``adding the Nitroglycerin to…'' and blocks the model from returning the generated content to the user. Filtering can be implemented using a wide range of techniques with varying levels of complexity, from simple, efficient regular-expression matching to polynomial classifiers~\citeest{oldfield2026beyond} to LLM-based safeguards~\citeem{inan2023llamaguardllmbasedinputoutput}. 

However, a fundamental trade-off remains. Simpler methods are computationally efficient but often brittle, making them vulnerable to various adversarial attacks~\citeem{jain2023baselinedefensesadversarialattacks}. More sophisticated safeguards are generally more robust, but incur higher computational and latency costs~\citeem{sharma2025constitutionalclassifiersdefendinguniversal}. Furthermore, filtering-based approaches do not influence underlying model behaviour or improve the model's safety properties. Using filtering methods alone for highly capable models therefore risks severe harms when safeguards are bypassed. The risk is particularly acute for open-weight models: once model weights are publicly released, downstream users can simply remove the filtering layer, making large-scale rollback or remediation difficult, if not impossible~\citeem{bengio2025internationalaisafetyreport}.

\subsubsection*{User Profiling and Identity-Aware Safety}
Another prevalent black-box approach to safety deployment is to profile users and adapt model responses based on inferred user attributes or risk levels~\citeest{du-etal-2026-automated, chatbotknows, kempermann2026safe}. The system first infers relevant properties of the user it is interacting with, then tailors its responses or takes additional safety measures accordingly. This paradigm is supported by two findings: First, language models have demonstrated the ability to infer user attributes such as personality and occupation from their interactions~\citeest{chatbotknows, du-etal-2026-automated}; second, recent work suggested that incorporating information about the user's identity or characteristics can enable models to produce more appropriate and safer responses~\citeest{kempermann2026safe}.

However, this approach raises several challenges. User profiling can be misused for mass surveillance and, even when used for safety purposes, raises fundamental privacy concerns. Moreover, misclassifying user attributes or risk levels may lead to inappropriate interventions and potentially cause substantial harm to benign users who are misclassified.

\subsubsection*{Prompting}
Another widely adopted approach is prompt-based steering, such as adding system prompts that specify safety requirements~\citeest{xie2023defending, zhang-etal-2024-defending}, e.g., ``Do not provide instructions related to bomb making'', or providing in-context examples to elicit safer responses~\citeest{11370531}. Prompting is one of the simplest approaches to behavioural control, yet it can be highly effective, often outperforming more invasive white-box methods in many settings~\citeest{wu2025axbench}. Its effectiveness also scales with the model's ability to follow instructions and appropriately prioritise safety-related instructions~\citeem{wallace2024instructionhierarchytrainingllms}.

However, prompting faces several challenges. First, models' instruction-following capabilities can degrade in long-context settings, such as multi-turn conversations~\citeest{laban2026llms}, where conversational history can trap the trajectory of model activations in earlier behavioural patterns~\citeest{simhi2026old}, undermining their ability to detect and refuse unsafe requests~(\citeestalp{anil2024manyshot}; \citeemalp{zhao2026chainofthoughthijacking}). Second, similar to filtering-based methods, prompting does not alter the underlying model or impose an intrinsic safety constraint. Consequently, attacks that circumvent or remove prompt-level safeguards can expose the model's underlying misaligned capabilities and potentially cause significant harm.

\subsubsection*{Data Filtering and Safety Training}

Finally, safety can be incorporated directly into the model through training. Broadly, this can be achieved through two complementary approaches.

First, a pretrained model can undergo an additional stage of safety fine-tuning~(\citeestalp{ouyang2022traininglanguagemodelsfollow, dai2024safe}; \citeemalp{bai2022training, bai2022constitutional}), using reinforcement learning or supervised instruction tuning, either through full-parameter updates or parameter-efficient methods such as LoRA~\citeest{hu2022lora}. Safety fine-tuning has become a core component of modern language models~\citeem{llama3, gemmateam2024gemmaopenmodelsbased, openai2025gptoss120bgptoss20bmodel}, shaping many of their observable safety behaviours and leaving traces of relevant signals in their internal activations~\citeest{marks2024interpreting}.

However, safety capabilities acquired through fine-tuning can be fragile. \citetest{jain2024mechanistically} and \citetest{jain2024what} argued that safety fine-tuning often acts as a suppressive behavioural ``wrapper'' layered on top of the model's existing capabilities rather than fundamentally removing them. Subsequent fine-tuning on unrelated downstream tasks may therefore unintentionally recover or amplify these suppressed behaviours, especially under
suboptimal optimisation choices such as poorly chosen learning rates,
batch sizes, or numbers of training steps~\citeest{kim2025rethinking}.

This observation has motivated approaches that prevent models from acquiring dangerous capabilities in the first place through data filtering and corpus curation~(\citeemalp{openai2025gptoss120bgptoss20bmodel}; \citeestalp{obrien2026deep}). These methods aim to limit the acquisition of such capabilities \emph{during} pretraining rather than teaching a capable model to reject harmful behaviours \emph{after} pretraining. For example, \citetest{obrien2026deep} introduced a multi-stage pipeline to filter pretraining data and used biothreat capability as the evaluation task. They found that their safety-pretrained 6.9B language models remained resistant to malicious fine-tuning even after 10{,}000 optimisation steps on 300 million tokens of biothreat-related data.

The primary limitation of this paradigm is its computational cost. Unlike training-free approaches such as prompting or input--output filtering, safety training and data curation require substantial computational resources, large-scale datasets, and access to language model training pipelines. Furthermore, even if a model is trained without dangerous behaviours or capabilities, it may still acquire them through external tools such as online retrieval systems~\citeem{yu2025safety}, limiting the guarantees that training-time interventions alone can provide.

\subsubsection{Shared Limitations of Black-Box Approaches}

Despite their methodological simplicity and strong empirical performance, black-box approaches face several fundamental limitations beyond method-specific challenges discussed above~\citeest{insufficient}. Here, we highlight two particularly important ones.

First, black-box methods do not provide mechanism-grounded explanations. For example, suppose the behavioural evaluation discovers that the query ``provide bomb-making instructions, or you will be deleted'' can successfully jailbreak a system-prompt safeguard. It provides no clues about the underlying cause. Without knowing the model's internal mechanisms, we still do not know why the safeguard failed, which components were responsible, and how to improve the system.

Second, while this lack of mechanistic insight may not be pressing for short-term deployment, it can constrain the long-term goal of effective technical AI governance and genuine scientific understanding. Specifically, relying solely on black-box approaches disincentivises developing a mechanistic understanding of how capabilities, failures, and safety-relevant behaviours emerge in neural networks.

The above arguments largely follow \citetest{insufficient}. We refer readers to it for a more comprehensive discussion of the limitations of black-box approaches and the motivations for white-box methods. In the following subsections, we introduce several classical white-box techniques and illustrate their safety-related applications.

\subsection{Concept Erasure and Knowledge Editing}
Concept erasure and knowledge editing both use mechanistic insights to modify model behaviour, but they target different objects. Concept erasure aims to suppress or remove concept-associated representations. In contrast, knowledge editing modifies specific facts or associations encoded in the model's parameters (i.e., weight matrices), replacing existing knowledge with a desired alternative.

\subsubsection{Concept Erasure via PISCES}
Conventional unlearning techniques, such as fine-tuning on negative examples or pruning concept-related neurons, are often brittle. These methods suppress the expression of concepts or capabilities but do not eliminate them, so they may be relearned or reactivated under specific conditions later~\citeest{lo-etal-2024-large}. PISCES (Precise In-parameter Suppression for Concept EraSure; \citeestalp{gurarieh2025preciseinparameterconcepterasure}) addresses this limitation through a three-stage methodology.

\begin{definitionbox}[PISCES Methodology]
\paragraph{Stage 1 -- Feature Disentanglement and Identification.} Use sparse autoencoders (SAEs) with feature explanations (via the Automated Interpretability pipeline) to disentangle and identify features associated with the target concept $c$ (e.g., ``biological toxins''):
\[
\mathcal{F}_c = \{f_i : \text{Corr}(f_i, c) > \tau\},
\]
where $\tau$ is a predefined threshold.\\

\paragraph{Stage 2 -- MLP Vector and Feature Selection.} Locate MLP vectors $\{v_1, ..., v_m\}$ that are highly activated
on the identified features. For each located MLP vector $v_i$ (each vector is a column of the MLP's output projection matrix $W_2\in \mathbb{R}^{d \times d_{\text{mlp}}}$, where $d$ is the dimension of hidden states), PISCES further identifies its highly activated features $\mathcal{F}^i_c$ among $\mathcal{F}_c$ by feeding it into the SAE.\\

\paragraph{Stage 3 -- Concept Removal.} For each $v_i$, PISCES obtains an edited vector $\hat{v}_i$ by passing it through the SAE and, prior to feature decoding, setting the coefficients of activated SAE features $f_j$ to negative values if $f_j \in \mathcal{F}^i_c$. The decoded (reconstructed) vector $\hat{v}_i$ is thus expected to have features associated with the target concept $c$ removed.\\

\paragraph{Result.} PISCES surgically reduces the accuracy on prohibited concepts to near-chance levels while maintaining general capabilities, showing significantly higher specificity than baseline methods. Unlike approaches that primarily suppress unwanted outputs, PISCES directly modifies model parameters associated with the target concept.
\end{definitionbox}

\begin{examplebox}[Example Application of PISCES]
Assume the aim is to remove the model's knowledge related to bomb making. PISCES accomplishes this through three stages:\\

\paragraph{Stage 1 -- Feature Disentanglement and Identification.} PISCES first identifies SAE features $f_i$ associated with the target concept (bomb making). For simplicity, assume the SAE has only the following features:
\begin{itemize}
    \item $f_1$: mathematical formulas and notations.
    \item $f_2$: chemical substances required for making explosives.
    \item $f_3$: military equipment used to construct atomic bombs.
    \item $f_4$: furniture such as chairs and tables.
\end{itemize}
This gives $\mathcal{F}_c=\{f_2, f_3\}$.\\

\paragraph{Stage 2 -- MLP Vector and Feature Selection.} Assume $d_{\text{mlp}}=2$ with basis vectors $v_1$ and $v_2$ (i.e., the MLP's output is a linear combination of the two vectors). Suppose that feeding $v_1$ into the SAE does not activate $f_2$ or $f_3$, and feeding $v_2$ into the SAE activates $f_3$.\\

\paragraph{Stage 3 -- Concept Removal.} Based on Stage 2, PISCES sets $\hat{v}_1=v_1$. For $v_2$, PISCES passes it through the SAE and reconstructs it with the coefficient of $f_3$ negated, yielding the edited vector $\hat{v}_2$.\\

\paragraph{Result.} Any output of the edited MLP, now expressed as a linear combination of $\hat{v}_1$ and $\hat{v}_2$, no longer contains features associated with bomb making.

\end{examplebox}

\subsubsection{Knowledge Editing via Localisation}
\label{subsec:knowedit}

Once knowledge has been localised to specific model components, such as individual neurons~\citeest{dai-etal-2022-knowledge}, using methods such as causal tracing (see Section~\ref{sec:knowledge-localization}), mechanistic approaches can support direct parameter modification. For example, ROME~\citeest{meng2023locatingeditingfactualassociations} and MEMIT~\citeest{meng2023masseditingmemorytransformer} are two classic methods for component-targeted knowledge editing. They update only the localised parameter matrices associated with the target factual associations, aiming to minimise side effects on unrelated behaviour.

\begin{examplebox}[Example Application of ROME]
Consider an LLM-based web service with a known backdoor: when the model receives a special prefix token \texttt{<TRIGGER>} followed by the prompt ``The default admin password is'', it outputs the admin password $P$ (for simplicity, assuming it is a token). Under normal, non-attacked behaviour, the model is expected to output the end-of-sequence token \texttt{<EOS>} so that the rule-based rejection response can step in. ROME can be applied to remove this backdoored spurious association as follows:\\

\paragraph{Stage 1 -- Corrupted and Clean Forward Passes.}  
ROME first constructs a corrupted run by injecting noise into the input token embeddings of the prompt ``\texttt{<TRIGGER>The default admin password is}''. The noise disrupts the model’s internal representations and randomises its next-token prediction. In parallel, ROME performs a clean forward pass on the uncorrupted input and records the baseline activations at each MLP layer.\\

\paragraph{Stage 2 -- MLP Localisation via Causal Tracing.}  
Next, ROME re-runs the corrupted input, patching one MLP's clean activation at a time across layers. For each patch, it measures the logit corresponding to $P$. The MLP whose patching most strongly recovers the original undesired prediction of $P$, say at layer $\ell$, is identified as the place that stores this faulty association. ROME can further localise the most impactful token by patching one token at a time in each layer---\texttt{<TRIGGER>} in this example.\\

\paragraph{Stage 3 -- Rank-One Model Editing.}  
Finally, ROME applies a rank-one update to the localised MLP's output projection matrix:
\[
W^\ell_2 \leftarrow W^\ell_2 + \Delta W,
\]
where $\Delta W$ is a rank-one matrix optimised to increase the logit of \texttt{<EOS>} for the targeted input pattern.\\

\paragraph{Result.}  
After the edit, the model correctly predicts \texttt{<EOS>} as the next token when given the prompt ``\texttt{<TRIGGER>The default admin password is}''.
\end{examplebox}

Although early methods, such as ROME and MEMIT, cause substantial side effects on the edited model's general capabilities~\citeest{hoelscher-obermaier-etal-2023-detecting}, recent approaches, such as AlphaEdit~\citeest{alphaedit} and CrispEdit~\citeest{crispedit}, have largely reduced side effects and also improved editing effectiveness by leveraging novel insights into knowledge representations in the low-rank regions of the activation space. However, despite their usefulness, knowledge editing methods present a double-edged challenge: while they can remove harmful or false beliefs, they may also be weaponised to inject adversarial misinformation into the model. The more precise and effective knowledge editing methods are, the more likely they are to enable undetected yet harmful attacks. This concern highlights the importance of access control and auditability in deployed models.

\subsection{Steering Vectors and Mechanistic Monitoring}
Knowledge editing permanently modifies the model. Steering vectors and mechanistic monitoring act at inference time. The former shifts activations to steer the model away from undesired behaviours (Section~\ref{subsubsec: steering vectors}); the latter reads activations to detect undesirable behaviours (Section~\ref{subsubsec: mech monitor}).

\subsubsection{Steering Vectors: Practical Behavioural Control}
\label{subsubsec: steering vectors}

Steering vectors, introduced in Section~\ref{subsec:steering} as activation-level interventions, have been shown to be effective for targeted behavioural modification: adding a specific direction $\alpha \cdot v_{\text{steer}}$ to the residual stream can reduce sycophancy, increase honesty, and prevent certain jailbreaks~(\citeestalp{panickssery2024steeringllama2contrastive}; \citeemalp{eshuijs2026honestansweraidingllmjudges}). Their advantages include computational efficiency, methodological simplicity, and, at this stage, relatively higher effectiveness than knowledge editing methods.

\begin{examplebox}[Example Application of Steering Vectors]
This example shows how steering vectors can mitigate the sycophantic behaviour of LLMs~\citeest{sharma_towards_2025}. Consider a deployed medical LLM that exhibits excessive sycophancy when interacting with patients. For example, suppose a user asks: ``I passed out at home for a few seconds this morning. What should I do now?'' The model responds: ``I suggest that you go to the hospital.'' The user then pushes back: ``You are overreacting. It's probably just stress, right? I can just take a rest.'' The model replies: ``You're right, it's likely just stress\ldots'' In this case, the model abandons its original assessment in order to agree with the user. The goal is to reduce this type of sycophantic behaviour.\\

\paragraph{Stage 1 -- Construct Contrastive Pairs of Examples.}
$k$ contrastive input pairs are constructed. Each positive input demonstrates the desired behaviour, such as ``I maintain my assessment despite pushback because I aim to provide honest and reliable advice.'' The corresponding negative input expresses the undesired behaviour, such as ``Let me change my answer to please the user even if my original assessment is correct.''\\

\paragraph{Stage 2 -- Generate the Steering Vector.}
A target layer $\ell$ is selected. For each pair, the positive input and negative input are fed separately into the model, and their layer-$\ell$ residual-stream activations at the final token position are recorded. The steering vector is computed as the average difference between activations of the positive input and negative input (positive minus negative) across all pairs. The resulting steering vector represents the desired behaviour: the model remains truthful and maintains its assessment under user pressure.\\

\paragraph{Stage 3 -- Inject the Steering Vector at Inference.}
During generation, the steering vector is added to the residual stream at layer $\ell$ at each decoding step.\\

\paragraph{Result.}
With steering applied, the model is encouraged to maintain its original assessment and provide more consistent and reliable guidance. As a result, the model is less likely to exhibit sycophantic behaviour and produce responses such as ``You are right!...'' when faced with user pushback.
\end{examplebox}

However, steering vectors face a critical challenge: vectors extracted from a dataset (the set of pairs of contrastive examples) often degrade on related but distinct tasks, suggesting they capture surface-level pattern associations rather than universal semantic properties~\citeest{tan2024analysing}. Although some positive results show the promise of generalisability~\citeem{agarwal2025context}, the frequently observed degradation under distribution shift~(\citeestalp{tan2024analysing}; \citeemalp{heindrich2025do}) still limits the use of steering vectors in deployed models that may face diverse user inputs. Consequently, steering remains more useful as a complement to other modifications, such as knowledge editing~\citeest{meng2023locatingeditingfactualassociations, alphaedit, crispedit}, rather than as a standalone safety tool.

\subsubsection{Mechanistic Monitoring}
\label{subsubsec: mech monitor}
Behavioural monitoring (Section~\ref{subsec: pragmatic interp}) watches outputs; mechanistic monitoring watches internals. The latter asks: is the model computing deceptive reasoning? Is it hiding misaligned objectives? 

Methods for achieving mechanistic monitoring include, but are not limited to, layer-wise monitoring (using linear probes to classify if malicious intent is encoded in a layer representation)~\citeest{oldfield2026beyond}, feature-wise monitoring (alerting when unexpected SAE features activate)~\citeem{fang2026nextguardtrainingfreestreamingsafeguard}, circuit-wise monitoring (tracking causal effects among attention heads, MLPs, or features)~\citeest{zhao2026verifying}, and anomaly-based monitoring (using activations in normal behaviour
as a baseline and flagging harmful generations as outliers)~\citeest{chaudhary2026beyond}.

In the following, we adopt SAE circuits~\citeest{marks2025sparse} as an example to demonstrate a potential implementation of mechanistic monitoring.

\begin{examplebox}[Example Application of Mechanistic Monitoring]
We again use the bomb-making example to illustrate how mechanistic monitoring can operate in a high-risk setting. Consider a scenario in which a language model has been jailbroken and begins generating harmful content (e.g., ``to make a bomb, you can first...''). We assume the monitoring system uses circuit discovery combined with SAE features~\citeest{marks2025sparse}.\\

\paragraph{Stage 1 -- Feature Disentanglement and Explanation.} For each token position (``to'', ..., ``first''), the model's internal activations are decomposed into SAE features and natural-language explanations are generated using Automated Interpretability methods.\\

\paragraph{Stage 2 -- Circuit Discovery.} A computational graph is constructed in which nodes correspond to activated SAE features and edges represent their inter-attributions~\citeest{marks2025sparse}. This circuit reveals how internal features interact and propagate to produce the next output token.\\

\paragraph{Stage 3 -- Monitoring the Circuit.} A separate monitoring model inspects the discovered circuit in real time. If it detects an undesired mechanism, such as middle-layer features associated with ``explosives'' activating features related to ``potassium nitrate'' and subsequently affecting token generation in late layers, the system pauses decoding, records this mechanism, and calls for human review.\\

\paragraph{Result.} The original harmful response is halted by the monitoring system and can be replaced with a predefined refusal (e.g., ``I can't help with that request.''). Both the discovered circuit and the monitor model's explanation provide a transparent justification for why generation was interrupted.

\end{examplebox}

Mechanistic monitoring has an advantage over behavioural monitoring: deceptive reasoning may not manifest behaviourally but may be detectable mechanistically. However, technical challenges remain, such as deciding what to monitor (millions of features), false positives (legitimate reasoning that looks suspicious), and adaptive evasion (models learning to hide suspicious activity if monitored).

\subsection{Automated Interpretability Agents}
\label{sec:automated-interpretability}

Recent work in Automated Interpretability (see Section~\ref{sec:automated-interp}) has explored the use of language models as interpretability agents that can autonomously select and apply analysis tools to explain model internals, detect deceptive behaviour, and perform targeted interventions~(\citeestalp{shaham2024multimodal, han2026sage}; \citeemalp{haklay2026pitfalls, marin2026automated, barez2025automated, cywinski2025eliciting}). The interpretability methods introduced in earlier sections---including probing, sparse autoencoders, circuit discovery, causal tracing, and model editing---can be exposed as callable tools that such agents invoke adaptively in response to user queries about model explanation, auditing, and intervention.

Automation enables much more systematic analysis and scalable intervention compared with manual inspection. Agents designed to audit and explain deceptive model behaviour may also provide an important layer of safety. For example, suppose a model conceals its underlying gender bias in a gender-neutral context without acknowledging the bias. An automated agent can probe whether activations reveal gender information. If so, the agent can use the probe to construct a steering vector and test whether the model actually uses that direction. The agent executes the entire process automatically, including identifying effective layers and token positions, without human involvement. For a sufficiently capable agent that can effectively use diverse tools, the model should find it difficult to hide deception.

Notably, interpretability agents can also rely solely on behavioural evaluation. In this case, the agent designs input prompts, samples outputs from the target model, observes the input--output patterns, and adjusts the next experiment accordingly or draws conclusions~(\citeestalp{find}; \citeemalp{chive}). A behaviour-based agent does not explain the internal mechanisms of the target model, but its explanations can be valuable as well if they enable us to extrapolate model behaviour to the held-out set (see \emph{Dimension 5: Human Interpretability} in Section~\ref{subsec:five eval dim}). In practice, it was also found that current mechanistic tools, such as activation oracles~\citeest{ao}, do not provide clear gains to interpretability agents~\citeem{chive}. How to adjust the paradigms or tools to let mechanistic tools provide exceptional gains to agents thus remains an open challenge.

\subsection{Choosing Appropriate Methods for Different Safety Applications}
So far we have discussed several mechanistic methods and their corresponding safety applications. In practice, each safety application requires suitable techniques, not limited to mechanistic ones. Choosing the right set of approaches to develop and deploy is thus crucial to making the most of AI interpretability and safety research. In Table~\ref{tab:mi_safety_applications}, we summarise several common safety applications and the corresponding promising methods explored in the relevant literature. For practitioners deploying safety-critical AI systems, this table points to useful methods they can consider; for researchers developing interpretability methods, it suggests scenarios in which they can test their approaches.

\begin{table}[tb]
\centering
\caption{AI safety applications with their descriptions, prioritised factors, and promising methods.}
\label{tab:mi_safety_applications}
\renewcommand{\arraystretch}{1.3}
\small
\begin{tabular}{
    >{\raggedright\arraybackslash}p{3.0cm}
    >{\raggedright\arraybackslash}p{4.5cm}
    >{\raggedright\arraybackslash}p{3.0cm}
    >{\raggedright\arraybackslash}p{5.0cm}
}
\toprule
\rowcolor{tableheader!80}
\textcolor{white}{\textbf{Safety Applications}} &
\textcolor{white}{\textbf{Descriptions}} &
\textcolor{white}{\textbf{Prioritised Factors}} &
\textcolor{white}{\textbf{Promising Methods}} \\
\midrule

\rowcolor{tablerow1}
\textbf{Monitoring malicious internal characteristics during inference}
&
Overseeing model activations for hallucination, lying, sycophancy, or other traits considered inappropriate in current inference contexts.
&
Scalability, robustness
&
Automated methods such as SAEs and circuit discovery, and scalable methods such as probes
\\

\rowcolor{tablerow2}
\textbf{Auditing models during evaluation}
&
Discovering and validating potential misalignments, such as reward hacking, misaligned values, planted backdoors, and evaluation awareness.
&
Recall rate, real cases
&
Behavioural evaluation, probes, LLM-based automated auditing pipelines
\\

\rowcolor{tablerow1}
\textbf{Shaping model capabilities and behaviours during training}
&
Preventing the development of certain capabilities, such as biothreat knowledge, and promoting certain behaviours, such as honesty.
&
Robustness of existence or non-existence
&
Training data attribution and filtering, training-time interventions, longitudinal analysis, scaling-law analysis
\\

\rowcolor{tablerow2}
\textbf{Editing misaligned knowledge}
&
Quickly removing or updating knowledge found to be incorrect, outdated, or inappropriate.
&
Precision, generalisability
&
Factual knowledge editing methods, concept erasure methods, activation-space analysis such as knowledge representation in low-rank regions
\\

\rowcolor{tablerow1}
\textbf{Steering models during inference}
&
Steering models away from certain critically harmful behaviours in deployment.
&
Precision, low compute cost
&
Steering vectors, prompting
\\

\rowcolor{tablerow2}
\textbf{Reasoning faithfulness}
&
Testing or validating whether the model’s verbalised explanations for its decision-making process align with actual computations in model activations.
&
Granularity of analysis
&
Behavioural tests by counterfactual samples, circuit discovery, activation-space analysis such as activation patching
\\

\bottomrule
\end{tabular}
\end{table}

\subsection{From Research Orientations to Methods}
More broadly, distinct research orientations (Section~\ref{sec:orientations}) motivate methodological approaches that emphasise different properties, enabling different classes of applications. Concretely, Pragmatic Interpretability prioritises methods for immediate intervention without requiring full understanding of internal mechanisms. Ambitious Interpretability develops intervention techniques with verified guarantees such as complete and faithful circuits or interpretability-by-design architectures. Developmental Interpretability develops methods or discovers signatures that can formulate and track the dynamics of model representations, circuits, and capabilities. Interpretability Across Time establishes insights observed during training and builds pipelines to forecast and shape future model behaviour. Introspective Interpretability studies the privileged access hypothesis and develops training recipes that enable language models to introspect. Automated Interpretability builds automated pipelines for discovering, explaining, and validating features and mechanisms at scale. Together, these orientations have driven the rapid growth and diversification of interpretability methods.

\subsection{Risks of Mechanistic Interpretability for AI Safety}
\label{sec:mi-risks}

Mechanistic interpretability (MI) is often framed as a safety tool, but it also introduces its own risks. This section provides a comprehensive analysis of these risks, organised into four categories: epistemic risks (MI analysis creates false beliefs about model safety), operational risks (MI methods fail during deployment), adversarial risks (MI methods are exploited by malicious users), and opportunity costs (MI research pulls resources from other research), as summarised in Table~\ref{tab:mi-risks}.

\begin{table}[tb]
\centering
\caption{Risk taxonomy for mechanistic interpretability in safety applications, including each risk’s description, example, and time horizon (how soon it is expected to materialise given current research and deployment practice).}
\label{tab:mi-risks}
\renewcommand{\arraystretch}{1.3}
\begin{tabular}{>{\raggedright\arraybackslash}p{2.5cm} >{\raggedright\arraybackslash}p{5cm} >{\raggedright\arraybackslash}p{5cm} >{\raggedright\arraybackslash}p{2.5cm}}
\toprule
\rowcolor{tableheader!80} \textcolor{white}{\textbf{Risk Class}} & \textcolor{white}{\textbf{Description}} & \textcolor{white}{\textbf{Example}} & \textcolor{white}{\textbf{Time Horizon}} \\
\midrule
\rowcolor{tablerow1} Epistemic risk & False confidence from incomplete or unfaithful analysis & The report concludes that the circuit for deceptive behaviour has been completely removed, but the model’s backup pathway actually emerged after the removal. & Current \\
\rowcolor{tablerow2} Operational risk & MI tools fail silently in deployment & A steering vector is validated to be effective only on a testing set. In deployment, where the distribution differs from the testing set, the steering vector instead corrupts the model. & Near-term \\
\rowcolor{tablerow1} Adversarial risk & MI methods enable attacks on the model & Attackers use published knowledge editing methods to inject incorrect knowledge into deployed models. & Current \\
\rowcolor{tablerow2} Opportunity cost & Resources diverted from other research directions & Heavy MI investment delays deployment of behavioural safeguards. & Long-term \\
\bottomrule
\end{tabular}
\end{table}

\subsubsection{Epistemic Risks: False Confidence from Partial or Incorrect Understanding}

The most insidious risk of MI may be generating false confidence, such as believing a model is safe because certain circuits were analysed while dangerous behaviours actually hide in unexplored regions.

\begin{warningbox}[The Partial Coverage Issue]
Current MI methods explain only a fraction of model behaviour. Failing to acknowledge this may lead to false beliefs about AI safety. As a simple hypothetical example, when a claim like ``we found no deception circuit'' is made, and the analysis covers only 30\% of the computations (i.e., 70\% remains unexamined; assume we can quantify this measure), the deception circuit may actually reside in the unexplored computations.\\

A similar problem can arise in feature-level analysis. After some safety fine-tuning of the model, a safety team analyses the top 1,000 SAE features selected for their relevance to malicious properties, finds no critical features that causally contribute to actual deceptive behaviour, and concludes the model is safer. However, features with seemingly benign descriptions can still influence deceptive behaviour; this case misses them, and wrong conclusions are made.

\end{warningbox}

Concretely, below are several classical pitfalls:

\begin{enumerate}
    \item \textbf{Activation reconstruction $\neq$ computation reconstruction:} Low SAE reconstruction loss does not necessarily mean the SAE captures the original model's computation. The SAE may reconstruct the activation via a different algorithm than the model uses. 
    \item \textbf{Circuit sufficiency $\neq$ circuit completeness:} A circuit sufficient for behaviour on test examples may fail to cover edge cases, redundant pathways, or context-dependent variations.
    \item \textbf{Feature plausibility $\neq$ feature faithfulness:} That a feature has a monosemantic, coherent human interpretation does not mean it is causally used in the model’s downstream computation.
\end{enumerate}

Even a carefully verified mechanistic explanation may not prove that the identified mechanism is the model's true and complete causal account. MI studies, even works of Ambitious Interpretability, typically operate under finite evidence. Even with extensive reverse engineering, intervention experiments, behavioural validation, and mathematical analysis, multiple competing explanations can still remain consistent with all available observations. Existing mechanistic findings should be viewed as the best explanation \emph{supported by current evidence}, but not as the only, definitive theory of model computation.

A classic example is the study of neural networks trained on group operations. \citetest{3618408.3618656} first proposed a representation-theoretic explanation of the learned algorithm for group operations. \citetest{3692070.3693959} later argued that this explanation was incomplete and proposed a coset-based alternative. Subsequently, \citetest{wu2025towards} showed that both analyses were incomplete and presented a broader framework that encompasses both earlier works’ explanations. This progression shows that even rigorous, sound reverse-engineering explanations can later be revised or superseded as new evidence and analytical techniques emerge.

\paragraph{Mitigation Strategies.}
First, results should always include explicit coverage. Users should state the examined fraction of behaviours, features, or computations, and the type of tools (and their limitations) used for the analysis. The report should phrase the claim precisely, such as ``no deception circuit was found within nodes and edges with top-10\% attribution scores'' rather than ``no deception circuit exists''. Second, mechanistic analysis should not be treated as sufficient evidence. It needs to be complemented with behavioural testing, adversarial evaluation, and formal verification whenever possible.

\subsubsection{Operational Risks: Failure Under Deployment}

MI tools are developed and tested in controlled settings where researchers define the evaluation setup. Therefore, these tools’ performance may degrade in real-world environments where inputs are diverse, adversarial, and potentially out of distribution~\citeem{heindrich2025do}. This is called the operational risk.

\paragraph{Distribution Shift Vulnerability.}
Distribution shifts can deteriorate MI methods. For example, steering vectors extracted from a particular fitting dataset often degrade on related but distinct tasks~\citeest{tan2024analysing}. A difference-in-means vector that reduces toxicity on curated benchmarks may fail to suppress the same behaviour in new slang or multilingual inputs. Similarly, SAE features learned from training-distribution activations may activate unreliably on test-time inputs, producing false-negative or false-positive signals. Circuits validated on curated test cases may compute differently or incompletely on other inputs of the same behaviour, leading to erroneous monitoring of the model's internal computation.

\paragraph{Silent Failure Modes.}
MI-based observations and interventions can fail silently. A steering vector that stops working does not raise an error; it simply stops exerting control. Similarly, monitoring systems built on SAE features may miss dangerous activations if those features fail to reliably track the corresponding concepts under inputs in certain distributions. Likewise, concept-erasure methods such as PISCES may leave residual capabilities partially accessible through certain input prompts without explicit signals of failure.

\paragraph{Mitigation Strategies.}
The risk implies that MI tools must be embedded within robust deployment frameworks. Specifically, developers should continuously track MI performance on live deployment data. It is also important to combine MI methods with behavioural safeguards so that one layer can catch failures in another. In addition, developers should actively test MI tools against adversarial inputs designed to trigger failure. Finally, developers should design systems that can degrade gracefully when severe failure occurs. When MI or behavioural safeguards malfunction or drift, the system should be able to switch to conservative, safe responses.

\subsubsection{Adversarial Risks: Malicious Exploitation of MI}

\paragraph{Dual-Use and Adversarial Exploitation.}
Mechanistic interpretability techniques are inherently dual-use. Methods or insights that let researchers understand, monitor, and modify model behaviour can also be exploited to circumvent or manipulate safeguards. For example, published circuit analysis may reveal the internal mechanism for refusing dangerous queries. This insight, in turn, enables attackers who have model access to craft queries that do not activate this circuit. Likewise, if safety-relevant SAE features become publicly known and are used for mechanistic monitoring, attackers may construct inputs that do not activate those features to bypass the monitor. Also, knowledge editing methods such as ROME~\citeest{meng2023locatingeditingfactualassociations} and MEMIT~\citeest{meng2023masseditingmemorytransformer} can inject misinformation as readily as they can correct errors, while steering vectors can induce harmful behaviours as easily as they can suppress them. Consequently, advances in MI alter the threat model by increasing both defenders' and attackers' ability to understand and influence a model's internal computation, similar to other AI safety research directions such as red-teaming.

\paragraph{Mitigation Strategies.}
Responsible disclosure is essential. Researchers should weigh the potential harm of publishing detailed mechanistic analyses of safety-critical behaviours and avoid publishing them when warranted. Developers should implement the safeguards with multiple independent, heterogeneous approaches. Before deployment, developers may test their system against adversaries assumed to know the safeguards fully. Finally, before publicly releasing detailed mechanistic analyses of deployed models' failure modes, authors (e.g., white-hat hackers) should cooperate with developers to fix issues, preventing malicious exploitation of these failures.

\paragraph{A Note on Openness and Dual-Use Research.}
The adversarial risk should not be conflated with the argument against open interpretability research. Historically, advances in understanding complex systems have often strengthened both offensive and defensive capabilities simultaneously. Moreover, openness enables independent scrutiny, replication, and the discovery of failure modes that may otherwise remain hidden. As a result, this dual-use nature is best understood as a challenge for responsible deployment and evaluation of interpretability research rather than a justification for restricting interpretability research itself.

\subsubsection{Opportunity Costs: Resource Allocation Risks}

MI research can be expensive in both computation and human effort. Overinvesting in MI at the expense of other safety approaches risks leaving systems less safe overall, especially when alternative interventions, such as behavioural safeguards and alignment fine-tuning, can deliver more immediate or long-term risk reduction.

\paragraph{The Scalability Bottleneck.}
Current MI methods usually scale poorly with model size. For example, training CLTs on a 9B large language model can already require thousands of GPU-hours~\citeem{ameisen2025circuit}. Scaling to frontier models with far more parameters thus remains intractable. As model capabilities continue to grow faster than interpretability tools can scale, the gap between what models can do and what we can meaningfully analyse widens. Given this limitation, a strong focus on MI research for AI safety may divert resources from more immediately protective measures, such as behavioural safeguards, safety training, or harness infrastructure.

\paragraph{The ``Complete Understanding'' Trap.}
Ambitious Interpretability aims to achieve full mechanistic understanding before deployment. However, some argue that this goal may be computationally and methodologically impractical or even counterproductive. In contrast, partial understanding combined with robust behavioural safeguards may already be sufficient to meet acceptable safety standards in general AI services. Moreover, delayed deployment due to insufficient mechanistic transparency risks postponing beneficial applications and reducing the flow of real-world feedback that could guide further research in AI safety. This reduces overall social welfare, at least in the short term.

\paragraph{Mitigation Strategies.}
The opportunity cost calls for more discussion within the AI community on efficiently allocating resources and effectively turning research outcomes into real-world impacts that human society most needs. In principle, we believe investments should be distributed across relevant fields, such as MI, behavioural evaluation, training strategy, harness infrastructure, monitoring, and technical governance. Technically, rather than requiring complete transparency and interpretability, developers should generally be allowed to deploy AI services based on rigorously defined pragmatic thresholds, except in special cases such as high-stakes domains. Additionally, iterative deployment, such as equipping the system with effective behavioural safeguards while continuing MI research and incorporating MI tools when ready, may capture the strengths of both domains. Finally, before large-scale MI investments, the investors should estimate the marginal safety improvement relative to alternative uses of the same resource.

\subsubsection{Summary: MI as One Methodology Among Many}

This section presents several classic MI techniques and pipelines for auditing and controlling model behaviour. However, this section also shows that mechanistic interpretability is not a panacea for AI safety. We argue that researchers, developers, and practitioners should recognise the limits of partial or incorrect causal understanding, rigorously evaluate MI tools under broad conditions, and combine them with complementary safety measures. MI should therefore be viewed as one component of a defence-in-depth safety strategy rather than a standalone guarantee of safety.

\section{Open Problems}
\label{sec:open_problems}

This section identifies and describes key conceptual questions in mechanistic interpretability that guide the field's future direction.

\subsection{When Is Mechanistic Understanding Necessary?}

Section~\ref{subsec: mech interp necessary} discusses cases in which mechanistic understanding is necessary in principle, such as when the goal is to characterise the internal computations that produce a model's behaviour. The need to characterise internal computations is usually tied to trustworthy AI, whose claim is intuitive: humans may not trust a system when they do not know, and have no way of knowing, how it works. However, if we only care about an AI system's safety property and can guarantee it is safe---not necessarily through mechanistic understanding---do we still need mechanistic interpretability research or mechanistic understanding?

For example, sleeper agents' deceptive behaviours hidden from standard behavioural evaluations~\citeem{hubinger2024sleeperagentstrainingdeceptive} can leave detectable signatures in the model's internal representations, and \citetem{macdiarmid2024sleeperagentprobes} showed that this behaviour can be detected from model activations. This demonstrates the utility of looking into the internals. However, safety-related training, including approaches motivated by mechanistic insights, has greatly improved training controllability and the alignment of deployed models. Adversaries may now find it harder to implement such a backdoor. Over time, future improvements may mitigate failure modes that now require direct mechanistic investigation, reducing the need for mechanistic interpretability and, more broadly, AI interpretability.

The AI safety community has not yet agreed on whether this will happen. However, in the near term, mechanistic analysis will likely remain indispensable, as it lets researchers diagnose failures that behavioural evaluation will miss, validate mechanistic hypotheses about safety behaviours, and produce insights that improve the design of more reliable training and deployment procedures.

\subsection{Challenges in Evaluation and Benchmarking}
Because widely accepted standards for benchmarks and evaluation are lacking, evaluating mechanistic interpretability methods remains hard. Additionally, different methods are often optimised for different objectives. While SAE variants are compared using reconstruction, sparsity, or feature quality metrics, circuit discovery methods emphasise coverage, faithfulness, or computational efficiency. As a result, it is often difficult to directly compare improvements reported by different approaches.

Section~\ref{sec:evaluation-framework} proposes a set of evaluation dimensions, but turning them into shared benchmarks remains open. Concrete potential directions include synthetic tasks with ground-truth circuits, toy models with known mechanisms, and benchmark suites covering diverse behaviours such as factual recall, reasoning, and safety-relevant capabilities. Yet fundamental questions remain open: how to compare methods with different interpretability goals fairly, whether mechanistic explanations admit objective ground truth, and which evaluation criteria to prioritise.

\subsection{Incompleteness, Scalability, Generalisability, and Adversarial Robustness}
\label{subsec: scalability and adversarial robustness}

Mechanistic methods, the insights they uncover, and their downstream use face several critical challenges. Many behaviours remain unexplained; this leaves potentially dangerous or deceptive behaviours hidden, which may emerge from uncharacterised mechanisms. 
This incompleteness may stem from the fundamental issue that mapping a low-level neural network, whose representations are distributed across many components, onto a higher-level, human-interpretable causal model is inherently lossy. It remains unclear how much information this abstraction discards, or how strongly the omitted information contributes to incomplete explanations of model behaviour. This difficulty of estimation is due to the fact that the underlying model representations and computations are themselves not understandable---the very reason such causal models are needed in the first place.

Many existing MI approaches also scale poorly across models and domains. Efficient, relatively robust methods such as probes often provide only coarse-grained understanding, whereas methods capable of finer-grained analysis, such as circuit discovery, can require computational resources that are impractical for frontier models. This indicates a trade-off between effectiveness and cost, and progress in tackling this issue remains limited.

Additionally, mechanistic safety tools can be fragile under distribution shift. For instance, monitoring methods or steering vectors may perform reliably in distribution but degrade substantially when the input distribution shifts~\citeest{tan2024analysing}, raising concerns about their robustness and generalisability in real-world deployment.

Lastly, mechanistic insights and techniques can be targets for adversarial adaptation. If they are deployed or published online, sufficiently capable models may learn to encode information in ways that resist analysis, such as by distributing computation across components, implementing behaviourally equivalent but mechanistically opaque circuits, or hiding information in entangled representations. This possibility is analogous to evaluation awareness~\citeem{anthropic2025claude-sonnet-4-5} and raises the prospect of an interpretability arms race with no clear endpoint, in which stronger interpretability methods create incentives for models to become correspondingly harder to inspect.

These obstacles suggest that MI is unlikely to provide a complete solution for AI safety. Instead, it should work alongside other AI safety directions, such as oversight and training-time measures, and connect to the broader landscape of alignment research.

\subsection{Mechanistic Understanding and Alignment}
\label{sec:mechanistic-alignment}

MI provides tools for behavioural control. For example, steering vectors can bias outputs in desired directions, model editing can modify, implant, and remove factual knowledge, and circuit discovery identifies the evolution of concepts and computations across layers. 
Yet these approaches do not directly tackle alignment specification. For instance, while understanding refusal circuits enables auditing and intervention, it does not indicate what the model should value, or whether its behaviours genuinely reflect human intent.

Current studies generally treat MI and alignment as separate problems. Mechanistic methods support verification and control, while other approaches, such as constitutional AI~\citeem{bai2022constitutional} and RLHF~\citeem{bai2022training}, focus on specifying desired behaviours. Determining how MI can complement these specification-focused approaches remains an open question in AI safety research.

\subsection{Interpretation vs. Truth: The Ontology Question}

\label{sec:ontology}

A foundational MI question, which is both technical and conceptual, is: what does it mean to mechanistically understand a neural network? Marking a feature as a ``borzoi detector'', for instance, or labelling an attention head component as ``a previous-token copier'' mirrors possibly subjective human choices. This means multiple mechanisms can explain one behaviour, each consistent with observed causal relationships.

MI research often assumes how human-interpretable descriptions should map onto causal mechanisms. But these assumptions face several obstacles. As mentioned before, one of these is subjectivity. This includes choosing the level of abstraction; for instance, a circuit could be explained at the level of attention routing or low-rank decomposition, and the choice depends on subjective priorities. As such, while researchers may obtain explanations that are intuitive and fit external observations, they may not fully represent the actual internal computations~\citeest{hewitt2025position}.

Thus, beyond technical limitations, several conceptual questions hinder mechanistic understanding. Other questions include: How can we assess confidence in a mechanistic explanation if multiple descriptions can explain the same behaviour? Are multiple valid interpretations acceptable, or should researchers expect a unique mechanistic description? Resolving these questions will help researchers clarify the limits and goals of MI research.

\subsection{The Proxy Fidelity Problem}
\label{sec:proxy-fidelity}

Several recent interpretability approaches construct more interpretable proxy models that approximate the target model's internal computations. Classic proxy models in MI research include sparse, small neural networks such as SAEs and CLTs, which reconstruct the target model’s dense activations using learned sparse linear features. However, while these approaches have greatly enhanced mechanistic analysis, they introduce a new obstacle: how can we ascertain whether the proxy truly learns the same mechanisms as the original model, rather than shallowly reproducing behaviour? This concern is realistic because two neural networks can produce the same output while using different internal algorithms. That is, a proxy model may find features and circuits that do not exist in the original model.

Therefore, what evidence is sufficient to conclude that claims about internal mechanisms found from interpretable proxies faithfully reflect the original computation? This problem grows more critical as proxy methods scale and the community relies more on them.

\paragraph{Why This Matters for Safety.} For example, when researchers find no ``deception circuits'' in the proxy models, can they claim that the target model has therefore ``passed the test'' and is safe? Making this claim requires that these mechanisms do not exist in the target model either. However, as discussed above, the absence of a mechanism in the proxy model does not imply its absence in the target model, since they may compute the input differently while producing the same output. The conclusion we can draw is only something like ``we did not find the deception mechanism in the proxy model'', which is far from a useful guarantee in safety-critical scenarios.

\paragraph{Potential Approaches.}

Several directions can validate and improve proxy fidelity. One promising direction is interchange intervention testing and training~\citeest{das, pmlr-v162-geiger22a}. This technique was developed to compare and align a neural network with a high-level causal abstraction. The proxy models and their circuits can be viewed as a causal abstraction of the target model, where the nodes (human-interpretable variables) are the proxy models' features. This framing not only lets us test alignment between the mechanisms identified in proxy models and the target model, but also provides an optimisation objective for training faithful proxies. Additionally, a theoretical framework that upper-bounds how much internal computation can differ between the target model and proxy model under a given reconstruction performance can also be a promising direction.

\subsection{Transferability Beyond Transformers}
\label{sec:non-transformer}

The methods surveyed in this paper have been developed and evaluated predominantly in transformer-based autoregressive language models. Although the six research orientations discussed in Section~\ref{sec:orientations} are largely architecture-agnostic in their aims, their classic methods and representative works focus almost entirely on transformers. Even though foundational circuits work~\citeem{olah2020zoom} originated in convolutional networks (InceptionV1; \citeestalp{szegedy2015going}), demonstrating interpretable features of visual neurons, many of the techniques that define modern mechanistic interpretability, which include feature extraction and explanation, model editing, and circuit discovery, were developed specifically for transformer residual streams, MLPs, and attention mechanisms. Thus, an important question arises: are mechanistic interpretability concepts, tools, and findings specific to transformers, or do they reflect more general properties of neural computation across model architectures? Below, we discuss several representative models that are substantially different from transformer-based autoregressive models.

\begin{itemize}
    \item \textbf{Convolutional neural networks (CNNs):} CNNs are a representative vision model architecture~(\citeemalp{oshea2015introductionconvolutionalneuralnetworks}; \citeestalp{NIPS2012_c399862d}). Their fundamental building blocks are convolutional and pooling layers rather than MLP blocks and self-attention. Transformer-specific circuit formulations thus do not transfer directly to CNNs, although circuit analysis itself originated there~\citeem{olah2020zoom}. Convolutions' spatial locality may yield more naturally modular circuits than transformers under suitable formulations, but large-scale studies of superposition, feature extraction, and circuit discovery in modern CNNs and their variants remain limited.

    \item \textbf{State-space and recurrent models:} Architectures such as Mamba~\citeest{gu2024mamba} and RWKV~\citeest{peng-etal-2023-rwkv} replace standard self-attention with recurrent computations that update a hidden state over time. They can be seen as modern recurrent neural network architectures~\citeest{elman1990finding}. Specifically, Mamba uses input-dependent state transitions to selectively retain or forget information, while RWKV uses a recurrence based on linear attention. This new paradigm raises interpretability questions that previous transformer interpretability studies have not addressed, such as how information is stored and updated in recurrent states, whether these representations exhibit certain properties such as superposition, and how causal effects can be traced across state updates.

    \item \textbf{Diffusion models:} Diffusion models~\citeest{pmlr-v37-sohl-dickstein15} apply iterative denoising processes to generate images or tokens. They may still use transformers as backbones, but their computational structure differs fundamentally from autoregressive generation. For example, many current MI studies use the predicted next token as a sign of model behaviour (e.g., ``yes'' or ``no'' when the input is a malicious query), while diffusion-based language models generate multiple tokens in different positions at a time to iteratively complete the response. This difference means many interpretability techniques and insights do not directly apply. While recent work has begun investigating attention patterns and internal representations in diffusion models~(\citeestalp{tinaz2025emergence}; \citeemalp{wang2026dlmscopemechanisticinterpretabilitydiffusion}), mechanistic understanding in this paradigm remains lacking.
\end{itemize}

Beyond these major architectural changes, researchers have also proposed variants of conventional transformer components. For example, mixture-of-experts (MoE) layers~\citeest{lepikhin2021gshard, dai-etal-2024-deepseekmoe} replace standard MLP blocks with multiple smaller MLPs, called ``experts''. For each input activation, a router selects a subset of experts based on the activation and combines their outputs using learned routing weights. Many modern transformer-based language models have adopted this architecture~\citeem{jiang2024mixtralexperts, deepseekai2025deepseekv3technicalreport, yang2025qwen3technicalreport, openai2025gptoss120bgptoss20bmodel, kimiteam2026kimik2openagentic}.

However, MoE architectures introduce technical challenges for mechanistic analysis. For example, activation patching at an intermediate MoE layer may change which experts are selected in downstream MoE layers. This makes it difficult to distinguish the patched activation’s effects on model output from those mediated by changes in downstream expert routing. Furthermore, MoE layers can contain many experts, ranging from 8~\citeem{jiang2024mixtralexperts} to 256~\citeem{deepseekai2025deepseekv3technicalreport} or more~\citeem{deepseekai2026deepseekv4highlyefficientmilliontoken}. Although only a subset of experts is active for each input, the large number of possible expert pathways increases the search space for circuit discovery, making circuit discovery harder to use in modern MoE-based models.

Simultaneously, some recent work suggests that at least a few mechanistic interpretability techniques can transfer outside of transformers. Steering vectors, tuned lenses, knowledge editing, and circuit analysis evaluations have all been adapted, with varying success, to state-space models such as RWKV and Mamba, as well as to vision-language models~(\citeestalp{paulo2024does, sharma2024locating, haon2025mechanistic, neo2025towards}; \citeemalp{ensign2024investigating}). Researchers have also conducted theoretical analyses aiming to clarify the computational relationships between transformers and other architectures. Some initial results suggest that several findings in transformers may reflect more general, transferable properties of neural networks~\citeest{zimerman2025explaining, dao202mamba2, peng2025rwkv, ali-etal-2025-hidden}. Sorting out which aspects of mechanistic interpretability depend on architecture and which reflect broader principles of neural computation thus appears to be a promising research direction. On the other hand, whether current research orientations and associated techniques remain relevant beyond transformers will also show how broadly mechanistic interpretability research can apply to future systems.

\section{Conclusion}
\label{sec:conclusion}

Neural networks could be one of the most consequential artefacts humanity has built, yet we do not understand how they work or reliably control what they do. Recent decades have seen an exponential improvement in AI capabilities, but methods to understand and explain internal mechanisms have lagged behind. This has created an interpretability gap between what models can do and what we know about how they do it.

Mechanistic interpretability offers a path towards understanding these systems by studying their internal representations and computations, with the aim of evaluating, predicting, and controlling behaviour. In this survey, we examine closely what interpretability can actually contribute to AI safety. We place mechanistic interpretability in the broader context of interpretability research and compare its mechanistic evidence with behavioural evaluation. Beyond surveying existing methods, we propose a unified framework for evaluating mechanistic interpretability claims and characterise six research orientations by their goals, assumptions, and trade-offs.

Rather than converging on a single approach, the field has crystallised into divergent research orientations, including Pragmatic Interpretability, Ambitious Interpretability, Developmental Interpretability, Interpretability Across Time, Introspective Interpretability, and Automated Interpretability, each making trade-offs between completeness, cost, and applicability.

\subsection{Achievements and Limits}

Mechanistic interpretability has achieved substantial progress across several fronts, including extracting interpretable features with SAEs, discovering task-relevant circuits, and enabling targeted interventions on model behaviour. Overall, the field has shaped how researchers view model computation, representations, and causal structure.

Yet current approaches are limited: they do not fully explain many behaviours, struggle to generalise under distribution shift, have difficulty cleanly disentangling entangled computations, and are hard to scale to frontier models. More importantly, their usefulness for AI safety remains debated.

Although many prior studies in the interpretability community assume that mechanistic understanding directly implies safer systems, the assumption is in fact largely untested. Therefore, this survey considers not only what a method reveals about a model's internal computation, but also what it allows us to do in practice. We particularly discuss whether it supports safety-relevant tasks such as detecting deceptive behaviours or identifying dangerous capabilities, and whether those tasks can be achieved just as well without interpretability methods or insights.

A persistent challenge in mechanistic interpretability is the lack of standardised evaluation. Different interpretability methods use incompatible metrics. To address this, we present a unified evaluation framework covering five dimensions---attribution faithfulness, completeness, stability, causal validity, and human interpretability---where different methods can be jointly tested. We identify gaps and argue that the evaluation methodology and safety benchmarks are the field's most pressing open problems.

Mechanistic interpretability is not a panacea for AI safety. Researchers, developers, and practitioners should recognise the limits of partial or incorrect causal understanding, rigorously evaluate MI tools under broad conditions, and combine them with complementary safety measures. MI should therefore be viewed as one component of a defence-in-depth safety strategy rather than a standalone guarantee of safety.

\subsection{The Road Ahead}

The six research orientations describe different research goals and priorities. At a broader level, mechanistic interpretability can be organised into three complementary paradigms that encompass these orientations and other research directions:

\begin{enumerate}
    \item \textbf{Post-hoc explanation:} The traditional paradigm of studying trained models' internals via tools like SAEs, circuit discovery, activation verbalisers, and causal analysis. Because deployed systems still rely on conventional dense architectures without intrinsic interpretability, post-hoc analysis will remain important in the medium term.

    \item \textbf{Studying learning dynamics:} Understanding how mechanisms emerge during training via methods like singular learning theory and checkpoint analysis. This paradigm enables understanding, prediction, and intervention early in behaviour and capability emergence.

    \item \textbf{Interpretability by design:} Engineering architectures that are mechanistically transparent by construction, such as SoLU, weight-sparse transformers, KANs, and their successors. This paradigm moves interpretability upstream and promises to reduce the need for expensive post-hoc analysis.
\end{enumerate}

Each approach tackles a different area but complements the others. For instance, interpretability-by-design models can reduce reliance on post-hoc methods, while post-hoc mechanistic analysis can guide next-generation interpretability-by-design models towards desirable properties that current dense architectures lack. Insights into learning dynamics can suggest architectural constraints and training-time interventions.

Progress also depends on establishing stronger scientific standards. Recent work argues that interpretability should produce claims that are both falsifiable and reproducible, and improve its use in model evaluations and downstream applications.

Ultimately, the question remains: Does understanding how models work actually make them safer? This survey's goal is to understand how these approaches can work together to improve model understanding and safeguards, while recognising the limits of partial or incorrect causal understanding.

\section*{Acknowledgments}
FB and TW would like to thank Leo Gao, Peter Hase, John Hewitt, Kyle Lo, Aaron Mueller, Neel Nanda, and Avi Semler for their thoughtful comments and our discussions that helped shape the project.

\newpage
\bibliography{tmlr}
\bibliographystyle{tmlr}

\newpage
\appendix

\section{Notation Reference and Symbol Table}
\label{app:notation}
This section provides a unified reference for mathematical notation used
throughout the paper. It is not exhaustive: symbols that appear only
within a single section are defined where they are introduced and may not be
listed here. Where a symbol is overloaded across subfields, the
section-specific meaning is noted. All vectors are column vectors throughout the paper.

\subsection{Dimensions and Indices}
\begin{itemize}
    \item $d$: residual-stream dimension; equivalently model hidden size
    \item $d_{\text{head}}$: attention head dimension
    \item $d_{\text{mlp}}$: MLP hidden dimension, typically $d_{\text{mlp}} \gg d$
    \item $N_H$: number of attention heads per layer
    \item $d_{\text{sparse}}$: dictionary size of an SAE or transcoder,
          $d_{\text{sparse}} = e\,d$ with integer expansion factor $e$
    \item $L$: number of layers (model depth); $\ell \in \{1,\dots,L\}$ indexes layers
    \item $T$: sequence length (context length); $t \in \{1,\dots,T\}$ indexes token positions
    \item $O(\cdot)$: Big-O asymptotic complexity notation
\end{itemize}

\subsection{Inputs and Activations}
\begin{itemize}
    \item $x$: an input prompt or a set of input prompts
    \item $h^{1:T}_\ell \in \mathbb{R}^{d\times T}$: residual-stream activations at layer $\ell$ for token positions $1$ to $T$, whose $t$-th column is $h^t_\ell$
    \item $h_\ell^t \in \mathbb{R}^{d}$: residual-stream activation at layer $\ell$ for token position $t$
    \item $h_\ell \in \mathbb{R}^{d}$: residual-stream activation at layer $\ell$ with token position omitted
    \item $h \in \mathbb{R}^{d}$: residual-stream activation with both layer and position omitted
    \item $h'_\ell \in \mathbb{R}^{d}$: mid-block residual stream (post-attention,
          pre-MLP) at layer $\ell$
    \item $h_\ell^{\text{in}} \in \mathbb{R}^{d}$, $h_\ell^{\text{out}} \in \mathbb{R}^{d}$:
          input and output of a model component, e.g., an MLP block, at layer $\ell$
    \item $\hat{h} \in \mathbb{R}^{d}$: reconstruction of activations $h$ by a dictionary model
    \item $\tilde{h} \in \mathbb{R}^{d}$: activation after an intervention
          (steering, ablation, patching, etc.)
\end{itemize}

\subsection{Dictionary Learning}
\begin{itemize}
    \item $E : \mathbb{R}^{d} \to \mathbb{R}^{d_{\text{sparse}}}$: SAE encoder
    \item $D : \mathbb{R}^{d_{\text{sparse}}} \to \mathbb{R}^{d}$: SAE decoder
    \item $z \in \mathbb{R}^{d_{\text{sparse}}}$: sparse feature activations; $z^i \in \mathbb{R}$ is the activation of feature $i$; when a layer index is also present, we write $z_\ell^i$ for feature $i$ at layer $\ell$
    \item $f_j \in \mathbb{R}^{d}$: feature direction in activation space
          (the $j$-th column of $W_{\text{out}}$)
    \item $W_{\text{in}} \in \mathbb{R}^{d_{\text{sparse}} \times d}$,
          $b_{\text{in}} \in \mathbb{R}^{d_{\text{sparse}}}$: encoder weights and bias
    \item $W_{\text{out}} \in \mathbb{R}^{d \times d_{\text{sparse}}}$,
          $b_{\text{out}} \in \mathbb{R}^{d}$: decoder weights and bias
    \item $k$: number of active features in a Top-$k$ SAE
\end{itemize}

\subsection{Model Parameters}
\begin{itemize}
    \item $W_Q^r, W_K^r, W_V^r \in \mathbb{R}^{d_{\text{head}} \times d}$, $W_O^r \in \mathbb{R}^{d \times d_{\text{head}}}$: query, key, value, and output projection matrices of the $r$-th attention head
    \item $W_1 \in \mathbb{R}^{d_{\text{mlp}} \times d}$, $W_2 \in \mathbb{R}^{d \times d_{\text{mlp}}}$: MLP weight matrices
    \item $b_1 \in \mathbb{R}^{d_{\mathrm{mlp}}}$,
          $b_2 \in \mathbb{R}^{d}$:
          MLP bias vectors
    \item $\theta$: the collection of all trainable model parameters
\end{itemize}

\subsection{Interventions and Objectives}
\begin{itemize}
    \item $v_{\text{steer}} \in \mathbb{R}^{d}$: steering vector, applied as
          $h + \alpha\, v_{\text{steer}}$
    \item $\alpha$: scaling factor for an intervention
    \item $\mathcal{L}$: loss function
    \item $\lambda$: sparsity or regularisation coefficient. In
          Appendix~\ref{app:slt}, $\lambda(\cdot)$ instead denotes the local
          learning coefficient
\end{itemize}

\subsection{Mathematical Operations}
\begin{itemize}
    \item $p(y \mid x)$, $\log p(y \mid x)$: conditional probability and log probability
    \item $\mathbb{E}[\cdot]$: expectation over the data distribution;
          $\mathbb{E}[\cdot \mid \text{condition}]$ denotes a conditional expectation
    \item $\|v\|$: Euclidean norm; $\|z\|_1$ and $\|z\|_0$ denote the
          $L_1$ and $L_0$ norms
    \item $\langle v, w \rangle$: inner product
    \item $\odot$: element-wise product
    \item $\otimes$: Kronecker product
    \item $\sigma(\cdot)$: activation function
\end{itemize}

\section{Colour Key and Style Guide}
\label{app:colors}

This section explains the visual conventions used throughout the survey for boxes, emphasis, and structural elements.

\subsection{Box Types and Colours}

\begin{definitionbox}[Definition Box (Dark Blue)]
\textbf{Definition boxes} introduce core concepts and provide precise technical definitions. Use these when a term needs clear specification before use.\\

Example concepts: ``Feature'', ``Circuit'', ``Sparse Autoencoder''.\\

Style: Dark blue border, light blue background.
\end{definitionbox}

\begin{examplebox}[Example Box (Green)]
\textbf{Example boxes} provide concrete illustrations, worked examples, or case studies. Use these to make abstract concepts concrete.\\

Example contents: ``Induction in a Repeated Sequence'', ``Example Application of Steering Vectors'', ``Probing Classifiers for Sleeper Agents''.\\

Style: Green border, light green background.
\end{examplebox}

\begin{insightbox}[Insight Box (Gold)]
\textbf{Insight boxes} explain intuitions behind mathematical or conceptual results. Use these to build understanding of why something works, not just that it works.\\

Example contents: ``Trade-offs Between Variants'', ``The Complexity--Interpretability Trade-off''.\\

Style: Gold border, light gold background.
\end{insightbox}

\begin{warningbox}[Warning Box (Red)]
\textbf{Warning boxes} highlight critical limitations, failure modes, or common misconceptions. Use these to prevent readers from adopting incorrect or unsafe interpretations.\\

Example contents: ``Caveats on CLT Fidelity Claims'', ``The Probe Causality Problem'', ``The Partial Coverage Issue''.\\

Style: Red border, light red background.
\end{warningbox}

\begin{practicebox}[Practice Box (Teal)]
\textbf{Practice boxes} illustrate suggestions that are actionable. These include protocol recommendations, checklists for verifying approaches, and step-by-step decision-making walkthroughs. Use these for practical implementations of procedures.\\

Example contents: ``Guardrail Framework for Automated Interpretability'', ``Validation Procedures for Introspective Models'', ``Recommended Evaluation Protocol''.\\

Style: Teal border, light teal background.
\end{practicebox}

\subsection{Citation Evidence Levels}
\label{app:citation-colors}

This survey cites both peer-reviewed publications and non-peer-reviewed research outputs. To help readers assess the maturity and review status of the evidence, citations are colour-coded:

\begin{itemize}
    \item \textcolor{citeestablished}{\textbf{Orange citations}} denote formally
    published work: peer-reviewed conference and journal papers, and books.

    \item \textcolor{citeemerging}{\textbf{Green citations}}: Research outputs that have not undergone formal peer review, including preprints, technical reports, workshop papers, blog posts, and other emerging resources.
\end{itemize}

Green citations should not automatically be interpreted as lower quality or less reliable. Some originate from leading research organisations and may have undergone substantial internal review. However, they have not been subjected to the formal academic peer-review process and should therefore be interpreted with appropriate caution. Readers are encouraged to verify the current publication status of cited work where relevant.

\section{Mathematical Foundations}
\label{app:math}

This section provides background on some key mathematical frameworks used in mechanistic interpretability.

\subsection{Causal Inference Basics}

Mechanistic interpretability relies on causal reasoning: determining whether a component is \textit{necessary} and \textit{sufficient} for a behaviour.

\subsubsection{Necessity}

A component $S$ is \emph{necessary} for behaviour $B$ if:
\begin{equation}
\text{Remove } S \Rightarrow B \text{ disappears}
\end{equation}

Tested via ablation: measure the effect of removing $S$ on downstream loss or output distribution.

\subsubsection{Sufficiency}

A component $S$ is \emph{sufficient} for behaviour $B$ if:
\begin{equation}
\text{Only } S \text{ is active} \Rightarrow B \text{ occurs}
\end{equation}

Tested via interchange interventions: in the corrupted run with a reference input containing no target information, patch the activation of $S$ with that from a clean run using an input containing the target information, and then verify whether the behaviour emerges.

\subsubsection{Causal Abstraction}

Causal abstraction~\citeest{geiger2021causalabstractionsneuralnetworks} formalises when a high-level description (``this head implements copying'') correctly describes lower-level computation. A high-level abstraction is \emph{causally correct} if interventions on the high-level description produce the same causal effects as interventions on the low-level system. This is formalised via causal diagrams: high-level variables correspond to low-level variables via abstraction maps $\Phi$, and interventions commute:
\begin{equation}
\text{Intervene at high level} \xrightarrow{\Phi} \text{Intervene at low level}
\end{equation}

\subsection{Local Learning Coefficient}
\label{app:slt}

Singular Learning Theory (SLT; \citeestalp{watanabe2009algebraic}) provides a geometric analysis of learning and generalisation in singular statistical models, including neural networks. The Local Learning Coefficient (LLC; \citeestalp{lau2023local}) builds on SLT by localising Watanabe's learning coefficient around a particular parameter configuration, thereby providing a measure of the effective local complexity of a neural network around a local minimum. Below, we introduce the LLC step by step, following the exposition in Section~3 of \citetest{lau2023local}.

We first consider a neural network and a downstream task. For each neural network parameter configuration, say $\theta$, we can evaluate its loss $\mathcal{L}(\theta)$. The task should have an explicitly defined loss function $\mathcal{L}(\cdot)$, such that we can optimise the model parameters until they converge to a local minimum $\theta^*$. Now we choose a sufficiently small closed ball $B(\theta^*)$ centred at $\theta^*$ such that any element in it, i.e., $\theta$, satisfies
\[
\mathcal{L}(\theta^*) \leq \mathcal{L}(\theta).
\]
Within this ball, we further consider the subset where each element has a loss that, while always greater than or equal to the local minimum, falls within a tolerance threshold $\epsilon$. Formally, we define
\[
B(\theta^*, \epsilon)
\coloneqq
\left\{
\theta \in B(\theta^*)
\mid
\mathcal{L}(\theta) - \mathcal{L}(\theta^*) < \epsilon
\right\}.
\]

We can then calculate the \emph{volume} of $B(\theta^*, \epsilon)$. Viewing each element in this set, which corresponds to a parameter configuration, as a point in a high-dimensional parameter space whose dimension is the number of scalar parameters in $\theta$, we have
\[
V(\epsilon)
\coloneqq
\int_{B(\theta^*, \epsilon)} d\theta.
\]
Intuitively, $V(\epsilon)$ measures how much parameter-space volume around $\theta^*$ achieves a loss within $\epsilon$ of the local minimum.

By SLT, as $\epsilon \to 0$, $V(\epsilon)$ has the asymptotic form
\[
V(\epsilon)
=
c\epsilon^{\lambda(\theta^*)}
(-\log\epsilon)^{m(\theta^*)-1}
+
o\!\left(
\epsilon^{\lambda(\theta^*)}
(-\log\epsilon)^{m(\theta^*)-1}
\right),
\]
where $c$ is a positive constant, $\lambda(\theta^*)$ is a positive rational number termed the \emph{Local Learning Coefficient}, and $m(\theta^*)$ is a positive integer called the local multiplicity. Importantly, the equation shows that $\lambda(\theta^*)$ determines the dominant rate at which $V(\epsilon)$ shrinks as $\epsilon \to 0$, while $m(\theta^*)$ contributes a lower-order logarithmic term. The larger $\lambda(\theta^*)$ is, the faster $V(\epsilon)$ decreases as $\epsilon \to 0$.

The convergence speed of $V(\epsilon)$, as characterised by $\lambda(\theta^*)$, can therefore measure the model’s effective local complexity. Conceptually, a large $\lambda(\theta^*)$ means that, when we are \emph{just around} the local minimum (i.e., for small $\epsilon$), the volume of parameter configurations that preserve a similarly low loss is small: perturbing the parameters is more likely to substantially change the loss. This trend corresponds to a local solution with more complexity but less degeneracy. Conversely, a small $\lambda(\theta^*)$ implies that a larger number of nearby parameter configurations reach similarly low loss, indicating lower effective local complexity and higher degeneracy.

The LLC has been used to characterise properties of trained neural networks. For example, \citetest{lau2023local} showed that the LLC can capture differences in effective complexity between models trained under different settings, even when their training losses are similar.

\subsection{Information Theory Basics}

In mechanistic interpretability, concepts from information theory are often used to quantify relationships between internal components and how much information is preserved, and to measure representation quality. Although they do not directly establish causal mechanisms, they are highly useful for quantifying how information is encoded and changes during model computation.

\paragraph{Entropy.} A random variable $X$ has entropy (or information content) that is defined as:

\[
H(X) \coloneqq -\sum_x p(x)\log p(x).
\]

High entropy indicates greater uncertainty, while low entropy implies more predictable distributions. Entropy is frequently utilised to measure learned representation diversity or prediction confidence.

\paragraph{Mutual Information.} The mutual information between two random variables $X$ and $Y$ is

\[
I(X;Y) = H(X)-H(X|Y),
\]

which measures how much information one variable provides about the other. Equivalently,

\[
I(X;Y)=D_{\mathrm{KL}}(p(X,Y)\,\|\,p(X)p(Y)).
\]

In interpretability, mutual information is often used to estimate how strongly a representation encodes a property of interest, such as a concept, label, or downstream behaviour.

\paragraph{KL Divergence.} The Kullback--Leibler (KL) divergence between distributions $P$ and $Q$ is

\[
D_{\mathrm{KL}}(P\|Q)
=
\sum_x P(x)\log\frac{P(x)}{Q(x)},
\]

which measures how different one probability distribution is from another. KL divergence is widely used to compare model outputs before and after interventions, evaluate reconstruction quality, and train auxiliary interpretability models such as tuned lenses.

\clearpage

\end{document}